\documentclass{article}

\usepackage[accepted]{icml2024}

\usepackage[utf8]{inputenc} 
\usepackage[T1]{fontenc}    
\usepackage[hidelinks]{hyperref}       
\usepackage{url}            
\usepackage{booktabs}       
\usepackage{amsfonts}       
\usepackage{nicefrac}       
\usepackage{microtype}      
\usepackage{xcolor}         
\usepackage{enumitem}
\usepackage[pdftex]{graphicx}
\usepackage{grffile}
\usepackage{subcaption}
\usepackage{float}
\usepackage{stfloats}
\usepackage[frozencache,cachedir=.]{minted}

\usepackage{amsmath, amsthm, amsfonts, amssymb}
\usepackage{siunitx}
\usepackage{tikz}
\usepackage{mathtools}
\usepackage[setmargin=true,marginparwidth=0.5in,hide=true]{marginalia}
\usepackage{stmaryrd}
\usepackage{dsfont}

\usepackage{mleftright}
\mleftright

\usepackage[%
        all=normal,
        paragraphs=tight,
        ]{savetrees}

\DeclareRobustCommand{\circled}[1]{%
  \tikz[baseline=(char.base)]{
    \node[shape=circle,draw,inner sep=1pt,outer sep=-10pt] (char) {#1};
  }%
}
\usetikzlibrary{shapes.geometric}
\newcommand{\squared}[1]{%
  \tikz[baseline=(char.base),square/.style={regular polygon,regular polygon sides=4}]{
    \node[square,draw,inner sep=1pt,outer sep=-10pt] (char) {#1};
  }%
}

\def\R{\ensuremath\mathbb{R}}

\DeclareMathOperator*{\argmin}{arg\,min}

\newcommand{\denote}[1]{\ensuremath\llbracket #1 \rrbracket}

\newcommand{\set}[1]{\left\{#1\right\}}

\graphicspath{{./}}

\theoremstyle{plain}

\theoremstyle{definition}

\theoremstyle{remark}

\icmltitlerunning{Learning to Compile Programs to Neural Networks}

\begin{document}

\twocolumn[
  \icmltitle{Learning to Compile Programs to Neural Networks}

\begin{icmlauthorlist}
\icmlauthor{Logan Weber}{mit}
\icmlauthor{Jesse Michel}{mit}
\icmlauthor{Alex Renda}{mit}
\icmlauthor{Michael Carbin}{mit}
\end{icmlauthorlist}

\icmlaffiliation{mit}{MIT CSAIL, Cambridge, MA}

\icmlcorrespondingauthor{Logan Weber}{loganweb@mit.edu}

\icmlkeywords{Hypernetworks, Neural Surrogates, Programs, Compiler}

\vskip 0.3in
]

\printAffiliationsAndNotice{}

\begin{abstract}
  A \textit{neural surrogate of a program} is a neural network that mimics the behavior of a program.
  Researchers have used these neural surrogates to automatically tune program inputs, adapt programs to new settings, and accelerate computations.
  Researchers traditionally develop neural surrogates by training on input-output examples from a single program.
  Alternatively, language models trained on a large dataset including many programs can consume program text, to act as a neural surrogate.
  Using a language model to both generate a surrogate and act as a surrogate, however, leading to a trade-off between resource consumption and accuracy.
  We present \textit{neural surrogate compilation}, a technique for producing neural surrogates directly from program text without coupling neural surrogate generation and execution.
  We implement neural surrogate compilers using hypernetworks trained on a dataset of C programs and find that they produce neural surrogates that are $1.9$-$9.5\times$ as data-efficient, produce visual results that are $1.0$-$1.3\times$ more similar to ground truth, and train in $4.3$-$7.3\times$ fewer epochs than neural surrogates trained from scratch.
\end{abstract}


\vspace{-1em}
\section{Introduction}

A \emph{neural surrogate} is a neural network that models a subset of the observable behavior of a program~\citep{ProgrammingNeuralSurrogates2021}.
Neural surrogates have been used to automatically configure image signal processing units and CPU simulators~\citep{DiffTuneForGraphics2019, DiffTune2020}, improve the accuracy of manufacturing and physics simulations~\citep{InjectionMoldingSurrogateAdaptation2018,InertialConfinementFusionSurrogateAdaptation2020}, accelerate the computer architecture design process~\citep{FastSimulation2006}, and accelerate computations in signal processing, robotics, 3D games, compression, machine learning, and image processing~\citep{Parrot2012}.

\paragraph{Neural Surrogate Training.}
The research community has developed a variety of techniques to train neural surrogates.
The traditional approach is to train a neural surrogate of a single program by collecting and curating a dataset of input-output pairs and then training a neural network to predict the program's output given an input~\citep{ProgrammingNeuralSurrogates2021}.

Another point in the spectrum is to amortize the cost of training neural surrogates by training a \textit{universal neural surrogate}: a neural network that directly consumes the text of a program and predicts the program's output for a given input~\citep{zaremba2015learning,Scratchpad2022,gu2024cruxeval}.
A key benefit of universal neural surrogates is that one only needs to create a dataset once.
Once trained, a universal neural surrogate can act as the neural surrogate of a given program without the need to curate a dataset of program-specific, input-output pairs.

However, universal neural surrogates \NA{necessarily} use the same model to process the program text as is used to predict the program output, and accurate prediction may require multiple forward passes~\citep{Scratchpad2022,wei2022chain}.
These limitations pose challenges for deploying such a model as a neural surrogate because small models may not be able to emulate complex programs~\citep{zaremba2015learning} and large models~\citep{openai2023gpt4} may not be able to execute in the resource-constrained environments where neural surrogates have been used~\citep{Parrot2012,mendis_thesis_2020,munk2022probabilistic}.

\paragraph{Our Approach: Neural Surrogate Compilation.}

To maintain the benefits of universal neural surrogates while bypassing the above limitations, we propose to use a \textit{neural surrogate compiler}.
A neural surrogate compiler is a system that accepts a program's text as input and produce an initial neural surrogate of the program, which can vary in behavioral quality.
Similarly to a traditional compiler, a neural surrogate compiler requires a significant upfront cost that is amortized over the generation of initializations for many neural surrogates.
We demonstrate in this work that when compared to the traditional approach of training a neural surrogate from a random initialization, neural surrogates produced by neural surrogate compilers can be finetuned to closely mimic the behavior of the program at a lower cost, as measured in data efficiency and training time.

\paragraph{Contributions.}
To implement a neural surrogate compiler, we adapt the BERT architecture~\cite{BERTTiny2019} into a \textit{hypernetwork}---a hypernetwork is a neural network that produces the parameters of another neural network~\cite{Hypernetworks2017}.
We name the resulting architecture \textit{\textsc{CompNet}}.

To train neural surrogate compilers, we develop \textsc{ExeStack}, a dataset of $69{,}083$ executable C programs collected from The Stack~\cite{TheStack2022}, a large corpus of source code.
To train \textsc{CompNet}s, we refine \textsc{ExeStack} into \textsc{ExeStackCPN}, a dataset of $37{,}772$ programs that is compatible with our chosen hypernetwork architecture.
We then evaluate neural surrogates initialized via \textsc{CompNet} on \textsc{ExeStackCPN} and \textsc{ParrotBenchCPN}, the latter being a set of benchmarks from prior work in approximate computing~\cite{Parrot2012}.

Surrogates trained from \textsc{CompNet} initializations achieve $1.9$-$9.5\times$ lower error than neural surrogates trained from scratch, with the same amount of data; on a color quantization task, they produce images that are $1.0$-$1.3\times$ more similar to images produced by an exact implementation than images produced by surrogates trained from random initialization; and they achieve a target error with $4.3$-$7.3\times$ fewer epochs than neural surrogates trained from scratch.






\comment{
Recent work investigates whether Transformer-based models can learn to execute programs~\cite{Scratchpad2022,TransformersLearnShortcuts2022,LoopedTransformers2023,TeachingTransformersArithmetic2023}.
In these works, the models act as neural interpreters for programs.
\NA{Our work introduces a complementary question: can Transformer-based models learn to compile programs to neural networks?}
}







\section{Neural Surrogate Compilation}\label{sec:neural_surrogate_compilers}
\begin{figure*}
  \centering
  \includegraphics*[width=0.85\linewidth]{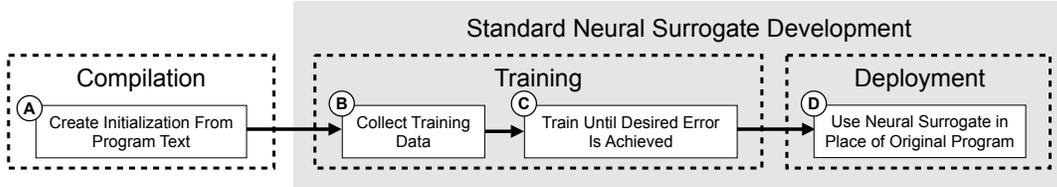}
  \vspace*{-0.75em}
  \caption{Neural surrogate development with neural surrogate compilation}\label{fig:neural_surrogate_compilation_workflow}
  \vspace*{-0.25em}
\end{figure*}

A \emph{neural surrogate compiler} is a system that is specialized to a family of neural surrogate architectures to accept a program's text as input and produce an initial neural surrogate of the program.
Figure~\ref{fig:neural_surrogate_compilation_workflow} presents the neural surrogate compilation workflow alongside the traditional workflow for developing a neural surrogate.
In a traditional neural surrogate development workflow,
one collects training data (\circled{B}), trains the neural surrogate until its \NA{error meets the desired threshold} (\circled{C}), and then uses it in place of the original program (\circled{D}).
Neural surrogate compilation (\circled{A})  introduces a new, initial step in the neural surrogate compilation workflow in which a neural surrogate compiler maps the program text to a neural network initialization for use in the training of the neural surrogate.
The typical strategy to train a neural surrogate is through supervised learning of a neural network with a curated dataset of input-output pairs from the program~\citep{ProgrammingNeuralSurrogates2021}.

In this section, we formalize the problem of efficiently training a neural surrogate and introduce a new approach to solving this problem using a neural surrogate compiler.

\subsection{The Efficient Surrogate Training Problem}
We first formalize the problem of training a neural surrogate.
We assume we are given a program text $p: \mathcal{P}$ that denotes a function $\denote{p} : \mathcal{I}_p \to \mathcal{O}_p$,\footnote{
  $\denote{\cdot} : \mathcal{P} \to (\mathcal{I}_p \to \mathcal{O}_p)$ is notation used in programming language theory to refer to the function a program implements.
} where $\mathcal{P}$ is the space of programs under consideration, $\mathcal{I}_p$ is the type of values $p$ accepts as input and $\mathcal{O}_p$ is the type of values $p$ produces as output.
We also assume a target neural surrogate architecture $a : \mathbb{R}^d \to \mathcal{I}_p \to \mathcal{O}_p$, which takes a set of parameters $\theta : \mathbb{R}^d$ and produces a surrogate function from $\mathcal{I}_p$ to $\mathcal{O}_p$.
The goal is to find a set of parameters $\theta : \mathbb{R}^d$ such that the neural surrogate $a(\theta) : \mathcal{I}_p \to \mathcal{O}_p$ has low approximation error: \[
  \forall i : \mathcal{I}_p.\,
  a(\theta)\left(i\right) \approx \denote{p}\left(i\right)
\]
To measure the quality of surrogate outputs, we use a loss function $\ell : \mathcal{O}_p \times \mathcal{O}_p \to \mathbb{R}_{\geq 0}$ that measures the difference between the output of the program and the output of the surrogate.
To measure overall surrogate quality, we use the expected loss over a distribution of inputs: 
\begin{equation}
  \mathcal{L}(a(\theta), p)= \mathbb{E}_{i \sim \mathcal{I}_p}[\ell(a(\theta)(i), \denote{p}(i))]  
  \label{eq:expected_loss}
\end{equation}
As with most learning problems, a challenge in training neural surrogates is that the error of a surrogate depends on the budget dedicated to collecting training data (input-output pairs of the program) and the number of epochs used to train the surrogate.
We formalize these costs by defining a \emph{training procedure} $t_a : \mathcal{P} \times \mathbb{R}_{\geq 0} \times \mathbb{N} \to \R^d$ for a given surrogate architecture $a$ as a random function that takes program text $p$, a training data budget $b : \R_{\geq 0}$, and training time budget $n : \R_{\geq 0}$ and produces a set of parameters $\theta : \R^d$ for the surrogate.

We then define the \emph{efficient surrogate training problem} as finding a training procedure $t_a$ for a given program $p$, architecture $a$, sample budget $b$, training time budget $n$, and loss function $\ell$ that minimizes the expected loss of the resulting surrogate: \[
  \argmin_{t_a}
  \mathbb{E}_{\theta \sim t_a(p, b, n)}
  \left[
    \mathcal{L}(a(\theta), p)
  \right],
\]
The standard approach to training a neural surrogate is to randomly initialize the parameters of the surrogate and then use a gradient-based optimization algorithm to minimize the loss against a dataset of input-output pairs from the program~\citep{ProgrammingNeuralSurrogates2021}.

\subsection{Neural Surrogate Compilation}

A neural surrogate compiler is a system $\phi: (p: \mathcal{P}) \to \R^{d_p}$ that accepts program text $p$ and produces parameters $\theta \in \R^{d_p}$ for a neural surrogate architecture $a_p$ depending on the program $p$.
We use a neural surrogate compiler to solve the efficient surrogate training problem.

We formalize the development of a neural surrogate compiler as an optimization problem.
The goal is to develop a system $\phi$ such that for every program $p$, the surrogate $f = a_p\left(\phi\left(p\right)\right)$ can be trained efficiently.
Optimizing for a system that generates surrogates that can be trained efficiently is challenging.
As a simple proxy, we optimize for a system that generates surrogates that achieve low loss:
\[
  \vspace{0em}
  \argmin_{\phi \in \mathcal{P} \to \R^d} \mathbb{E}_{p \sim \mathcal{P}}\left[
    \mathcal{L}(a_p\left(\phi\left(p\right)\right), p)
    \right].
\]
\\
\vspace{-3em}

\section{\textsc{CompNet}}\label{sec:compnet}
The \textsc{CompNet} architecture is an implementation of a neural surrogate compiler using hypernetworks.
We explain the architecture, how to train it, then how to extract neural surrogates from its outputs.

\begin{figure*}[ht]
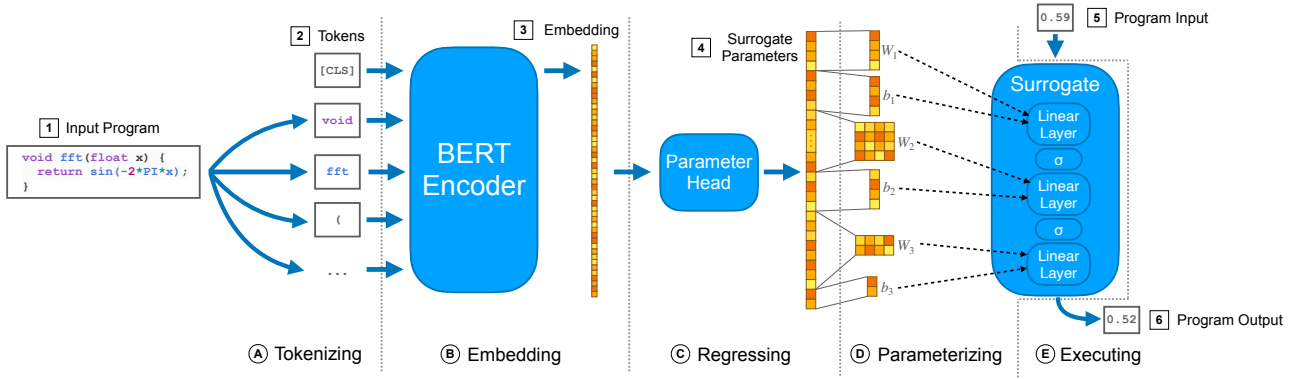

  \begin{center}
  \includegraphics[width=0\linewidth]{figures/hybertnet_system_diagram.pdf}
  \includegraphics[width=\linewidth]{figures/hybertnet_system_diagram.pdf}
  \end{center}
  \caption{
    System diagram describing the \textsc{CompNet} architecture, comprising five phases:
    (A) tokenizing an input program,
    (B) embedding the program using a BERT encoder,
    (C) regressing the embeddings to a parameter vector using a parameter head,
    (D) parameterizing a neural network using the parameter vector, and
    (E) executing the neural network surrogate.
  }\label{fig:hypernet_arch}
\end{figure*}

\subsection{Architecture}
Figure~\ref{fig:hypernet_arch} presents the design of a \textsc{CompNet}.
A \textsc{CompNet} accepts program text $p: \mathcal{P}$ as input and produces parameters $\theta \in \R^{d}$ for a neural surrogate architecture $a : \R^d \to \mathcal{I} \to \mathcal{O}$ with as many inputs as the largest architecture one wishes to compile to and a single output.
We call this architecture a \textit{covering architecture}.


\circled{A} First, \textsc{CompNet} tokenizes an input program (\squared{1}), resulting in a sequence of tokens (\squared{2}) including the distinguished BERT \emph{classification token} \texttt{[CLS]}.

\circled{B} \textsc{CompNet} then uses a BERT encoder~\citep{BERT2019} to embed the sequence of tokens, resulting in an embedding per token.
The output of this step is the embedding of the classification token (\squared{3}); \textsc{CompNet} discards the embeddings of the other tokens.

\circled{C} Next, \textsc{CompNet} uses a \textit{parameter head}, implemented as a single linear layer, to map the classification token embedding to a neural surrogate parameter vector~(\squared{4}).

\circled{D} Then, \textsc{CompNet} interprets the vector of parameters as the weights and biases of the covering architecture.
The output of this step is a neural surrogate of the input program.

\circled{E} Finally, \textsc{CompNet} executes the neural surrogate with the interpreted parameters on a program input (\squared{5}) to produce a prediction of the program output (\squared{6}).

\subsection{Training}\label{sec:training}
Training a \textsc{CompNet} requires a dataset of programs and input-output pairs for each program.
Note that this dataset is not considered as part of the budget in the efficient surrogate training problem, since it is amortized over all programs the \textsc{CompNet} is used to compile.


Each step of training proceeds by selecting a batch of programs and input-output pairs for those programs, generating neural surrogate parameters for each program, interpreting the neural surrogate parameters as parameters for the covering architecture, executing each neural surrogate with the batch of inputs, then calculating the loss between the neural surrogates' predicted outputs and the true outputs.
To match the signature of the covering architecture, the batch of inputs is padded out to match the number of inputs for the covering architecture (e.g., if a covering architecture has $9$ inputs and a program has $3$ inputs, the compiled architecture for that program is fed $9$ inputs).
For padding, we use inputs drawn from the same distribution as the program inputs (see Appendix~\ref{sec:padding_for_variable_inputs} for details).

Backpropagation proceeds as usual, except that one does not update the parameters of the neural surrogates, since each generated neural surrogate is ephemeral.
Instead, backpropagation only updates the parameters of the \textsc{CompNet}.
Appendix~\ref{sec:compnet_training_details}~contains~additional~training~details.

\begin{figure*}[ht]
  \centering
  \includegraphics*[width=\textwidth]{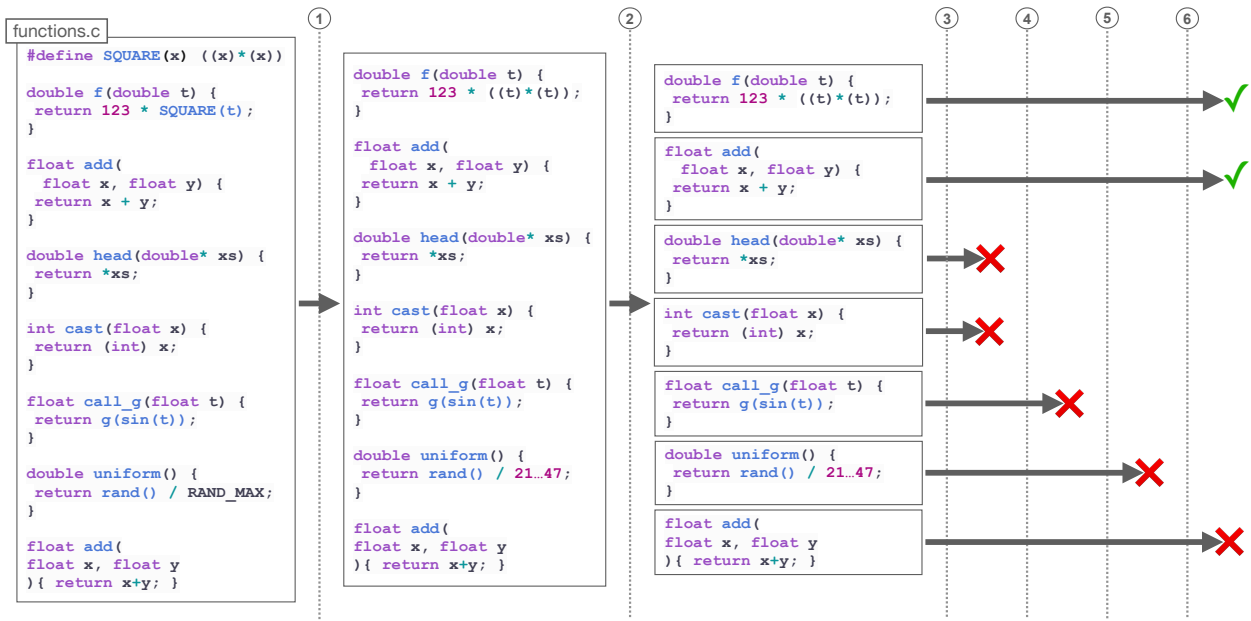}
  \vspace{-2.6em}
  \caption{
    The \textsc{ExeStack} generation pipeline.
    Starting with C source files from The Stack, we apply a sequence of maps followed by a sequence of filters.
    The steps are \circled{1} run the C preprocessor, \circled{2}  extract functions from the source file, \circled{3} remove functions with pointers in their type signature and nonnumeric functions, \circled{4} remove nonexecutable functions and collect input-output pairs, \circled{5} remove nondeterministic functions, and \circled{6} remove any duplicate programs.
    Red ``X''s denote that a function does not pass a filter and green checkmarks denote that a function passes all filters.
  }\label{fig:exestack_creation}
  \vspace{-1em}
\end{figure*}

\subsection{Surrogate Extraction}\label{sec:surrogate_extraction}
The output of a \textsc{CompNet} is parameters for the covering architecture, which might not match the number of inputs and outputs of the program being compiled.
To adapt the covering architecture to the target number of inputs, one finetunes the resulting architecture on data where the excess inputs are set to zero, allowing one to then remove the weights in the input layer corresponding to the excess inputs (see Appendix~\ref{sec:padding_for_variable_inputs} for details on this choice).
To adapt the covering architecture to the target number of outputs, one clones the weights for the single output in the output layer for each new output that is needed (see Appendix~\ref{sec:variable_output_strategies} for details on this choice).
To align the program text with the training distribution (i.e., single-output programs), one also modifies the input program to produce a single output (e.g,. the first output of the original program).
When neither the number of inputs nor the number of outputs matches the covering architecture, all of the above modifications are applied in the same finetuning run.

\section{\textsc{ExeStack}}\label{sec:exe_stack}

The strategy we presented in Section~\ref{sec:compnet} for learning a neural surrogate compiler requires a dataset of programs and input-output examples describing the behavior of each program.
To meet this requirement, we developed \textsc{ExeStack}, a dataset of $69{,}083$ pointer-free, numerical, executable, deterministic C programs and corresponding input-output examples.
\textsc{ExeStack} is based on The Stack~\cite{TheStack2022}, a dataset of 3 TB of permissively licensed source code written in various programming languages scraped from GitHub.

Figure~\ref{fig:exestack_creation} summarizes the process of generating \textsc{ExeStack} (see Appendix~\ref{sec:exestack_exec_appendix} for details).
The restriction to pointer-free functions simplifies the \textsc{ExeStack} generation methodology, and yet, a model trained on \textsc{ExeStack} could still handle programs using statically-sized data structures containing numeric data (e.g., arrays), as they can be transformed into functions with a fixed number of arguments.

\section{Evaluation}\label{sec:evaluation}
To evaluate the claim that neural surrogate compilation lowers the development cost of neural surrogates, we answer the following research questions.
\vspace{0.5em}

\textbf{RQ 1:} Does a neural surrogate initialized by a \textsc{CompNet} converge to a lower test loss than a neural surrogate initialized randomly, for a fixed training set size?

\textbf{RQ 2:} Does a neural surrogate initialized by a \textsc{CompNet} produce better results in an application than a neural surrogate initialized randomly, for a fixed training set size?

\textbf{RQ 3:} Does a neural surrogate initialized by a \textsc{CompNet} converge to a target test loss in fewer epochs than a neural surrogate initialized randomly?

Our results demonstrate that \textsc{CompNet}s lead to improvements in data efficiency (Section~\ref{sec:data_efficiency}), perceptual quality (Section~\ref{sec:e2e_results}), and training time (Appendix~\ref{sec:training_time}).







\begin{table*}
  \small
  \centering
  \renewcommand{\arraystretch}{1.4}
  \begin{tabular}{lp{3cm}p{3.65cm}p{2.6cm}ll}
  \toprule
      Benchmark & Description & Train Inputs & Test Inputs & \#Inputs & \#Outputs \\
      \midrule
      \texttt{fft} & Radix-2 Cooley-Tukey fast Fourier transform & 32,768 random floating point numbers & 2,048 random floating point numbers & 1 & 2 \\
      \texttt{invk2j} & Inverse kinematics for 2-joint arm & 10,000 random (x,~y) coordinates & 10,000 random (x,~y) coordinates & 2 & 2 \\
      \texttt{kmeans} & $k$-means clustering & 50,000 random (r,~g,~b) values & 220x200 color image & 6 & 1 \\
      \texttt{sobel} & Sobel edge detector & One 512x512 color image & 220x200 color image & 9 & 1 \\
      \bottomrule
  \end{tabular}
  \renewcommand{\arraystretch}{1}
  \caption{The programs from \textsc{ParrotBench} we include in \textsc{ParrotBenchCPN}~\cite{Parrot2012}.}\label{tbl:parrot_benchmarks}
  \vspace*{-1em}
\end{table*}

\comment{
We develop \textsc{CompNet}s capable of targeting \textsc{ParrotBenchCPN} (Section~\ref{sec:compnet_development}).
We then use these \textsc{CompNet}s to produce initializations of neural surrogates for \textsc{ExeStack} test programs and \textsc{ParrotBenchCPN} programs, then finetune them to achieve higher accuracy.
We explain our methodology for doing so in Section~\ref{sec:eval_methodology}.
To quantify improvements in the neural surrogate development process, we measure data efficiency and training time.
}

\subsection{Methodology}\label{sec:compnet_development}
To develop and evaluate \textsc{CompNet}s, we select a BERT architecture for the neural surrogate compiler and a multilayer perceptron for the covering architecture, we produce datasets that \textsc{CompNet}s can be trained and evaluated on, we introduce alternative initialization methods to compare against, and we finetune surrogates produced by each of the initialization methods.

\subsubsection{\textsc{CompNet} Architecture}

We use the BERT-Tiny architecture~\cite{BERTTiny2019} for the BERT encoder in \textsc{CompNet}, and we adapt a neural surrogate architecture from~\citet{Parrot2012} into a covering architecture for this \textsc{CompNet}.

The architecture used by~\citet{Parrot2012} is a multilayer perceptron consisting of a single input, a hidden layer of 4 neurons, another hidden layer of 4 neurons, and 2 outputs, and it uses a sigmoid activation function. 
For their evaluation, the authors introduce a suite of benchmarks, \textsc{ParrotBench}, consisting of numerical programs from various domains.
The authors apply their techniques to the architecture above on a fast Fourier transform benchmark in \textsc{ParrotBench} and achieve a $3.6\times$ speedup.
This architecture therefore places a floor on the system speedup that motivates our investigation of Parrot, in that the architectures \citet{Parrot2012} use for all other programs in \textsc{ParrotBenchCPN} are at least as computationally expensive as the one we choose.
We adapt this architecture to take in 9 inputs and produce 1 output, so it can be used to compile programs with up to 9 inputs, and so it is compatible with \textsc{ExeStack}.

As the \textsc{CompNet} loss function, we use mean squared error (MSE) between predicted and true outputs.

\subsubsection{Datasets}

We evaluate the effectiveness of \textsc{CompNet}s on test programs from \textsc{ExeStackCPN} and programs from \textsc{ParrotBenchCPN} (see Table~\ref{tbl:parrot_benchmarks}).
These datasets are refinements of \textsc{ExeStack} and \textsc{ParrotBench} that are compatible with the instantiation of the \textsc{CompNet} architecture described above. 

\paragraph{\textsc{ExeStackCPN}.} 
We produce \textsc{ExeStackCPN} by applying additional filters to \textsc{ExeStack}, resulting in $37{,}772$ programs.
See Appendix~\ref{sec:exestackcpn_generation} for details.

From the full set of programs, we create a training, validation, and testing set using an $80$/$10$/$10$ split.
Each program has input-output examples, so we additionally create a training and testing set for these examples using a $50$/$50$ split.
In Sections~\ref{sec:data_efficiency} and Appendix~\ref{sec:training_time}, we evaluate performance on \textsc{ExeStackCPN} using $1{,}000$ programs from the testing set.

\paragraph{\textsc{ParrotBenchCPN}.}
\textsc{ParrotBenchCPN} programs come from a diverse set of application domains, they are all written in C, each consists of a single function, and they are numeric in nature, making them suitable for evaluating \textsc{CompNet}s.
Table~\ref{tbl:parrot_benchmarks} shows the programs in \textsc{ParrotBenchCPN}, including descriptions of the computations and input datasets.
In Appendix~\ref{sec:parrot_benchmark_details}, we explain how we chose these programs, we list the program source, and we explain how we generated input datasets.

\paragraph{Downcasting Error.}
The covering architecture we chose uses a single-precision floating-point data type, but some programs in \textsc{ExeStackCPN} use double-precision floating-point data types.
In Appendix~\ref{sec:downcasting_error_justification}, we explain why compiling programs with double-precision data types to the single-precision covering architecture incurs negligible error.

\subsubsection{Alternative Initialization Methods}
Besides random initialization, we compare \textsc{CompNet}s to two alternative initialization methods: model-agnostic meta learning~\cite{MAML2017} and pretrained initializations.
Neither initialization method conditions on program text, so they both result in constant initializations that one uses for every program.
We briefly describe these techniques here and how we train them, and we provide shorthand for referencing each initialization method.
In Appendix~\ref{sec:related_work}, we survey related work in this area in detail.

\paragraph{Model-Agnostic Meta Learning.}
Model-agnostic meta learning (MAML) is a meta-learning technique for producing neural network initializations that can be quickly finetuned to achieve low error on a given task.
One trains MAML initializations by sampling tasks from some space of training tasks, finetuning on them, and backpropagating through the finetuning process into the initialization.
\vspace{-0.5em}

\paragraph{Pretrained Neural Surrogates.}
A simpler alternative to MAML is to train a single neural surrogate on the union of all input-output examples from programs in a dataset such as \textsc{ExeStackCPN}.
We call initializations trained in this way \textit{pretrained neural surrogates}.

\paragraph{Training.}
We train 3 instances of each initialization method on \textsc{ExeStackCPN} training programs using the same covering architecture as \textsc{CompNet}s.
See Appendices~\ref{sec:maml_training_details},~\ref{sec:pts_training_details},~\ref{sec:padding_for_variable_inputs},~and~\ref{sec:variable_output_strategies} for details on MAML training, pretrained surrogate training, variable-input support, and variable-output support, respectively.

\paragraph{Initialization Method Shorthand.}
We use shorthand names for each initialization method in figures.
We refer to \textsc{CompNet}s as ``CPN'', MAML as ``MAML'', pretrained surrogates as ``PTS'', and random initialization as ``RND''.

\subsubsection{Finetuning Surrogates}
Here, we collect the finetuning methodology for surrogates in this evaluation, including optimization methods, hyperparameters, random seed behavior, and how we measure the improvements achieved by these surrogates.

For all surrogates produced by the initialization methods we consider, we use the following finetuning methodology.
We use the Adam optimizer with no weight decay, a learning rate of 0.01, and MSE as the loss function.
The only difference between our methodology and the methodology of Esmaeilzadeh et al. \yrcite{Parrot2012} is that we use the Adam optimizer instead of stochastic gradient descent, and we use the He initialization method~\cite{HeInit2015}---they do not specify how they initialize their neural surrogates.

We use $9$ trials with different random seeds for every configuration in the experiments of Section~\ref{sec:data_efficiency} and Appendix~\ref{sec:training_time}.
Note that, for \textsc{CompNet}, MAML, and pretrained surrogate initializations, changing random seeds only changes the training data order, since the initialization is deterministic.

For data efficiency and training time, we quantify results using geometric mean improvements over random initialization.
These are only relative measures, so in Appendix~\ref{sec:absolute_error_justification}, we demonstrate the neural surrogates we train achieve sufficiently low absolute error for downstream applications.

\begin{figure*}[t]
  \centering
  \scalebox{0.8}{
    \begin{tabular}{lllll}
      \toprule
      Statistic & CPN & MAML & PTS \\
      \midrule
      0th & $\num{6.36e-08}\times$ & $\num{4.68e-06}\times$ & $\mathbf{1.35 \cdot 10^{-4}\times}$ \\
      25th & $\mathbf{1.23\times}$ & $0.87\times$ & $0.76\times$ \\
      50th & $\mathbf{5.84\times}$ & $1.17\times$ & $1.28\times$ \\
      75th & $\mathbf{54.36\times}$ & $1.71\times$ & $2.66\times$ \\
      100th & $\mathbf{4.43 \cdot 10^7 \times}$ & $\num{8.52e+03}\times$ & $\num{7.14e+04}\times$ \\
      \midrule
      MPI & \textbf{21}st & 35th & 37th \\
      \midrule
      GM & $\mathbf{9.50\times}$ & $1.09\times$ & $1.08\times$ \\
      \bottomrule
    \end{tabular}
  }
    \hspace{1em}
    \begin{tabular}{lllll}
      \toprule
      Dataset Size & CPN & MAML & PTS \\
      \midrule
      $0\%$ & $\mathbf{84.40\times}$ & $1.42\times$ & $2.63\times$ \\
      $0.1\%$ & $\mathbf{10.43\times}$ & $0.91\times$ & $0.87\times$ \\
      $1\%$ & $\mathbf{2.90\times}$ & $0.51\times$ & $0.90\times$ \\
      $10\%$ & $\mathbf{4.12\times}$ & $1.54\times$ & $0.88\times$ \\
      $100\%$ & $\mathbf{6.67\times}$ & $1.53\times$ & $0.76\times$ \\
      \bottomrule
      \end{tabular}

  \caption{
    Geometric mean test loss improvement over random initialization on $1{,}000$ \textsc{ExeStackCPN} test programs, taken over all programs and dataset sizes (left) and grouped by dataset sizes (right).
    The table on the left reports improvements at a sample of percentiles from 0th (performance that is the worst compared to random initialization) to 100th (performance that is the best compared to random initialization), reports the minimum percentile at which an initialization method improves over random initialization (MPI), and reports overall geometric mean improvements (GM).
  }\label{fig:test_programs_data_efficiency}
\end{figure*}

\begin{figure*}[t]
  \centering
  \scalebox{0.8}{
  \begin{tabular}{lllll}
    \toprule
    Stat. & CPN & MAML & PTS \\
    \midrule
    0th & $0.22\times$ & $\mathbf{0.28\times}$ & $0.23\times$ \\
    25th & $\mathbf{0.88\times}$ & $0.82\times$ & $0.75\times$ \\
    50th & $\mathbf{1.23\times}$ & $0.97\times$ & $0.97\times$ \\
    75th & $\mathbf{2.96\times}$ & $1.14\times$ & $1.26\times$ \\
    100th & $\mathbf{106.91\times}$ & $1.99\times$ & $38.18\times$ \\
    \midrule
    MPI & \textbf{36}th & 54th & 54th \\
    \midrule
    GM & $\mathbf{1.91\times}$ & $0.93\times$ & $1.05\times$ \\
    \bottomrule
  \end{tabular}}
  \hspace{0.5em}
  \begin{tabular}{lllll}
    \toprule
    Program & CPN & MAML & PTS \\
    \midrule
    \texttt{fft} & $\mathbf{1.47\times}$ & $0.98\times$ & $0.61\times$ \\
    \texttt{invk2j} & $1.01\times$ & $\mathbf{1.07\times}$ & $1.05\times$ \\
    \texttt{kmeans} & $\mathbf{7.85\times}$ & $0.68\times$ & $2.24\times$ \\
    \texttt{sobel} & $\mathbf{1.14\times}$ & $1.06\times$ & $0.85\times$ \\
    \bottomrule
  \end{tabular}
  \hspace{0.5em}
  \begin{tabular}{lllll}
    \toprule
    \% Data & CPN & MAML & PTS \\
    \midrule
    $0\%$ & $\mathbf{1.81\times}$ & $0.90\times$ & $1.56\times$ \\
    $0.1\%$ & $\mathbf{1.98\times}$ & $0.94\times$ & $0.98\times$ \\
    $1\%$ & $\mathbf{1.77\times}$ & $0.93\times$ & $0.79\times$ \\
    $10\%$ & $\mathbf{2.38\times}$ & $1.11\times$ & $1.23\times$ \\
    $100\%$ & $\mathbf{1.68\times}$ & $0.81\times$ & $0.86\times$ \\
    \bottomrule
  \end{tabular}
  \caption{
    Geometric mean test loss improvement over random initialization on \textsc{ParrotBenchCPN}, taken over all programs and dataset sizes (left), grouped by programs (middle), and grouped by dataset sizes (right).
    The top table reports improvements at a sample of percentiles from 0th to 100th, reports the minimum percentile at which an initialization method improves over random initialization (MPI), and reports overall geometric mean improvements (GM).
  }\label{fig:parrot_programs_data_efficiency}
\end{figure*}

\subsection{Data Efficiency Improvements}\label{sec:data_efficiency}
To assess whether \textsc{CompNet}s improve data efficiency, we use \textsc{CompNet}s to initialize neural surrogates, finetune on subsets of training data of various sizes, and then compare the results to those of other initialization methods.
We detail the methodology of this experiment then present results.
\vspace{-0.1em}
\subsubsection{Methodology}\label{sec:data_efficiency_methodology}\label{sec:eval_methodology}
We now describe the configurations we sweep over and the methodology we use to finetune surrogates.
\vspace{-0.1em}
\paragraph{Experiment Configurations.}
In this experiment, we sweep over configurations consisting of a program, a dataset size, and an initialization method (e.g., a \textsc{CompNet}).
Each dataset size specifies the percentage of the training data to train neural surrogates on.
We sweep over the following percentages: $\set{0\%, 0.1\%, 1\%, 10\%, 100\%}$.

\vspace{-0.1em}
\paragraph{Dataset Selection.}
Given a configuration consisting of a program, a dataset size percentage $c \in [0, 1]$, and an initialization method, we select a random subset $\mathcal{D}_{\text{sub}}$ of the training data $\mathcal{D}_{\text{train}}$ of size $c|\mathcal{D}_{\text{train}}|$.
We use an $80/20$ split to divide $\mathcal{D}_{\text{sub}}$ into train and validation sets $\mathcal{D}_{\text{sub train}}$ and $\mathcal{D}_{\text{sub val}}$.
We sample $9$ different subsets of this size and use a different training seed for each subset, yielding $9$ trials total.
\vspace{-0.1em}
\paragraph{Finetuning.}
For each trial, we initialize a neural surrogate according to the initialization method.
We then train on $\mathcal{D}_{\text{sub train}}$ for $5{,}000$ epochs.
The final test loss we report for a trial is the test loss at the epoch closest to the epoch with the lowest validation error.\footnote{We only compute test loss before training, after every $3$ epochs of training, and after training.}
When the dataset size is $0\%$, we use the test loss at the final epoch.

\paragraph{Quantifying Improvements.}
We define the improvement for a given configuration (consisting of an initialization method, program, and dataset size) as the ratio of the test loss achieved by random initialization on that configuration and the test loss achieved by that configuration.
We average all test losses over trials and instances of an initialization methods (using arithmetic mean) prior to computing ratios.
For example, we train $3$ instances of \textsc{CompNet}s, and for each instance, we perform $9$ surrogate finetuning trials, so we compute an average over $27$ items.
For each initialization method, we report the geometric mean of the improvements grouped by program, grouped by dataset size, and overall.
For some programs and initialization methods, the resulting surrogates achieve losses of $0$.
We discard these results before computing the geometric mean\footnote{
  We discard $2.4\%$ of entries total for \textsc{ExeStackCPN} programs and $0\%$ of entries total for \textsc{ParrotBenchCPN} programs.
  For \textsc{ExeStackCPN} programs, we discard $3.5\%$ of \textsc{CompNet} entries, $0\%$ of MAML entries, $4.5\%$ of pretrained surrogate entries, and $0\%$ of randomly initialized surrogate entries.
}.

In some figures, we present the improvement at various percentiles---from $0$th to $100$th---as well as the minimum percentile of improvement (MPI).
The percentiles from $0$th to $100$th show the performance that is the worst compared to random initialization up to performance that is the best compared to random initialization.
The MPI is the minimum percentile at which an initialization method improves over random initialization.

\subsubsection{Results}\label{sec:data_efficiency_results}

Figures~\ref{fig:test_programs_data_efficiency} and~\ref{fig:parrot_programs_data_efficiency} show finetuning results for a sample of 1,000 \textsc{ExeStackCPN} test programs and \textsc{ParrotBenchCPN}, respectively.
See Appendix~\ref{sec:data_efficiency_extended} for the test losses used to compute improvements.

\paragraph{\textsc{ExeStackCPN} Test Programs.}

\textsc{CompNet}s achieve the best results on average, with a $9.50\times$ improvement over random initialization, whereas MAML and pretrained surrogates achieve only a $1.09\times$ and $1.08\times$ improvement on average.
\textsc{CompNet}s improve over random initialization in as low as the $21$st percentile of configurations, whereas MAML and pretrained surrogates improve over random initialization after the $35$th and $37$th percentiles, respectively.

\textsc{CompNet}s improve on \textsc{ExeStackCPN} test programs most prominently in the zero-shot regime, where the improvement is $84.40\times$ over random initialization, whereas MAML and pretrained surrogates achieve improvements of $1.42\times$ and $2.63\times$, respectively.
The zero-shot regime is also the only regime where pretrained surrogates show an improvement.
The worst performance for both \textsc{CompNet}s and MAML is in the middle of the dataset sizes we evaluated, at a dataset size of $1\%$, where they achieved $2.90\times$ and $0.51\times$, respectively.
The worst performance for pretrained surrogates, however, is at a dataset size of $100\%$, where they achieve a $0.76\times$ improvement.

\begin{figure*}
  \centering
  \includegraphics[width=0\textwidth]{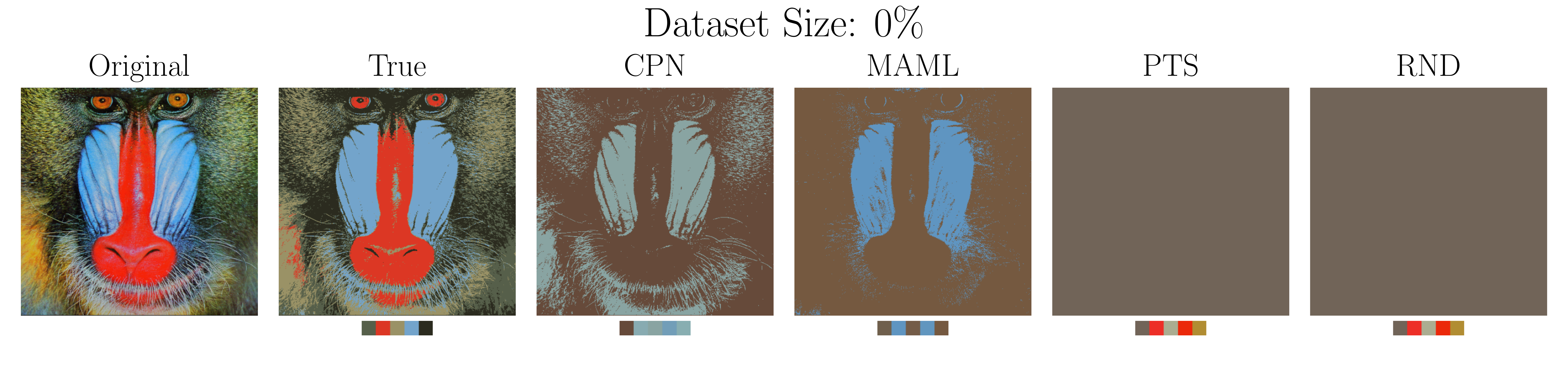}
  \\
  \includegraphics[width=\linewidth]{figures/color_palette/color_quantization_baboon_dataset_size_0_0.pdf}
  \\
  \vspace*{-1em}
  \includegraphics[width=\linewidth]{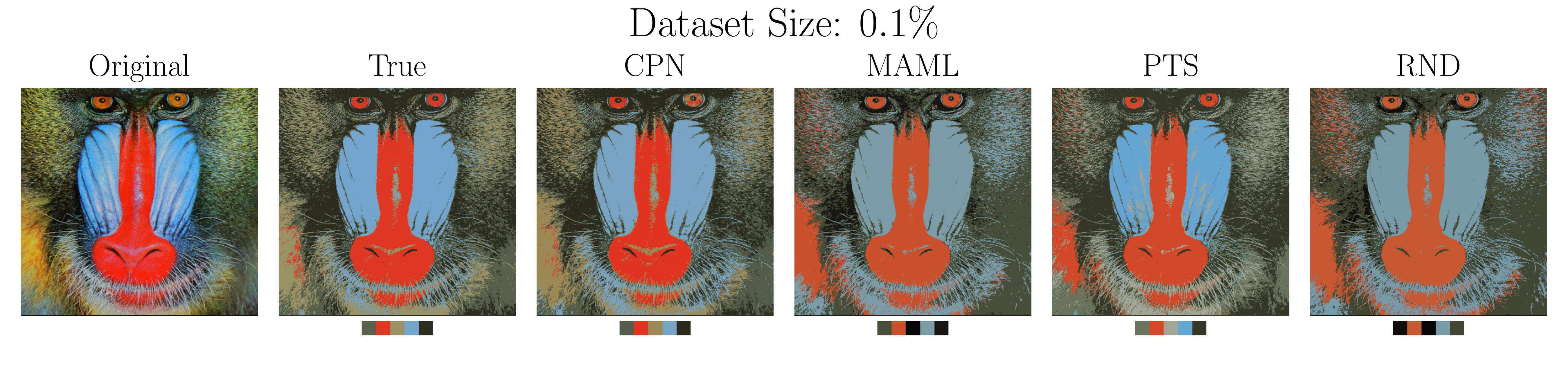}
  \\
  \vspace*{-1em}
  \caption{
  Color quantization results for a ground-truth NumPy implementation (``True'') vs. approximate implementations.
  The original image of a baboon is on the left, followed by images transformed to adhere to a palette of $5$ colors.
  }
  \vspace{-0.5em}
  \label{fig:kmeans-baboon}
\end{figure*}


\paragraph{\textsc{ParrotBenchCPN} Programs.}
\textsc{CompNet}s achieve the best results on average, achieving a $1.91\times$ improvement over random initialization, whereas MAML worsened performance ($0.93\times$) and pretrained surrogates slightly improved performance ($1.05\times$).
\textsc{CompNet}s improve over random initialization in as low as the $36$th percentile of configurations, whereas MAML and pretrained surrogates both improve over random initialization after the $54$th percentile.

\textsc{CompNet}s improve or do not worsen data efficiency on each \textsc{ParrotBenchCPN} program, with the smallest improvement on \texttt{invk2j} ($1.01\times$) and the largest improvement on \texttt{kmeans} ($7.85\times$).
MAML shows the largest improvement on \texttt{invk2j} ($1.07\times$) but worsens performance on \texttt{fft} and \texttt{kmeans}, achieving $0.98\times$ and $0.68\times$, respectively.
Pretrained surrogates show the largest improvement on \texttt{kmeans} ($2.24\times$), but they worsen performance on \texttt{fft} and \texttt{sobel}, achieving $0.61\times$ and $0.85\times$, respectively.

Unlike the results for \textsc{ExeStackCPN}, the improvement due to \textsc{CompNet}s is greatest near the middle of the dataset sizes we evaluated over.
The greatest improvement of $2.38\times$ occurs at $10\%$, and the smallest improvement of $1.68\times$ occurs at $100\%$.
MAML worsens performance at most dataset sizes, except at $10\%$, where it achieves a $1.11\times$ improvement over random initialization.
Pretrained surrogates worsen performance at most dataset sizes except $0\%$ and $10\%$, where they achieve $1.56\times$ and $1.23\times$, respectively.

Since \textsc{CompNet}s improve data efficiency over random initialization on both \textsc{ExeStackCPN} and \textsc{ParrotBenchCPN}, we answer yes to RQ 1.

\begin{figure*}
  \centering

  \begin{tabular}{lllll}
    \toprule
    Dataset Size & CPN & MAML & PTS & RND \\
    \midrule
    $0\%$ & $\mathbf{2.67 \cdot 10^3} \pm 541.$ & $\num{2.79e+03} \pm 347.$ & $\num{3.04e+03} \pm \num{63.}$ & $\num{3.05e+03} \pm \num{0.00}$ \\
    $0.1\%$ & $\mathbf{984.} \pm 733.$ & $\num{1.79e+03} \pm 554.$ & $\num{1.73e+03} \pm 725.$ & $\num{1.43e+03} \pm 544.$ \\
    $1\%$ & $\mathbf{528.} \pm 219.$ & $782. \pm 300.$ & $760. \pm 256.$ & $619. \pm 249.$ \\
    $10\%$ & $\mathbf{452.} \pm 222.$ & $717. \pm 212.$ & $690. \pm 195.$ & $782. \pm 307.$ \\
    $100\%$ & $\mathbf{504.} \pm 220.$ & $766. \pm 189.$ & $699. \pm 171.$ & $655. \pm 121.$ \\
    \bottomrule
  \end{tabular}
  \\[-0.5em]
  \vspace{1em}
  \begin{tabular}{lllll}
    \toprule
    Dataset Size & CPN & MAML & PTS & RND \\
    \midrule
    $0\%$ & $\mathbf{0.33} \pm 0.11$ & $0.26 \pm 0.03$ & $0.25 \pm 0.02$ & $0.25 \pm 0.0$ \\
    $0.1\%$ & $\mathbf{0.61} \pm 0.15$ & $0.45 \pm 0.12$ & $0.47 \pm 0.16$ & $0.53 \pm 0.09$ \\
    $1\%$ & $\mathbf{0.72} \pm 0.12$ & $0.64 \pm 0.11$ & $0.65 \pm 0.10$ & $0.70 \pm 0.11$ \\
    $10\%$ & $\mathbf{0.76} \pm 0.12$ & $0.64 \pm 0.09$ & $0.64 \pm 0.08$ & $0.63 \pm 0.08$ \\
    $100\%$ & $\mathbf{0.73} \pm 0.13$ & $0.62 \pm 0.08$ & $0.64 \pm 0.05$ & $0.65 \pm 0.06$ \\
    \bottomrule
  \end{tabular}


  \caption{
    Quantitative comparison of end-to-end results produced by various initialization methods on color quantization for a $5$-color palette.
    \textbf{(Top)} The average MSE of the image produced by each initialization method compared to the image produced by a ground-truth implementation of the \texttt{kmeans} kernel (lower is better).
    \textbf{(Bottom)} The average SSIM of the image produced by each initialization method compared to the image produced by a ground-truth implementation of the \texttt{kmeans} kernel (higher is better).
  }\label{fig:kmeans}
  \vspace{-1em}
\end{figure*}

\subsection{Neural Surrogates for Color Quantization}\label{sec:e2e_results}
To assess whether a \textsc{CompNet} can improve the quality of results in an end-to-end application, we use a trained \textsc{CompNet} to initialize a neural surrogate used for \textit{color quantization} and compare it to other initialization methods~\cite{Kanungo2002AnEK}.
Color quantization is the process of reducing the number of distinct colors in an image.
For example, Figure~\ref{fig:kmeans-baboon} depicts an image of a baboon color quantizated to five colors.


\subsubsection{Methodology}
We follow the methodology for color quantization from \citet{Kanungo2002AnEK} who apply $k$-means clustering to the (R, G, B) vectors representing the colors of pixels of an image and select the cluster centroids as the colors in the palette.
We run $k$-means clustering for $40$ iterations or until the distance between the old centroids and new centroids is less than $\num{1e-5}$.
Each pixel color is then remapped to the closest color in the palette.

We use the Euclidean distance function to compute the distance between two RGB vectors.
We consider both a reference NumPy implementation and approximate implementations given by neural surrogates of the \texttt{kmeans} kernel in \textsc{ParrotBenchCPN}~\cite{numpy-2020}.

We use surrogates from the data efficiency evaluation of Section~\ref{sec:data_efficiency}.
For visual comparisons, we choose a single surrogate for each dataset size and initialization method.
Since here we evaluate on a distinct image from the testing set of the \texttt{kmeans} kernel, using the testing set as a validation set does not constitute data leakage, so we choose the surrogates with the lowest test losses.
For quantitative comparisons, we aggregate over all surrogates and no selection criterion is necessary.

We quantify the similarity between NumPy-quantized images and surrogate-quantized images using both MSE and the structural similarity index measure (SSIM), the latter of which provides a quantitative model for the percieved similarity of images~\cite{SSIM}.

\subsubsection{Results}
Figure~\ref{fig:kmeans-baboon} depicts the result of applying $5$-color quantization to an image of a baboon using surrogates trained on dataset sizes of $0\%$ and $0.1\%$ of the training set.
Figure~\ref{fig:kmeans} shows quantitative results comparing initialization methods on $5$-color quantization at various dataset sizes.
Each entry shows the average and standard deviation of a metric over all trials and instances of an initialization method.
See Appendix~\ref{sec:kmeans_results_extended} for more dataset sizes and color palette sizes.
\vspace{-0.5em}

\paragraph{Visual Results.}
At a dataset size of $0\%$, \textsc{CompNet}- and MAML-initialized surrogates are the only surrogates that produce images with detail.
The image produced by a \textsc{CompNet} surrogate shows more detail than the image produced by a MAML surrogate, which primarily captures details on the nose.
At a dataset size of $0.1\%$, all initialization methods produce images that resemble the original image.
Images produced by \textsc{CompNet}-initialized and pretrained surrogates have a higher contrast than images produced by MAML-initialized and randomly initialized surrogates.
\vspace{-0.5em}

\paragraph{Quantitative Results.}
At all dataset sizes, \textsc{CompNet}-initialized surrogates have the lowest MSE and the highest SSIM on average.
Among the other initialization methods, there is no consistent winner across dataset sizes.

The variance for the MSE results is comparable across all initialization methods and is high enough that there is overlap among methods.
For example, at a dataset size of $0\%$, an MSE result that is one standard deviation below the mean for MAML is lower than the mean for \textsc{CompNet}s.
However, for all other dataset sizes the mean MSE for \textsc{CompNet}s is lower than the mean MSE for MAML, even after subtracting a single standard deviation.

For the SSIM results, at smaller dataset sizes, the variance is high enough that there is overlap among methods.
At larger dataset sizes though, the results are more clearly separated, with \textsc{CompNet}s having the highest mean SSIM, even when one adds a single standard deviation to the mean SSIM for each of the other initialization methods.

\section{Conclusion}
In this paper, we presented the concept of a neural surrogate compiler and demonstrated how a neural surrogate compiler can be implemented with \textsc{CompNet}s.
We provided a dataset, \textsc{ExeStack}, that one can use to learn neural surrogate compilers.
We demonstrated the effectiveness of \textsc{CompNet}s on \textsc{ExeStackCPN} programs and \textsc{ParrotBenchCPN}, a suite of numerical benchmarks.
Specifically, we showed \textsc{CompNet}-initialized surrogates achieve losses that are $1.9$-$9.5\times$ lower than randomly initialized surrogates, they produce color-quantized images that are $1.0$-$1.3\times$ more similar to images produced by an exact implementation than images produced by randomly initialized surrogates, and they train in $4.3$-$7.3\times$ fewer epochs than randomly initialized surrogates.

The key insight of our work is that a programming language can condition the space of neural network initializations.
In the limit, a neural surrogate compiler could produce initializations requiring no training to achieve low error.
More broadly, neural surrogate compilers could be used to encode programmatically specified behaviors in neural networks, potentially accelerating training for more general tasks.

\section*{Impact Statement}
This paper presents work whose goal is to advance the field of Machine Learning.
There are many potential societal consequences of our work, none of which we feel must be specifically highlighted here.

\section*{Acknowledgements}
We would like to thank Ellie Cheng, Charles Yuan, and the anonymous reviewers for their helpful comments and suggestions.
This work was supported in-part by the National Science Foundation (CCF-1918839 and CCF-2217064) and an Intel Research Fellowship.

\bibliography{refs}
\bibliographystyle{icml2024}

\clearpage
\appendix

\section{Related Work}\label{sec:related_work}
Neural surrogate compilation is inspired by literature on neural surrogates of programs and meta-learning.
In the following sections, we survey these fields and describe other efforts to compile programs to neural networks.

\subsection{Neural Surrogates of Programs}
A common approach to developing neural surrogates of programs is to train a program-specific neural surrogate\footnote{
    In this section, we emphasize when neural surrogates are program-specific, to contrast with universal neural surrogates.
} on a dataset of input-output examples~\cite{ProgrammingNeuralSurrogates2021}, or more recently, to train a universal neural surrogate on a dataset that includes many programs~\cite{zaremba2015learning,Scratchpad2022}.
Our work presents an alternative method for training neural surrogates of numerical programs that maintains the speed of program-specific neural surrogates but incorporates the data efficiency benefits of universal neural surrogates.

\paragraph{Program-Specific Neural Surrogates.}
Researchers across scientific disciplines have used neural surrogates of numerical programs to accelerate computations, adapt to new settings, and enable gradient-based optimization.
\citet{Parrot2012} demonstrate that neural surrogates of numerical programs can improve performance for computations in signal processing, robotics, 3D games, compression, machine learning, and image processing. To accelerate optical metasurface design,~\citet{MetasurfaceDesign} use neural surrogates of numerical simulators and~\citet{PDEsurrogates} use neural surrogates of partial differential equations.~\citet{InjectionMoldingSurrogateAdaptation2018} and~\citet{InertialConfinementFusionSurrogateAdaptation2020} use neural surrogates of numerical simulators for plastic injection molding and inertial confinement fusion, respectively, to facilitate data-efficient finetuning on real physical data.~\citet{ShimaSolarCell} accelerate numerical simulations for solar cells using neural surrogates, and they use transfer learning to quickly adapt neural surrogates when simulator configurations change.~\citet{AtilimSurrogates} use neural surrogates of non-differentiable, numerical physical simulators, to enable gradient-based optimization of simulator parameters.

Researchers have used nonnumerical surrogates to optimize and explore discrete configuration spaces.
\citet{DiffTuneForGraphics2019} and \citet{DiffTune2020} develop neural surrogates of a black-box image signal processing unit and a cycle-accurate CPU simulator, respectively; both techniques enable gradient-based optimization of program inputs, to match some desired input-output behavior.
\citet{HLSSurrogate2020} develop a neural surrogate of a high-level synthesis pipeline for hardware.
Using this surrogate, they lower the cost of predicting the performance and cost of hardware configurations, and they use transfer learning to lower the cost of developing neural surrogates for new configuration spaces.

\paragraph{Universal Neural Surrogates.}
Researchers have developed universal neural surrogates using a variety of architectures.
Early work in this area uses long short-term memory networks to predict the results of executing simple, synthetic Python programs~\citep{zaremba2015learning}.
Later work uses graph neural networks that model program structure in a similar evaluation setup~\cite{LearningToExecGNN2020}.
More recently, researchers have trained Transformer-based models on synthetic datasets of programs or large datasets that include programs~\cite{LLMProgramSynthesis2021, Scratchpad2022,openai2023gpt4,SparksOfAGI2023,gu2024cruxeval}.

\subsection{Meta-Learning}
Meta-learning can improve data efficiency and transfer learning when there is task-agnostic knowledge that can be extracted from a family of tasks~\cite{MetaLearningSurvey2020}.
For example, in the setting we consider, the knowledge of how to execute programs is not specific to any one program but is useful for compiling each program.
We describe the technique we employ, hypernetworks~\cite{Hypernetworks2017}, as well as another meta-learning technique, MAML (model-agnostic meta-learning)~\cite{MAML2017}.
The most noteworthy difference between the two is that, in the former, the parameter space of the meta-learner and the learners differ, whereas, in the latter, these spaces are the same.


\paragraph{Hypernetworks.}
Hypernetworks were first proposed by Ha et al. and achieve state-of-the-art results on sequence modeling tasks~\cite{Hypernetworks2017}.
More recent work by Jin et al. proposes a system, N$^3$, that adapts Transformers to function as hypernetworks 
that condition on text for few-shot learning on image classification tasks~\yrcite{LangToNet2020}.

\comment{
The interface to a neural surrogate compiler is most similar to that of N$^3$, since N$^3$ accepts text as input and produces weights in an offline fashion.
However, they differ in their task specifications---a neural surrogate compiler accepts code as input and N$^3$ accepts natural language descriptions of objects to classify---and they differ in their implementations---our system generates weights directly and their system generates weights as an update to a set of pretrained parameters.
}




\paragraph{Model-Agnostic Meta-Learning.}
MAML is a framework for developing neural network initializations that can be finetuned to new tasks with a small amount of data and a few iterations of SGD~\cite{MAML2017}.
Some authors have noted, however, that MAML couples the task space complexity to the complexity of the individual tasks~\cite{HyperTransformer2022}, making the parameter space a bottleneck as the task space grows.
Our technique does not suffer from this issue because the hypernetwork can be larger than the generated neural surrogate.



\subsection{Compiling Programs to Neural Networks}
There exists prior work on compiling programs to neural networks, though usually as a means of understanding neural network architectures, rather than producing neural surrogates of programs.

Lindner et al. present a compiler, \textit{Tracr}, from the \textit{RASP programming language} to Transformer weights.
The Restricted Access Sequence Processing (RASP) Language is a language with operations developed in analogy to the attention and feedforward operations in a Transformer; notably, RASP is not Turing-complete.
Tracr was designed for the purpose of conducting interpretability experiments and evaluating interpretability methods~\cite{tracr-2023,rasp-2021}.
Since Tracr was not designed with model efficiency in mind, the resulting models are much larger than a roughly equivalent model trained from gradient descent would be, as evidenced by their evaluation.
The \textsc{CompNet} architecture, however, can be trained to target any size of architecture.

Giannou et al. present the \textit{looped Transformer}, a Transformer-based architecture that functions as a programmable computer~\cite{looped-transformers-2023}.
To execute a program, one expresses the program as commands in their instruction set, encodes this sequence of commands as the Transformer input, then executes the Transformer in a loop until it reaches a halt command.
Their instruction set is Turing-complete, and they use it to implement a calculator, linear algebra library, and in-context learning algorithms.
This architecture can be thought of as a universal neural surrogate, in contrast to a neural surrogate compiler.

\section{\textsc{ExeStack} Generation (Extended)}\label{sec:exestack_exec_appendix}
Here we provide a detailed explanation of each step in generating \textsc{ExeStack}, following the flow of Figure~\ref{fig:exestack_creation}.

\paragraph{\circled{1} Preprocessing.}
We pull the functions in \textsc{ExeStack} from files that may contain preprocessor directives, which may affect the ability for these functions to be executed in isolation, if left unexpanded.
We run the C preprocessor on source files until no more lines begin with ``\texttt{\#}'', or we have run it twice, or an invocation fails.

\paragraph{\circled{2} Extracting Functions.}
Recognize and collect all functions from each source file.

\paragraph{\circled{3} Filtering for Pointer-Free Numeric Functions.}
To filter for numeric functions in C programs, we only include C functions that use exclusively \texttt{float} and \texttt{double} data types in the function signature.
Due to the possibility of dynamically sized inputs in the presence of pointers and the ambiguity of whether a pointer represents an input or output, we do not allow pointer types.
Consequently, we also do not allow \texttt{void} as an output type.
If checking a file for the above conditions takes longer than 8 seconds, we discard it.
Note that these filters still allow integral and pointer data types to be used within the function.

\paragraph{\circled{4} Filtering for Executable Functions and Collecting Outputs.}
To simultaneously check for executability and collect outputs from a function, we first generate $2{,}048$ sets of inputs by sampling from the uniform distribution $\mathcal{U}([-1, 1]^n)$, where $n$ is the maximum number of desired inputs, and we use the same sets of inputs for all programs.
We embed these inputs in a C program that includes the function source, as well as an execution harness for collecting outputs.
When a program has fewer inputs than the maximum of $n$, we truncate the inputs we embed to the number of inputs the program has.
When a program has more inputs than the maximum of $n$, we discard it.
We compile the harness with the C standard math library included, since many numerical functions in C make use of this library.
If there are any errors during compilation or execution of a function, we discard the function.
Figure~\ref{fig:exec_harness_example} shows an example of the execution harness instantiated for a function.





\paragraph{\circled{5} Filtering for Deterministic Functions.}
Since a neural surrogate is often a deterministic function of its inputs and weights (e.g., multilayer perceptrons), we filter nondeterministic functions from our dataset.
We check for determinism by running a function 5 times on the same inputs, all sampled from $\mathcal{U}(-1, 1)$, and observing whether the output differs on any execution.
For neural surrogate architectures that are not deterministic, this step can be omitted.

\paragraph{\circled{6} Deduplication.}
We use a whitespace-invariant tokenizer to remove duplicate tokenized programs.

\begin{figure}
\begin{minted}{c}
#include <stdlib.h>
#include <stdio.h>
#include <math.h>
    
float inputs[1024][1] = {
  {0.10740153873327762},
  ...
};

float fftSin_Output0(float x) {
    return sin(-2 * 3.1415 * x);
}

int main() {
    for (int i = 0; i < 1024; i++) {
        float arg0 = inputs[i][0];
        float out = fftSin(arg0);
        printf("%f,", out);
        printf("\n");
    }
    return 0;
}
\end{minted}
\caption{Source code template used for checking executability and collecting outputs, instantiated with the source of the \texttt{fft} kernel in \textsc{ParrotBenchCPN}.}
\label{fig:exec_harness_example}
\end{figure}

\section{\textsc{ExeStackCPN} Generation}\label{sec:exestackcpn_generation}
To produce \textsc{ExeStackCPN}, we apply the following additional filters to \textsc{ExeStack}:
\vspace*{-0.5em}
\begin{itemize}[leftmargin=*,itemsep=0pt]
\item  \textbf{Filtering Long Programs.}
Since BERT-Tiny has a maximum context length of 512 tokens, we remove functions with more than 512 tokens.
We first strip comments from all programs to allow more programs to fit within the context.

\item \textbf{Filtering Large Outputs.}
Large or NaN outputs can lead to training instability for neural networks, so we additionally remove functions with any outputs with an absolute magnitude of 10 or larger or a NaN value.

\item \textbf{Decontaminating Against \textsc{ParrotBenchCPN}.}
It is possible that \textsc{ExeStack} contains similar programs to those in \textsc{ParrotBenchCPN}.
If we trained a \textsc{CompNet} on these programs, improvements over random initialization could be due to memorization.
To address this problem, we remove any programs from \textsc{ExeStack} that are syntactically similar to programs in \textsc{ParrotBenchCPN}.

\end{itemize}
\vspace*{-0.5em}
For the evaluation in Section~\ref{sec:evaluation}, we allow programs with a maximum of $9$ inputs in the execution filter of \textsc{ExeStack}, since this is the number of inputs in the covering architecture we choose.

\begin{figure*}
  \includegraphics*[width=\textwidth]{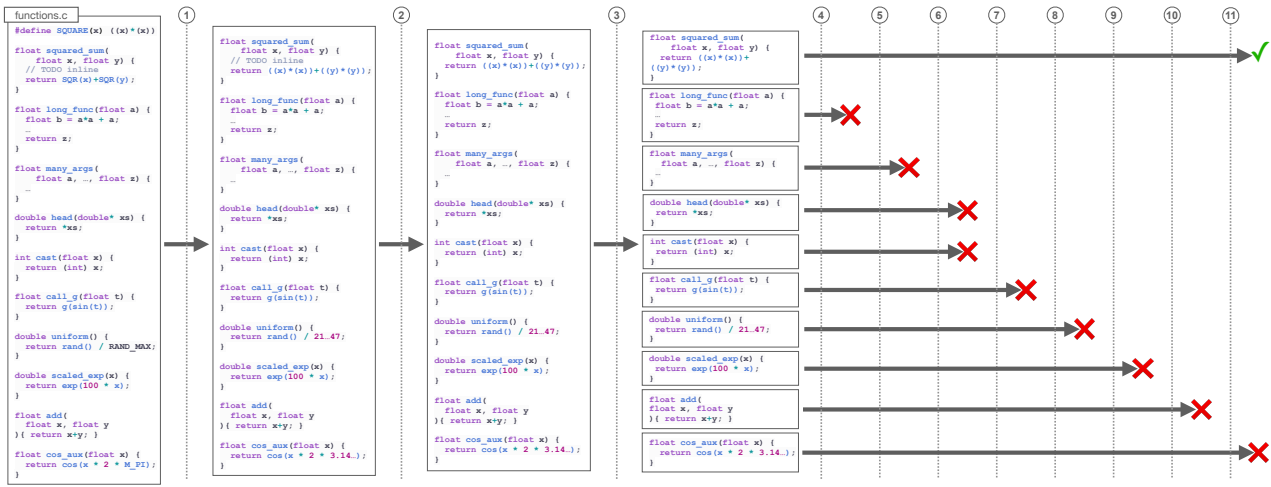}
  \caption{
    The \textsc{ExeStackCPN} generation pipeline (i.e., \textsc{ExeStack} tailored to \textsc{CompNet}s).
    Starting with C source files from The Stack, we apply a sequence of maps followed by a sequence of filters.
    The steps are \circled{1} run the C preprocessor, \circled{2} remove comments, \circled{3} extract functions from the source file, \circled{4} remove functions with more tokens than a user-specified threshold (e.g., the maximum context length), \circled{5} remove functions with more inputs than the target topology, \circled{6} remove functions with pointers in their type signature and nonnumeric functions, \circled{7} remove nonexecutable functions and collect input-output pairs, \circled{8} remove nondeterministic functions, \circled{9} remove functions with any outputs larger than a user-specified threshold, when run on the set of input-output pairs, \circled{10} remove any duplicate programs, and \circled{11} remove any programs syntactically similar to programs in \textsc{ParrotBenchCPN}.
    Red ``X''s denote that a function does not pass a filter and green checkmarks denote that a function passes all filters.
  }\label{fig:exestack_hbn_creation}
\end{figure*}

\begin{figure}[h]
  \centering
  \includegraphics[width=\linewidth]{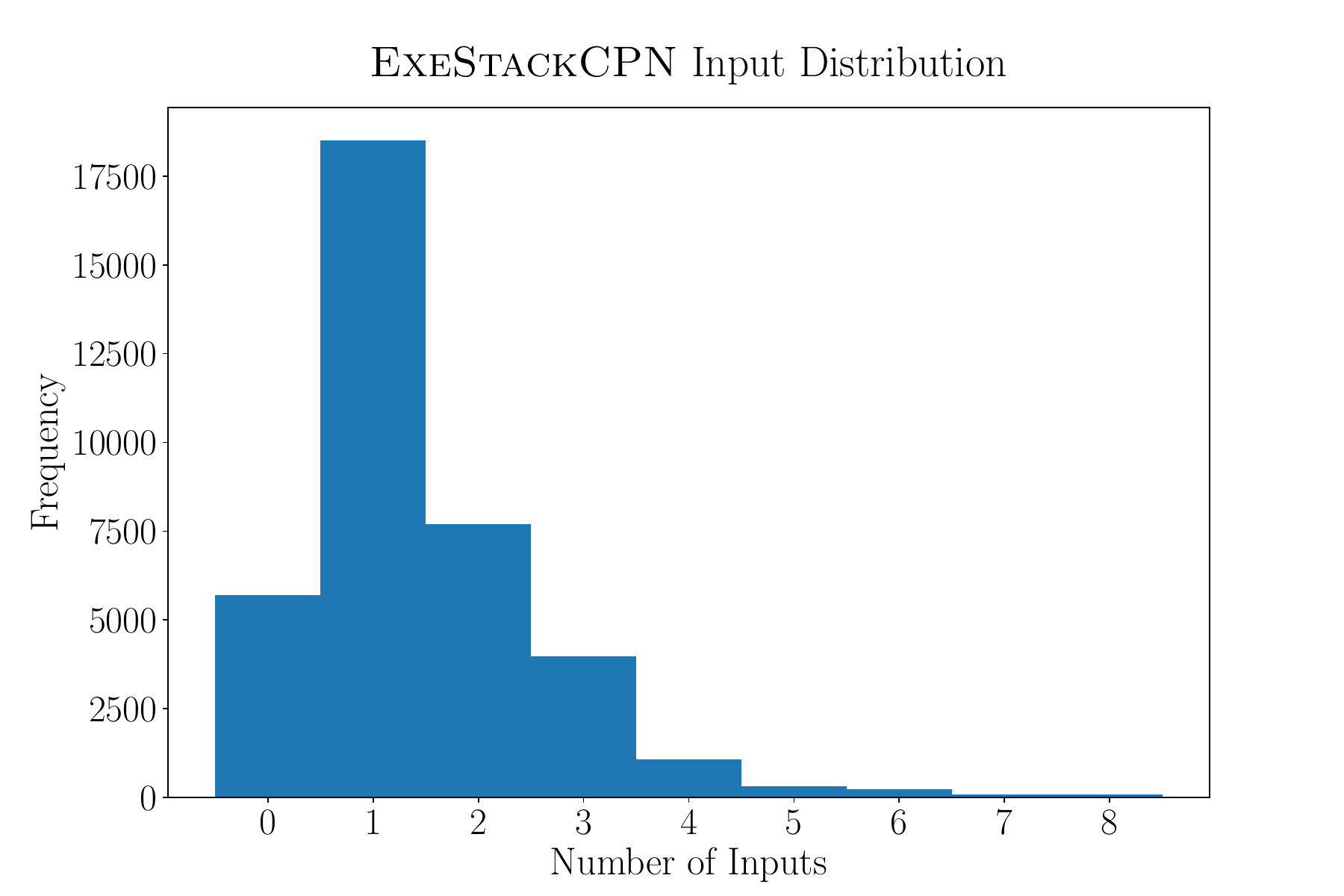}
  \caption{
    Distribution of the number of program inputs for programs in \textsc{ExeStackCPN}.
  }\label{fig:exestack_input_distribution}
\end{figure}

Figure~\ref{fig:exestack_hbn_creation} depicts the entire pipeline for generating \textsc{ExeStackCPN}, Figure~\ref{fig:exestack_cpn_characteristics} shows a summary of the characteristics of \textsc{ExeStackCPN}, and Figure~\ref{fig:exestack_input_distribution} contains a histogram showing the distribution of arity among \textsc{ExeStackCPN} programs.
For the remainder of this section, we detail the decontamination step.

\begin{figure}[h]
  \centering
  \begin{tabular}{ll}
  \toprule
  Characteristic & Value \\
  \midrule
  Max Program Length (In Tokens) & 512 \\
  Tokenizer Vocab Size & 30,522 \\
  \# Programs in Dataset & 37,772 \\
  \# Tokens in Dataset & 1,728,304 \\
  \# I/O Pairs Per Program & 2,048 \\
  \bottomrule
  \end{tabular}
  \caption{
    Summary of \textsc{ExeStackCPN} characteristics.
  }\label{fig:exestack_cpn_characteristics}
\end{figure}

\subsection{\textsc{ExeStackCPN} Decontamination}\label{sec:exestack_hbn_decontam}
To ensure the improvements observed in Section~\ref{sec:evaluation} are not due to memorization, the final step of \textsc{ExeStackCPN} generation is decontamination against \textsc{ParrotBenchCPN} programs.
A prevailing decontamination methodology is to remove any syntactic matches up to whitespace~\cite{Starcoder2022, Starcoder2024}.
Though \textsc{ExeStack} is not contaminated with \textsc{ParrotBenchCPN} programs according to this methodology, we strengthen our methodology to additionally remove syntactically similar programs.
This decontamination consists of bespoke syntactic analyses---one for each \textsc{ParrotBenchCPN} program.
For the remainder of this section, we present each of these syntactic analyses and a sample of the near-duplicate programs they detect.
In total, decontamination removes 375 functions.

\subsubsection{FFT (Output 0)}
Recall, the source for the \texttt{fft (0)} kernel in \textsc{ParrotBenchCPN} is
\begin{minted}{c}
float fftSin_Output0(float x) {
    return sin(-2 * 3.1415 * x);
}
\end{minted}
To decontaminate \textsc{ExeStackCPN} against this program, we search for programs satisfying all conditions below:
\begin{itemize}
    \item Contains ``sin''
    \item Contains either ``3.14'' or ``M\_PI''
    \item Is at most 5 (non-empty) lines long
    \item Has one input
\end{itemize}
This methodology surfaces 29 matches.
Below, we include a sample of 5 of these matches:
\begin{minted}{c}
float seno(float x) {
  return sin(x * M_PI / 180);
}

float exponential(float value) {
  return sin(value * 3.14f / 2);
}

float easeOutSine(float time) {
  return sin(time * M_PI / 2);
}

double sine(double t) {
  return sin(2 * M_PI * t);
}

double cosine(double t) {
  return cos(2 * M_PI * t);
}
\end{minted}

\subsubsection{FFT (Output 1)}
Recall, the source for the \texttt{fft (1)} kernel in \textsc{ParrotBenchCPN} is
\begin{minted}{c}
float fftSin_Output1(float x) {
    return cos(-2 * 3.1415 * x);
}
\end{minted}
To decontaminate \textsc{ExeStackCPN} against this program, we search for programs satisfying all conditions below:
\begin{itemize}
    \item Contains ``cos''
    \item Contains either ``3.14'' or ``M\_PI''
    \item Is at most 5 (non-empty) lines long
    \item Has one input
\end{itemize}
This methodology surfaces 20 matches.
Below, we include a sample of 5 of these matches:
\begin{minted}{c}
float coss(float x) {
  return cos(x * M_PI / 180);
}

double cosine(double t) {
  return cos(2 * M_PI * t);
}

float hamming(float x) {
  return 0.54-0.46*cos(2*M_PI*x);
}

float easeInSine(float time) {
  return 1 - cos(time * M_PI / 2);
}

float easeInOutSine(float time) {
  return 0.5 * (1 - cos(M_PI * time));
}
\end{minted}

\subsection{InverseK2J (Output 0)}
Recall, the source for the \texttt{invk2j (0)} kernel in \textsc{ParrotBenchCPN} is
\begin{minted}{c}
float inversek2j_Output0(
    float x, float y) {
  float l1 = 0.5 ;
  float l2 = 0.5 ;
  float theta2 = (float) acos(
    ((x * x) + (y * y) -
      (l1 * l1) -
      (l2 * l2)) /
    (2 * l1 * l2)
  );
  return (float) asin(
    (y * (l1 + l2 * cos(theta2)) -
      x * l2 * sin(theta2)) /
    (x * x + y * y)
  );
}
\end{minted}
To decontaminate \textsc{ExeStackCPN} against this program, we search for programs satisfying all conditions below:
\begin{itemize}
    \item Contains ``asin'', ``acos'', ``sin'', and ``cos''
    \item Contains either ``.5'' or (``/'' and ``2'')
    \item Is at most 7 (non-empty) lines long
    \item Has two inputs
\end{itemize}
This methodology surfaces 0 matches.

\subsection{InvK2J (Output 1)}
Recall, the source for the \texttt{invk2j (1)} kernel in \textsc{ParrotBenchCPN} is
\begin{minted}{c}
float inversek2j_Output1(
    float x, float y) {
  float l1 = 0.5 ;
  float l2 = 0.5 ;
  return (float) acos(
    ((x * x) + (y * y) -
      (l1 * l1) - (l2 * l2)) / 
    (2 * l1 * l2)
  );
}
\end{minted}
To decontaminate \textsc{ExeStackCPN} against this program, we search for programs satisfying all conditions below:
\begin{itemize}
    \item Contains ``acos''
    \item Contains either ``.5'' or (``/'' and ``2'')
    \item Is at most 6 (non-empty) lines long
    \item Has two inputs
\end{itemize}
This methodology surfaces 0 matches.

\subsubsection{KMeans}
Recall, the source for the \texttt{kmeans} kernel in \textsc{ParrotBenchCPN} is
\begin{minted}{c}
float euclideanDistance(
    float p_0, float p_1, float p_2,
    float c1_0, float c1_1, float c1_2) {
  float r;

  r = 0;
  r += (p_0 - c1_0) * (p_0 - c1_0);
  r += (p_1 - c1_1) * (p_1 - c1_1);
  r += (p_2 - c1_2) * (p_2 - c1_2);

  return sqrt(r);
}
\end{minted}
To decontaminate \textsc{ExeStackCPN} against this program, we search for programs satisfying all conditions below:
\begin{itemize}
    \item Contains ``sqrt'', ``*'', ``+'', and ``-''
    \item Has 6 inputs
\end{itemize}
This methodology surfaces 10 matches.
Below, we include a sample of 5 of these matches:
\begin{minted}{c}
float len(
  float x0, float y0, float z0,
  float x1, float y1, float z1 ){
    return sqrt(
      (x1-x0)*(x1-x0) +
      (y1-y0)*(y1-y0) +
      (z1-z0)*(z1-z0)
    );    
}

float dist(
    float x1, float y1,float z1,
    float x2,float y2,float z2) {
  return sqrt(
    (x1-x2)*(x1-x2) + 
    (y1-y2)*(y1-y2) +
    (z1-z2)*(z1-z2)
  );
}

float calc_dist(
    float x0, float y0, float z0,
    float x1, float y1, float z1) {
  float dx   = (x1 - x0);
  float dy   = (y1 - y0);
  float dz   = (z1 - z0);
  float dist = sqrtf(
    (dx * dx) +
    (dy * dy) +
    (dz * dz)
  );
  return dist;
}

double dist(
    double x0, double y0, double z0,
    double x1, double y1, double z1) {
  return sqrt(
    (x1 - x0) * (x1 - x0) +
    (y1 - y0) * (y1 - y0) +
    (z1 - z0) * (z1 - z0)
  );
}

double dist(
    double ax, double ay, double az,
    double bx, double by, double bz) {
  return sqrt(
    (ax - bx)*(ax - bx) +
    (ay - by)*(ay - by) +
    (az - bz)*(az - bz)
  );
}
\end{minted}

\subsubsection{Sobel}
Recall, the source for the \texttt{sobel} kernel is
\begin{minted}{c}
float sobel(
    float w00, float w01, float w02,
    float w10, float w11, float w12,
    float w20, float w21, float w22) {
  float sx = 0.0;
  sx += w00 * -1;
  sx += w10 * 0;
  sx += w20 * 1;
  sx += w01 * -2;
  sx += w11 * 0;
  sx += w21 * 2;
  sx += w02 * -1;
  sx += w12 * 0;
  sx += w22 * 1;

  float sy = 0.0;
  sy += w00 * -1;
  sy += w10 * -2;
  sy += w20 * -1;
  sy += w01 * 0;
  sy += w11 * 0;
  sy += w21 * 0;
  sy += w02 * 1;
  sy += w12 * 2;
  sy += w22 * 1;

  float s = sqrt(
    sx * sx + sy * sy);
  if (s >= (256 / sqrt(
      256 * 256 + 256 * 256)))
    s = 255 / sqrt(
      256 * 256 + 256 * 256);
  return s;
}
\end{minted}
To decontaminate \textsc{ExeStackCPN} against this program, we search for programs satisfying all conditions below:
\begin{itemize}
    \item Contains ``sqrt'', ``+'', ``*'', and ``/''
    \item Has 9 inputs
\end{itemize}
This methodology surfaces 0 matches.

\section{\textsc{ParrotBenchCPN} Generation}\label{sec:parrot_benchmark_details}

\begin{figure}
  \centering
  \begin{tabular}{lcc}
    \toprule
    Benchmark & Train Inputs & Test Inputs \\
    \midrule
    \texttt{fft} (0) & $32768$ & $2048$ \\
    \texttt{fft} (1) & $32768$ & $2048$ \\
    \texttt{invk2j} (0) & $10000$ & $10000$ \\
    \texttt{invk2j} (1) & $10000$ & $10000$ \\
    \texttt{kmeans} & $50000$ & $48400$ \\
    \texttt{sobel} & $18725$ & $17976$ \\
    \bottomrule
  \end{tabular}
  \caption{
    Number of training and testing inputs for each benchmark in \textsc{ParrotBenchCPN}.
  }\label{fig:parrotbenchshort_dataset_size}
\end{figure}

Here, we present the \textsc{ParrotBench} programs, explain the modifications we made to \textsc{ParrotBench} to produce \textsc{ParrotBenchCPN}, list the resulting source code, and describe how we generate inputs for these programs.




\subsection{\textsc{ParrotBench} Source}\label{sec:parrotbench_code}
The kernels in \textsc{ParrotBench} are \texttt{fft} (Figure~\ref{fig:orig_fft_code}), \texttt{inversek2j} (Figure~\ref{fig:orig_invk2j_code}), \texttt{jmeint} (Figure~\ref{fig:orig_jmeint_code}), \texttt{jpeg} (Figure~\ref{fig:orig_jpeg_code}), \texttt{kmeans} (Figure~\ref{fig:orig_kmeans_code}), and \texttt{sobel} (Figure~\ref{fig:orig_sobel_code}).
We obtained these kernels from \href{https://github.com/he-actlab/AxBench_old}{the AxBench repository}.
For brevity, we have referred to the \texttt{inversek2j} kernel as \texttt{invk2j} throughout this paper.

\begin{figure*}
\centering
\begin{minipage}{0.45\textwidth}
\begin{minted}{c}
void fftSinCos(float x, float* s, float* c) {
    *s = sin(-2 * PI * x);
    *c = cos(-2 * PI * x);
}
\end{minted}
\end{minipage}
\caption{Code for the \texttt{fft} benchmark in \textsc{ParrotBench}.}\label{fig:orig_fft_code}
\end{figure*}

\begin{figure*}
\centering
\begin{minipage}{0.45\textwidth}
\begin{minted}{c}
float l1 = 0.5 ;
float l2 = 0.5 ;

void inversek2j(float x, float y, float* theta1, float* theta2) {
	*theta2 = (float) acos(
    ((x * x) + (y * y) - (l1 * l1) - (l2 * l2)) /
    (2 * l1 * l2)) ;
	*theta1 = (float) asin(
    (y * (l1 + l2 * cos(*theta2)) - x * l2 * sin(*theta2)) /
    (x * x + y * y)) ;
}
\end{minted}
\end{minipage}
\caption{Code for the \texttt{invk2j} benchmark in \textsc{ParrotBench}.}\label{fig:orig_invk2j_code}
\end{figure*}

\begin{figure*}
\centering
\begin{minipage}{0.45\textwidth}
\begin{minted}[fontsize=\tiny]{c}
int tri_tri_intersect(float V0[3],float V1[3],float V2[3],
                      float U0[3],float U1[3],float U2[3])
{
  float E1[3],E2[3];
  float N1[3],N2[3],d1,d2;
  float du0,du1,du2,dv0,dv1,dv2;
  float D[3];
  float isect1[2], isect2[2];
  float du0du1,du0du2,dv0dv1,dv0dv2;
  short index;
  float vp0,vp1,vp2;
  float up0,up1,up2;
  float b,c,max;
  //int r;

  /* compute plane equation of triangle(V0,V1,V2) */
  SUB(E1,V1,V0);
  SUB(E2,V2,V0);
  CROSS(N1,E1,E2);
  d1=-DOT(N1,V0);
  /* plane equation 1: N1.X+d1=0 */

  /* put U0,U1,U2 into plane equation 1 to compute
  signed distances to the plane*/
  du0=DOT(N1,U0)+d1;
  du1=DOT(N1,U1)+d1;
  du2=DOT(N1,U2)+d1;

  /* coplanarity robustness check */
#if USE_EPSILON_TEST==true
  //printf("HERE\n");
  if(fabs(du0)<EPSILON) du0=0.0;
  if(fabs(du1)<EPSILON) du1=0.0;
  if(fabs(du2)<EPSILON) du2=0.0;
#endif
  du0du1=du0*du1;
  du0du2=du0*du2;


  if(du0du1>0.0f && du0du2>0.0f)
  { /* same sign on all of them + not equal 0 ? */
    //*output = 0 ;
    return 0;  /* no intersection occurs */
  }

  /* compute plane of triangle (U0,U1,U2) */
  SUB(E1,U1,U0);
  SUB(E2,U2,U0);
  CROSS(N2,E1,E2);
  d2=-DOT(N2,U0);
  /* plane equation 2: N2.X+d2=0 */

  /* put V0,V1,V2 into plane equation 2 */
  dv0=DOT(N2,V0)+d2;
  dv1=DOT(N2,V1)+d2;
  dv2=DOT(N2,V2)+d2;

#if USE_EPSILON_TEST==true
  //printf("THERE\n");
  if(fabs(dv0)<EPSILON) dv0=0.0;
  if(fabs(dv1)<EPSILON) dv1=0.0;
  if(fabs(dv2)<EPSILON) dv2=0.0;
#endif

  dv0dv1=dv0*dv1;
  dv0dv2=dv0*dv2;
        
  if(dv0dv1>0.0f && dv0dv2>0.0f)
  { /* same sign on all of them + not equal 0 ? */
    //*output = 1 ;
    return 0;  /* no intersection occurs */
  }
\end{minted}
\end{minipage}
\begin{minipage}{0.45\textwidth}
\begin{minted}[fontsize=\tiny]{c}
  /* compute direction of intersection line */
  CROSS(D,N1,N2);

  /* compute and index to the largest component of D */
  max=fabs(D[0]);
  index=0;
  b=fabs(D[1]);
  c=fabs(D[2]);
  if(b>max) max=b,index=1;
  if(c>max) max=c,index=2;

  /* this is the simplified projection onto L*/
  vp0=V0[index];
  vp1=V1[index];
  vp2=V2[index];
  
  up0=U0[index];
  up1=U1[index];
  up2=U2[index];

  /* compute interval for triangle 1 */
  COMPUTE_INTERVALS(
    vp0,vp1,vp2,
    dv0,dv1,dv2,
    dv0dv1,dv0dv2,
    isect1[0],isect1[1]);

  /* compute interval for triangle 2 */
  COMPUTE_INTERVALS(
    up0,up1,up2,
    du0,du1,du2,
    du0du1,du0du2,
    isect2[0],isect2[1]);

  SORT(isect1[0],isect1[1]);
  SORT(isect2[0],isect2[1]);

  if(isect1[1]<isect2[0] || isect2[1]<isect1[0])
  {
    //*output = 2 ;
    return 0;
  }
  //*output = 3 ;
  return 1;
}
\end{minted}
\end{minipage}
\caption{Code for the \texttt{jmeint} benchmark in \textsc{ParrotBench}.}\label{fig:orig_jmeint_code}
\end{figure*}

\begin{figure*}
\centering
\begin{minipage}{0.45\textwidth}
\begin{minted}[fontsize=\tiny]{c}
/* DCT for One block(8x8) */
void dct (INT16 *data)
{

  UINT16 i;
  INT32 x0, x1, x2, x3, x4, x5, x6, x7, x8;

  /*All values are shifted left by 10
  and rounded off to nearest integer */

  /* cos PI/16 * root(2)  */
  static const UINT16 c1=1420;
  /* cos PI/8 * root(2)  */
  static const UINT16 c2=1338;
  /* cos 3PI/16 * root(2)  */
  static const UINT16 c3=1204;
  /* cos 5PI/16 * root(2)  */
  static const UINT16 c5=805;
  /* cos 3PI/8 * root(2)  */
  static const UINT16 c6=554;
  /* cos 7PI/16 * root(2)  */
  static const UINT16 c7=283;

  static const UINT16 s1=3;
  static const UINT16 s2=10;
  static const UINT16 s3=13;

  for (i=8; i>0; i--)
  {
    x8 = data [0] + data [7];
    x0 = data [0] - data [7];

    x7 = data [1] + data [6];
    x1 = data [1] - data [6];

    x6 = data [2] + data [5];
    x2 = data [2] - data [5];

    x5 = data [3] + data [4];
    x3 = data [3] - data [4];

    x4 = x8 + x5;
    x8 -= x5;

    x5 = x7 + x6;
    x7 -= x6;

    data [0] = (INT16) (x4 + x5);
    data [4] = (INT16) (x4 - x5);

    data [2] = (INT16) ((x8*c2 + x7*c6) >> s2);
    data [6] = (INT16) ((x8*c6 - x7*c2) >> s2);

    data [7] = (INT16) (
      (x0*c7 - x1*c5 + x2*c3 - x3*c1) >> s2);
    data [5] = (INT16) (
      (x0*c5 - x1*c1 + x2*c7 + x3*c3) >> s2);
    data [3] = (INT16) (
      (x0*c3 - x1*c7 - x2*c1 - x3*c5) >> s2);
    data [1] = (INT16) (
      (x0*c1 + x1*c3 + x2*c5 + x3*c7) >> s2);

    data += 8;
  }

  data -= 64;
\end{minted}
\end{minipage}
\begin{minipage}{0.45\textwidth}
\begin{minted}[fontsize=\tiny]{c}
  for (i=8; i>0; i--)
  {
    x8 = data [0] + data [56];
    x0 = data [0] - data [56];

    x7 = data [8] + data [48];
    x1 = data [8] - data [48];

    x6 = data [16] + data [40];
    x2 = data [16] - data [40];

    x5 = data [24] + data [32];
    x3 = data [24] - data [32];

    x4 = x8 + x5;
    x8 -= x5;

    x5 = x7 + x6;
    x7 -= x6;

    data [0] = (INT16) ((x4 + x5) >> s1);
    data [32] = (INT16) ((x4 - x5) >> s1);

    data [16] = (INT16) ((x8*c2 + x7*c6) >> s3);
    data [48] = (INT16) ((x8*c6 - x7*c2) >> s3);

    data [56] = (INT16) ((x0*c7 - x1*c5 + x2*c3 - x3*c1) >> s3);
    data [40] = (INT16) ((x0*c5 - x1*c1 + x2*c7 + x3*c3) >> s3);
    data [24] = (INT16) ((x0*c3 - x1*c7 - x2*c1 - x3*c5) >> s3);
    data [8] = (INT16) ((x0*c1 + x1*c3 + x2*c5 + x3*c7) >> s3);

    data++;
  }
}

/* Multiply DCT Coefficients with Quantization table
and store in ZigZag location */
void quantization(
    INT16* const data, UINT16* const quant_table_ptr) {
  INT16 i;
  INT32 value;

  for (i = 63; i >= 0; i--) {
    value = data[i] * quant_table_ptr[i];
    value = (value + 0x4000) >> 15;

    Temp[zigzagTable[i]] = (INT16) value;
  }
}

// Kernel is:
//   dct(Y1);
//   quantization(Y1, ILqt);
\end{minted}
\end{minipage}
\caption{Code for the \texttt{jpeg} benchmark in \textsc{ParrotBench}.}\label{fig:orig_jpeg_code}
\end{figure*}

\begin{figure*}
\centering
\begin{minipage}{0.45\textwidth}
\begin{minted}{c}
float euclideanDistance(RgbPixel* p, Centroid* c1) {
	float r;

	r = 0;
	r += (p->r - c1->r) * (p->r - c1->r);
	r += (p->g - c1->g) * (p->g - c1->g);
	r += (p->b - c1->b) * (p->b - c1->b);

	return sqrt(r);
}
\end{minted}
\end{minipage}
\caption{Code for the \texttt{kmeans} benchmark in \textsc{ParrotBench}.}\label{fig:orig_kmeans_code}
\end{figure*}

\begin{figure*}
\centering
\begin{minipage}{0.45\textwidth}
\begin{minted}{c}
static float kx[][3] =
		{
			{ -1, -2, -1 },
			{  0,  0,  0 },
			{  1,  2,  1 }
		} ;

static float ky[][3] =
		{
			{ -1, 0, 1 },
			{ -2, 0, 2 },
			{ -1, 0, 1 }
		} ;

float convolve(float w[][3], float k[][3])
{
	float r ;
	r = 0.0 ;
	for( int j = 0 ; j < 3 ; j++ )
		for ( int i = 0 ; i < 3 ; i++ )
		{
			r += w[i][j] * k[j][i] ;
		}
	return r ;
}

float sobel(float w[][3])
{
	float sx ;
	float sy ;
	float s  ;

	sx = convolve(w, ky) ;
	sy = convolve(w, kx) ;
	s = sqrt(sx * sx + sy * sy) ;
	if (s >= (256 / sqrt(256 * 256 + 256 * 256)))
		s = 255 / sqrt(256 * 256 + 256 * 256);
	return s ;
}
\end{minted}
\end{minipage}
\caption{Code for the \texttt{sobel} benchmark in \textsc{ParrotBench}.}\label{fig:orig_sobel_code}
\end{figure*}

\subsection{\textsc{ParrotBench} Modifications}\label{sec:parrotbenchshort_code}
Due to methodological choices in \textsc{ExeStack} and architectural choices for \textsc{CompNet}s, we omit some \textsc{ParrotBench} benchmarks from \textsc{ParrotBenchCPN} and modify others.
We omit the \texttt{jmeint} and \texttt{jpeg} benchmarks in \textsc{ParrotBench} because they are significantly longer than the 512-token context length of a BERT-Tiny (1,192 and 1,250 tokens, respectively).
We modify the \texttt{fft} and \texttt{invk2j} benchmarks because they both use pointer arguments to store outputs, and our \textsc{CompNets} were not trained to support pointer arguments.
To make each function pointer-free, we split it into two functions, each function computing one of the outputs (Figures~\ref{fig:fft_code}~and~\ref{fig:invk2j_code}).
Additionally, the \texttt{sobel} benchmark uses pointer inputs, so we rewrite it to only use scalar inputs (Figure~\ref{fig:sobel_code}).
Finally, the \texttt{kmeans} benchmark uses custom structs to pass arguments, so we rewrite the benchmark to desugar these structs into their scalar components (Figure~\ref{fig:kmeans_code}).

\begin{figure*}
\centering
\begin{minipage}{0.45\textwidth}
\begin{minted}{c}
float fftSin_Output0(float x) {
    return sin(-2 * 3.1415 * x);
}

float fftSin_Output1(float x) {
    return cos(-2 * 3.1415 * x);
}
\end{minted}
\end{minipage}
\caption{Code for the \texttt{fft} benchmark in \textsc{ParrotBenchCPN}.}\label{fig:fft_code}
\end{figure*}

\begin{figure*}
\centering
\begin{minipage}{0.45\textwidth}
\begin{minted}{c}
float invk2j_Output0(float x, float y) {
  float l1 = 0.5 ;
  float l2 = 0.5 ;
  float theta2 = (float)acos(
    ((x * x) + (y * y) - (l1 * l1) - (l2 * l2)) /
    (2 * l1 * l2)) ;
  return (float)asin(
    (y * (l1 + l2 * cos(theta2)) - x * l2 * sin(theta2)) /
    (x * x + y * y)) ;
}

float invk2j_Output1(float x, float y) {
  float l1 = 0.5 ;
  float l2 = 0.5 ;
  return (float)acos(
    ((x * x) + (y * y) - (l1 * l1) - (l2 * l2)) /
    (2 * l1 * l2)) ;
}
\end{minted}
\end{minipage}
\caption{Code for the \texttt{invk2j} benchmark in \textsc{ParrotBenchCPN}.}\label{fig:invk2j_code}
\end{figure*}

\begin{figure*}
\centering
\begin{minipage}{0.45\textwidth}
\begin{minted}{c}
float euclideanDistance(
  float p_0, float p_1, float p_2,
  float c1_0, float c1_1, float c1_2) {
  float r;

  r = 0;
  r += (p_0 - c1_0) * (p_0 - c1_0);
  r += (p_1 - c1_1) * (p_1 - c1_1);
  r += (p_2 - c1_2) * (p_2 - c1_2);

  return sqrt(r);
}
\end{minted}
\end{minipage}
\caption{Code for the \texttt{kmeans} benchmark in \textsc{ParrotBenchCPN}.}\label{fig:kmeans_code}
\end{figure*}

\begin{figure*}
\centering
\begin{minipage}{0.45\textwidth}
\begin{minted}{c}
float sobel(
  float w00, float w01, float w02,
  float w10, float w11, float w12,
  float w20, float w21, float w22)
{
  float sx = 0.0;
  sx += w00 * -1;
  sx += w10 * 0;
  sx += w20 * 1;
  sx += w01 * -2;
  sx += w11 * 0;
  sx += w21 * 2;
  sx += w02 * -1;
  sx += w12 * 0;
  sx += w22 * 1;

  float sy = 0.0;
  sy += w00 * -1;
  sy += w10 * -2;
  sy += w20 * -1;
  sy += w01 * 0;
  sy += w11 * 0;
  sy += w21 * 0;
  sy += w02 * 1;
  sy += w12 * 2;
  sy += w22 * 1;

  float s = sqrt(sx * sx + sy * sy) ;
  if (s >= (256 / sqrt(256 * 256 + 256 * 256)))
    s = 255 / sqrt(256 * 256 + 256 * 256);
  return s ;
}
\end{minted}
\end{minipage}
\caption{Code for the \texttt{sobel} benchmark in \textsc{ParrotBenchCPN}.}\label{fig:sobel_code}
\end{figure*}

\subsection{\textsc{ParrotBenchCPN} Input Generation}\label{sec:parrot_input_generation}
We attempt to exactly replicate the dataset used by \citet{Parrot2012} for the subset of benchmarks we consider from \textsc{ParrotBench}.
To replicate their dataset, we analyze the source code in \href{https://github.com/he-actlab/AxBench_old}{the AxBench repository}, which contains all kernels in \textsc{ParrotBench}.
Figure~\ref{fig:parrotbenchshort_dataset_size} shows the size of the dataset produced by the methodology in the following sections.

\subsubsection{FFT}
To generate train inputs for \texttt{fft}, we generate $32{,}768$ inputs uniformly at random from $[0, 1/2]$.
To generate test inputs for \texttt{fft}, we generate $2{,}048$ inputs uniformly at random from $[0, 1/2]$, resampling as necessary whenever an input is generated that exists in the training set.

\subsubsection{InverseK2J}
To generate train inputs for \texttt{invk2j}, we generate $10{,}000$ inputs uniformly at random from $[-1/2, 1] \times [0, 1]$.
To generate test inputs for \texttt{invk2j}, we again generate $10{,}000$ inputs uniformly at random from $[-1/2, 1] \times [0, 1]$, but we resample whenever an input exists in the training set.

\subsubsection{KMeans}
To generate train inputs for \texttt{kmeans}, we generate $50{,}000$ inputs uniformly at random from $[0, 1]^6$.

\begin{figure}
  \centering
  \includegraphics[width=0.5\linewidth]{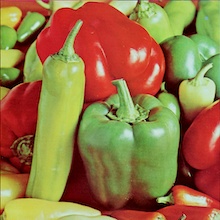}
  \caption{Image used to generate testing data for \texttt{kmeans}.}\label{fig:parrot_peppers}
  \vspace{-1em}
\end{figure}

To generate test inputs for \texttt{kmeans}, we use an image of peppers for RGB inputs (see Figure~\ref{fig:parrot_peppers}) and we generate $6$ centroids with uniformly random coordinates in $[0, 1]^3$, the number of centroids used by Esmaeilzadeh et al. (\href{https://github.com/he-actlab/AxBench_old/blob/1c3421004a84160fc4345b2fab254eb2f22bc032/apps/kmeans/src/kmeans.c#L72}{source}).
For each RGB input, we choose a random centroid to compute the \texttt{kmeans} kernel on, and we add the resulting I/O sample to the testing set.
This procedure results in a testing set containing $48{,}400$ inputs.

\subsubsection{Sobel}
To generate training and testing inputs for \texttt{sobel}, we read from files on the official repo of \citet{Parrot2012} (\href{https://github.com/he-actlab/AxBench_old/blob/1c3421004a84160fc4345b2fab254eb2f22bc032/apps/sobel/data/sobel_train.data}{here} and \href{https://github.com/he-actlab/AxBench_old/blob/1c3421004a84160fc4345b2fab254eb2f22bc032/apps/sobel/data/sobel_test.data}{here}, respectively).
These files contain $18{,}725$ and $17{,}976$ input-output pairs, respectively.

\section{\textsc{CompNet} Training Details}\label{sec:compnet_training_details}
\textsc{CompNet}s are controlled by the following hyperparameters: program batch size, input batch size, learning rate, number of training epochs, dataset program split, dataset input split, and the surrogate topology.
We swept over learning rates and chose fixed values for all other hyperparameters.
We selected the learning rate that achieved the best final loss on validation programs, averaged over trials, and we used all trials of the winning configuration as initialization methods.

\begin{figure*}
  \centering
  \includegraphics[width=0.9\linewidth]{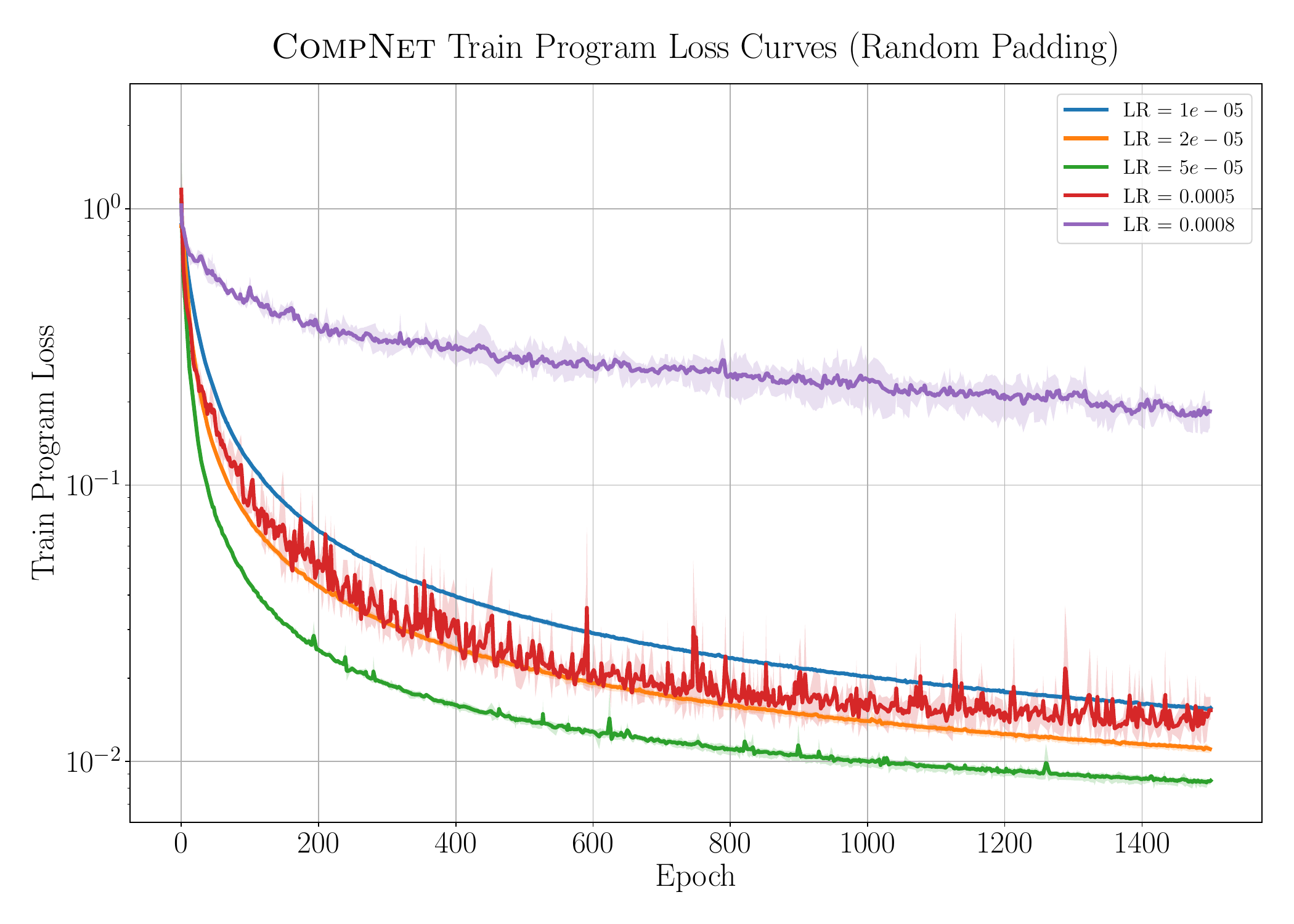}
  \includegraphics[width=0.9\linewidth]{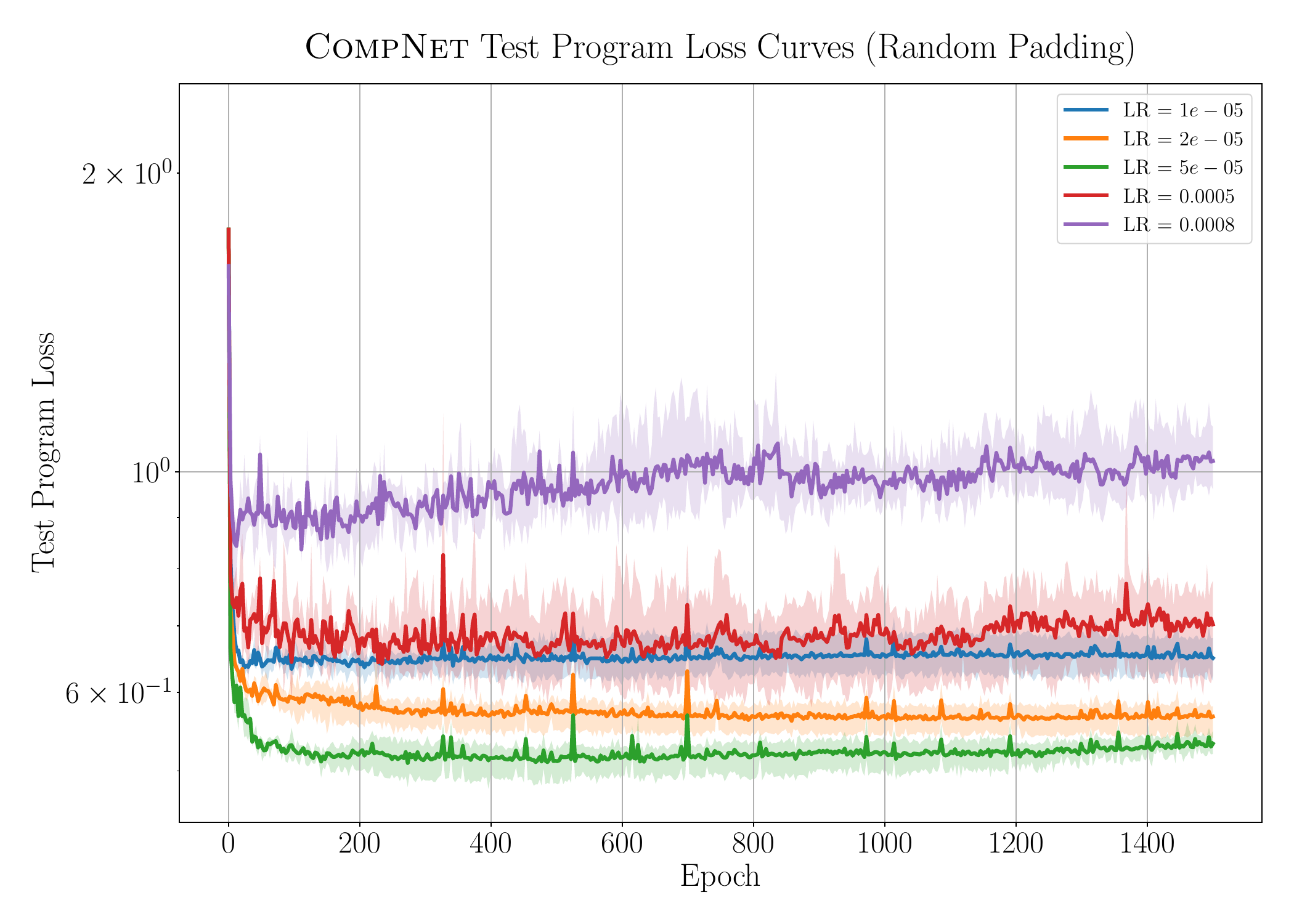}
  \caption{
    Loss curves for \textsc{CompNet} training hyperparameter sweep, using \textsc{ExeStackCPN} with random-padded inputs (see Appendix~\ref{sec:padding_for_variable_inputs}).
    Each curve represents a training trial.
    The test program loss includes validation programs as well.
  }\label{fig:compnet_random_pad_loss_curves}
\end{figure*}

\begin{figure*}
  \centering
  \includegraphics[width=0.9\linewidth]{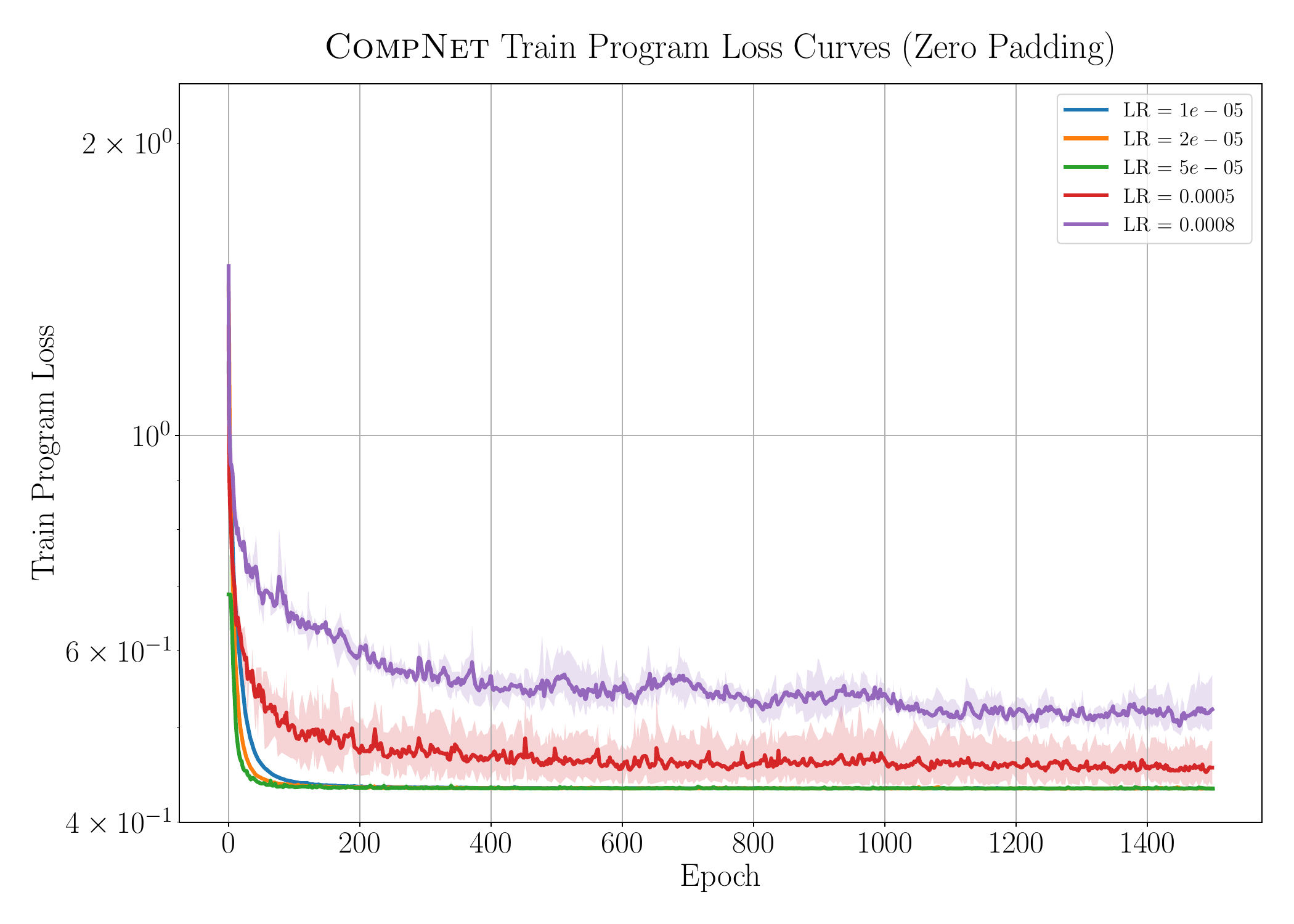}
  \includegraphics[width=0.9\linewidth]{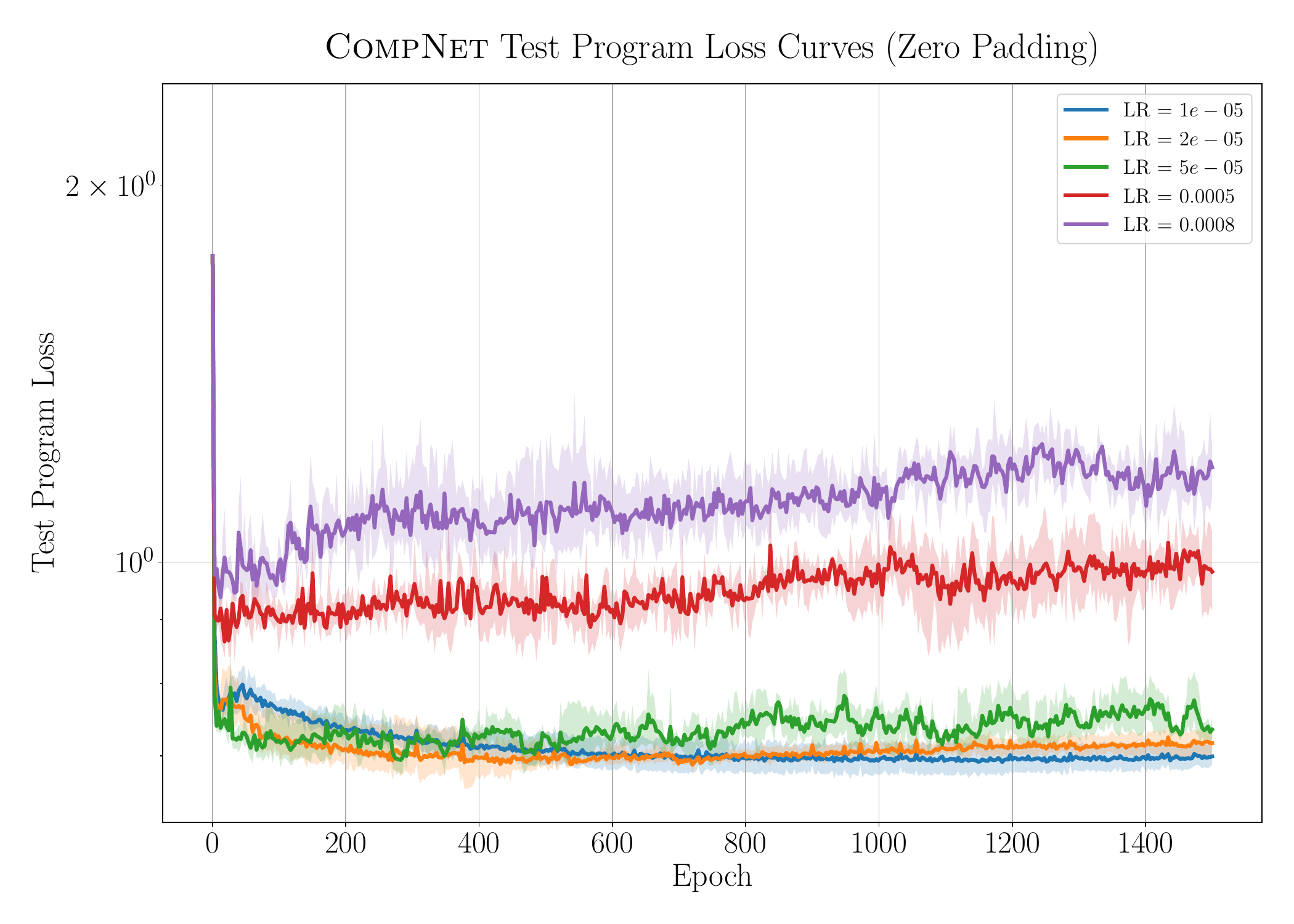}
  \caption{
    Loss curves for \textsc{CompNet} training hyperparameter sweep, using \textsc{ExeStackCPN} with zero-padded inputs (see Appendix~\ref{sec:padding_for_variable_inputs}).
    Each curve represents a training trial.
    The test program loss includes validation programs as well.
  }\label{fig:compnet_zero_pad_loss_curves}
\end{figure*}

We summarize the training configuration for \textsc{CompNet}s in Figure~\ref{fig:cpn_training_summary}.
Figures~\ref{fig:compnet_random_pad_loss_curves}~and~\ref{fig:compnet_zero_pad_loss_curves} show loss curves for \textsc{CompNet}s trained on \textsc{ExeStackCPN}, using both padding modes described in Appendix~\ref{sec:padding_for_variable_inputs}.

\begin{figure*}[h]
  \centering
  \begin{tabular}{ll}
  \toprule
  Setting & Value \\
  \midrule
  Architecture & BERT-Tiny \\
  Program Batch Size & $32$ \\
  Input Batch Size & $1024$ \\
  Learning Rate & $\in \set{1 \cdot 10^{-5}, \num{2e-5}, \mathbf{5 \cdot 10^{-5}}, \num{5e-4}, \num{8e-4}}$ \\
  \# Epochs & $1,500$ \\
  Dataset Program Split & $80/10/10$ \\
  Dataset Input Split & $50/0/50$ \\
  Surrogate Topology & $9 \rightarrow 4 \rightarrow 4 \rightarrow 1$ \\
  GPU & NVIDIA Tesla T4 16GB \\
  \# Trials & $3$ \\
  \bottomrule
  \end{tabular}
  \caption{
    Training configuration for \textsc{CompNet}s.
    We represent any values we sweep over as a set, and we bold the values that obtain the best final loss on test programs.
  }\label{fig:cpn_training_summary}
\end{figure*}

\section{MAML Training Details}\label{sec:maml_training_details}
MAML is controlled by the following hyperparameters: the meta batch size (number of tasks per batch), the input batch size (number of inputs per task), the number of epochs, the inner gradient update step size ($\alpha$), the outer gradient update step size ($\beta$), and the number of inner gradient update steps.
We modify the official MAML implementation\footnote{\href{https://github.com/cbfinn/maml}{https://github.com/cbfinn/maml}} to support training on \textsc{ExeStackCPN}.

We choose the meta batch size and the input batch size to align with how we train \textsc{CompNet}s (Appendix~\ref{sec:compnet_training_details}).
We decided to use the maximum number of epochs \citet{MAML2017} use in their applications ($70,000$), and we observed that in all applications, $\beta$ is fixed at $0.001$.
For the remaining parameters, $\alpha$ and the number of inner update steps, we perform a hyperparameter sweep, backing each configuration with 3 trials.
We choose the extents of each hyperparameter in the sweep as the minimum and maximum of hyperparameter settings observed in applications, and we add some points between these extents.
However, we limit the hyperparameter settings for $\alpha$ to a maximum of $0.2$, as previous experiments (not reported in this paper) showed training instability at higher values.

After each configuration finishes training, we finetune it for $20$ epochs for each of a sample of $5$ programs from the \textsc{ExeStackCPN} validation set.
We choose the hyperparameters with the lowest loss on the validation inputs at the end of finetuning, averaged over the sample of programs and trials.
We use all 3 trials of the winning configuration as initialization methods.

\begin{figure*}
  \centering
  \includegraphics[width=0.9\linewidth]{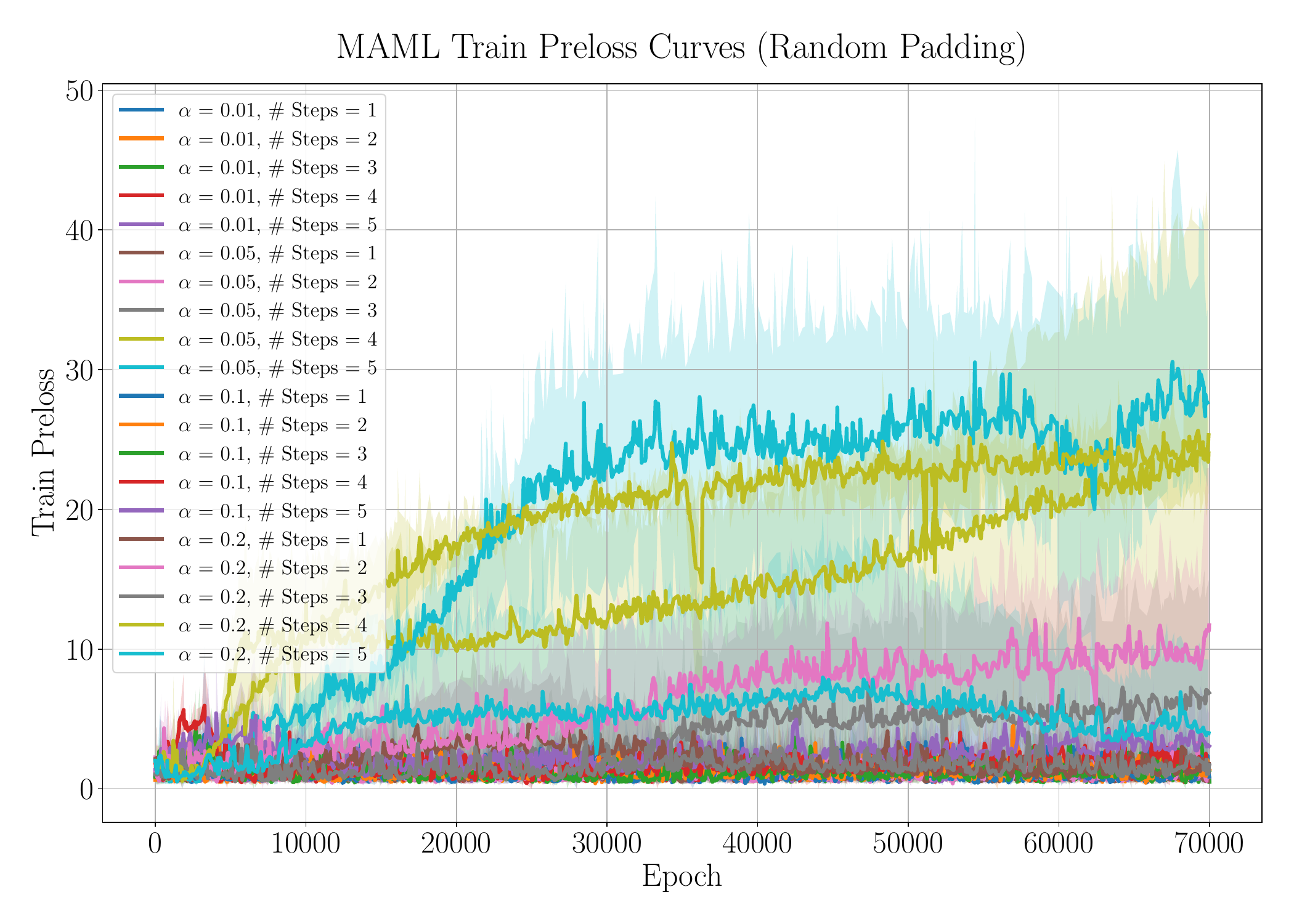}
  \includegraphics[width=0.9\linewidth]{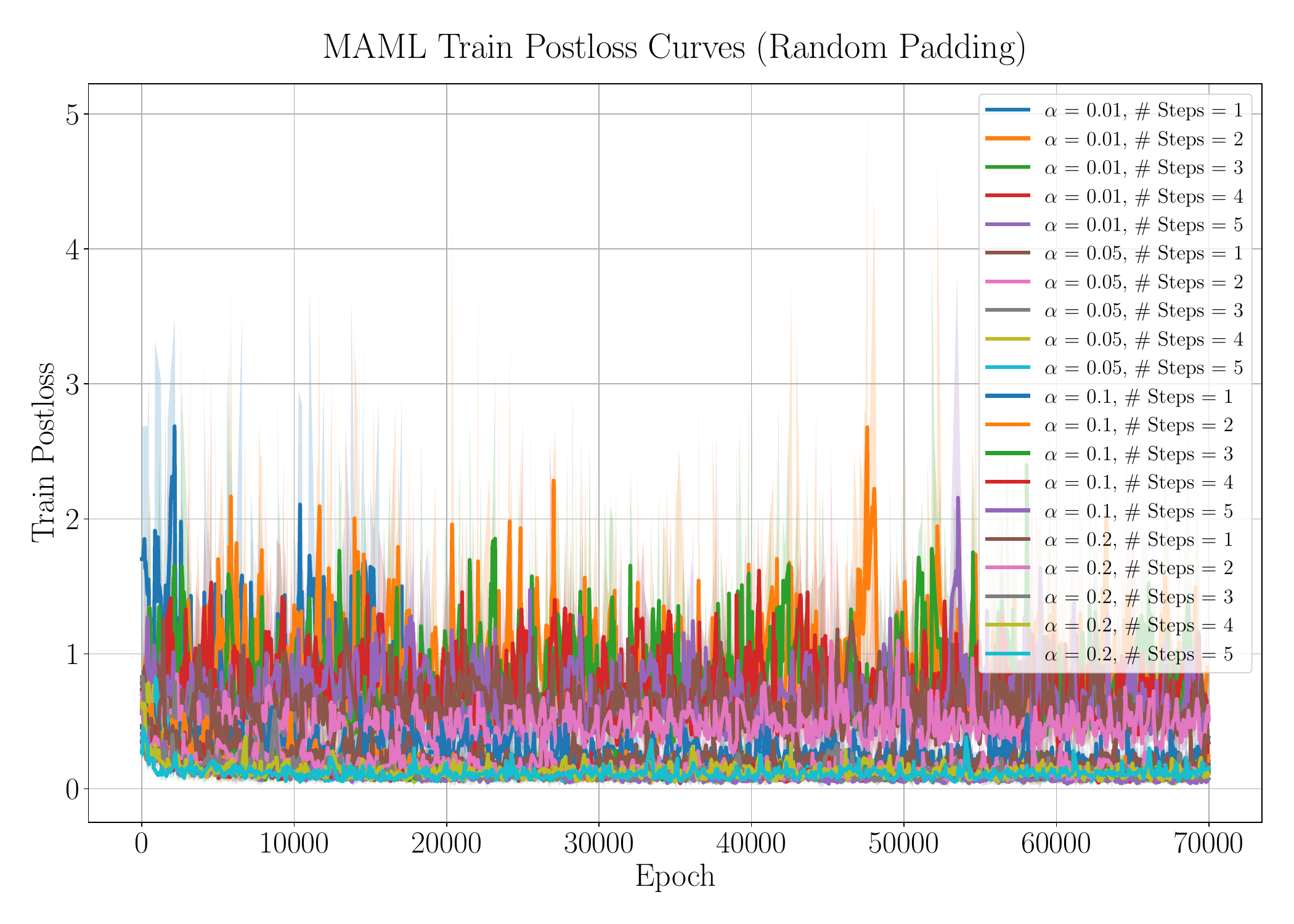}
  \caption{
    Loss curves for MAML training hyperparameter sweep, using \textsc{ExeStackCPN} with random-padded inputs (see Appendix~\ref{sec:padding_for_variable_inputs}).
    Each curve represents a training trial.
  }\label{fig:maml_random_pad_loss_curves}
\end{figure*}

\begin{figure*}
  \centering
  \includegraphics[width=0.9\linewidth]{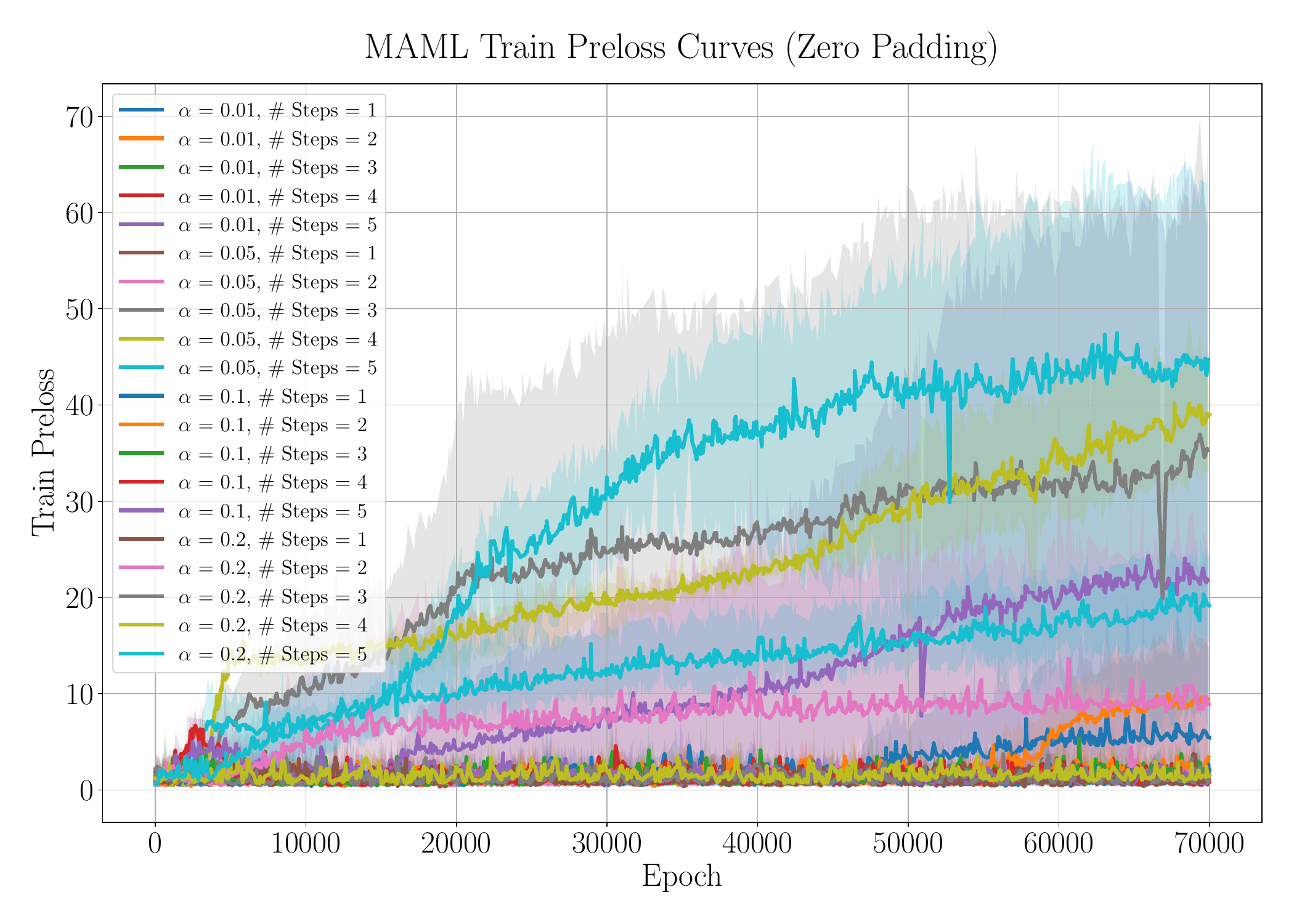}
  \includegraphics[width=0.9\linewidth]{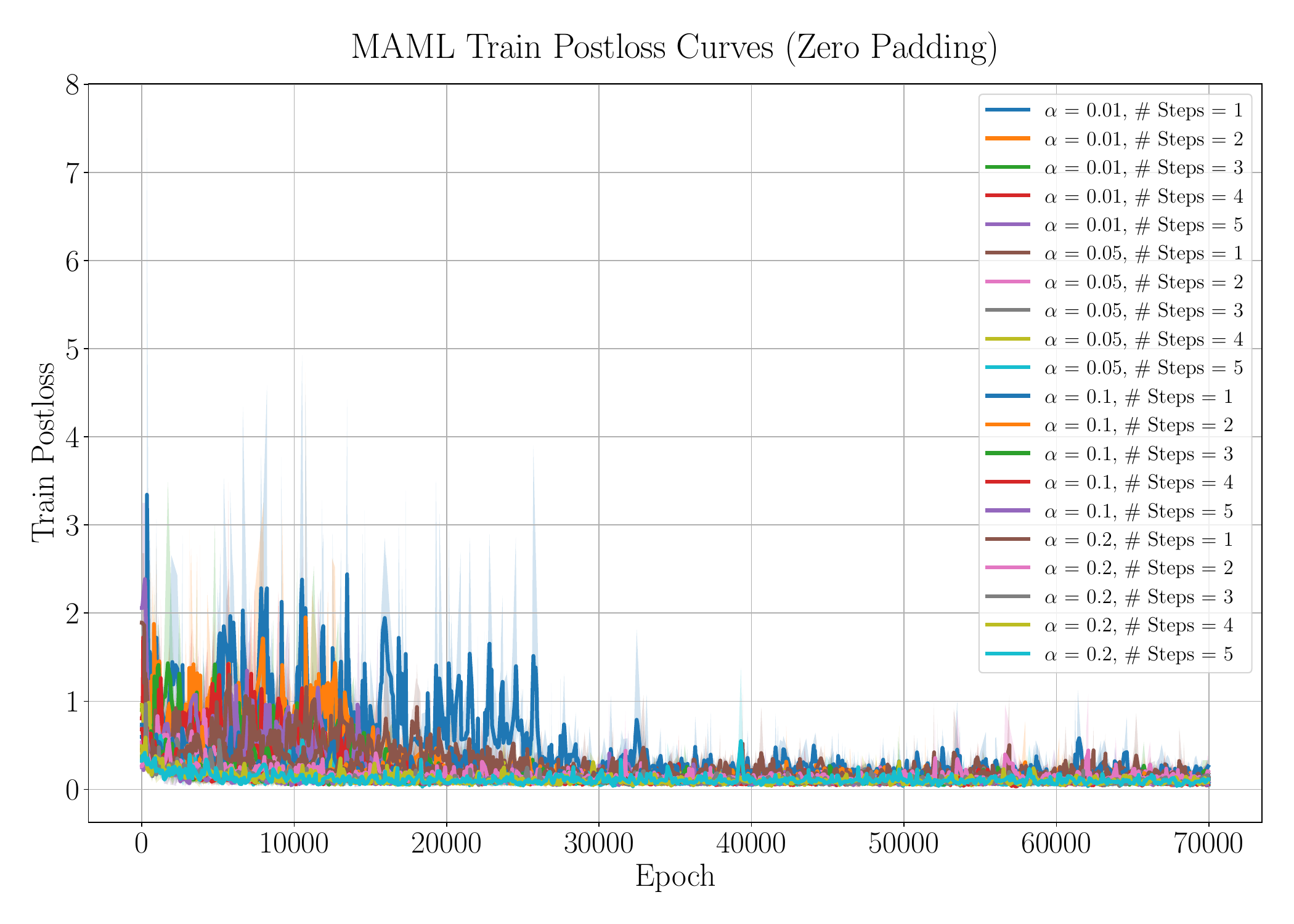}
  \caption{
    Loss curves for MAML training hyperparameter sweep, using \textsc{ExeStackCPN} with zero-padded inputs (see Appendix~\ref{sec:padding_for_variable_inputs}).
    Each curve represents a training trial.
  }\label{fig:maml_zero_pad_loss_curves}
\end{figure*}

We summarize the training configuration for MAML in Figure~\ref{fig:maml_training_summary}.
Figures~\ref{fig:maml_random_pad_loss_curves}~and~\ref{fig:maml_zero_pad_loss_curves} show loss curves for MAML initializations trained on \textsc{ExeStackCPN}, using both padding modes described in Appendix~\ref{sec:padding_for_variable_inputs}.
The curves include \textit{prelosses} and \textit{postlosses} for training programs.
In MAML training, the preloss is the loss of the current initialization when evaluated on a task, and the postloss is the loss of the initialization after finetuning for the number of inner gradient update steps.

\begin{figure}
  \centering
  \begin{tabular}{ll}
  \toprule
  Setting & Value \\
  \midrule
  Meta Batch Size & $32$ \\
  Input Batch Size & $1024$ \\
  \# Epochs & $70,000$ \\
  $\alpha$ & $\in \set{0.01, 0.05, 0.1, \mathbf{0.2}}$ \\
  $\beta$ & $0.001$ \\
  \# Inner Update Steps & $\in \set{1, 2, \mathbf{3}, 4, 5}$ \\
  \# Finetuning Epochs & $20$ \\
  \# Trials & $3$ \\
  Dataset Program Split & $80/10/10$ \\
  Dataset Input Split & $50/20/30$ \\
  Surrogate Topology & $9 \rightarrow 4 \rightarrow 4 \rightarrow 1$ \\
  GPU & NVIDIA Tesla T4 16GB \\
  \bottomrule
  \end{tabular}
  \caption{
    Training configuration for MAML.
    We represent any values we sweep over as a set, and we bold the values that obtain the best finetuning loss on validation programs.
  }\label{fig:maml_training_summary}
\end{figure}

\section{Neural Surrogate Pretraining Details}\label{sec:pts_training_details}
Similarly to \textsc{CompNet}s, pretrained surrogates are controlled by the following hyperparameters: program batch size, input batch size, learning rate, number of training epochs, dataset program split, dataset input split, and the surrogate topology.
We sweep over the same set of learning rates as for \textsc{CompNet}s, and we use the same values for other hyperparameters that we use for \textsc{CompNet}s.
We select the learning rate that achieves the best final loss on test programs, averaged over trials, and we use all trials of the winning configuration as initialization methods.

\begin{figure*}
  \centering
  \includegraphics[width=0.9\linewidth]{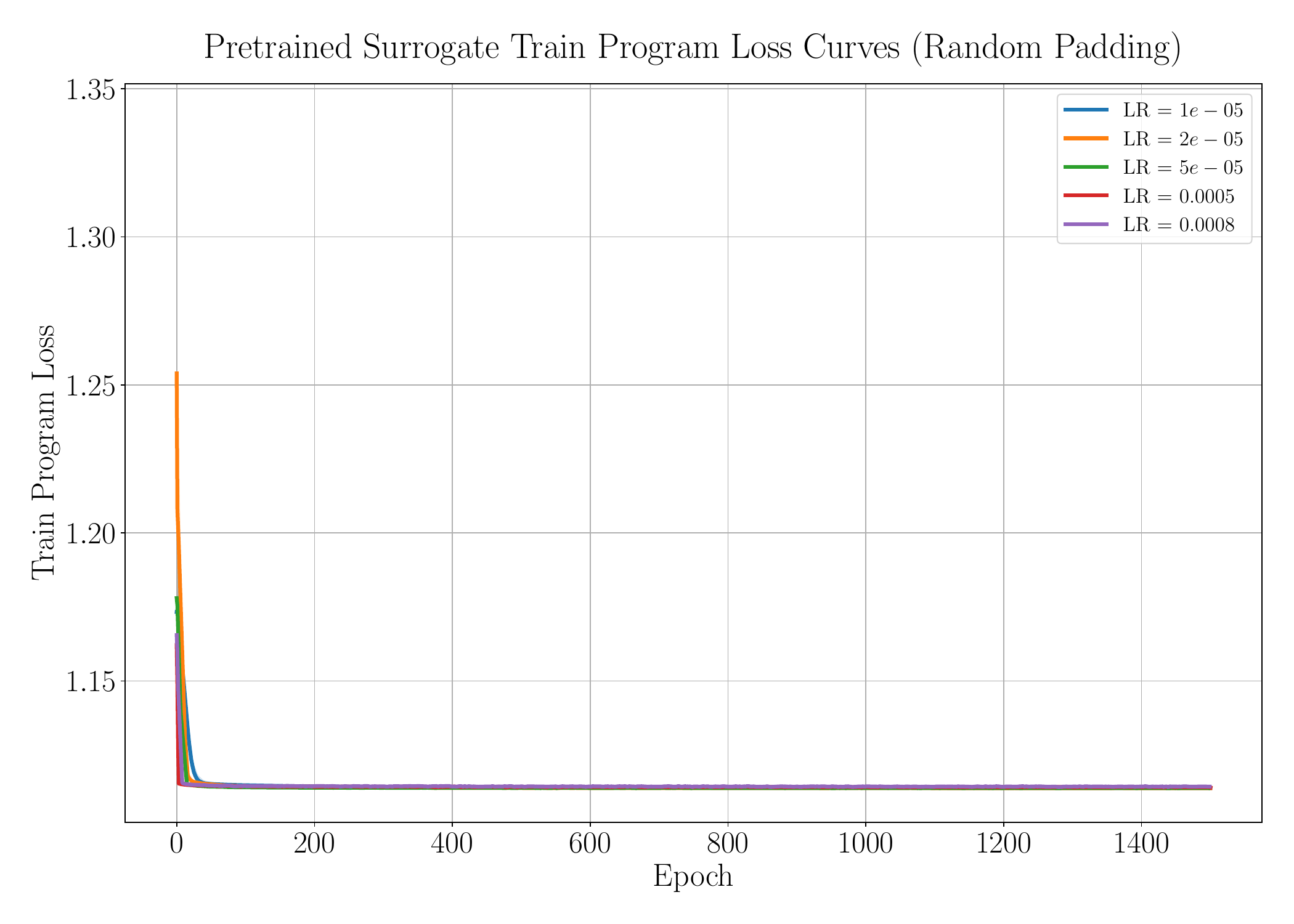}
  \includegraphics[width=0.9\linewidth]{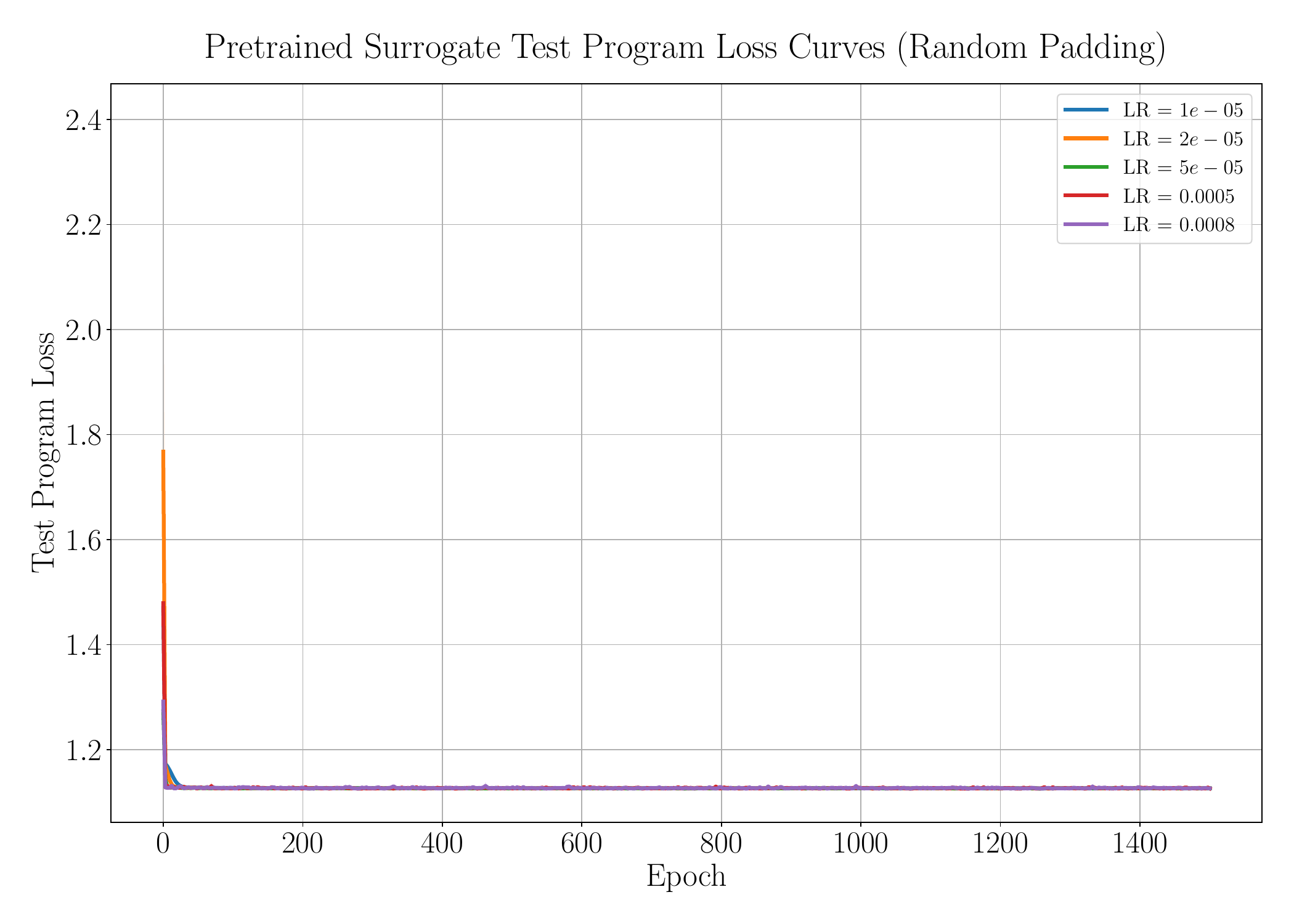}
  \caption{
    Loss curves for pretrained surrogate hyperparameter sweep, using \textsc{ExeStackCPN} with random-padded inputs (see Appendix~\ref{sec:padding_for_variable_inputs}).
    Each curve represents a training trial.
    The test program loss includes validation programs as well.
  }\label{fig:pts_random_pad_loss_curves}
\end{figure*}

\begin{figure*}
  \centering
  \includegraphics[width=0.9\linewidth]{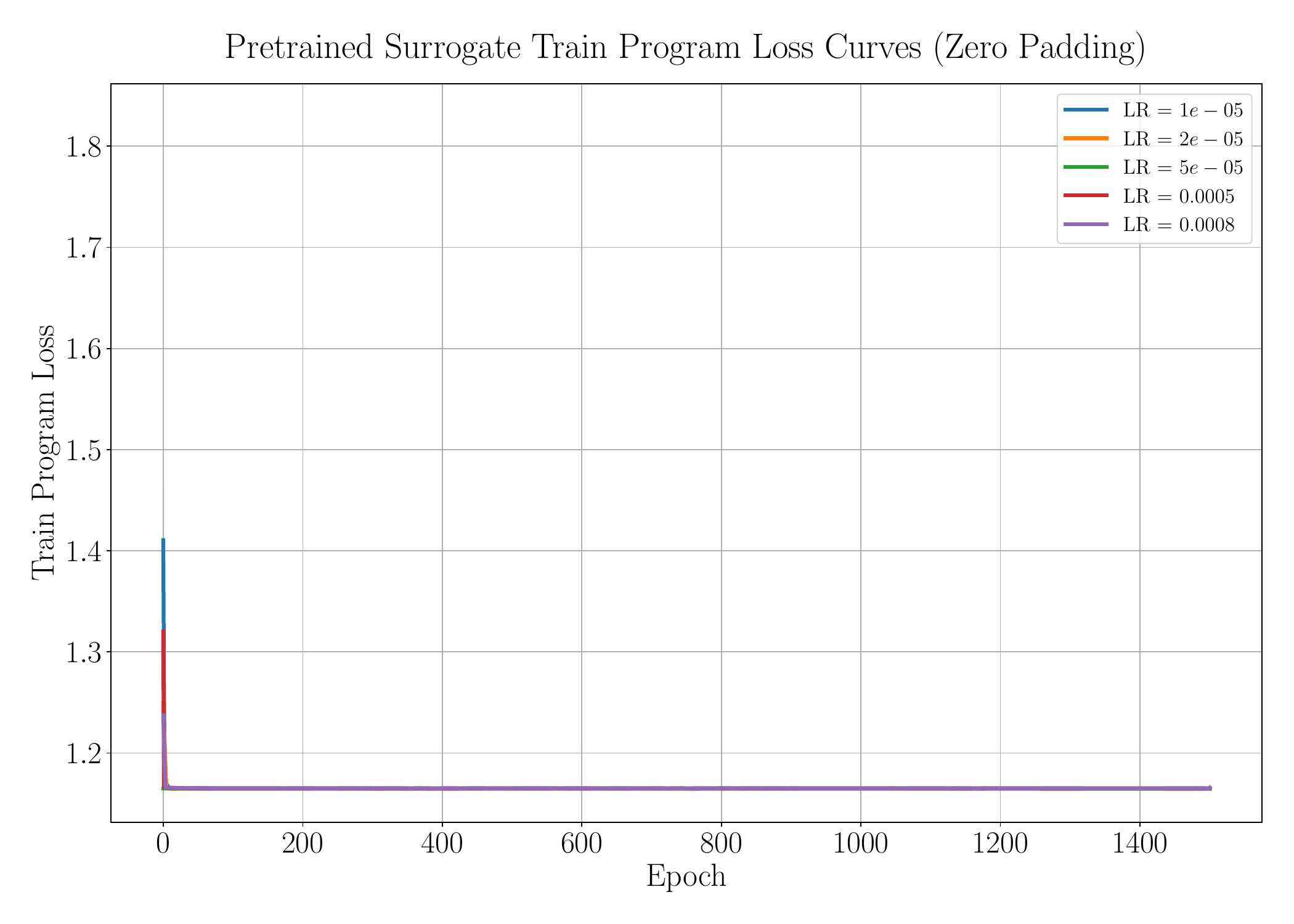}
  \includegraphics[width=0.9\linewidth]{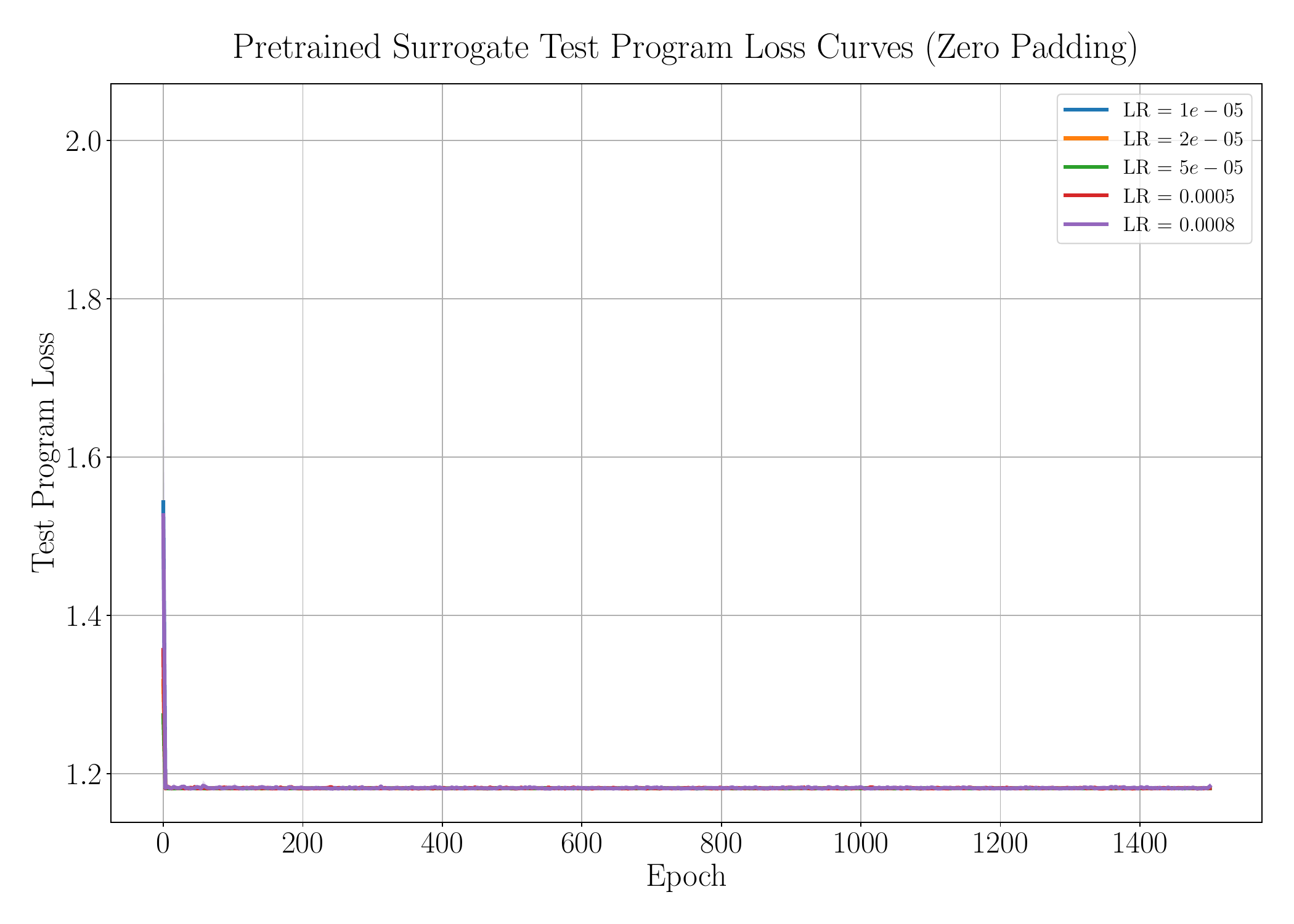}
  \caption{
    Loss curves for pretrained surrogate hyperparameter sweep, using \textsc{ExeStackCPN} with zero-padded inputs (see Appendix~\ref{sec:padding_for_variable_inputs}).
    Each curve represents a training trial.
    The test program loss includes validation programs as well.
  }\label{fig:pts_zero_pad_loss_curves}
\end{figure*}

We summarize the training configuration for pretrained surrogates in Figure~\ref{fig:pts_training_summary}.
Figures~\ref{fig:pts_random_pad_loss_curves}~and~\ref{fig:pts_zero_pad_loss_curves} show loss curves for surrogates pretrained on \textsc{ExeStackCPN}, using both padding modes described in Appendix~\ref{sec:padding_for_variable_inputs}.

\begin{figure*}
  \centering
  \begin{tabular}{ll}
  \toprule
  Setting & Value \\
  \midrule
  Program Batch Size & $32$ \\
  Input Batch Size & $1024$ \\
  Learning Rate & $\in \set{\mathbf{1 \cdot 10^{-5}}, \num{2e-5}, \num{5e-5}, \num{5e-4}, \num{8e-4}}$ \\
  \# Epochs & $1,500$ \\
  Dataset Program Split & $80/0/20$ \\
  Dataset Input Split & $50/0/50$ \\
  Surrogate Topology & $9 \rightarrow 4 \rightarrow 4 \rightarrow 1$ \\
  GPU & NVIDIA Tesla T4 16GB \\
  \# Trials & $3$ \\
  \bottomrule
  \end{tabular}
  \caption{
    Training configuration for pretrained surrogates.
    We represent any values we sweep over as a set, and we bold the values that obtain the best final loss on test programs.
  }\label{fig:pts_training_summary}
\end{figure*}


\section{Data Efficiency Improvements (Extended)}\label{sec:data_efficiency_extended}


We compute improvements in the data efficiency evaluation as a ratio of the test loss achieved by random initialization over the test loss achieved by an initialization method.
Here, we present the test losses we use to compute these ratios, as well as test loss improvements grouped at a finer granularity.

\begin{figure*}
  \centering
  \includegraphics[width=0.49\linewidth]{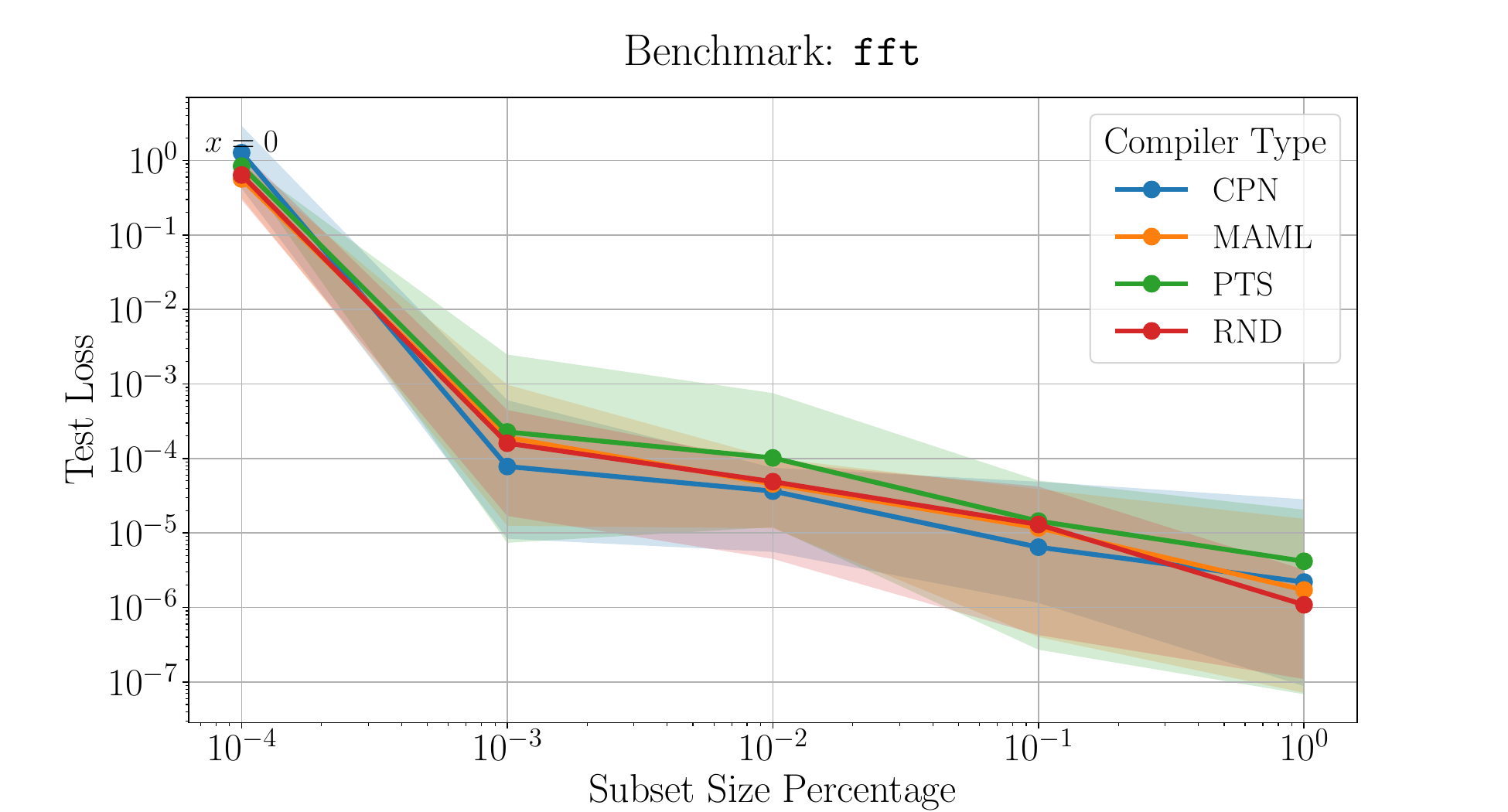}
  \includegraphics[width=0.49\linewidth]{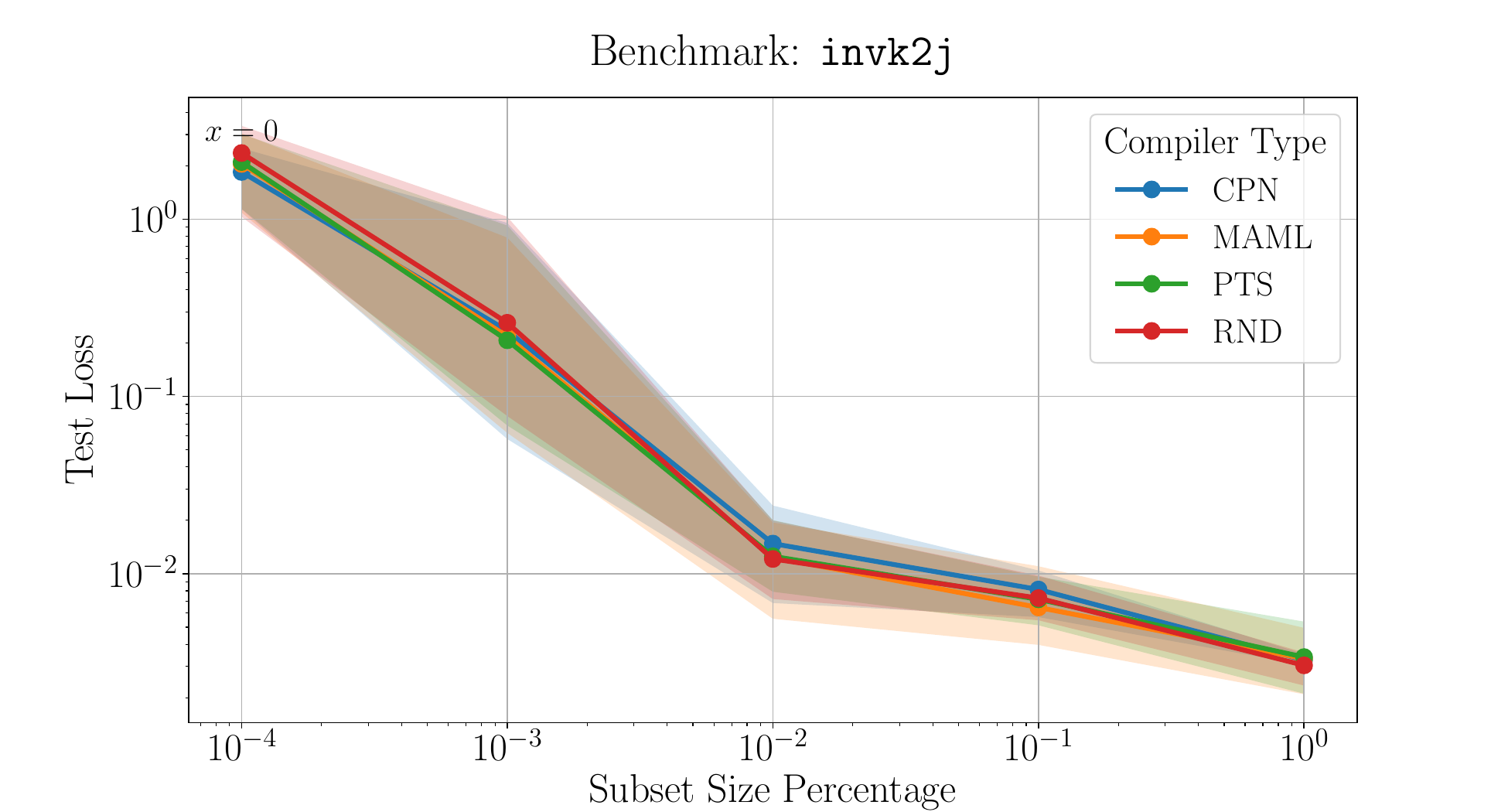}
  \includegraphics[width=0.49\linewidth]{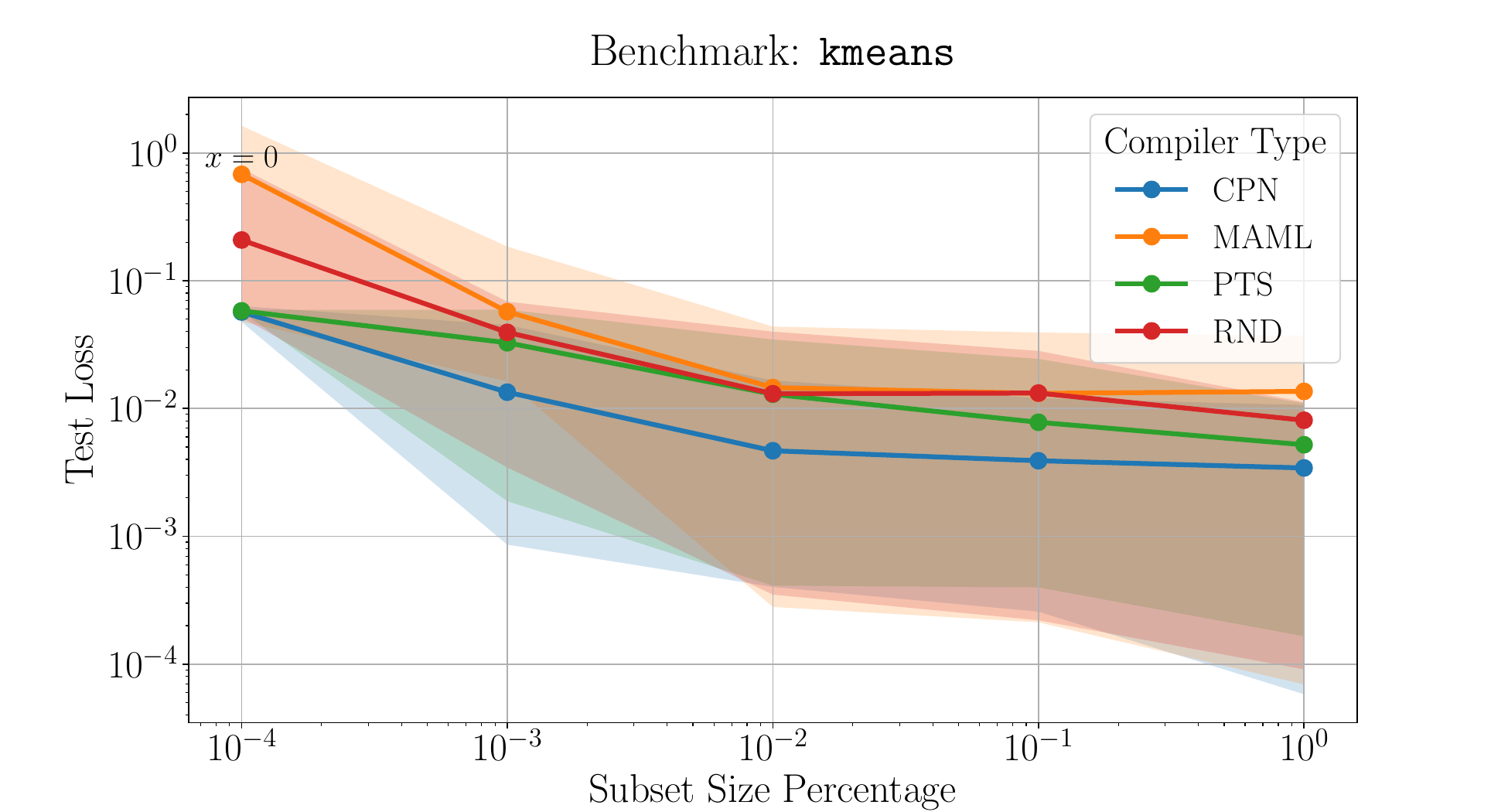}
  \includegraphics[width=0.49\linewidth]{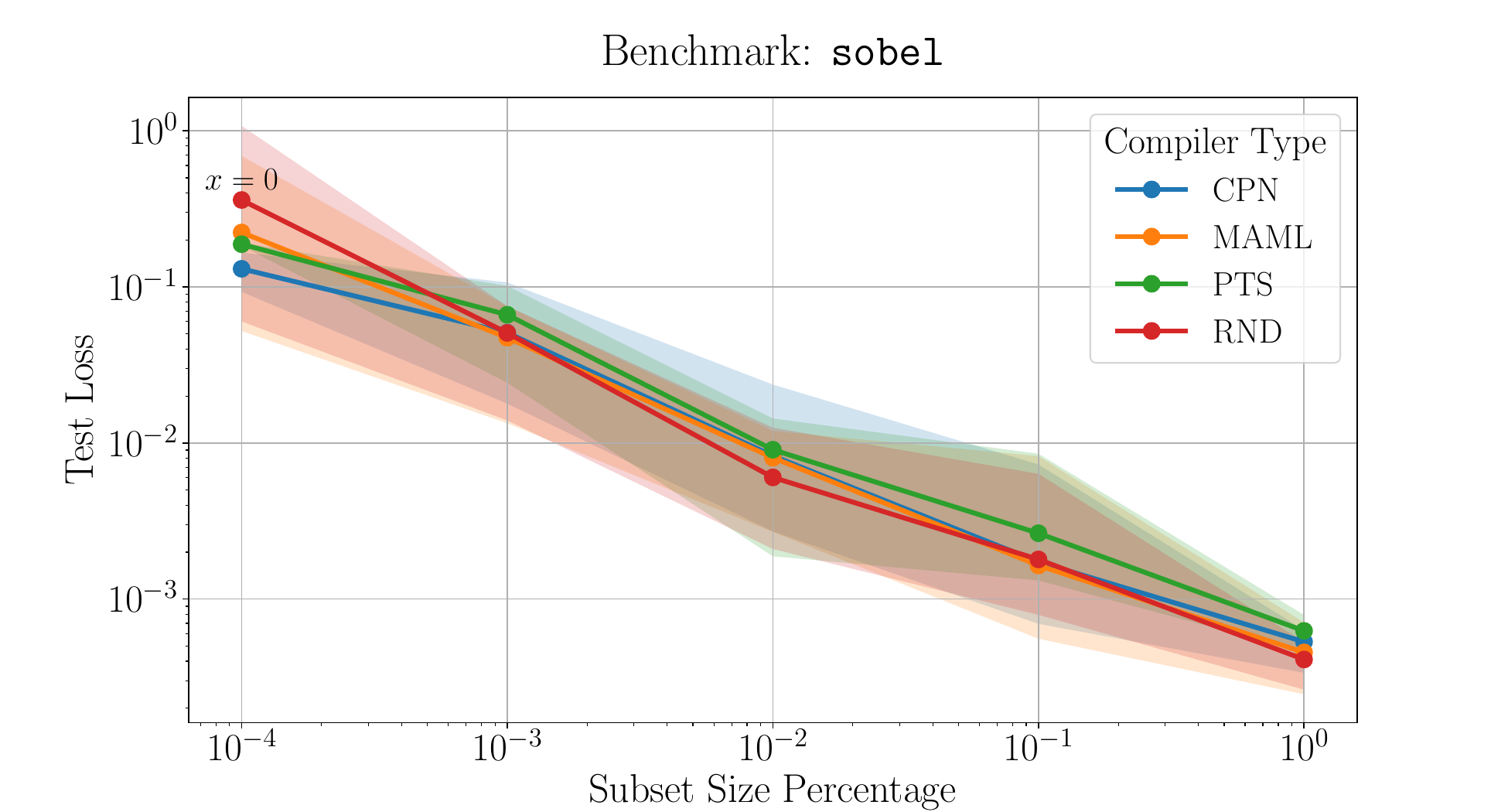}
  \caption{
    Log-log plot showing dataset size vs. test loss for each initialization method on each benchmark in \textsc{ParrotBenchCPN}.
    We present the test loss at the epoch with the lowest validation loss.
    When the validation set is empty, we use the test loss at the final epoch.
    Since zero is not a valid point on a logarithmic scale, we include the loss values for the empty dataset at a nonzero point on the $x$ axis that is also smaller than the smallest nonzero dataset size, and we label it with ``$x = 0$''.
  }\label{fig:dataset_size_vs_test_loss_parrot}
\end{figure*}

Figure~\ref{fig:dataset_size_vs_test_loss_parrot} shows test loss as a function of the dataset size.
For all initialization methods, performance improves the most on \texttt{fft} as more training data becomes available.
The difference between the average test loss at $0\%$ and $100\%$ is $\approx 6$ orders of magnitude for every initialization method, whereas it is only $\approx 2$-$3$ on other benchmarks.
The \texttt{kmeans} benchmark exhibits the least variation, with all initialization methods dropping by $\leq 2$ orders of magnitude on average, from $0\%$ to $100\%$ of training data.
The only benchmark where \textsc{CompNet}s dominate at all dataset sizes is \texttt{kmeans}.
For other benchmarks, \textsc{CompNet}s perform, on par, slightly better, or slightly worse, varying across dataset sizes.

\begin{figure*}
  \centering
  \includegraphics[width=0.48\linewidth]{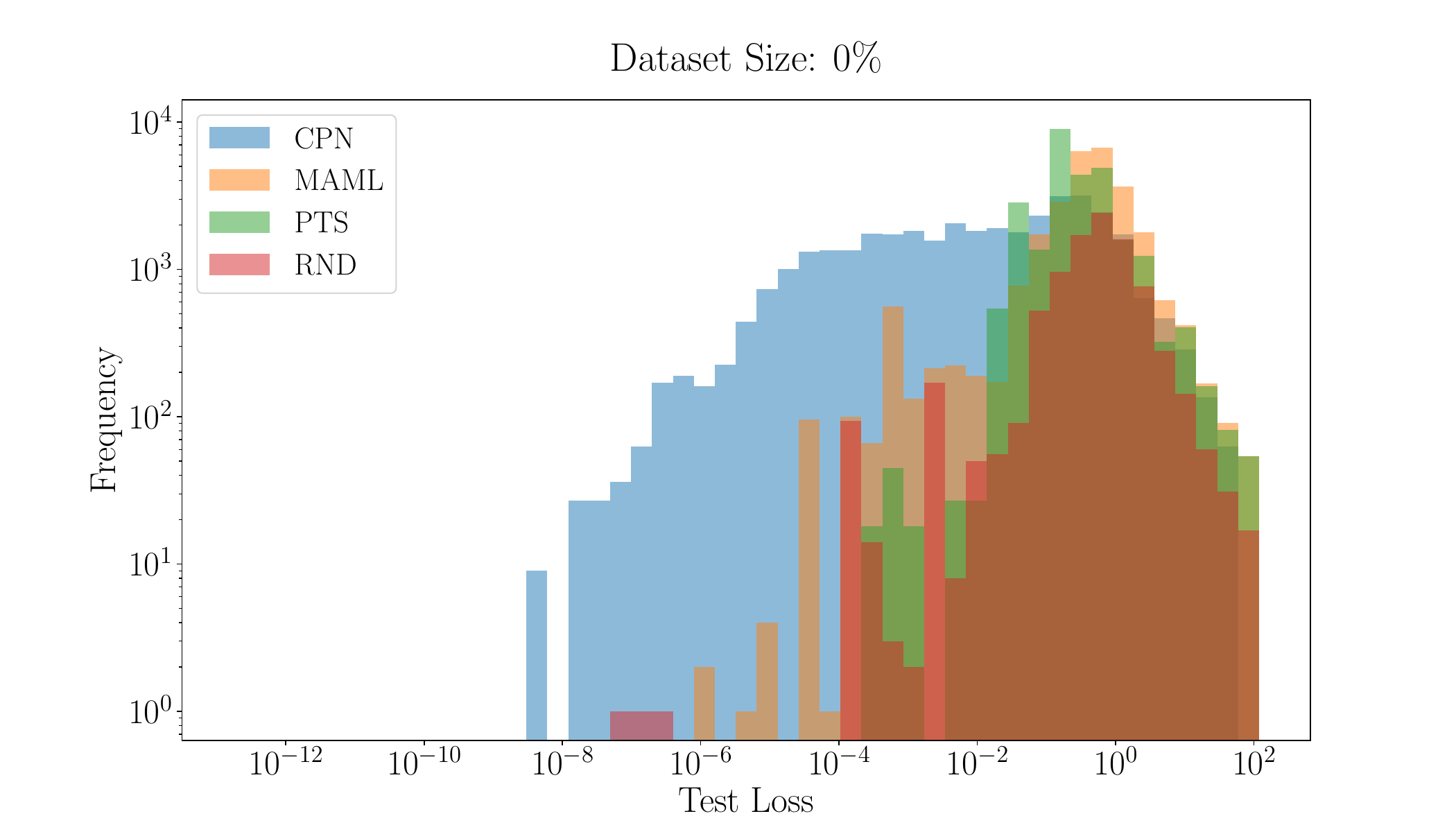}
  \includegraphics[width=0.48\linewidth]{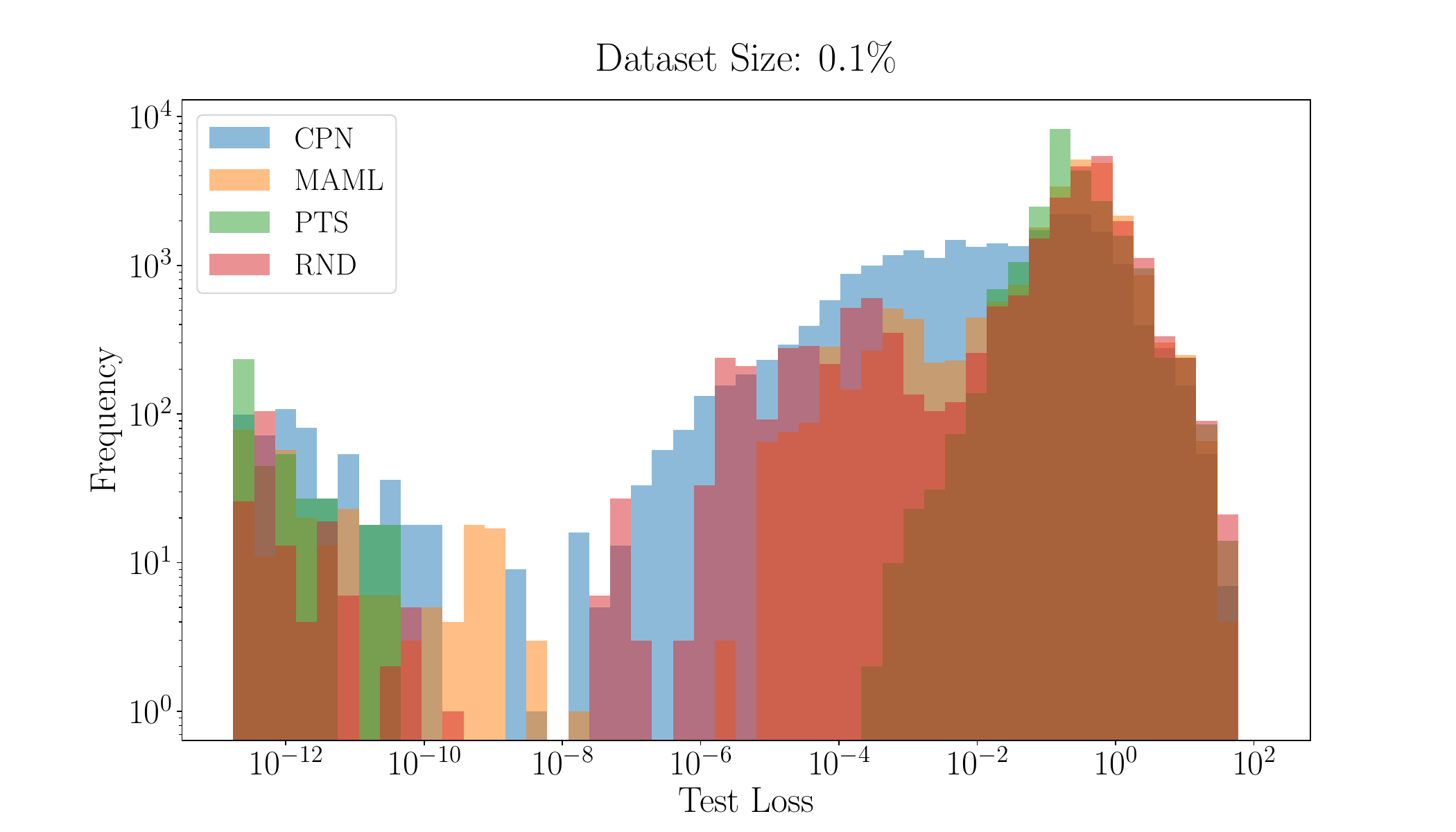}
  \includegraphics[width=0.48\linewidth]{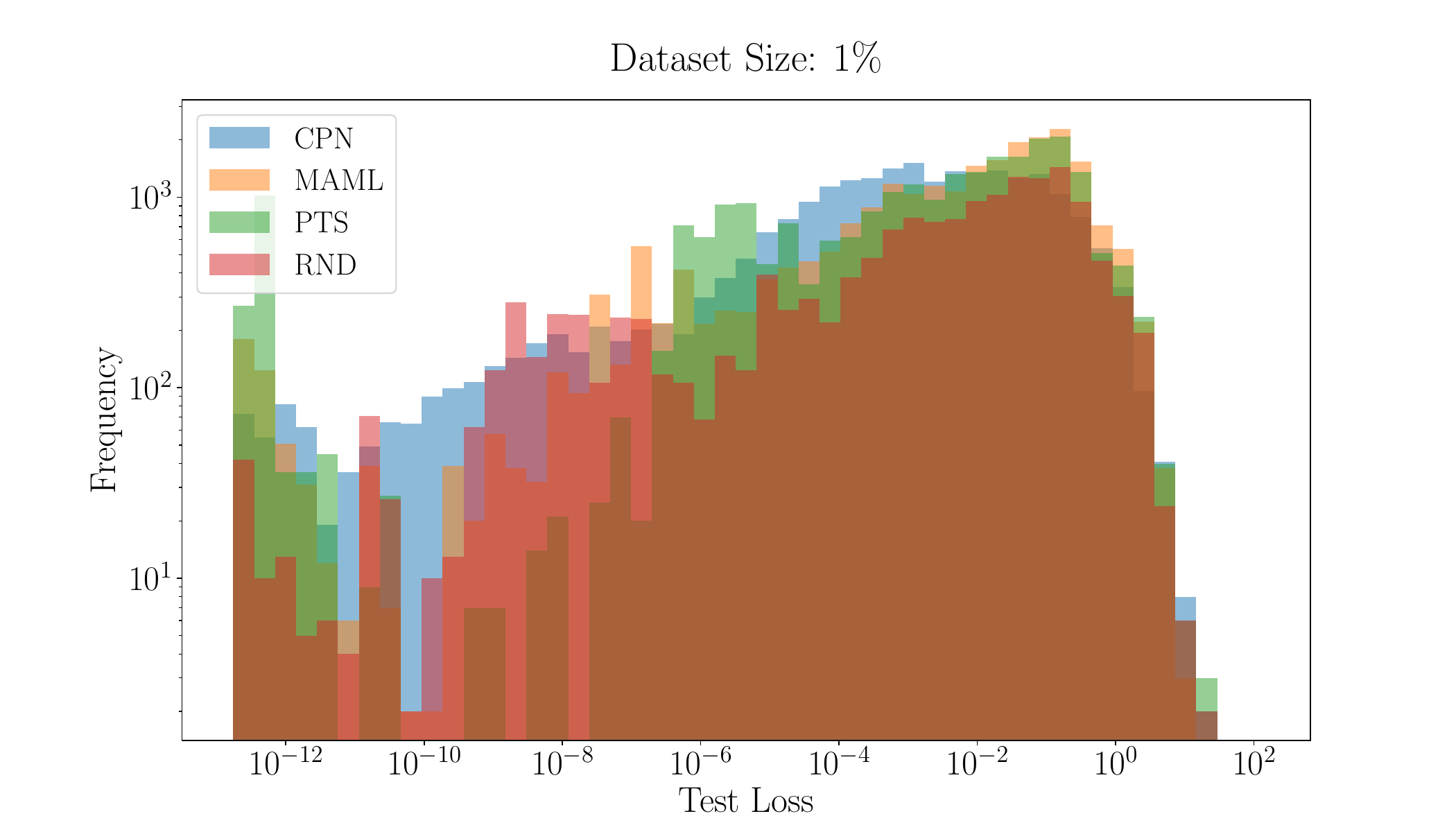}
  \includegraphics[width=0.48\linewidth]{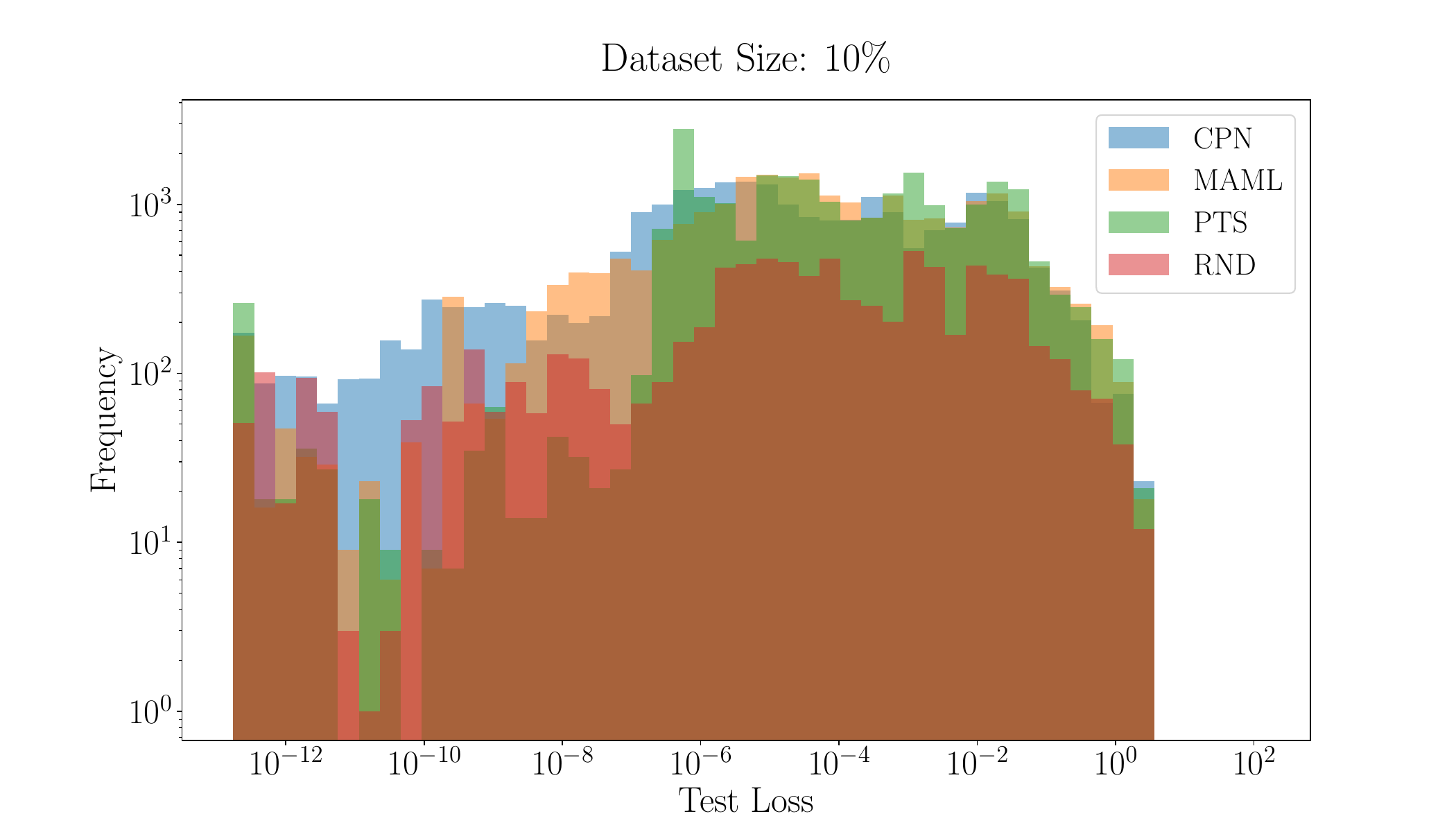}
  \includegraphics[width=0.48\linewidth]{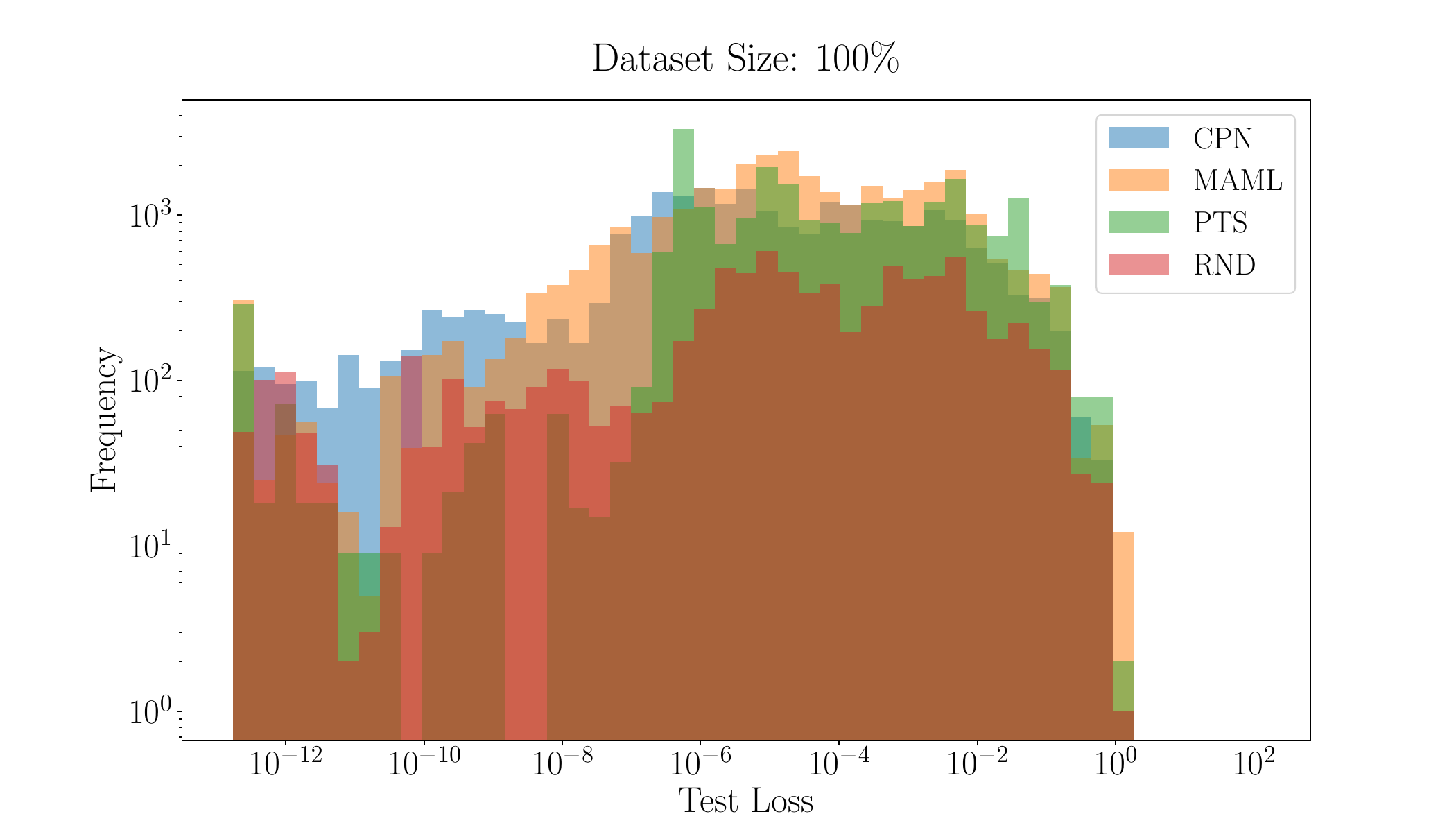}
  \caption{
    Histograms showing testing input losses of each initialization method at the epoch with the lowest validation loss for \textsc{ExeStackCPN} testing programs.
  }\label{fig:exestack_test_loss_best_val_epoch_data_efficiency}
\end{figure*}

Figure~\ref{fig:exestack_test_loss_best_val_epoch_data_efficiency} contains histograms showing, for a sample of $1{,}000$ \textsc{ExeStackCPN} test programs, the test loss each initialization method achieves at the epoch with the lowest validation loss.
At a dataset size of $0\%$, the distribution of \textsc{CompNet} losses is significantly skewed to smaller losses, relative to other initialization methods.
At a dataset size of $0.1\%$, every initialization method has roughly a bimodal loss distribution.
In the smaller-loss mode, each initialization method has comparable performance, but on the larger-loss mode, \textsc{CompNet}s still skew towards smaller losses.
As the dataset sizes increase, these modes begin to merge together, with \textsc{CompNet}s continuing to retain more mass in lower losses than other initialization methods.

Figure~\ref{fig:parrot_test_loss_best_val_epoch_data_efficiency} contains tables showing the test loss each initialization method achieves at the epoch with the lowest validation loss for \textsc{ParrotBenchCPN} programs.
At a dataset size of $0\%$, \textsc{CompNet}s have the worst loss on \texttt{fft}, but the best or among the best loss for all other benchmarks.
At a dataset size of $0.1\%$, \textsc{CompNet}s have the best or among the best loss for all benchmarks except for \texttt{sobel}, where MAML achieves the best loss by a significant margin.
At higher dataset sizes, none of the initialization methods consistently win for each benchmark.

Figure~\ref{fig:test_loss_improvement_by_fn_name_and_dataset_size_parrot} shows test loss improvements grouped by both programs and dataset sizes.
At a dataset size of $0\%$ for \texttt{fft}, \textsc{CompNet}s have among the worst test loss improvement, but for higher dataset sizes, \textsc{CompNet}s significantly outperform the other initialization methods, except at $100\%$, where MAML achieves a slightly better test loss improvement ($0.70\times$ vs. $0.75\times$).
At a dataset size of $0\%$ for \texttt{invk2j}, \textsc{CompNet}s have the best test loss improvement, but for higher dataset sizes, \textsc{CompNet}s achieve lower test loss improvements than other initialization methods, except at $100\%$, where \textsc{CompNet}s barely achieve the best test loss improvement ($0.93\times$ vs. $0.92\times$ for MAML and $0.90\times$ for pretrained surrogates).
At a dataset size of $0\%$ for \texttt{kmeans}, both \textsc{CompNet}s and pretrained surrogates achieve significant test loss improvements of greater than $3.5\times$, but at higher dataset sizes, \textsc{CompNet}s achieve better and better test loss improvements (maximum of $15.74\times$ at $100\%$ dataset size), whereas pretrained surrogates remain the same or worse.
At a dataset size of $0\%$ for \texttt{sobel}, all initialization methods achieve a test loss improvement of more than $1.6\times$, with \textsc{CompNet}s achieving the highest at $2.84\times$.
However, at higher dataset sizes, \textsc{CompNet}s and MAML alternate between improving, matching, or worsening test loss, and pretrained surrogates only worsen test loss, achieving a maximum of $0.77\times$ test loss improvement.


\begin{figure*}
  \centering
  \begin{tabular}{lllll}
    \toprule
    \multicolumn{5}{c}{Dataset Size $0\%$} \\
    \midrule
    Program & CPN & MAML & PTS & RND \\
    \midrule
    \texttt{fft} & $1.3 \pm 1.2$ & $0.6 \pm 0.2$ & $0.8 \pm 0.1$ & $0.6 \pm 0.3$ \\
    \texttt{invk2j} & $1.8 \pm 0.6$ & $2.0 \pm 0.6$ & $2.1 \pm 0.5$ & $2.4 \pm 0.8$ \\
    \texttt{kmeans} & $0.1 \pm \num{6.7e-03}$ & $0.7 \pm 0.4$ & $0.1 \pm \num{7.4e-04}$ & $0.2 \pm 0.2$ \\
    \texttt{sobel} & $0.1 \pm \num{3.0e-02}$ & $0.2 \pm 0.2$ & $0.2 \pm \num{3.5e-03}$ & $0.4 \pm 0.3$ \\
    \bottomrule
    \end{tabular}
    \\
    \vspace{1em}
    
    \begin{tabular}{lllll}
    \toprule
    \multicolumn{5}{c}{Dataset Size $0.1\%$} \\
    \midrule
    Program & CPN & MAML & PTS & RND \\
    \midrule
    \texttt{fft} & $\num{7.8e-05} \pm \num{1.1e-04}$ & $\num{1.9e-04} \pm \num{2.4e-04}$ & $\num{2.3e-04} \pm \num{4.6e-04}$ & $\num{1.6e-04} \pm \num{1.1e-04}$ \\
    \texttt{invk2j} & $0.2 \pm 0.2$ & $0.2 \pm 0.2$ & $0.2 \pm 0.2$ & $0.3 \pm 0.3$ \\
    \texttt{kmeans} & $\num{1.3e-02} \pm \num{1.5e-02}$ & $0.1 \pm \num{2.9e-02}$ & $\num{3.3e-02} \pm \num{1.6e-02}$ & $\num{3.9e-02} \pm \num{2.1e-02}$ \\
    \texttt{sobel} & $0.1 \pm \num{2.0e-02}$ & $\num{4.7e-02} \pm \num{2.1e-02}$ & $0.1 \pm \num{2.3e-02}$ & $0.1 \pm \num{1.9e-02}$ \\
    \bottomrule
    \end{tabular}
    \\
    \vspace{1em}
    
    \begin{tabular}{lllll}
    \toprule
    \multicolumn{5}{c}{Dataset Size $1\%$} \\
    \midrule
    Program & CPN & MAML & PTS & RND \\
    \midrule
    \texttt{fft} & $\num{3.6e-05} \pm \num{2.7e-05}$ & $\num{4.6e-05} \pm \num{2.2e-05}$ & $\num{1.0e-04} \pm \num{1.4e-04}$ & $\num{4.9e-05} \pm \num{2.8e-05}$ \\
    \texttt{invk2j} & $\num{1.5e-02} \pm \num{4.7e-03}$ & $\num{1.2e-02} \pm \num{3.7e-03}$ & $\num{1.3e-02} \pm \num{3.8e-03}$ & $\num{1.2e-02} \pm \num{3.9e-03}$ \\
    \texttt{kmeans} & $\num{4.7e-03} \pm \num{5.9e-03}$ & $\num{1.5e-02} \pm \num{1.1e-02}$ & $\num{1.3e-02} \pm \num{9.3e-03}$ & $\num{1.3e-02} \pm \num{1.3e-02}$ \\
    \texttt{sobel} & $\num{8.3e-03} \pm \num{4.9e-03}$ & $\num{8.1e-03} \pm \num{2.8e-03}$ & $\num{9.1e-03} \pm \num{3.2e-03}$ & $\num{6.0e-03} \pm \num{3.2e-03}$ \\
    \bottomrule
    \end{tabular}
    \\
    \vspace{1em}
    
    \begin{tabular}{lllll}
    \toprule
    \multicolumn{5}{c}{Dataset Size $10\%$} \\
    \midrule
    Program & CPN & MAML & PTS & RND \\
    \midrule
    \texttt{fft} & $\num{6.4e-06} \pm \num{9.0e-06}$ & $\num{1.2e-05} \pm \num{1.2e-05}$ & $\num{1.4e-05} \pm \num{1.5e-05}$ & $\num{1.3e-05} \pm \num{1.6e-05}$ \\
    \texttt{invk2j} & $\num{8.1e-03} \pm \num{1.4e-03}$ & $\num{6.5e-03} \pm \num{1.7e-03}$ & $\num{7.2e-03} \pm \num{1.3e-03}$ & $\num{7.3e-03} \pm \num{1.4e-03}$ \\
    \texttt{kmeans} & $\num{3.9e-03} \pm \num{5.0e-03}$ & $\num{1.3e-02} \pm \num{7.9e-03}$ & $\num{7.8e-03} \pm \num{7.0e-03}$ & $\num{1.3e-02} \pm \num{1.1e-02}$ \\
    \texttt{sobel} & $\num{1.7e-03} \pm \num{1.7e-03}$ & $\num{1.6e-03} \pm \num{1.4e-03}$ & $\num{2.6e-03} \pm \num{1.9e-03}$ & $\num{1.8e-03} \pm \num{1.6e-03}$ \\
    \bottomrule
    \end{tabular}
    \\
    \vspace{1em}
    
    \begin{tabular}{lllll}
    \toprule
    \multicolumn{5}{c}{Dataset Size $100\%$} \\
    \midrule
    Program & CPN & MAML & PTS & RND \\
    \midrule
    \texttt{fft} & $\num{2.2e-06} \pm \num{5.4e-06}$ & $\num{1.7e-06} \pm \num{3.1e-06}$ & $\num{4.2e-06} \pm \num{5.6e-06}$ & $\num{1.1e-06} \pm \num{1.1e-06}$ \\
    \texttt{invk2j} & $\num{3.3e-03} \pm \num{1.5e-04}$ & $\num{3.3e-03} \pm \num{6.5e-04}$ & $\num{3.4e-03} \pm \num{7.2e-04}$ & $\num{3.0e-03} \pm \num{4.8e-04}$ \\
    \texttt{kmeans} & $\num{3.4e-03} \pm \num{4.7e-03}$ & $\num{1.4e-02} \pm \num{9.5e-03}$ & $\num{5.2e-03} \pm \num{4.8e-03}$ & $\num{8.1e-03} \pm \num{4.3e-03}$ \\
    \texttt{sobel} & $\num{5.3e-04} \pm \num{7.9e-05}$ & $\num{4.6e-04} \pm \num{1.1e-04}$ & $\num{6.3e-04} \pm \num{8.5e-05}$ & $\num{4.1e-04} \pm \num{8.1e-05}$ \\
    \bottomrule
    \end{tabular}
  \caption{
    Average test loss achieved by each initialization method on the epoch with the best validation loss for \textsc{ParrotBenchCPN} programs.
    We include a table for each dataset size we evaluated on.
  }\label{fig:parrot_test_loss_best_val_epoch_data_efficiency}
\end{figure*}

\begin{figure*}
  \centering
  \begin{tabular}{lllll}
  \toprule
  \multicolumn{4}{c}{Benchmark: \texttt{fft}} \\
  \midrule
  Dataset Size & CPN & MAML & PTS \\
  \midrule
  $0\%$ & $0.76\times$ & $1.13\times$ & $0.76\times$ \\
  $0.1\%$ & $2.61\times$ & $0.84\times$ & $0.77\times$ \\
  $1\%$ & $1.94\times$ & $1.08\times$ & $0.54\times$ \\
  $10\%$ & $2.54\times$ & $1.17\times$ & $0.98\times$ \\
  $100\%$ & $0.70\times$ & $0.75\times$ & $0.26\times$ \\
  \bottomrule
  \end{tabular}
  \hspace{1em}
  \begin{tabular}{lllll}
  \toprule
  \multicolumn{4}{c}{Benchmark: \texttt{invk2j}} \\
  \midrule
  Dataset Size & CPN & MAML & PTS \\
  \midrule
  $0\%$ & $1.35\times$ & $1.15\times$ & $1.12\times$ \\
  $0.1\%$ & $1.14\times$ & $1.21\times$ & $1.26\times$ \\
  $1\%$ & $0.83\times$ & $0.98\times$ & $0.97\times$ \\
  $10\%$ & $0.90\times$ & $1.13\times$ & $1.02\times$ \\
  $100\%$ & $0.93\times$ & $0.92\times$ & $0.90\times$ \\
  \bottomrule
  \end{tabular}
  \\[1.5em]
  
  \begin{tabular}{lllll}
  \toprule
  \multicolumn{4}{c}{Benchmark: \texttt{kmeans}} \\
  \midrule
  Dataset Size & CPN & MAML & PTS \\
  \midrule
  $0\%$ & $3.68\times$ & $0.31\times$ & $3.58\times$ \\
  $0.1\%$ & $5.24\times$ & $0.70\times$ & $1.23\times$ \\
  $1\%$ & $8.22\times$ & $0.94\times$ & $1.08\times$ \\
  $10\%$ & $11.97\times$ & $1.04\times$ & $3.30\times$ \\
  $100\%$ & $15.74\times$ & $0.68\times$ & $3.59\times$ \\
  \bottomrule
  \end{tabular}
  \hspace{1em}
  \begin{tabular}{lllll}
  \toprule
  \multicolumn{4}{c}{Benchmark: \texttt{sobel}} \\
  \midrule
  Dataset Size & CPN & MAML & PTS \\
  \midrule
  $0\%$ & $2.84\times$ & $1.63\times$ & $1.92\times$ \\
  $0.1\%$ & $1.00\times$ & $1.08\times$ & $0.77\times$ \\
  $1\%$ & $0.75\times$ & $0.75\times$ & $0.67\times$ \\
  $10\%$ & $1.17\times$ & $1.11\times$ & $0.70\times$ \\
  $100\%$ & $0.77\times$ & $0.90\times$ & $0.66\times$ \\
  \bottomrule
  \end{tabular}

  \caption{
    Geometric mean testing loss improvement over randomly initialized surrogates on \textsc{ParrotBenchCPN} programs, grouped by both programs and dataset sizes.
  }\label{fig:test_loss_improvement_by_fn_name_and_dataset_size_parrot}
\end{figure*}

\section{Neural Surrogates for Color Quantization (Extended)}\label{sec:kmeans_results_extended}
In this section, we present visual results for all dataset sizes in the data efficiency evaluation, as well as quantitative results for $10$- and $15$-color palettes.

\subsection{Visual Results}
\begin{figure*}
  \includegraphics[width=0\textwidth]{figures/color_palette/color_quantization_baboon_dataset_size_0_0.pdf}
  \vspace*{-1em}
  \includegraphics[width=\textwidth]{figures/color_palette/color_quantization_baboon_dataset_size_0_0.pdf}
  \\
  \vspace*{-1em}
  \includegraphics[width=\textwidth]{figures/color_palette/color_quantization_baboon_dataset_size_0_001.pdf}
  \\
  \vspace*{-1em}
  \includegraphics[width=\textwidth]{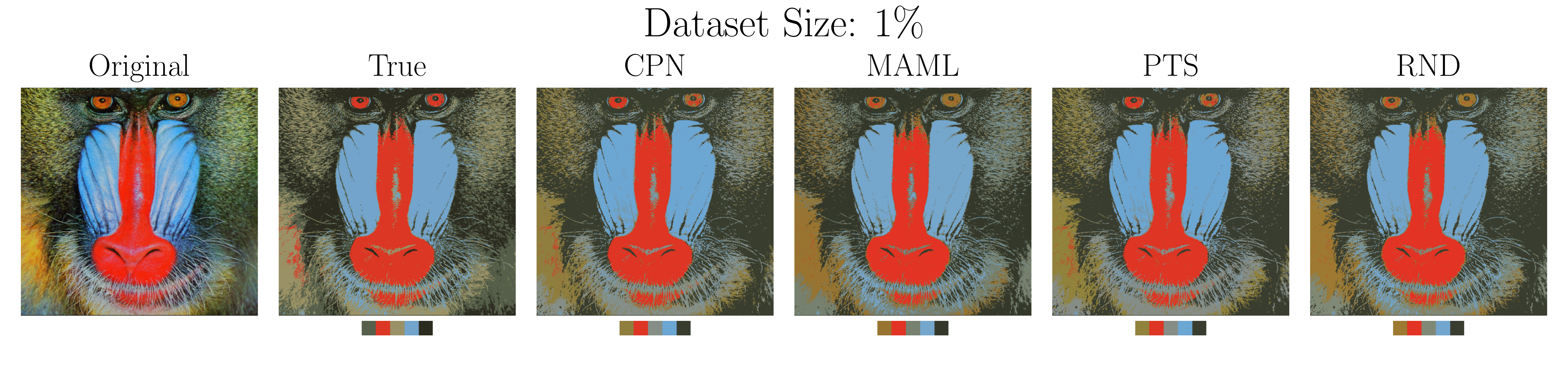}
  \\
  \vspace*{-1em}
  \includegraphics[width=\textwidth]{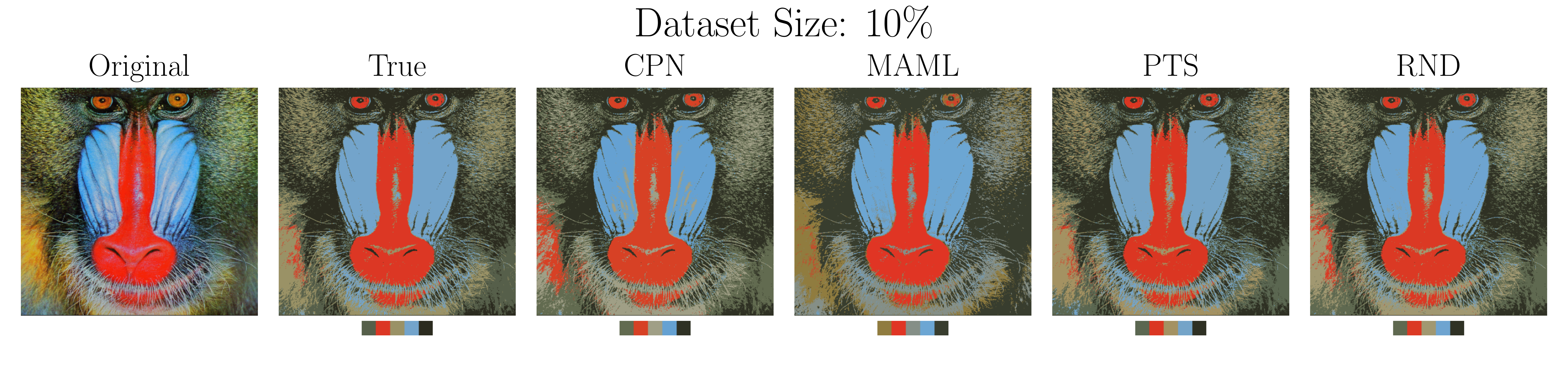}
  \\
  \vspace*{-1em}
  \includegraphics[width=\textwidth]{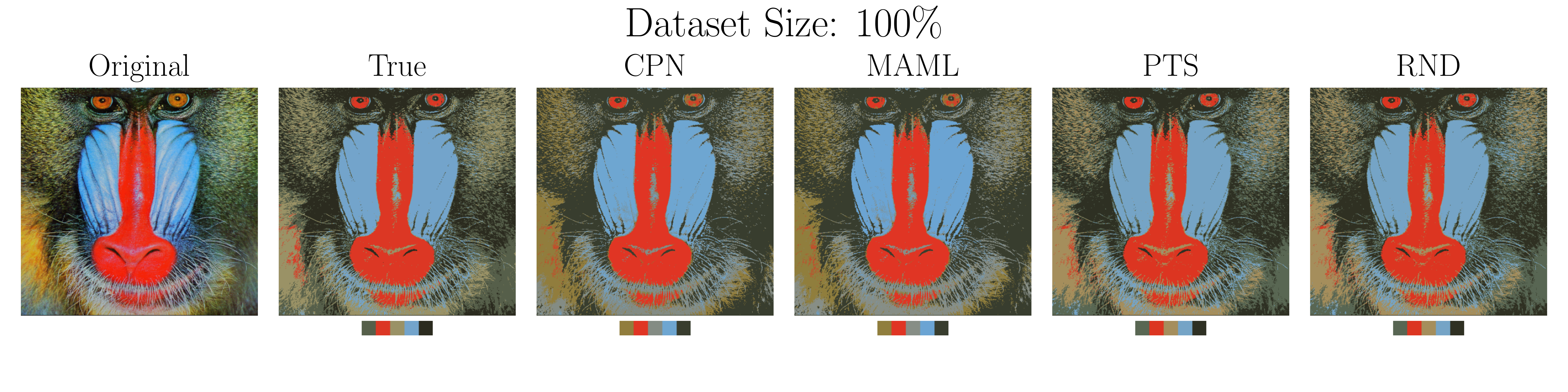}
  \\
  \vspace*{-1em}
  \caption{
    Color quantization results for a ground-truth NumPy implementation (``True'') vs. approximate implementations over various dataset sizes, each on a separate row.
    In each row, the original image of a baboon is on the left, followed by images transformed to adhere to a palette of $5$ colors.
  }\label{fig:full_kmeans_visual_results}
\end{figure*}

Figure~\ref{fig:full_kmeans_visual_results} contains visual results for surrogates trained on $0\%$, $0.1\%$, $1\%$, $10\%$, and $100\%$ of the \texttt{kmeans} training set.
\textsc{CompNet}s and MAML initializations are the only initialization methods that produce an image with detail at a dataset size of $0\%$.
Of the two, the \textsc{CompNet} result has more definition.
At a dataset size of $0.1\%$, all initialization methods produce images with detail.
MAML- and randomly initialized surrogates produce images with duller colors than \textsc{CompNet}-initialized and pretrained surrogates.
At larger dataset sizes, all initialization methods converge to images that look similar to the reference image.

\subsection{Quantitative Results}
Figures~\ref{fig:kmeans_10_color_palette_quant_results} and \ref{fig:kmeans_15_color_palette_quant_results} present quantitative results for $10$- and $15$-color palettes, respectively.
At all color palette sizes and dataset sizes, \textsc{CompNet}-initialized surrogates produce better results, in terms of both MSE and SSIM.




\begin{figure*}
  \centering
  \begin{tabular}{lllll}
    \toprule
    Dataset Size & CPN & MAML & PTS & RND \\
    \midrule
    $0\%$ & $\mathbf{2.87 \cdot 10^3} \pm 615.$ & $\num{3.03e+03} \pm 394.$ & $\num{3.28e+03} \pm 129.$ & $\num{3.30e+03} \pm \num{0.}$ \\
    $0.1\%$ & $\mathbf{979.} \pm 853.$ & $\num{1.90e+03} \pm 594.$ & $\num{1.72e+03} \pm 793.$ & $\num{1.46e+03} \pm 583.$ \\
    $1\%$ & $\mathbf{410.} \pm 194.$ & $677. \pm 267.$ & $631. \pm 226.$ & $615. \pm 298.$ \\
    $10\%$ & $\mathbf{401.} \pm 181.$ & $639. \pm 169.$ & $576. \pm 252.$ & $631. \pm 317.$ \\
    $100\%$ & $\mathbf{395.} \pm 184.$ & $627. \pm 237.$ & $498. \pm 229.$ & $510. \pm 154.$ \\
    \bottomrule
    \end{tabular}
    \\
    \vspace{1em}
    
    \begin{tabular}{lllll}
    \toprule
    Dataset Size & CPN & MAML & PTS & RND \\
    \midrule
    $0\%$ & $\mathbf{0.28} \pm 0.13$ & $0.20 \pm 0.04$ & $0.19 \pm 0.03$ & $0.19 \pm 0.00$ \\
    $0.1\%$ & $\mathbf{0.60} \pm 0.16$ & $0.42 \pm 0.12$ & $0.45 \pm 0.16$ & $0.49 \pm 0.09$ \\
    $1\%$ & $\mathbf{0.73} \pm 0.10$ & $0.63 \pm 0.11$ & $0.64 \pm 0.09$ & $0.66 \pm 0.12$ \\
    $10\%$ & $\mathbf{0.74} \pm 0.10$ & $0.62 \pm 0.07$ & $0.66 \pm 0.12$ & $0.65 \pm 0.14$ \\
    $100\%$ & $\mathbf{0.74} \pm 0.10$ & $0.63 \pm 0.11$ & $0.69 \pm 0.12$ & $0.68 \pm 0.09$ \\
    \bottomrule
  \end{tabular}

  \caption{
    Quantitative comparison of end-to-end results produced by various initialization methods on color quantization with a palette size of $10$ colors.
    \textbf{(Top)} The average mean squared error (MSE) of the image produced by each initialization method compared to the image produced by a ground-truth implementation of the \texttt{kmeans} kernel (lower is better).
    \textbf{(Bottom)} The average structural similarity index measure (SSIM) of the image produced by each initialization method compared to the image produced by a ground-truth implementation of the \texttt{kmeans} kernel (higher is better).
  }\label{fig:kmeans_10_color_palette_quant_results}
\end{figure*}

\begin{figure*}
  \centering
\begin{tabular}{lllll}
  \toprule
  Dataset Size & CPN & MAML & PTS & RND \\
  \midrule
  $0\%$ & $\mathbf{2.74 \cdot 10^3} \pm 506.$ & $\num{3.12e+03} \pm 386.$ & $\num{3.39e+03} \pm 81.$ & $\num{3.40e+03} \pm \num{0.}$ \\
  $0.1\%$ & $\mathbf{906.} \pm 782.$ & $\num{1.84e+03} \pm 633.$ & $\num{1.74e+03} \pm 801.$ & $\num{1.53e+03} \pm 783.$ \\
  $1\%$ & $\mathbf{417.} \pm 166.$ & $647. \pm 250.$ & $588. \pm 206.$ & $588. \pm 250.$ \\
  $10\%$ & $\mathbf{404.} \pm 163.$ & $578. \pm 155.$ & $545. \pm 207.$ & $577. \pm 279.$ \\
  $100\%$ & $\mathbf{392.} \pm 150.$ & $588. \pm 204.$ & $477. \pm 177.$ & $484. \pm 100.$ \\
  \bottomrule
  \end{tabular}
  \\
  \vspace{1em}
  
  \begin{tabular}{lllll}
  \toprule
  Dataset Size & CPN & MAML & PTS & RND \\
  \midrule
  $0\%$ & $\mathbf{0.31} \pm 0.11$ & $0.18 \pm 0.04$ & $0.16 \pm 0.02$ & $0.16 \pm 0.00$ \\
  $0.1\%$ & $\mathbf{0.61} \pm 0.15$ & $0.42 \pm 0.12$ & $0.44 \pm 0.16$ & $0.48 \pm 0.12$ \\
  $1\%$ & $\mathbf{0.71} \pm 0.07$ & $0.63 \pm 0.09$ & $0.65 \pm 0.08$ & $0.66 \pm 0.08$ \\
  $10\%$ & $\mathbf{0.72} \pm 0.08$ & $0.64 \pm 0.06$ & $0.66 \pm 0.08$ & $0.66 \pm 0.11$ \\
  $100\%$ & $\mathbf{0.73} \pm 0.07$ & $0.64 \pm 0.08$ & $0.69 \pm 0.07$ & $0.68 \pm 0.05$ \\
  \bottomrule
  \end{tabular}
  \caption{
    Quantitative comparison of end-to-end results produced by various initialization methods on color quantization with a palette size of $15$ colors.
    \textbf{(Top)} The average mean squared error (MSE) of the image produced by each initialization method compared to the image produced by a ground-truth implementation of the \texttt{kmeans} kernel (lower is better).
    \textbf{(Bottom)} The average structural similarity index measure (SSIM) of the image produced by each initialization method compared to the image produced by a ground-truth implementation of the \texttt{kmeans} kernel (higher is better).
  }\label{fig:kmeans_15_color_palette_quant_results}
\end{figure*}

\section{Training Time Improvements}\label{sec:training_time}
To assess whether \textsc{CompNet}s improve training time of neural surrogates, we use \textsc{CompNet}s to initialize neural surrogates, finetune on training data until they reach a target test loss, then compare the results to those of other initialization methods.
We first detail the methodology of this experiment, then present results.


\vspace*{-0.25em}
\subsubsection{Methodology}\label{sec:training_time_efficiency_methodology}
We now describe the methodology for setting a target test loss to use as a stopping condition, the configuration space we sweep over, how we quantify improvements, and how we visualize results.

\paragraph{Setting a Target Test Loss.}
We set a target test loss for each program by training $9$ randomly initialized surrogates for $5{,}000$ epochs.
The average final test loss is the target test loss for all initialization methods.

\paragraph{Experiment Configurations.}
In this experiment, we sweep over configurations consisting of a program and an initialization method.
Given a program and initialization method, we produce a neural surrogate initialization.
We then train the initialized neural surrogate on the training input set until it reaches the target test loss or until it reaches $15{,}000$ epochs.
We call whichever epoch comes first the \textit{finish epoch} for the trial.

\paragraph{Quantifying Improvements.}
We define the improvement for a given configuration (consisting of a program and initialization method) as the ratio of the finish epoch for random initialization and the finish epoch by the configuration's initialization method.
All finish epochs are averaged over trials (using arithmetic mean) prior to computing ratios.
For each initialization method, we report the geometric mean of the improvements grouped by program, grouped by dataset size, and overall.
For some programs and initialization methods, the resulting surrogates achieve losses of $0$.
We discard these results before computing the geometric mean.

There are a few subtleties in this methodology.
First, note that random initialization does not always have a finish epoch of $5{,}000$, because the target error set after $5{,}000$ epochs of training may have already been achieved earlier in training.
Also, since the timeout epoch ($15{,}000$) is $3\times$ the baseline finish epoch ($5{,}000$), the worst case slowdown for each initialization method is $\frac{1}{3}\times$.

\paragraph{Visualizing Results.}
Since we evaluate on many programs in \textsc{ExeStackCPN}, we plot the number of finished programs as a function of the number of epochs for each initialization method.
For each program and initialization method, we calculate the finish epoch for that program as the average finish epoch over all instances of the initialization method and all trials for that instance.

\subsubsection{Results}\label{sec:training_time_results}

\begin{figure}
  \centering
  \begin{tabular}{lllll}
    \toprule
    Statistic & CPN & MAML & PTS \\
    \midrule
    0th & $0.03\times$ & $\mathbf{0.06\times}$ & $0.03\times$ \\
    25th & $\mathbf{1.16\times}$ & $0.85\times$ & $0.61\times$ \\
    50th & $\mathbf{3.43\times}$ & $1.19\times$ & $1.03\times$ \\
    75th & $\mathbf{23.96\times}$ & $1.68\times$ & $1.56\times$ \\
    100th & $\mathbf{8.27 \cdot 10^3\times}$ & $26.54\times$ & $49.39\times$ \\
    \midrule
    MPI & \textbf{18}th & 36th & 48th \\
    \midrule
    GM & $\mathbf{7.28\times}$ & $1.16\times$ & $0.93\times$ \\
    \bottomrule
  \end{tabular}
  \caption{Geometric mean improvements and percentile improvements to training time over randomly initialized surrogates on a sample of $1{,}000$ \textsc{ExeStackCPN} test programs.
  MPI is the minimum percentile at which an initialization method improves over random initialization.
  }\label{fig:training_time_exestack_test_results_summary}
\end{figure}

\begin{figure}
  \centering
  \includegraphics[width=\linewidth]{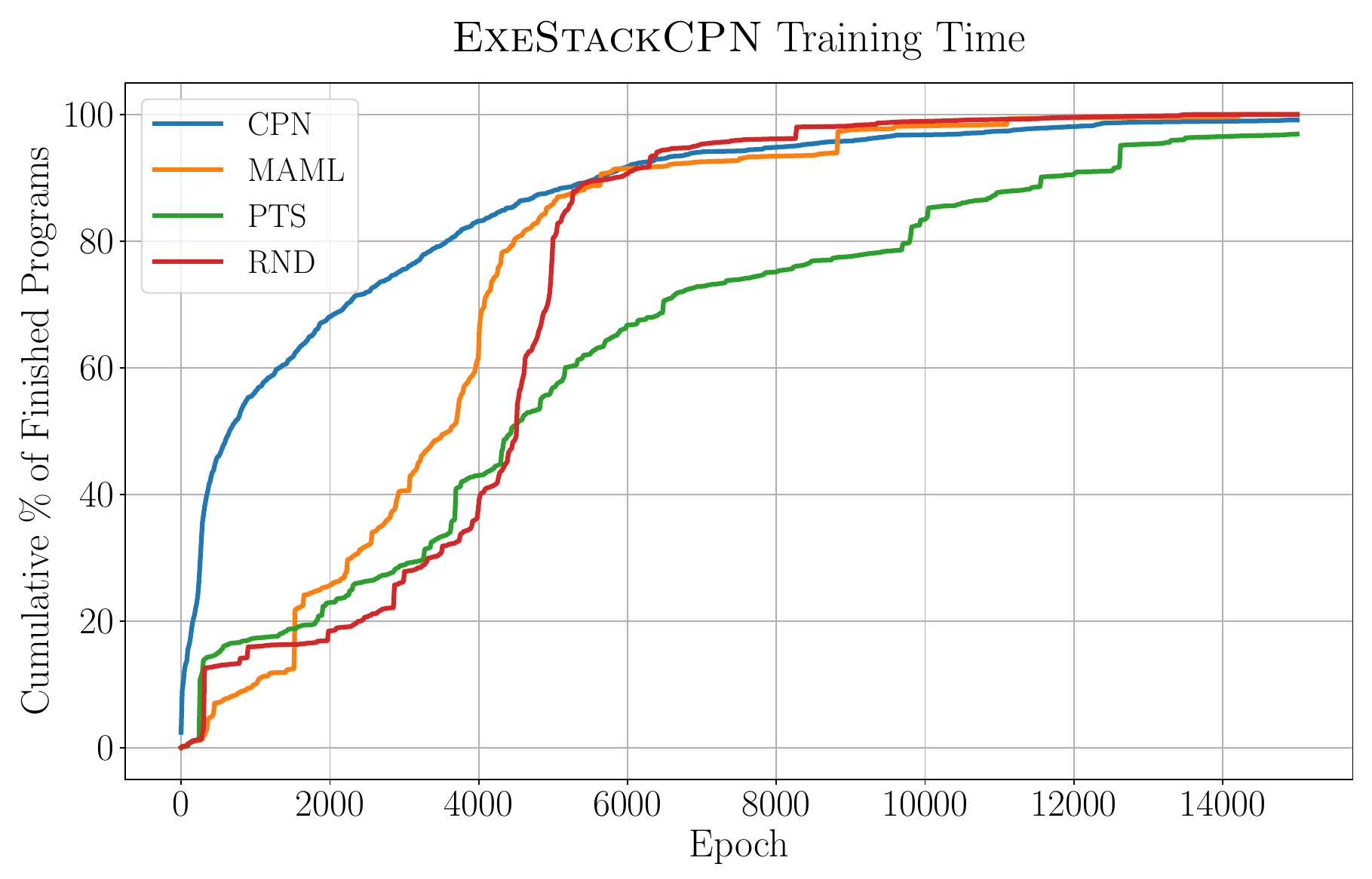}
  \caption{
    Epoch vs. percentage of \textsc{ExeStackCPN} programs that each initialization method finished at that epoch.
  }\label{fig:cumulative_exestack_training_time}
\end{figure}

\begin{figure}
  \centering
  \begin{tabular}{lllll}
    \toprule
    Statistic & CPN & MAML & PTS \\
    \midrule
    0th & $0.39\times$ & $0.38\times$ & $\mathbf{0.42\times}$ \\
    25th & $0.54\times$ & $\mathbf{0.63\times}$ & $0.57\times$ \\
    50th & $\mathbf{1.01\times}$ & $0.96\times$ & $0.83\times$ \\
    75th & $\mathbf{108.21\times}$ & $1.07\times$ & $7.91\times$ \\
    100th & $\mathbf{849.78\times}$ & $25.66\times$ & $278.11\times$ \\
    \midrule
    MPI & \textbf{50}th & 54th & 60th \\
    \midrule
    GM & $\mathbf{4.31\times}$ & $1.07\times$ & $2.35\times$ \\
    \bottomrule
  \end{tabular}
  \caption{
    Training time improvements at over random initialization on \textsc{ParrotBenchCPN}.
    We include percentiles from $0$th to $100$th, the minimum percentile of improvement (MPI), and the overall geometric mean improvement (GM).
  }
  \label{fig:training_time_parrot_results_summary_percentiles}
\end{figure}

\begin{figure}
  \hspace{1em}
  \begin{tabular}{lllll}
    \toprule
    Program & CPN & MAML & PTS \\
    \midrule
    \texttt{fft} & $\mathbf{1.43\times}$ & $0.83\times$ & $0.80\times$ \\
    \texttt{invk2j} & $0.49\times$ & $\mathbf{0.65\times}$ & $0.56\times$ \\
    \texttt{kmeans} & $\mathbf{674.47\times}$ & $2.15\times$ & $86.87\times$ \\
    \texttt{sobel} & $0.74\times$ & $\mathbf{1.14\times}$ & $0.79\times$ \\
    \bottomrule
  \end{tabular}
  \caption{
    Geometric mean training time improvements over random initialization on \textsc{ParrotBenchCPN}.
  }\label{fig:training_time_parrot_results_summary}
\end{figure}

The results are summarized in Figures~\ref{fig:training_time_exestack_test_results_summary}~and~\ref{fig:cumulative_exestack_training_time} for the sample of \textsc{ExeStackCPN} test programs and Figures~\ref{fig:training_time_parrot_results_summary_percentiles} and \ref{fig:training_time_parrot_results_summary} for \textsc \textsc{ParrotBenchCPN}.

\paragraph{\textsc{ExeStackCPN} Test Programs.}
\textsc{CompNet}s achieve the best results on average, with a $7.28\times$ improvement over random initialization, whereas MAML and pretrained surrogates achieve $1.16\times$ and $0.93\times$ improvements, respectively.
\textsc{CompNet}s improve over random initialization in as low as the $18$th perecentile, whereas MAML and pretrained surrogates improve over random initialization after the $36$th and $48$th percentile, respectively.

Until the $\approx5{,}000$th epoch, \textsc{CompNet}s finish training on strictly more programs than all other initialization methods.
At the $5{,}000$th epoch, \textsc{CompNet}s finish training for $\approx 90\%$ of programs.
For the remaining $10\%$ of programs, random initialization and MAML begin to overtake \textsc{CompNet}s, at epochs $\approx 6{,}250$ and $\approx 9{,}000$, respectively.

\paragraph{\textsc{ParrotBenchCPN} Programs.}
\textsc{CompNet}s achieve the best results on average, with a $4.31\times$ improvement over random initialization, whereas MAML and pretrained surrogates achieve $1.07\times$ and $2.35\times$ improvements, respectively.
\textsc{CompNet}s improve over random initialization after the $50$th perecentile, MAML improves over random initialization after the $54$th percentile, and pretrained surrogates improve over random initialization after the $48$th percentile.

\textsc{CompNet}s range between improvements of $0.49\times$ on \texttt{invk2j} to $674\times$ on \texttt{kmeans}.
The variance between other techniques is smaller, with MAML varying between $0.65\times$ on \texttt{invk2j} and $2.15\times$ on \texttt{kmeans}, and pretrained surrogates varying between $0.56\times$ on \texttt{invk2j} and $87\times$ on \texttt{kmeans}.
All initialization methods present slowdowns on \texttt{invk2j} and speedups on \texttt{kmeans}, so it is possible \texttt{ExeStackCPN} does not include similar computations to \texttt{invk2j} but does include similar computations to \texttt{kmeans}.
However, we use an extensive decontamination methodology (see Appendix~\ref{sec:exestack_hbn_decontam}), so we conclude these similarities are abstract in nature.


Since \textsc{CompNet}s improve training time over random initialization for programs in both \textsc{ExeStackCPN} and \textsc{ParrotBenchCPN}, we answer yes to RQ 2.

\begin{figure*}
  \centering
  \includegraphics[width=0.7\linewidth]{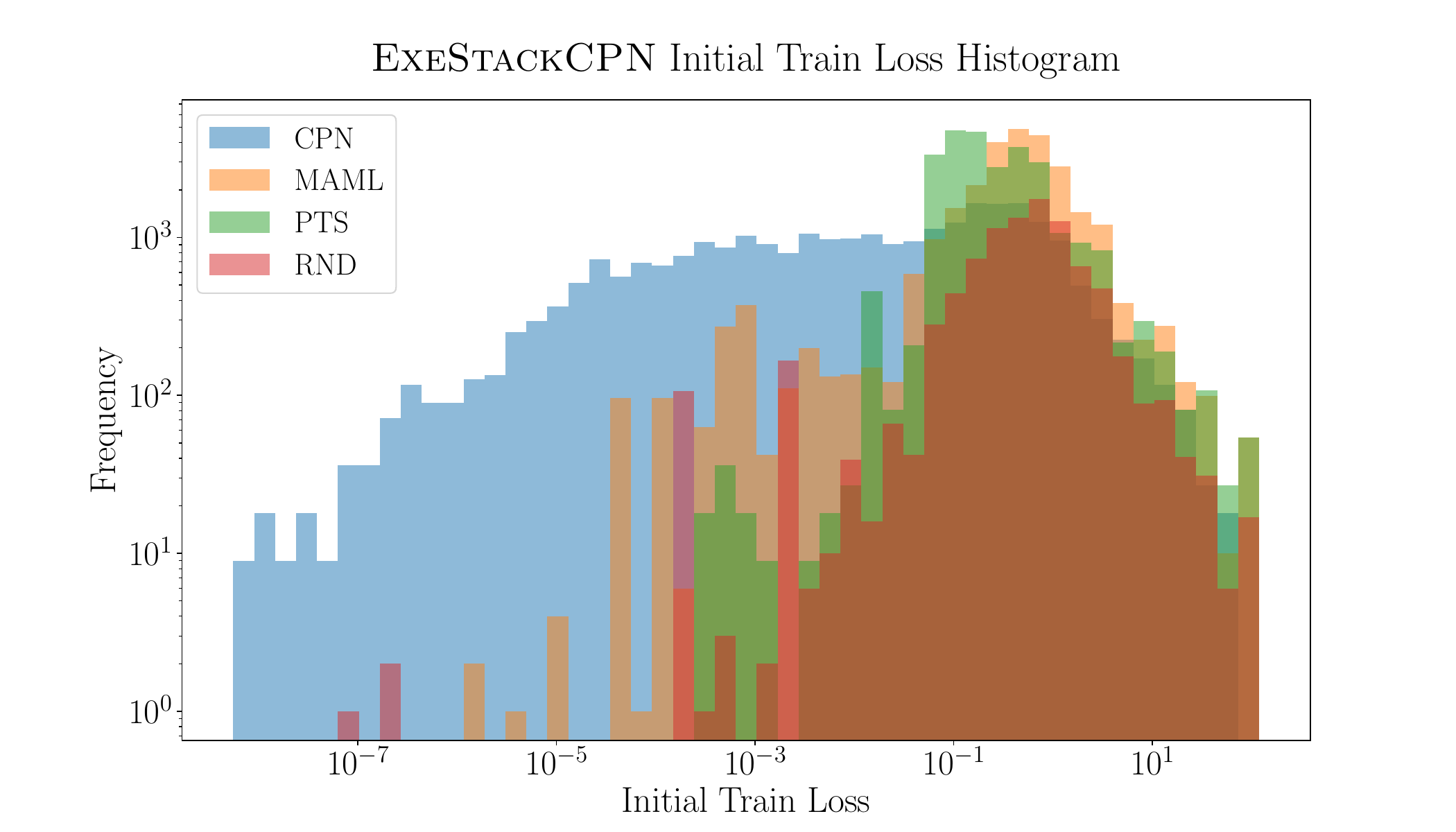}
  \includegraphics[width=0.7\linewidth]{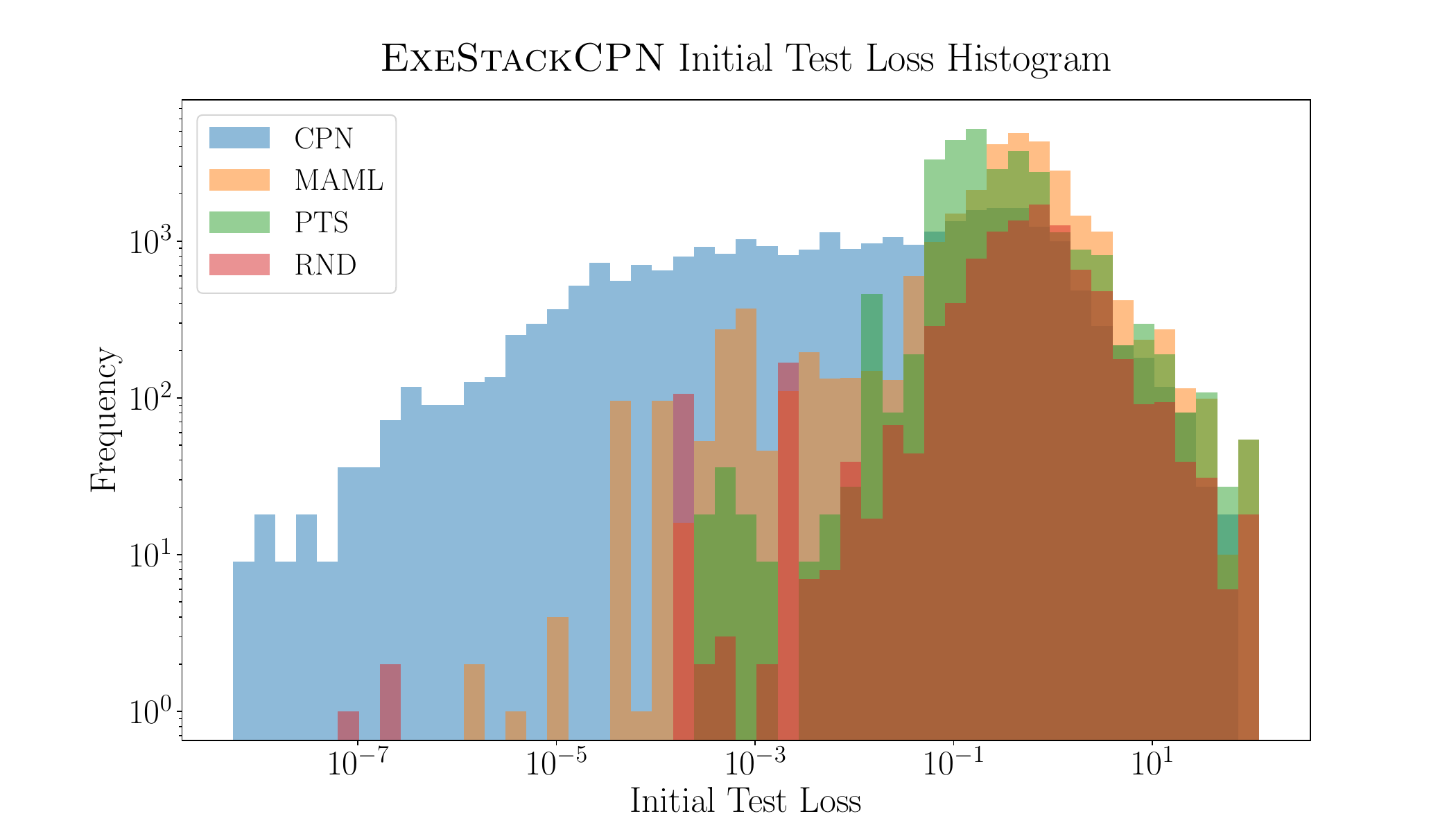}
  \includegraphics[width=0.7\linewidth]{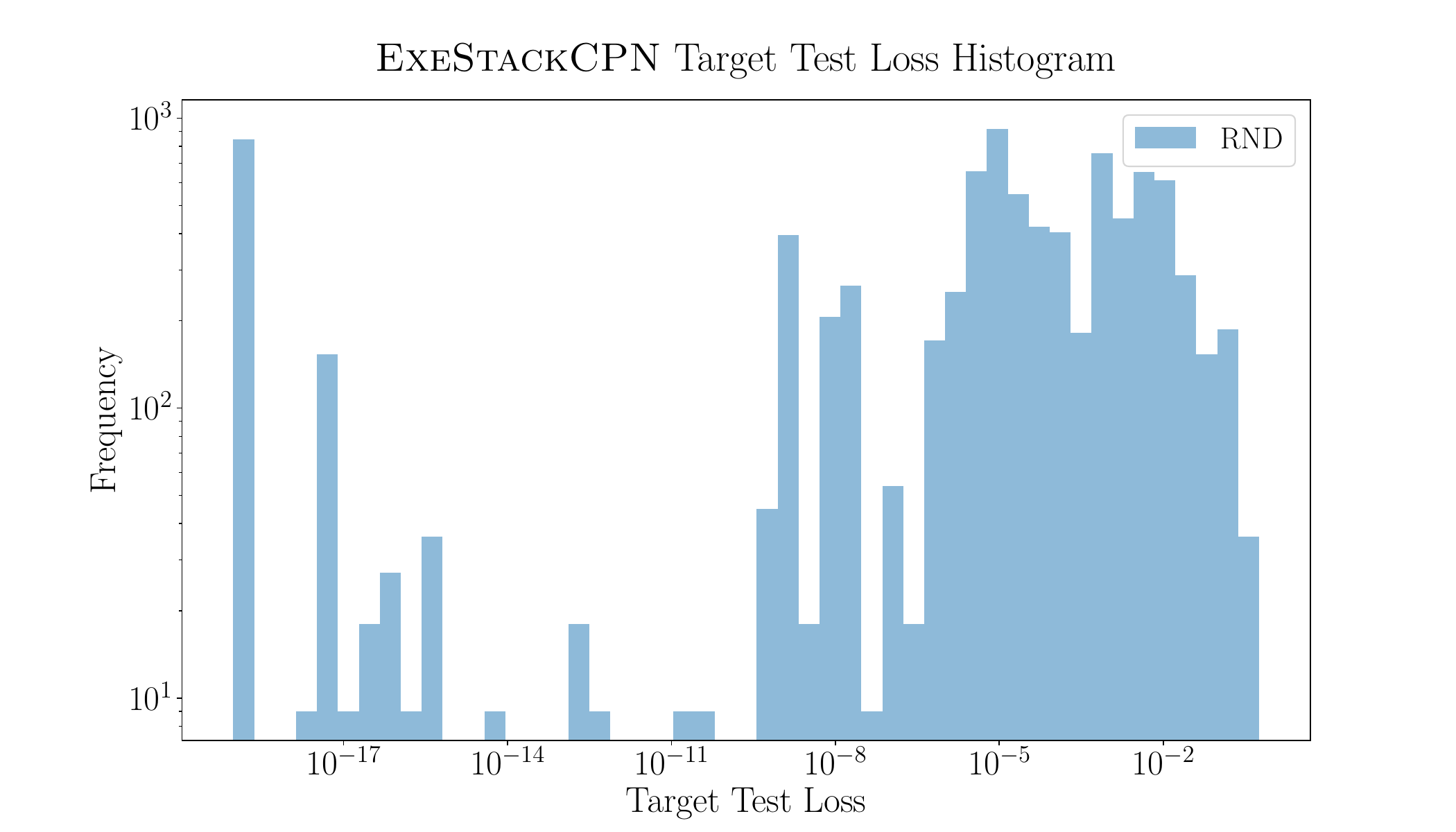}
  \caption{
    Histogram of initial training losses (top) and initial testing losses (bottom) for surrogates produced by each initialization method in the training time evaluation, as well as a histogram of the target testing losses set by random initialization after training for $5{,}000$ epochs.
    Losses are not averaged across instances of initialization methods and trials.
    Note that both the $x$ and $y$ axes are log-scale.
  }\label{fig:exestack_training_time_loss}
\end{figure*}

\paragraph{Additional Data.}
We present the initial train losses, initial test lossses, and target test losses for both \textsc{ExeStackCPN} (Figure~\ref{fig:exestack_training_time_loss}) and \textsc{ParrotBenchCPN} (Figures~\ref{fig:parrot_training_time_initial_train_loss}, \ref{fig:parrot_training_time_initial_test_loss}, and \ref{fig:parrot_training_time_target_test_loss}).
We also present the average finish epoch (Figure~\ref{fig:avg_finish_epoch_parrot}) for each initialization method and the number of timeouts (Figure~\ref{fig:num_timeouts_parrot}) on \textsc{ParrotBenchCPN} programs.

\begin{figure*}
  \centering
    \begin{tabular}{lllll}
    \toprule
    Program & CPN (0) & CPN (1) & CPN (2) & CPN \\
    \midrule
    \texttt{fft} & $0.48 \pm \num{1.81e-06}$ & $0.43 \pm \num{8.51e-06}$ & $2.94 \pm \num{6.80e-05}$ & $1.28 \pm 1.20$ \\
    \texttt{invk2j} & $1.14 \pm \num{5.67e-04}$ & $1.90 \pm \num{8.89e-04}$ & $2.54 \pm \num{1.52e-03}$ & $1.86 \pm 0.58$ \\
    \texttt{kmeans} & $0.12 \pm \num{1.44e-05}$ & $0.09 \pm \num{1.33e-05}$ & $0.08 \pm \num{1.20e-05}$ & $0.10 \pm 0.02$ \\
    \texttt{sobel} & $0.09 \pm \num{3.62e-04}$ & $0.13 \pm \num{4.21e-04}$ & $0.17 \pm \num{3.72e-04}$ & $0.13 \pm 0.03$ \\
    \bottomrule
    \end{tabular}
    
    \vspace{1em}
    \begin{tabular}{lllll}
    \toprule
    Program & MAML (0) & MAML (1) & MAML (2) & MAML \\
    \midrule
    \texttt{fft} & $0.57 \pm 0.18$ & $0.59 \pm 0.19$ & $0.53 \pm 0.14$ & $0.56 \pm 0.16$ \\
    \texttt{invk2j} & $2.08 \pm 0.64$ & $2.07 \pm 0.59$ & $2.02 \pm 0.56$ & $2.06 \pm 0.58$ \\
    \texttt{kmeans} & $0.76 \pm 0.46$ & $0.64 \pm 0.38$ & $0.79 \pm 0.52$ & $0.73 \pm 0.44$ \\
    \texttt{sobel} & $0.24 \pm 0.16$ & $0.18 \pm 0.12$ & $0.25 \pm 0.22$ & $0.22 \pm 0.17$ \\
    \bottomrule
    \end{tabular}
    
    \vspace{1em}
    \begin{tabular}{lllll}
    \toprule
    Program & PTS (0) & PTS (1) & PTS (2) & PTS \\
    \midrule
    \texttt{fft} & $0.83 \pm 0.07$ & $0.84 \pm 0.07$ & $0.84 \pm 0.07$ & $0.84 \pm 0.07$ \\
    \texttt{invk2j} & $2.11 \pm 0.57$ & $2.10 \pm 0.56$ & $2.12 \pm 0.59$ & $2.11 \pm 0.55$ \\
    \texttt{kmeans} & $0.10 \pm \num{1.92e-05}$ & $0.10 \pm \num{1.88e-05}$ & $0.10 \pm \num{1.87e-05}$ & $0.10 \pm \num{1.34e-03}$ \\
    \texttt{sobel} & $0.18 \pm \num{4.37e-04}$ & $0.19 \pm \num{4.40e-04}$ & $0.19 \pm \num{4.43e-04}$ & $0.19 \pm \num{3.55e-03}$ \\
    \bottomrule
  \end{tabular}
  \vspace{1em}

  \begin{tabular}{ll}
    \toprule
    Program & RND \\
    \midrule
    \texttt{fft} & $0.64 \pm 0.35$ \\
    \texttt{invk2j} & $2.37 \pm 0.80$ \\
    \texttt{kmeans} & $0.24 \pm 0.25$ \\
    \texttt{sobel} & $0.36 \pm 0.34$ \\
    \bottomrule
  \end{tabular}
  \caption{
    Average initial train loss on \textsc{ParrotBenchCPN} for surrogates produced by each initialization method.
    We include a column for each instance of an initialization method (e.g., ``CPN (0)'' is only one of the \textsc{CompNet}s we trained) as well as a column that averages over each instance (e.g., ``CPN'' is an average over all \textsc{CompNet}s we trained).
  }\label{fig:parrot_training_time_initial_train_loss}
\end{figure*}

\begin{figure*}
  \centering
    \begin{tabular}{lllll}
    \toprule
    Program & CPN (0) & CPN (1) & CPN (2) & CPN \\
    \midrule
    \texttt{fft} & $0.48 \pm \num{0.00e+00}$ & $0.42 \pm \num{0.00e+00}$ & $2.92 \pm \num{0.00e+00}$ & $1.27 \pm 1.18$ \\
    \texttt{invk2j} & $1.14 \pm \num{0.00e+00}$ & $1.89 \pm \num{0.00e+00}$ & $2.51 \pm \num{0.00e+00}$ & $1.84 \pm 0.57$ \\
    \texttt{kmeans} & $0.06 \pm \num{0.00e+00}$ & $0.06 \pm \num{0.00e+00}$ & $0.05 \pm \num{0.00e+00}$ & $0.06 \pm 0.01$ \\
    \texttt{sobel} & $0.09 \pm \num{0.00e+00}$ & $0.13 \pm \num{0.00e+00}$ & $0.17 \pm \num{0.00e+00}$ & $0.13 \pm 0.03$ \\
    \bottomrule
    \end{tabular}
    
    \vspace{1em}
    \begin{tabular}{lllll}
    \toprule
    Program & MAML (0) & MAML (1) & MAML (2) & MAML \\
    \midrule
    \texttt{fft} & $0.57 \pm 0.18$ & $0.59 \pm 0.19$ & $0.53 \pm 0.14$ & $0.56 \pm 0.17$ \\
    \texttt{invk2j} & $2.07 \pm 0.63$ & $2.06 \pm 0.59$ & $2.01 \pm 0.56$ & $2.05 \pm 0.57$ \\
    \texttt{kmeans} & $0.71 \pm 0.43$ & $0.60 \pm 0.37$ & $0.74 \pm 0.50$ & $0.68 \pm 0.42$ \\
    \texttt{sobel} & $0.24 \pm 0.16$ & $0.18 \pm 0.12$ & $0.25 \pm 0.22$ & $0.22 \pm 0.17$ \\
    \bottomrule
    \end{tabular}
    
    \vspace{1em}
    \begin{tabular}{lllll}
    \toprule
    Program & PTS (0) & PTS (1) & PTS (2) & PTS \\
    \midrule
    \texttt{fft} & $0.83 \pm 0.07$ & $0.84 \pm 0.07$ & $0.85 \pm 0.07$ & $0.84 \pm 0.07$ \\
    \texttt{invk2j} & $2.09 \pm 0.57$ & $2.09 \pm 0.56$ & $2.11 \pm 0.58$ & $2.10 \pm 0.55$ \\
    \texttt{kmeans} & $0.06 \pm \num{0.00e+00}$ & $0.06 \pm \num{0.00e+00}$ & $0.06 \pm \num{0.00e+00}$ & $0.06 \pm \num{7.58e-04}$ \\
    \texttt{sobel} & $0.18 \pm \num{0.00e+00}$ & $0.19 \pm \num{0.00e+00}$ & $0.19 \pm \num{0.00e+00}$ & $0.19 \pm \num{3.52e-03}$ \\
    \bottomrule
  \end{tabular}
  \vspace{1em}

  \begin{tabular}{ll}
    \toprule
    Program & RND \\
    \midrule
    \texttt{fft} & $0.64 \pm 0.36$ \\
    \texttt{invk2j} & $2.36 \pm 0.80$ \\
    \texttt{kmeans} & $0.21 \pm 0.23$ \\
    \texttt{sobel} & $0.36 \pm 0.34$ \\
    \bottomrule
  \end{tabular}
    
  \caption{
    Average initial test loss on \textsc{ParrotBenchCPN} for surrogates produced by each initialization method.
    We include a column for each instance of an initialization method (e.g., ``CPN (0)'' is only one of the \textsc{CompNet}s we trained) as well as a column that averages over each instance (e.g., ``CPN'' is an average over all \textsc{CompNet}s we trained).
  }\label{fig:parrot_training_time_initial_test_loss}
\end{figure*}

\begin{figure*}
  \centering
  \begin{tabular}{ll}
    \toprule
    Program & RND \\
    \midrule
    \texttt{fft} & $\num{3.96e-06}$ \\
    \texttt{invk2j} & $\num{2.83e-03}$ \\
    \texttt{kmeans} & $0.01$ \\
    \texttt{sobel} & $\num{4.16e-04}$ \\
    \bottomrule
  \end{tabular}
  \caption{
    Target test loss for each \textsc{ParrotBenchCPN} program, set by training randomly initialized surrogates for $5{,}000$ epochs over $9$ trials and using the average final test loss.
  }\label{fig:parrot_training_time_target_test_loss}
\end{figure*}

\begin{figure*}
  \centering
  \footnotesize
  \begin{tabular}{lllll}
    \toprule
    Program & CPN-R Z/Z (Clone) (0) & CPN-R Z/Z (Clone) (1) & CPN-R Z/Z (Clone) (2) & CPN-R Z/Z (Clone) \\
    \midrule
    \texttt{fft} & $379.7 \pm 31.5$ & $262.0 \pm 15.9$ & $1140.7 \pm 40.3$ & $594.1 \pm 398.0$ \\
    \texttt{invk2j} & $15000.0 \pm 0.0$ & $9200.7 \pm 885.0$ & $12581.3 \pm 1956.4$ & $12260.7 \pm 2700.6$ \\
    \texttt{kmeans} & $12.0 \pm 1.5$ & $6.0 \pm 0.0$ & $6.0 \pm 0.0$ & $8.0 \pm 3.0$ \\
    \texttt{sobel} & $11316.7 \pm 2774.9$ & $10637.0 \pm 2332.6$ & $4232.3 \pm 2168.6$ & $8728.7 \pm 4008.5$ \\
    \bottomrule
  \end{tabular}
  \vspace{1em}

  \begin{tabular}{lllll}
    \toprule
    Program & MAML-Z Z/Z (Reinit) (0) & MAML-Z Z/Z (Reinit) (1) & MAML-Z Z/Z (Reinit) (2) & MAML-Z Z/Z (Reinit) \\
    \midrule
    \texttt{fft} & $649.3 \pm 469.7$ & $824.0 \pm 527.4$ & $1069.0 \pm 1047.5$ & $847.4 \pm 722.4$ \\
    \texttt{invk2j} & $9837.3 \pm 5124.5$ & $5327.7 \pm 4358.9$ & $13949.0 \pm 3153.0$ & $9704.7 \pm 5464.3$ \\
    \texttt{kmeans} & $198.7 \pm 192.1$ & $13338.7 \pm 4984.0$ & $5018.0 \pm 7486.5$ & $6185.1 \pm 7449.2$ \\
    \texttt{sobel} & $5695.7 \pm 3764.1$ & $6572.7 \pm 5221.4$ & $3668.3 \pm 2193.8$ & $5312.2 \pm 3970.6$ \\
    \bottomrule
  \end{tabular}
  \vspace{1em}

  \begin{tabular}{lllll}
    \toprule
    Program & PTS (0) & PTS (1) & PTS (2) & PTS \\
    \midrule
    \texttt{fft} & $459.3 \pm 241.2$ & $1384.7 \pm 1098.1$ & $1024.3 \pm 1087.0$ & $956.1 \pm 950.3$ \\
    \texttt{invk2j} & $13794.3 \pm 3345.0$ & $12470.3 \pm 5072.0$ & $6632.7 \pm 5282.7$ & $10965.8 \pm 5477.0$ \\
    \texttt{kmeans} & $188.0 \pm 36.3$ & $58.7 \pm 42.3$ & $18.3 \pm 1.0$ & $88.3 \pm 80.0$ \\
    \texttt{sobel} & $9964.0 \pm 4035.9$ & $5436.7 \pm 120.1$ & $7555.7 \pm 485.2$ & $7652.1 \pm 2939.6$ \\
    \bottomrule
  \end{tabular}
  \vspace{1em}

  \begin{tabular}{ll}
    \toprule
    Program & RND \\
    \midrule
    \texttt{fft} & $693.0 \pm 689.0$ \\
    \texttt{invk2j} & $5835.7 \pm 5133.0$ \\
    \texttt{kmeans} & $5098.7 \pm 7429.2$ \\
    \texttt{sobel} & $5881.0 \pm 3900.7$ \\
    \bottomrule
  \end{tabular}
  \caption{
    Average epoch at which each initialization method achieves the target testing loss for the training time evaluation on \textsc{ParrotBenchCPN}.
    We include a column for each instance of an initialization method (e.g., ``CPN (0)'' is only one of the \textsc{CompNet}s we trained), as well as a column that averages over all instances of an initialization method (e.g., ``CPN'' is an average over all \textsc{CompNet}s we trained).
  }\label{fig:avg_finish_epoch_parrot}
\end{figure*}

\begin{figure*}
  \centering
  \footnotesize
    \begin{tabular}{lllll}
    \toprule
    Program & CPN (0) & CPN (1) & CPN (2) & CPN \\
    \midrule
    \texttt{fft} & $0 / 9$ & $0 / 9$ & $0 / 9$ & $0 / 27$ \\
    \texttt{invk2j} & $9 / 9$ & $0 / 9$ & $3 / 9$ & $12 / 27$ \\
    \texttt{kmeans} & $0 / 9$ & $0 / 9$ & $0 / 9$ & $0 / 27$ \\
    \texttt{sobel} & $0 / 9$ & $0 / 9$ & $0 / 9$ & $0 / 27$ \\
    \bottomrule
    \end{tabular}
    
    \vspace{1em}
    \begin{tabular}{lllll}
    \toprule
    Program & MAML (0) & MAML (1) & MAML (2) & MAML \\
    \midrule
    \texttt{fft} & $0 / 9$ & $0 / 9$ & $0 / 9$ & $0 / 27$ \\
    \texttt{invk2j} & $4 / 9$ & $1 / 9$ & $8 / 9$ & $13 / 27$ \\
    \texttt{kmeans} & $0 / 9$ & $8 / 9$ & $3 / 9$ & $11 / 27$ \\
    \texttt{sobel} & $1 / 9$ & $2 / 9$ & $0 / 9$ & $3 / 27$ \\
    \bottomrule
    \end{tabular}
    
    \vspace{1em}
    \begin{tabular}{lllll}
    \toprule
    Program & PTS (0) & PTS (1) & PTS (2) & PTS \\
    \midrule
    \texttt{fft} & $0 / 9$ & $0 / 9$ & $0 / 9$ & $0 / 27$ \\
    \texttt{invk2j} & $7 / 9$ & $7 / 9$ & $1 / 9$ & $15 / 27$ \\
    \texttt{kmeans} & $0 / 9$ & $0 / 9$ & $0 / 9$ & $0 / 27$ \\
    \texttt{sobel} & $3 / 9$ & $0 / 9$ & $0 / 9$ & $3 / 27$ \\
    \bottomrule
  \end{tabular}
  \vspace{1em}

  \begin{tabular}{ll}
    \toprule
    Program & RND \\
    \midrule
    \texttt{fft} & $0 / 9$ \\
    \texttt{invk2j} & $0 / 9$ \\
    \texttt{kmeans} & $3 / 9$ \\
    \texttt{sobel} & $1 / 9$ \\
    \bottomrule
  \end{tabular}

  \caption{
    Number of trials where each initialization method does not achieve the target test loss after training for $15{,}000$ epochs during the training time evaluation on \textsc{ParrotBenchCPN}.
    We include a column for each instance of an initialization method (e.g., ``CPN (0)'' is only one of the \textsc{CompNet}s we trained) as well as a column that sums over each instance (e.g., ``CPN'' is a sum over all \textsc{CompNet}s we trained).
  }\label{fig:num_timeouts_parrot}
\end{figure*}

\section{\textsc{ParrotBenchCPN} True vs. Predicted Functions}
In this appendix, we present graphs showing the function each \textsc{ParrotBenchCPN} program implements, as well as the approximations of the function each initialization method produces in the data efficiency evaluation of Section~\ref{sec:data_efficiency}.
To visualize the behavior of multivariate functions, we generate a graph for each argument, where we vary that argument and fix all other arguments to zero.
We include graphs for the training set at each of the dataset sizes we evaluated on in Section~\ref{sec:data_efficiency} (i.e., $\set{0\%, 0.1\%, 1\%, 10\%, 100\%}$).
Each line is an average over compilers of that type (e.g., all the \textsc{CompNet}s we trained for the CPN compiler type) and all trials for those compilers.
The fill-between for each compiler type shows the minimum and maximum predictions across each instance of that compiler type and each surrogate initialized by that instance.

\begin{figure*}
\centering
\includegraphics[width=0.49\textwidth]{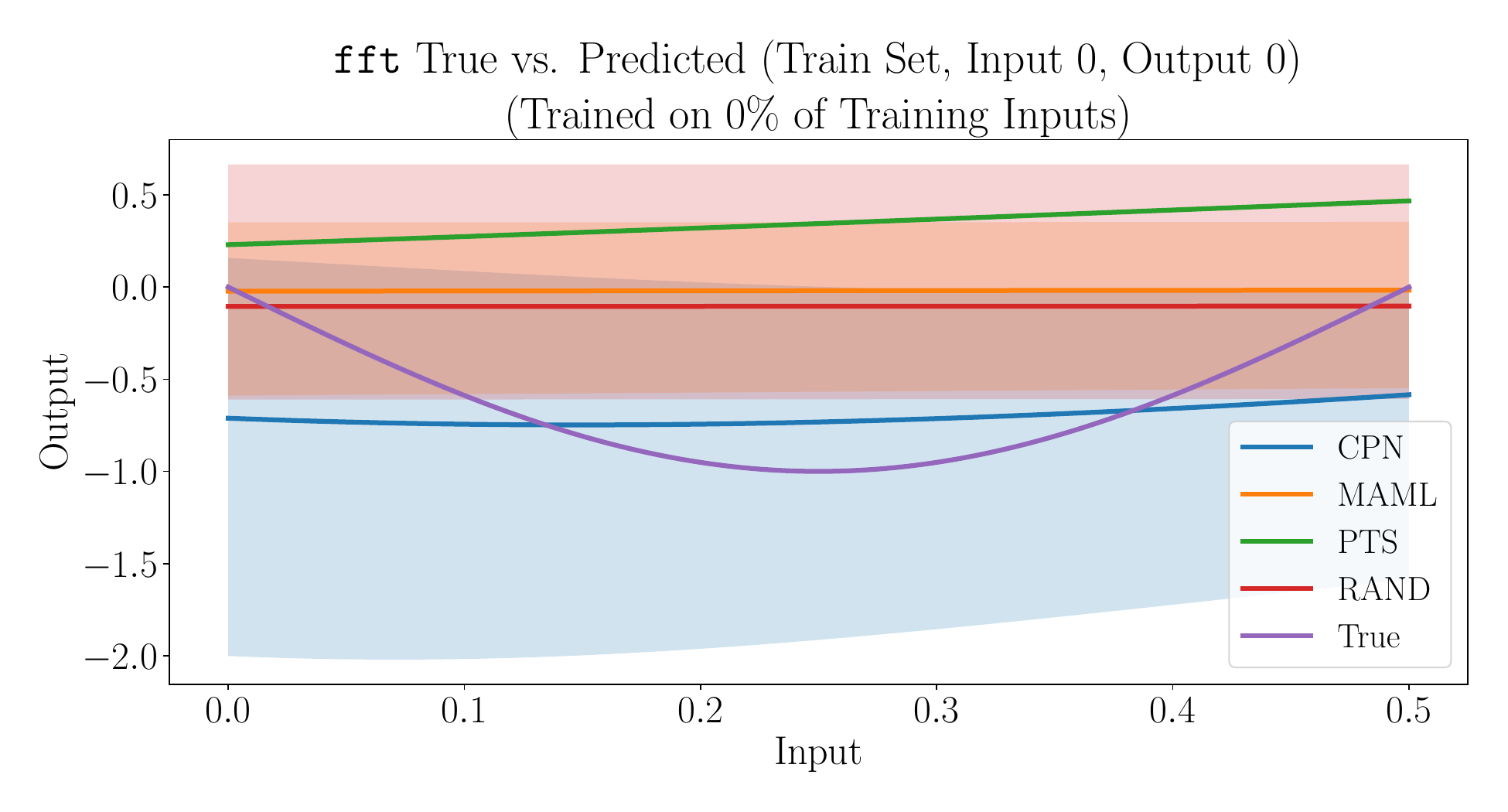}
\includegraphics[width=0.49\textwidth]{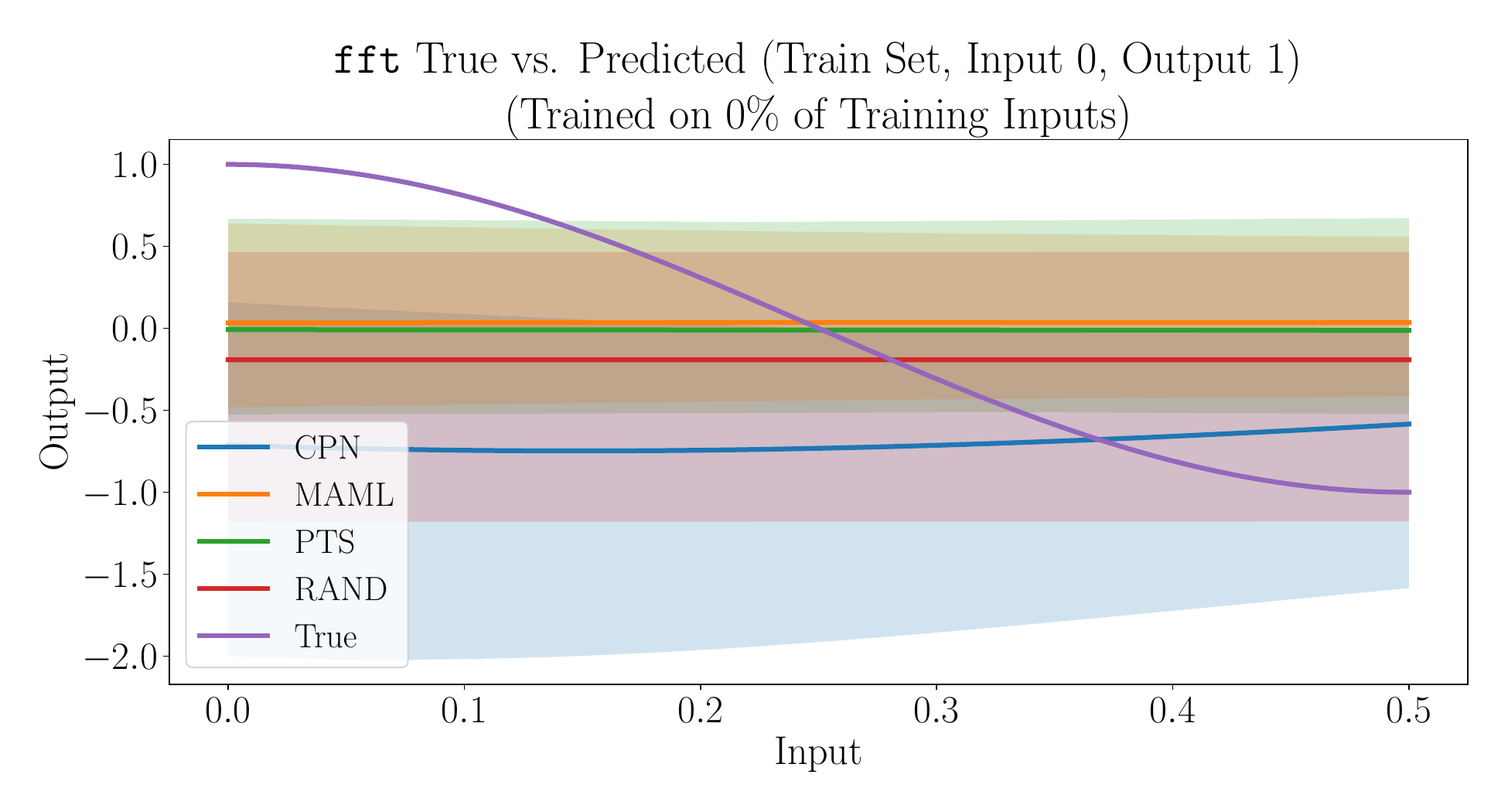}
\\
\vspace{-1.5em}
\includegraphics[width=0.49\textwidth]{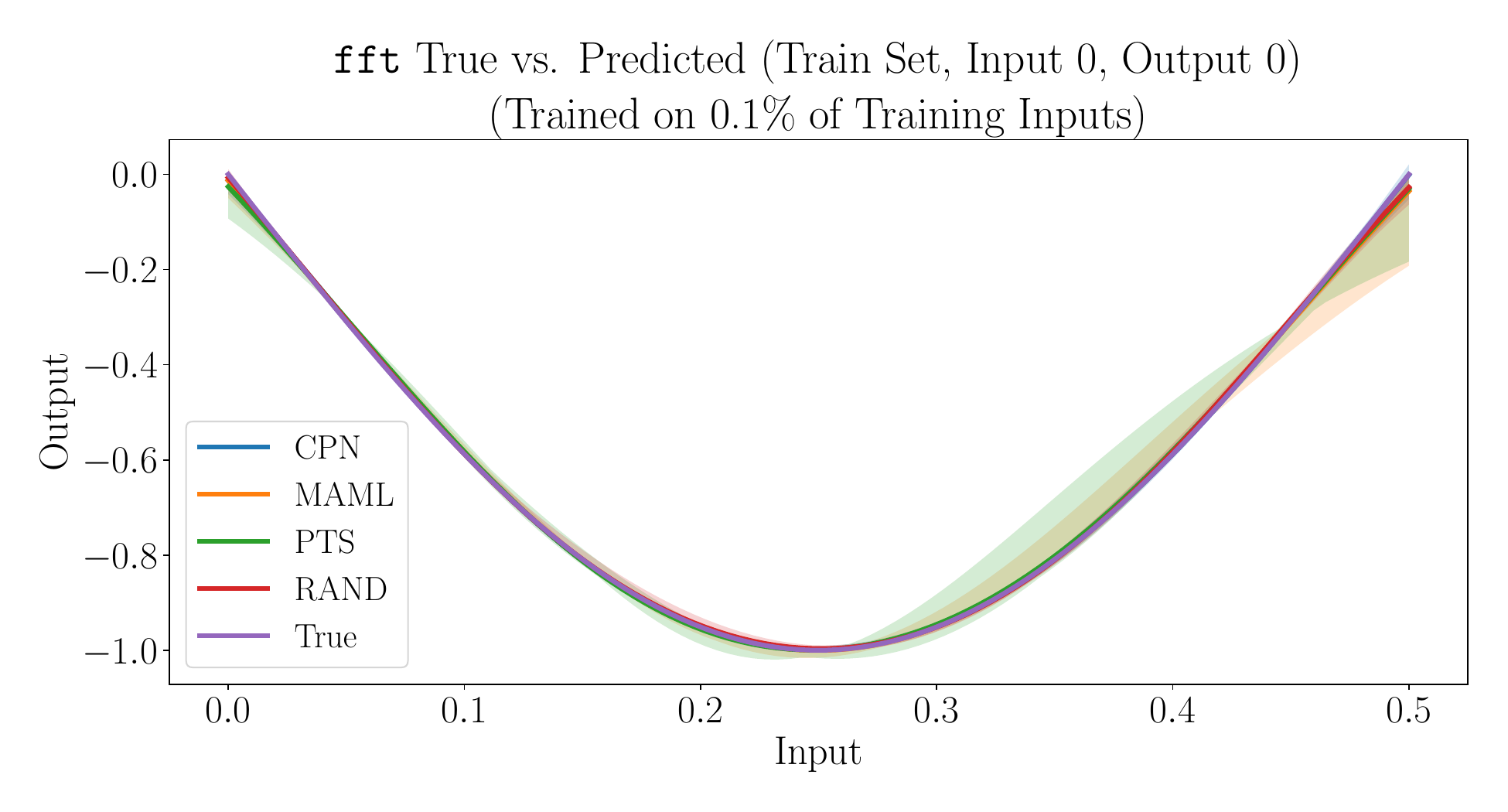}
\includegraphics[width=0.49\textwidth]{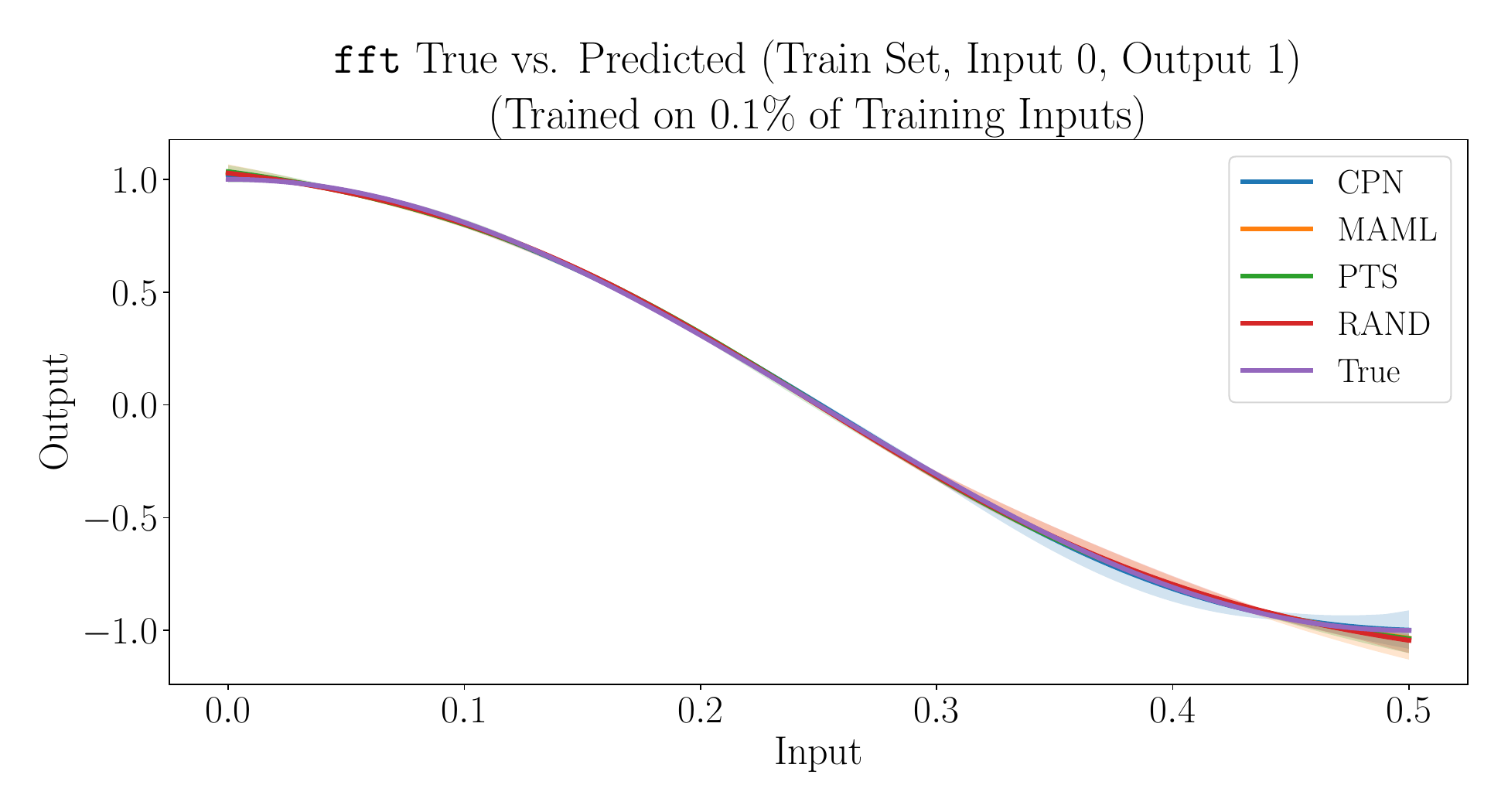}
\\
\vspace{-1.5em}
\includegraphics[width=0.49\textwidth]{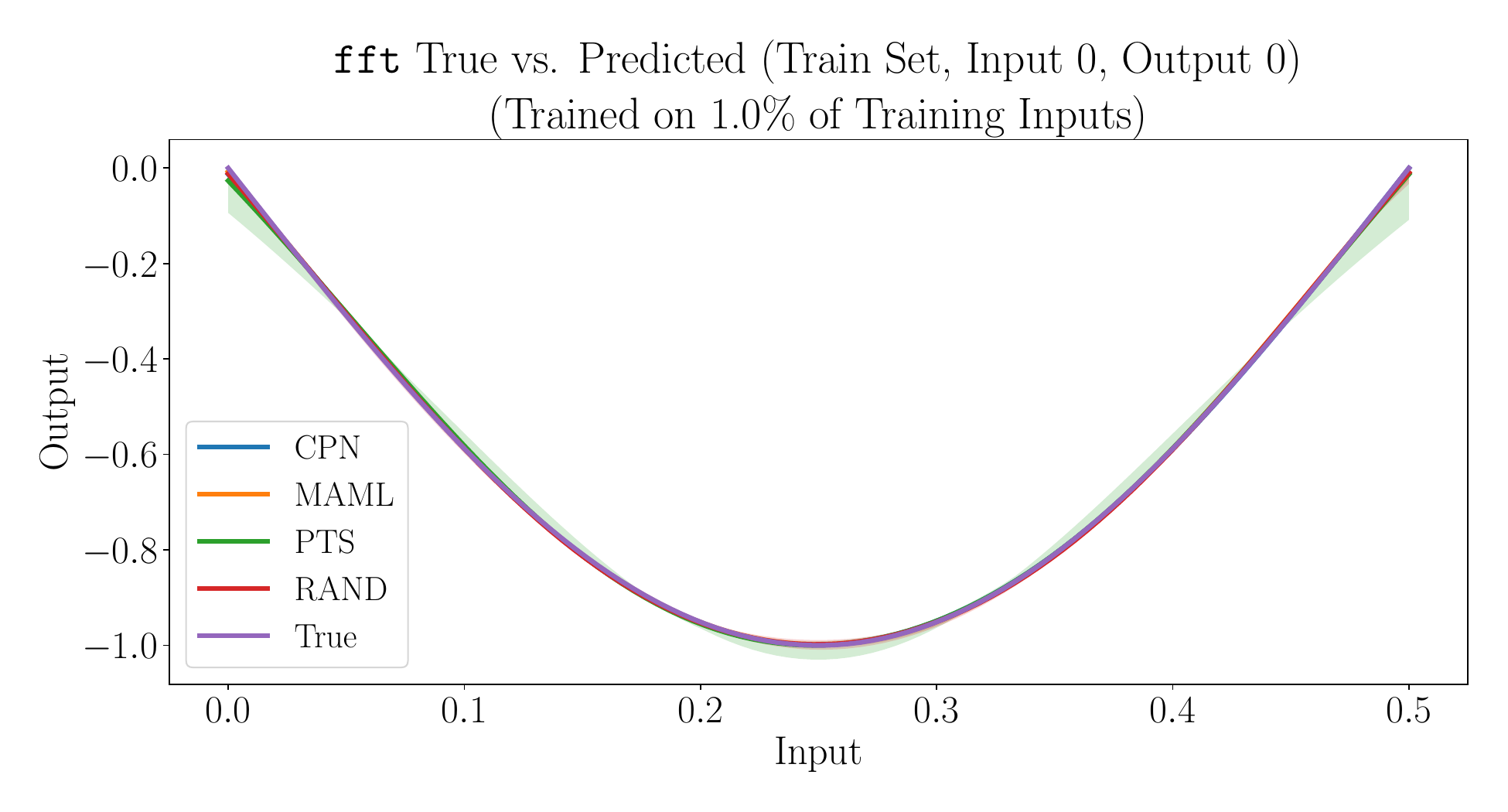}
\includegraphics[width=0.49\textwidth]{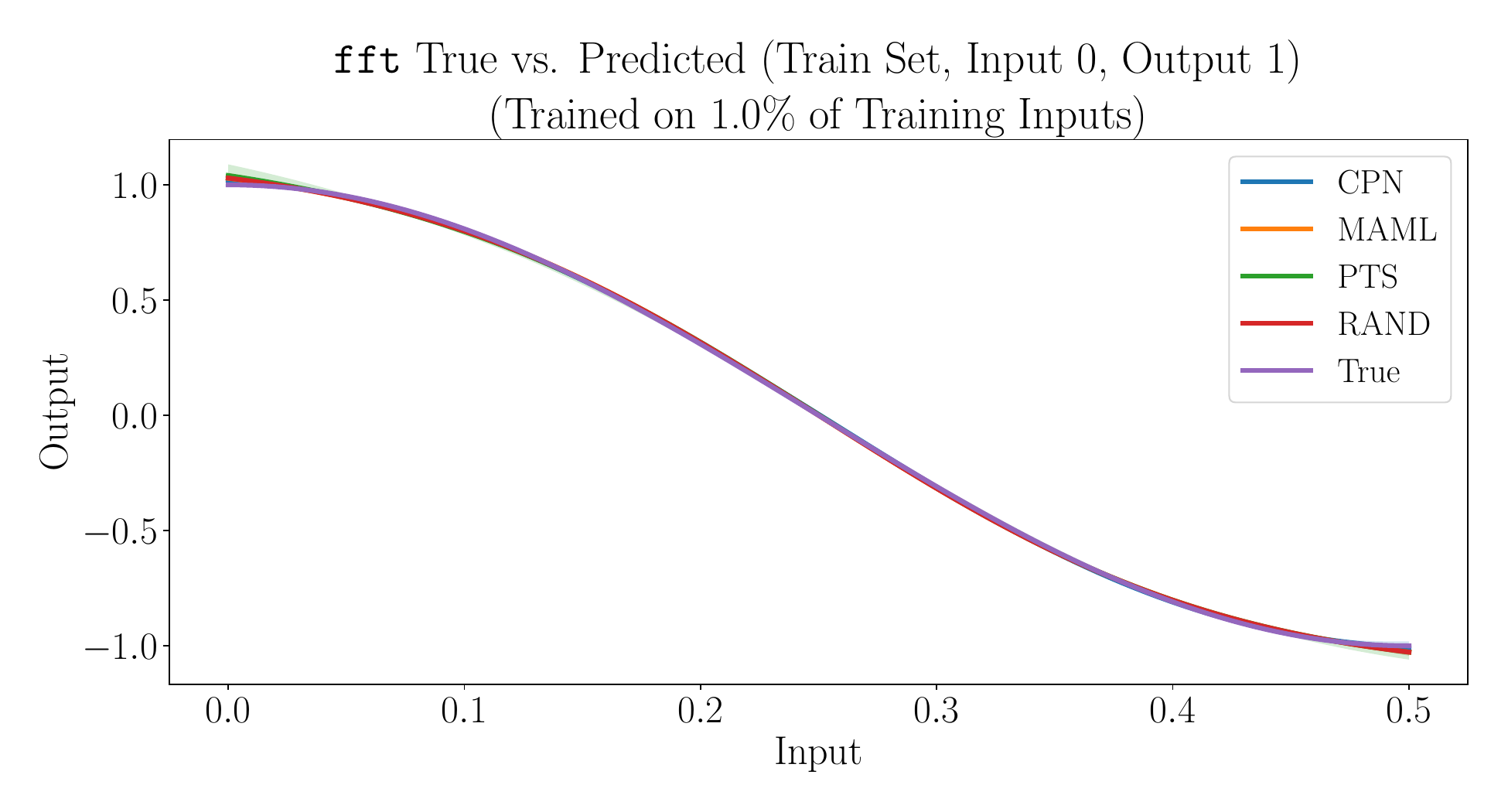}
\\
\vspace{-1.5em}
\includegraphics[width=0.49\textwidth]{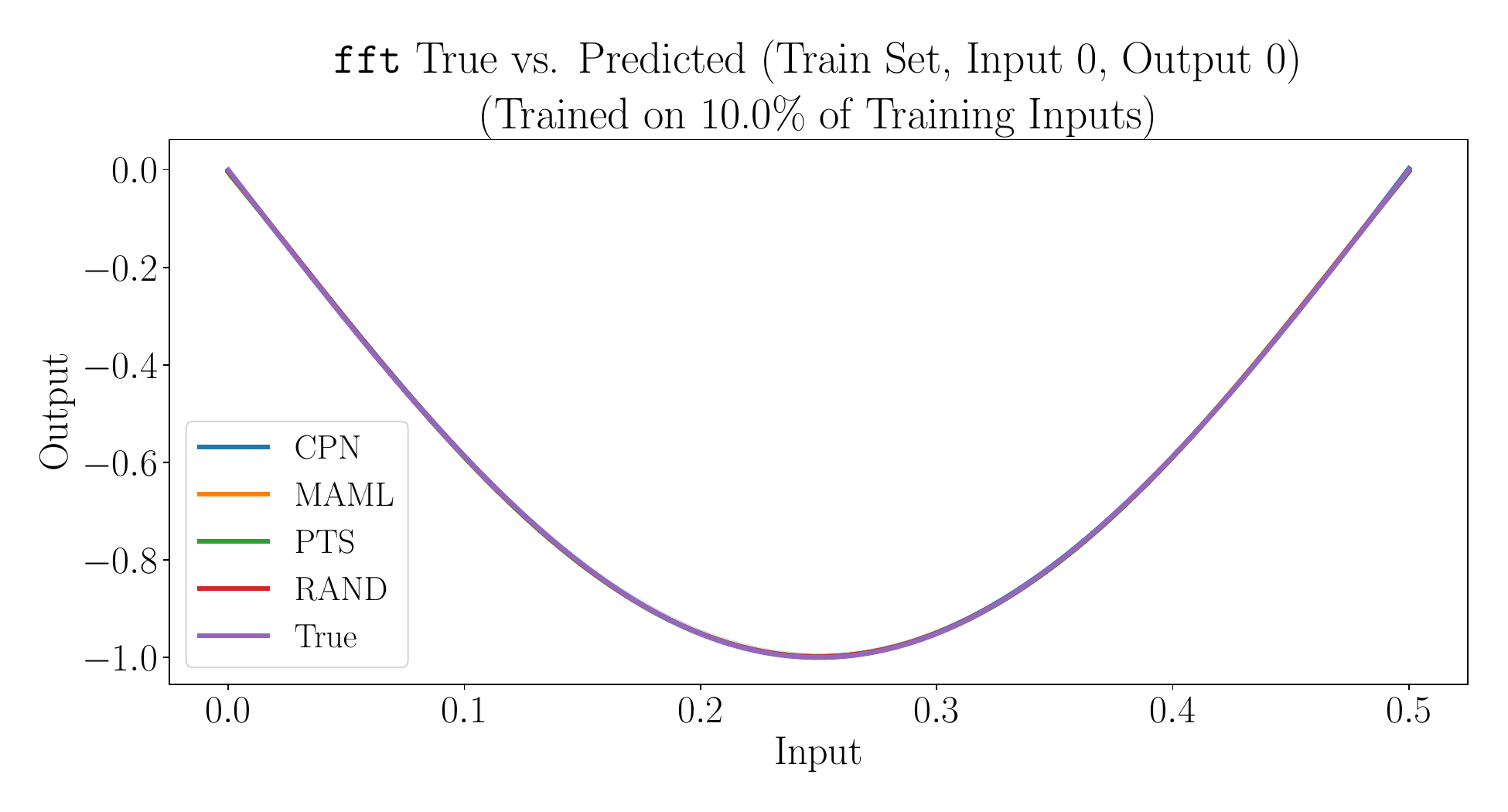}
\includegraphics[width=0.49\textwidth]{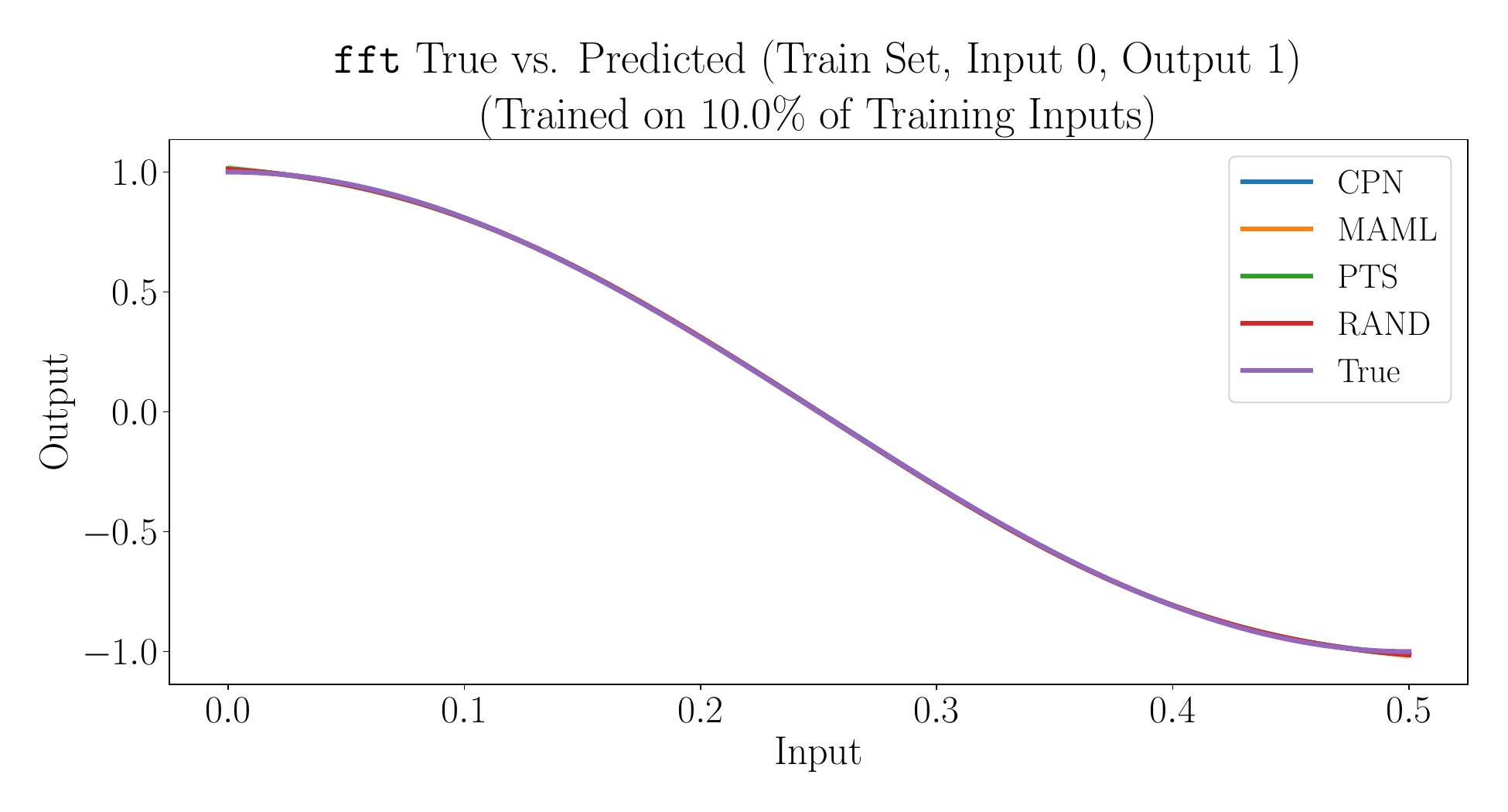}
\\
\vspace{-1.5em}
\includegraphics[width=0.49\textwidth]{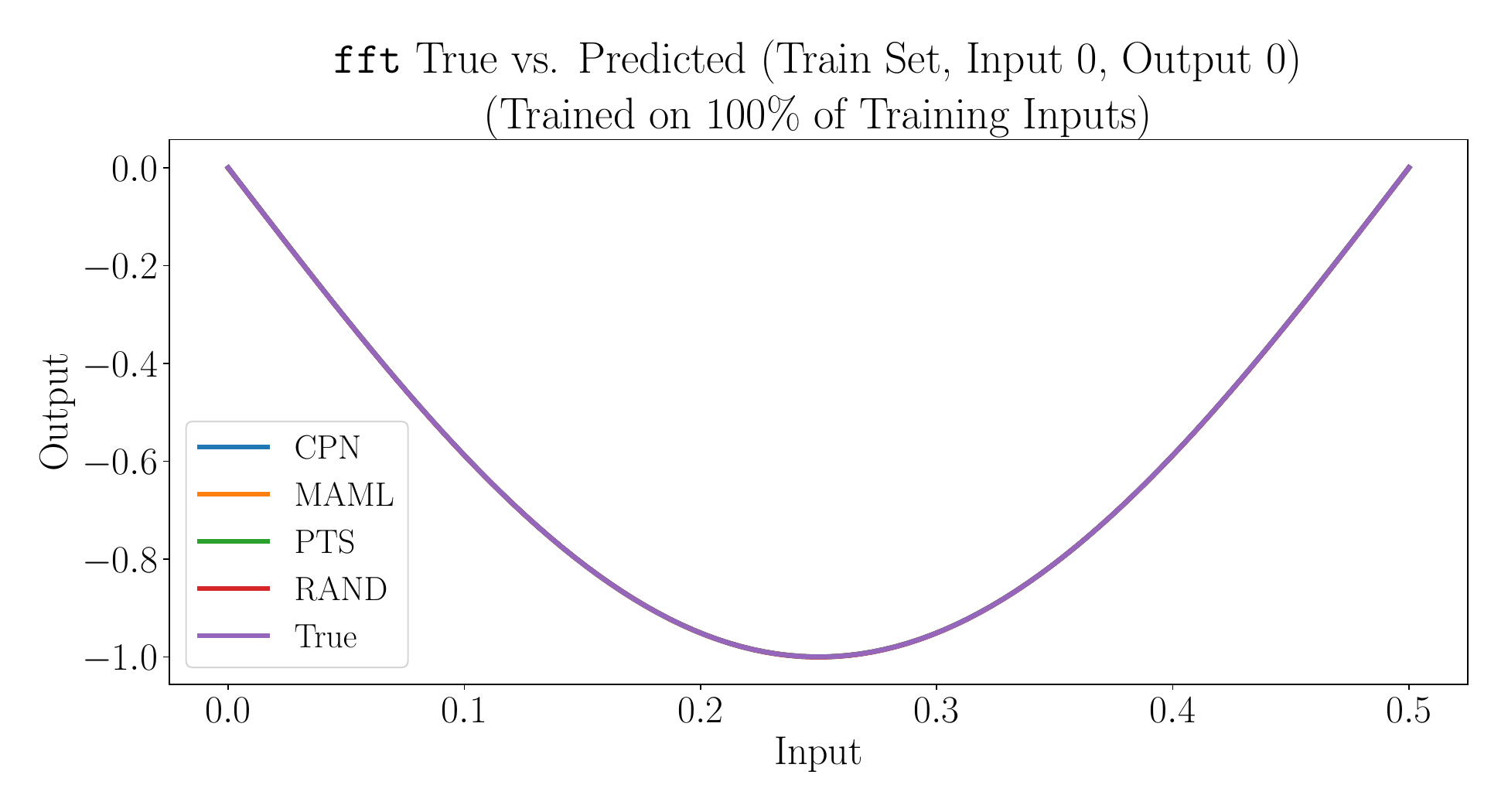}
\includegraphics[width=0.49\textwidth]{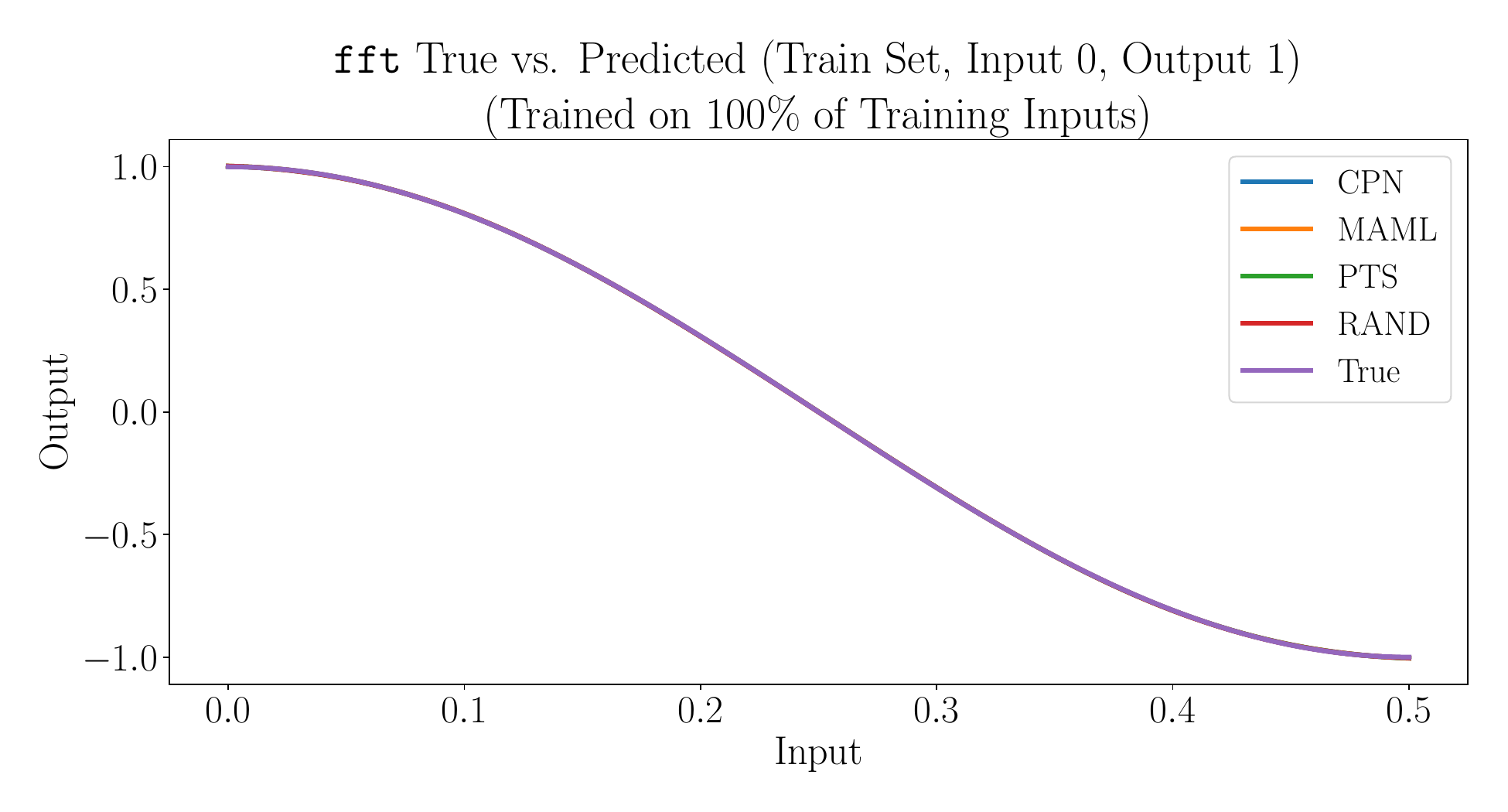}
  \caption{
    Visual comparisons of the ground-truth \texttt{fft} function from \textsc{ParrotBenchCPN} and neural surrogate approximations thereof.
    We include results for all dataset sizes evaluated in Section~\ref{sec:data_efficiency}, and we include plots for each output of the kernel when the input is varied.
  }\label{fig:true_vs_pred_fft}
\end{figure*}

\begin{figure*}
\centering
\includegraphics[width=0.49\textwidth]{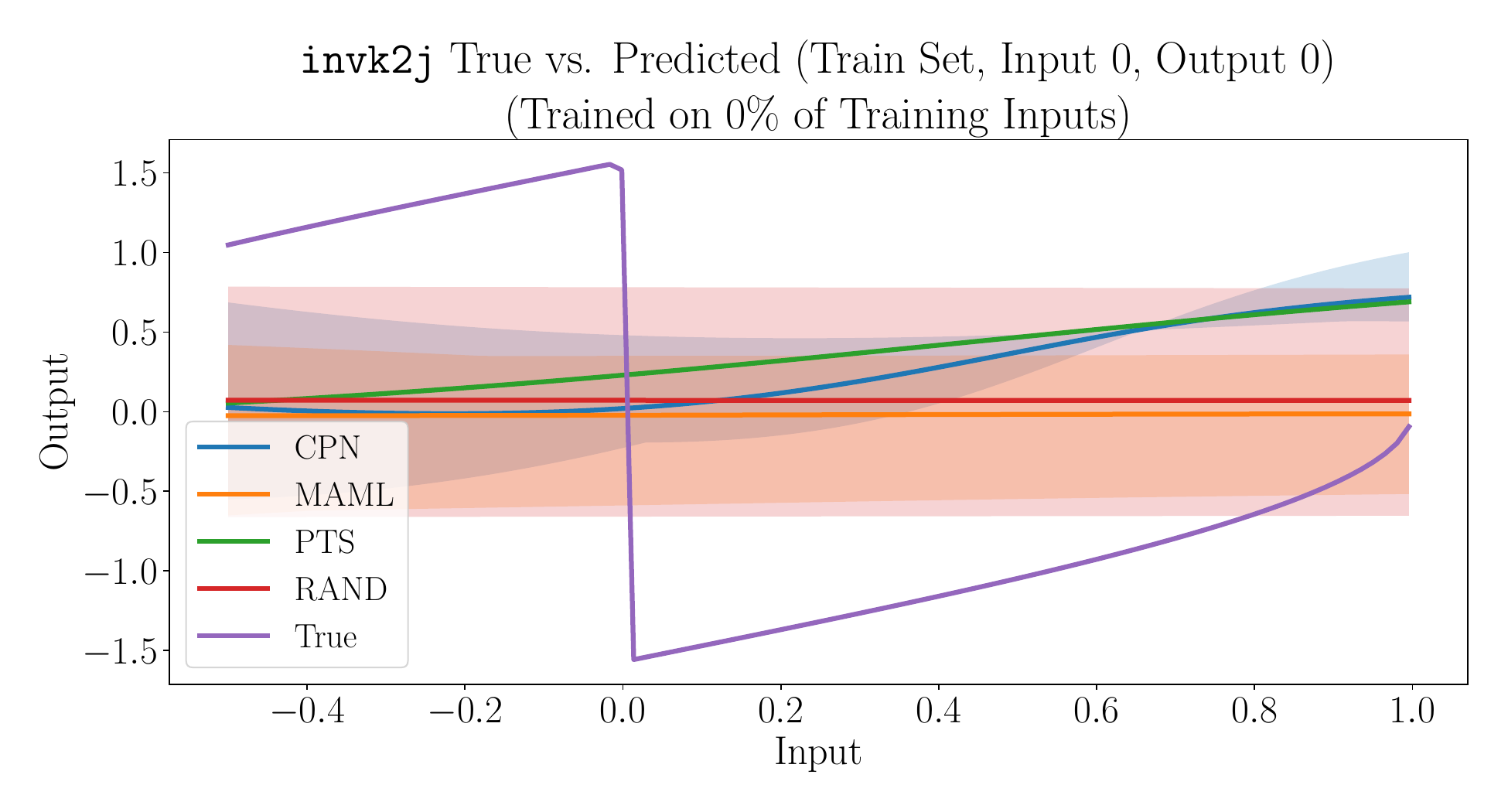}
\includegraphics[width=0.49\textwidth]{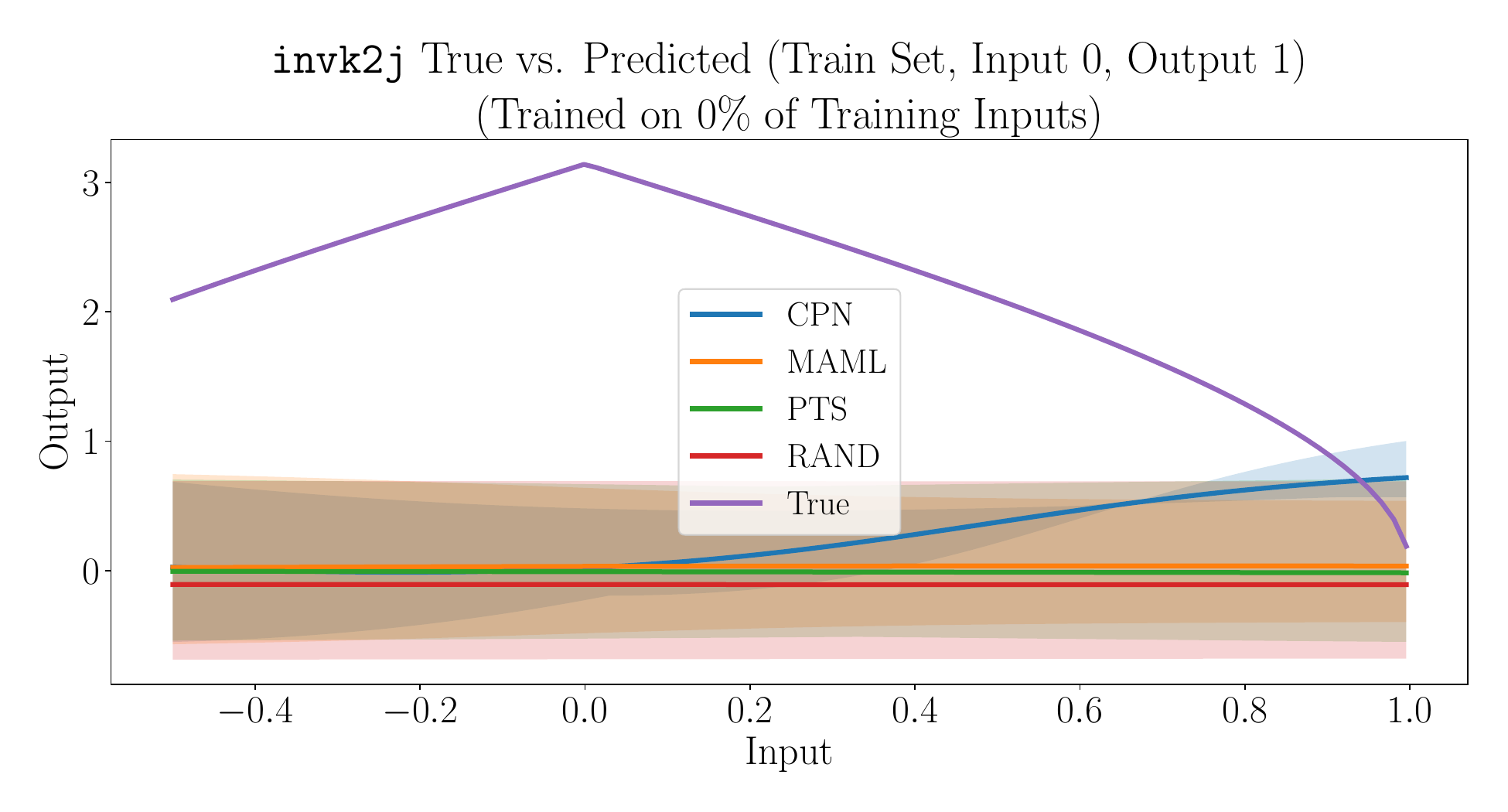}
\\
\vspace{-0.5em}
\includegraphics[width=0.49\textwidth]{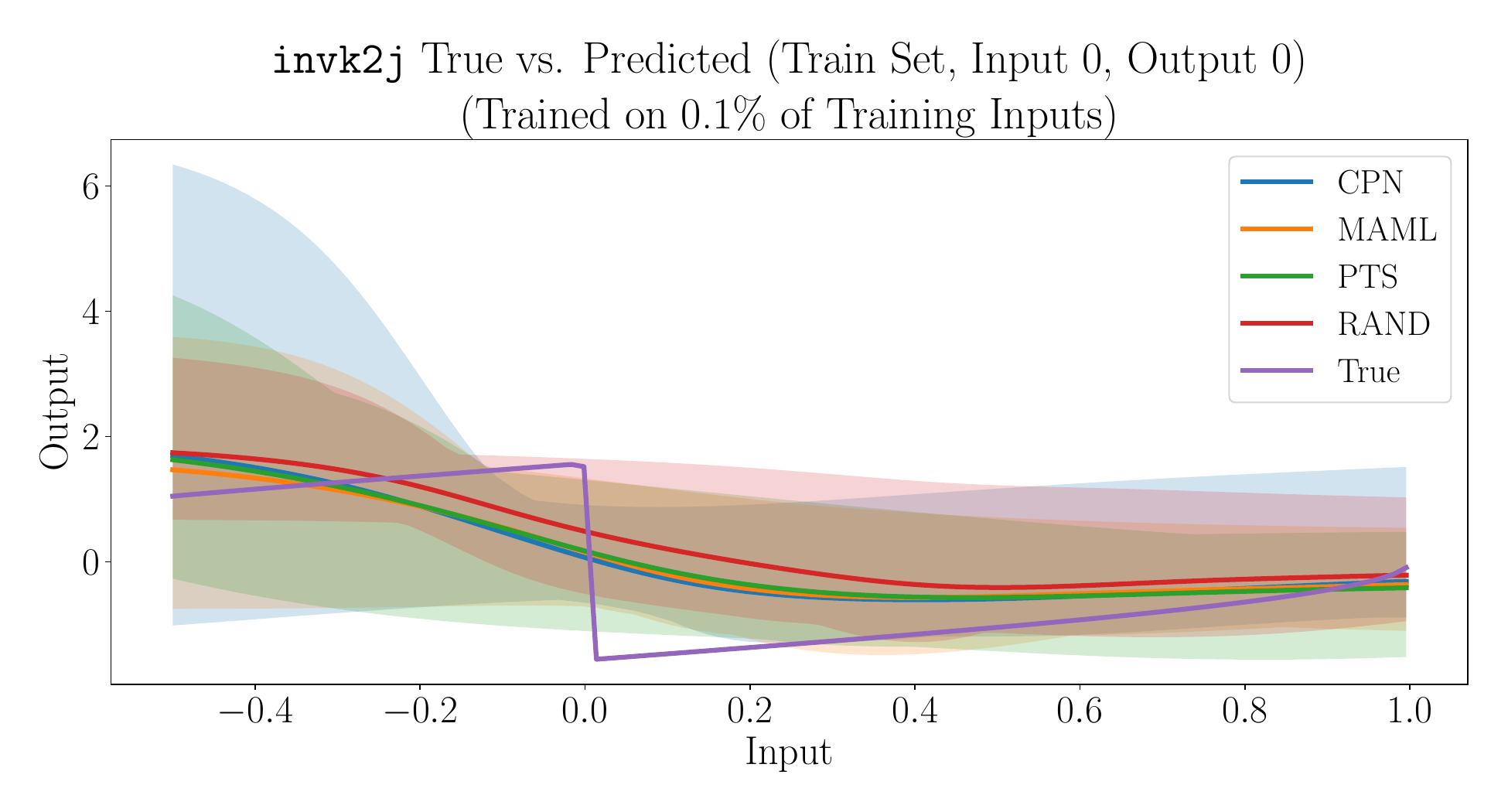}
\includegraphics[width=0.49\textwidth]{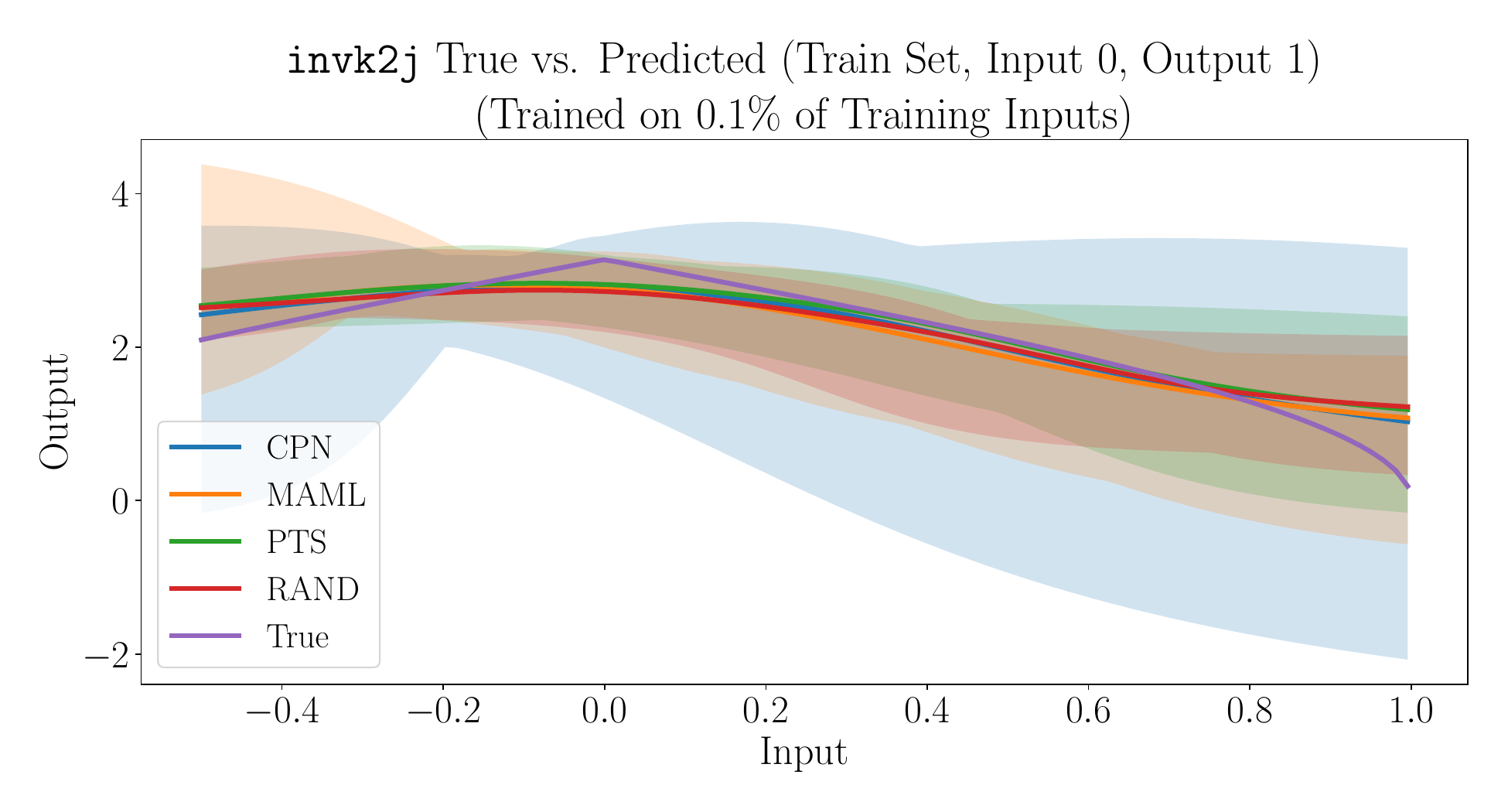}
\\
\vspace{-0.5em}
\includegraphics[width=0.49\textwidth]{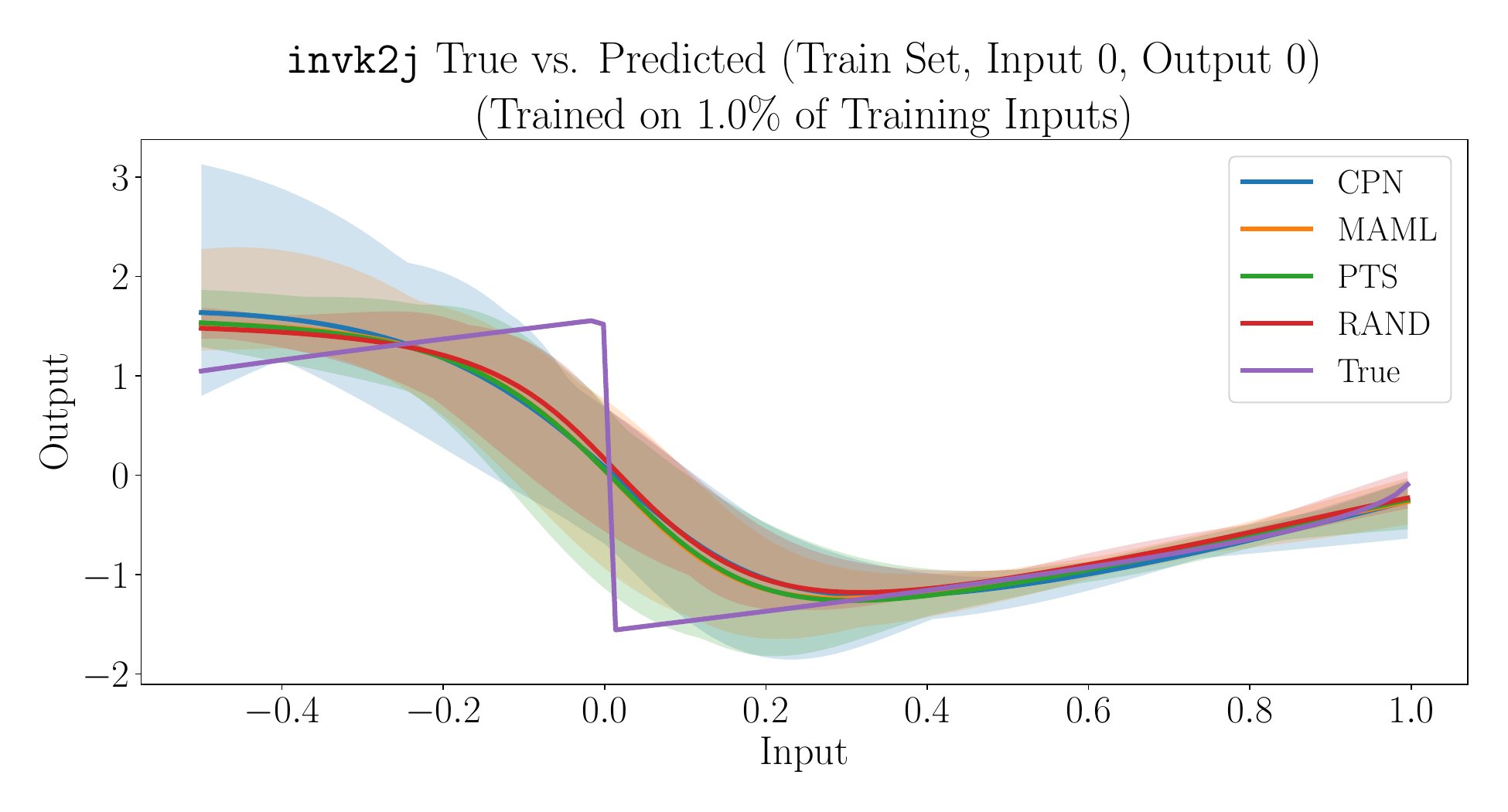}
\includegraphics[width=0.49\textwidth]{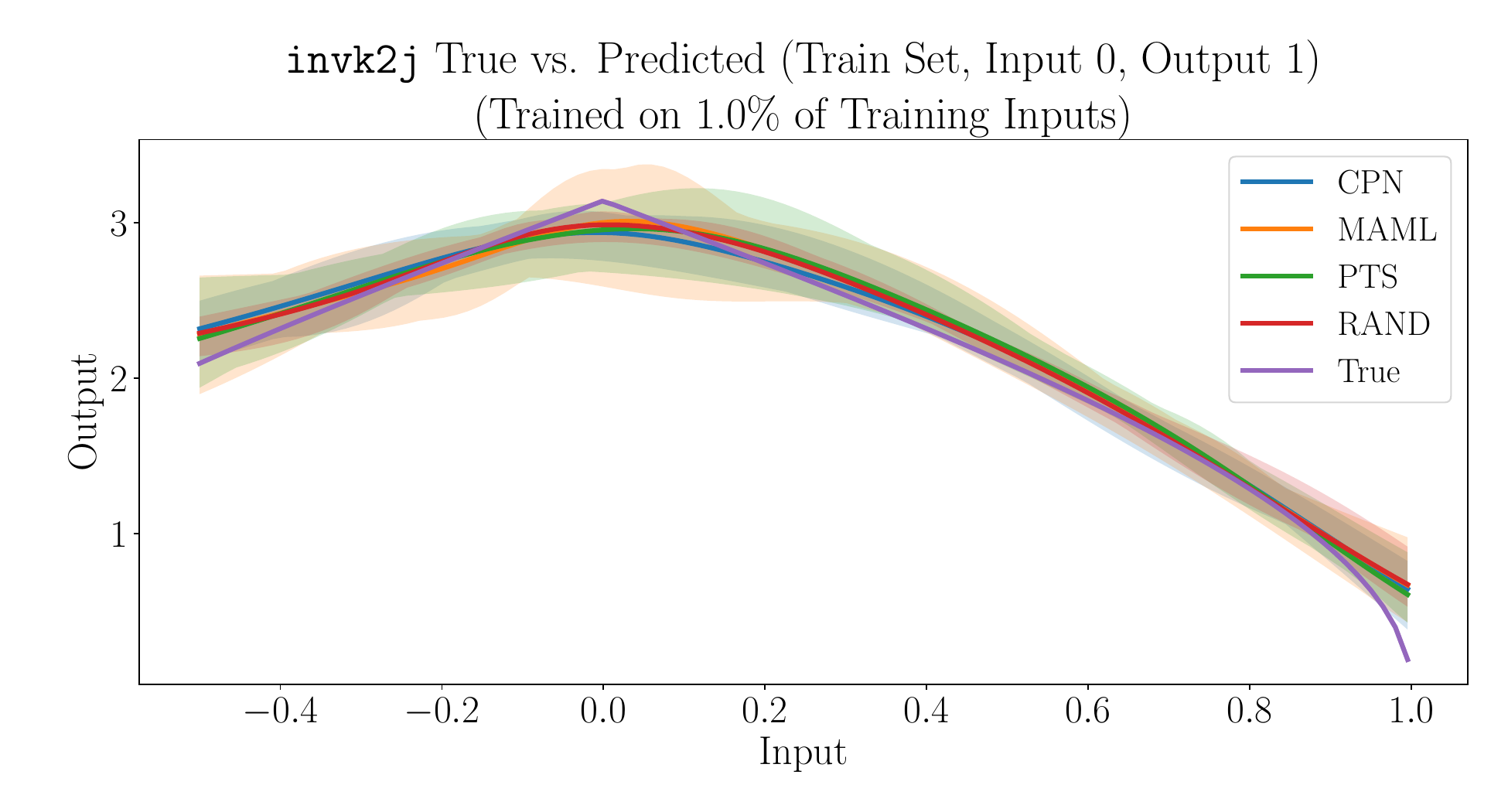}
\\
\vspace{-0.5em}
\includegraphics[width=0.49\textwidth]{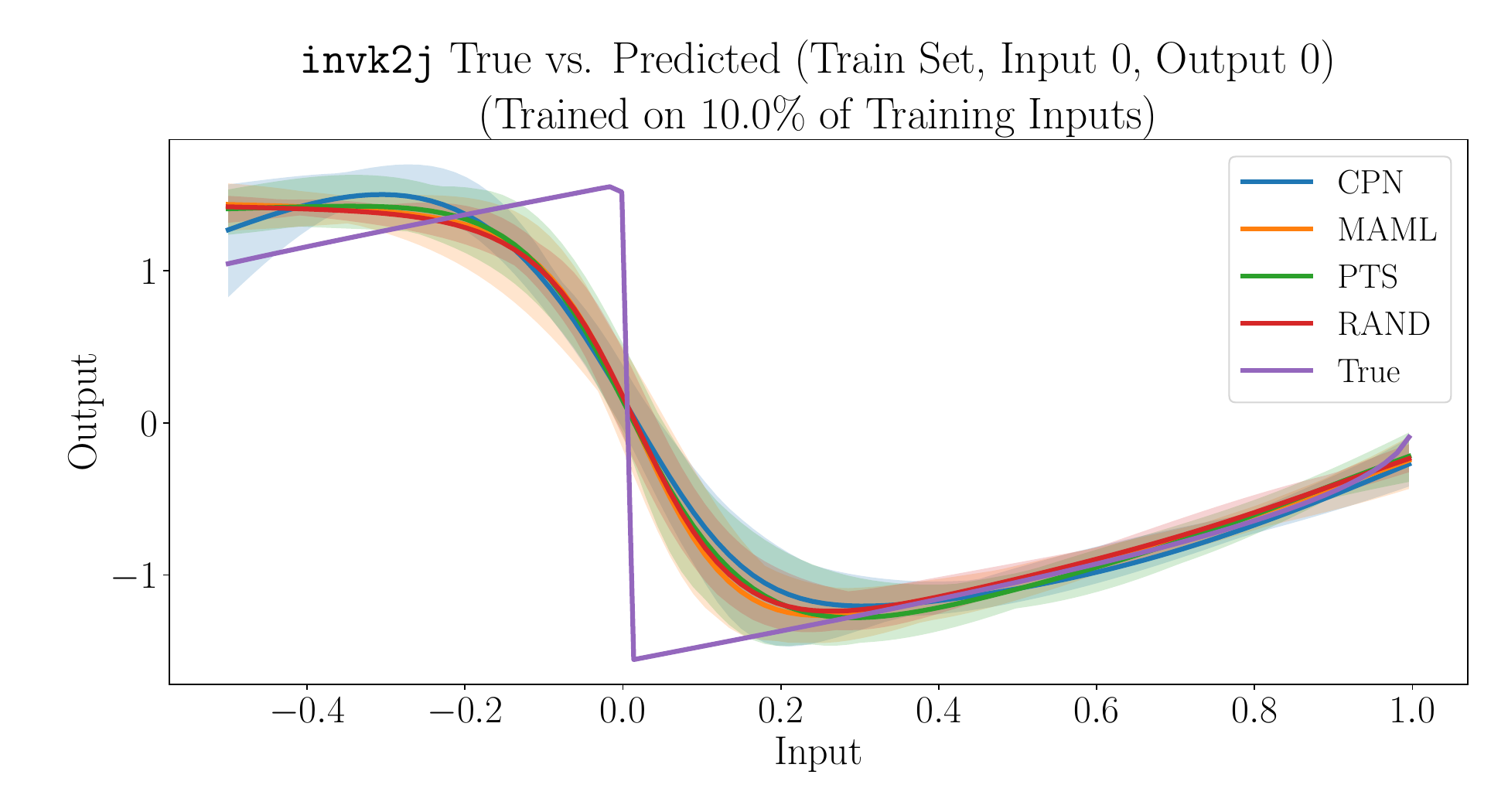}
\includegraphics[width=0.49\textwidth]{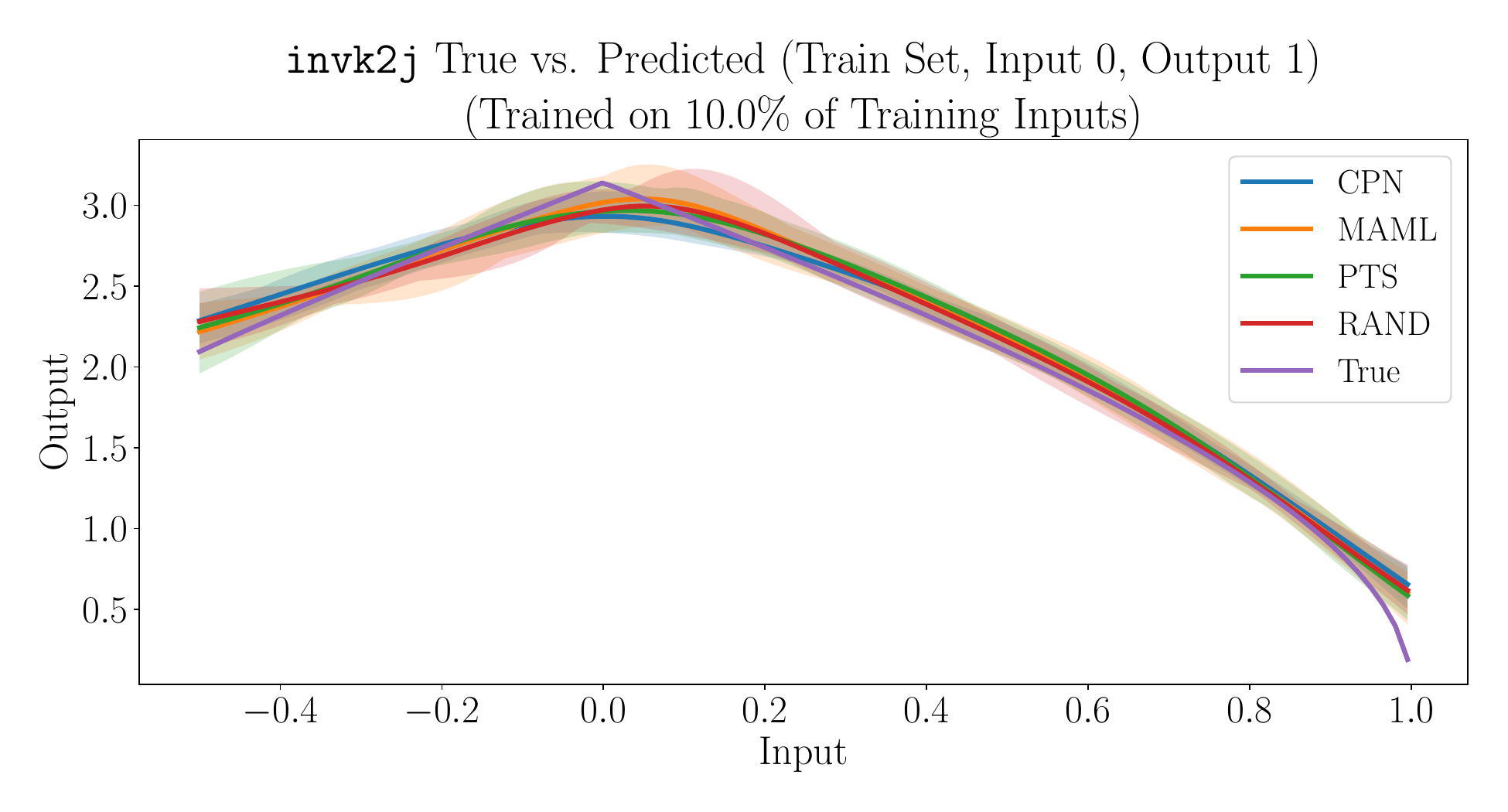}
\\
\vspace{-0.5em}
\includegraphics[width=0.49\textwidth]{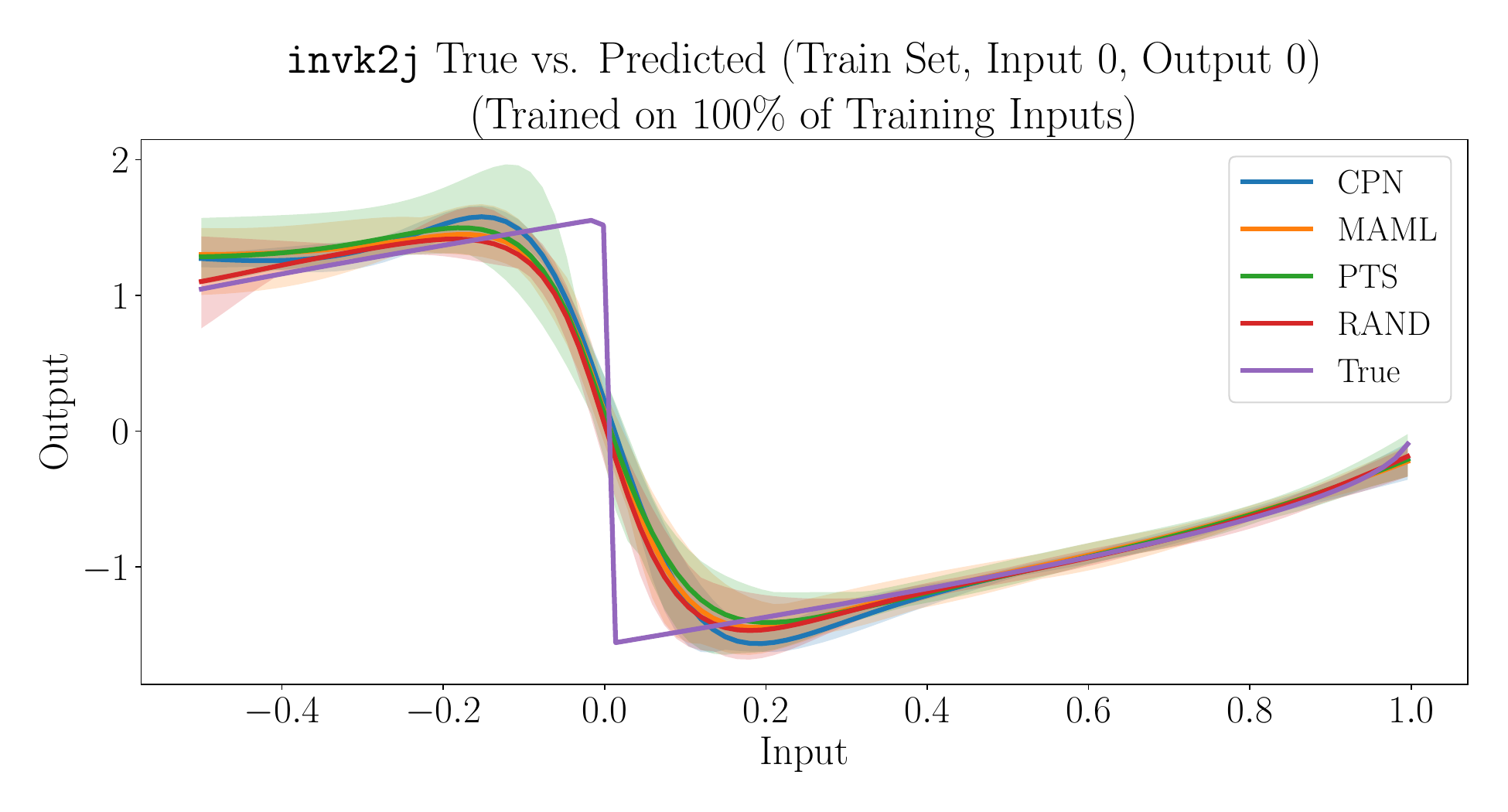}
\includegraphics[width=0.49\textwidth]{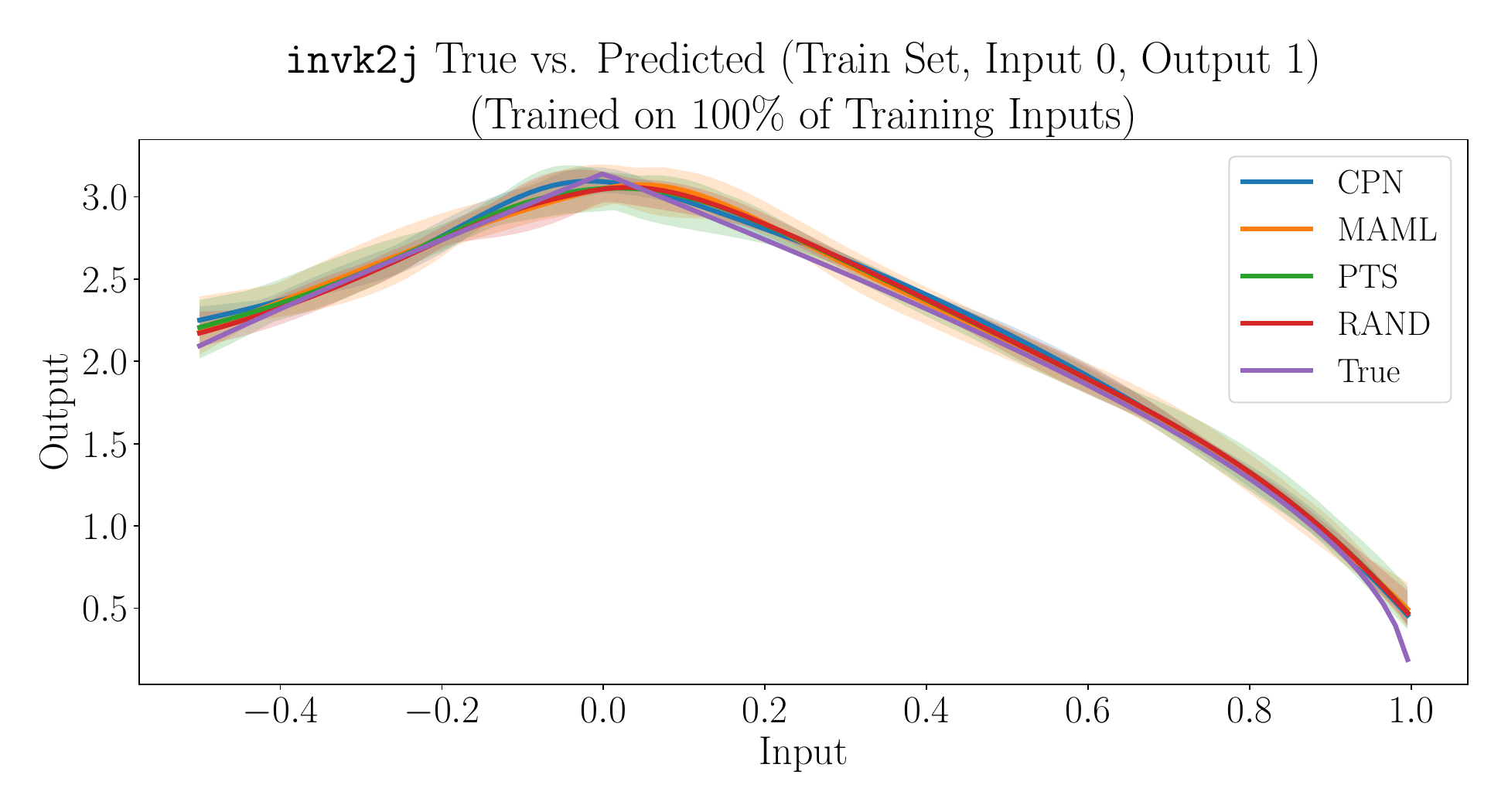}
\\
\vspace*{-1.2em}
  \caption{
    Visual comparisons of the ground-truth \texttt{invk2j} function from \textsc{ParrotBenchCPN} and neural surrogate approximations thereof, when the first input is varied.
    We include results for all dataset sizes evaluated in Section~\ref{sec:data_efficiency}, and we plot each output of the kernel when the input is varied.
  }\label{fig:true_vs_pred_invk2j_input_0}
\end{figure*}

\begin{figure*}
\centering
\includegraphics[width=0.49\textwidth]{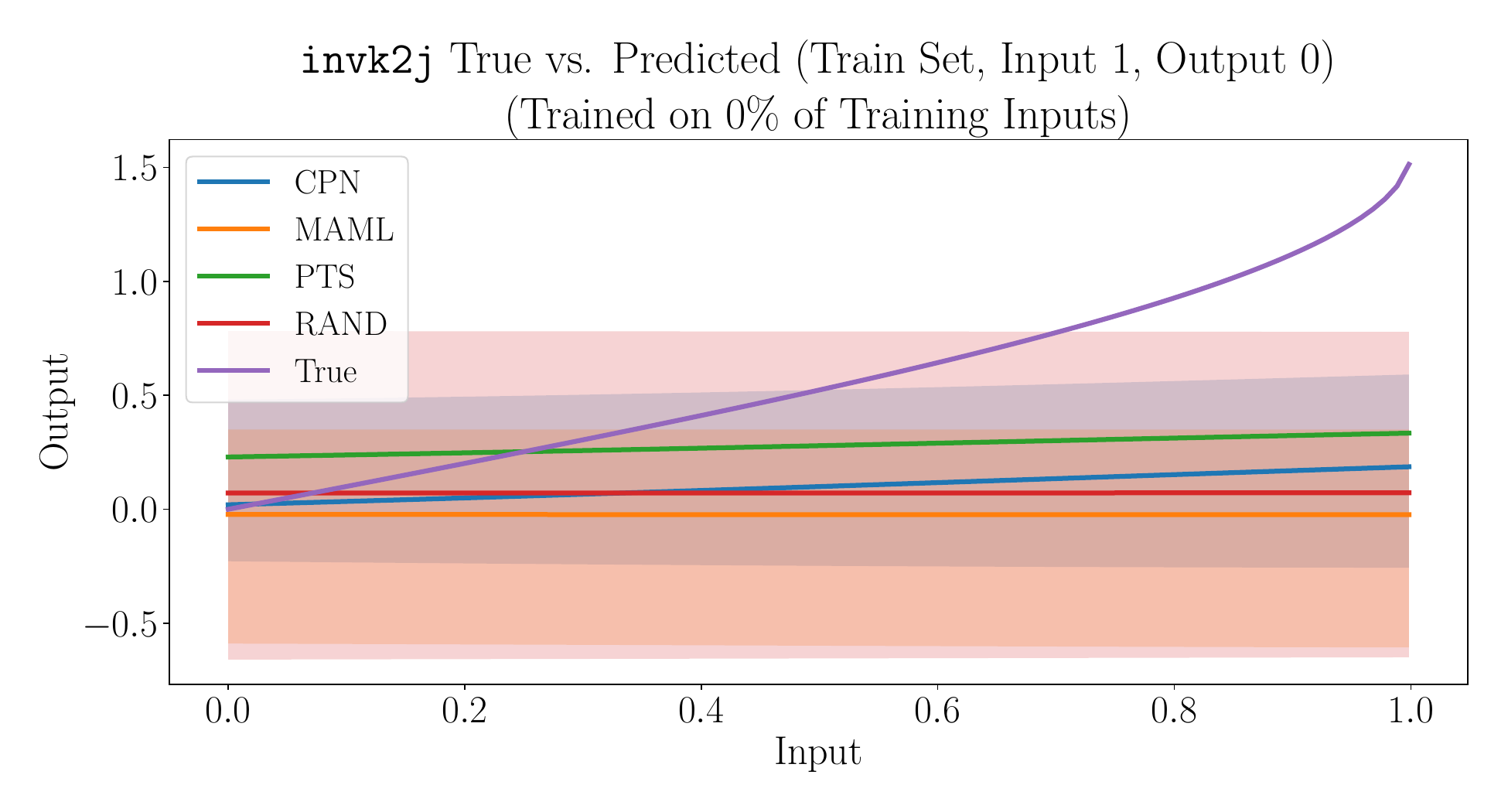}
\includegraphics[width=0.49\textwidth]{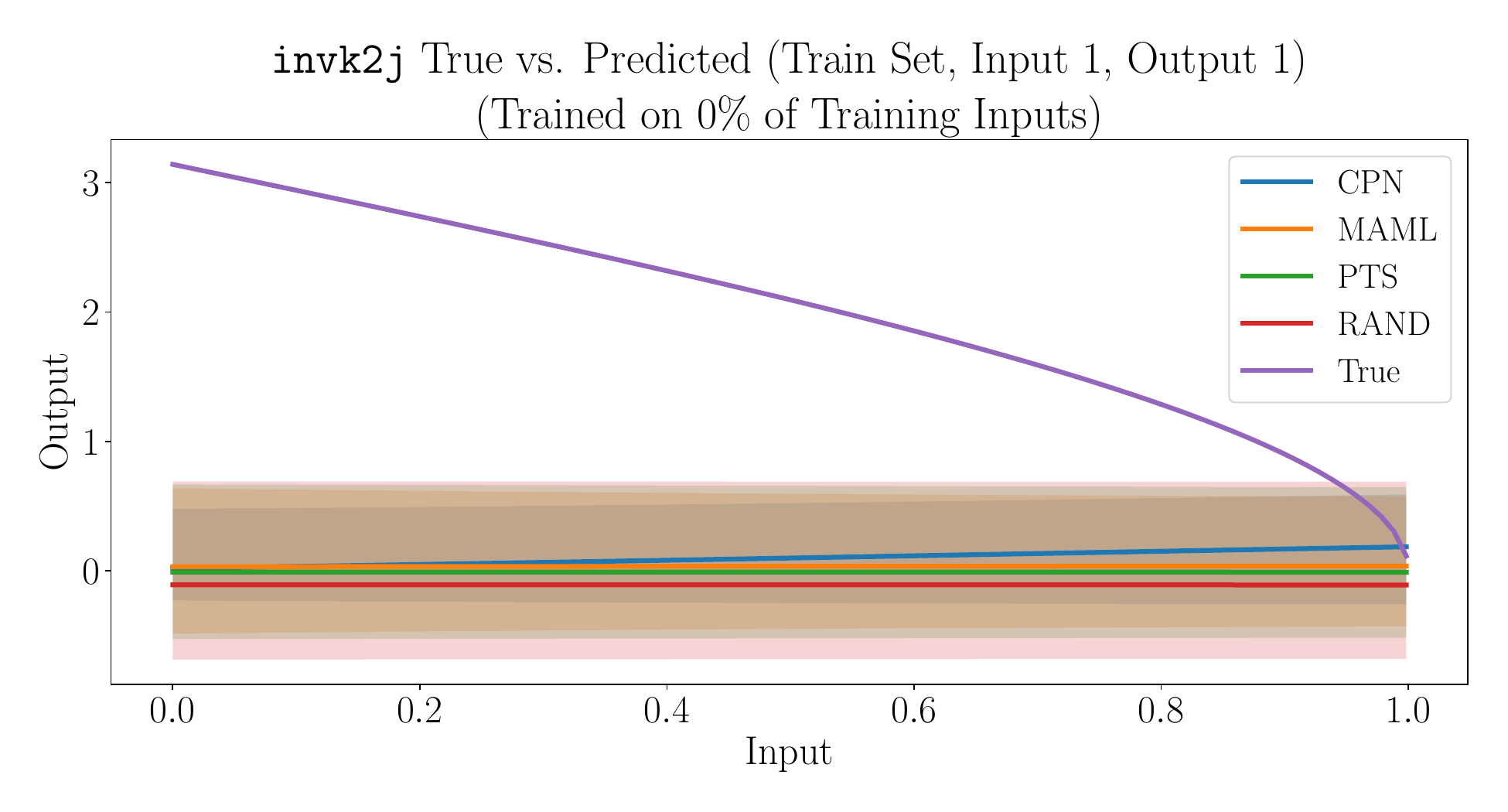}
\\
\vspace{-0.5em}
\includegraphics[width=0.49\textwidth]{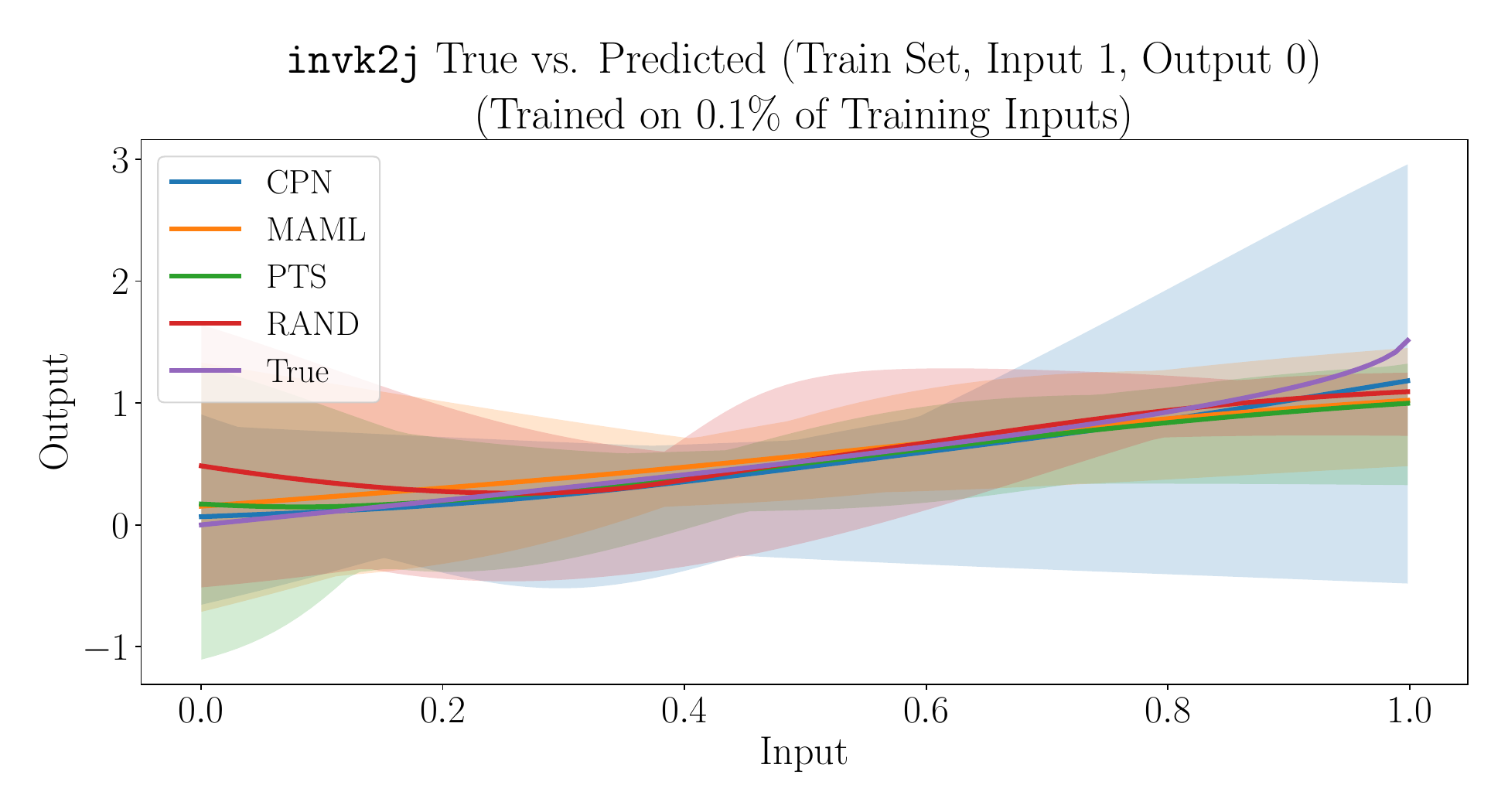}
\includegraphics[width=0.49\textwidth]{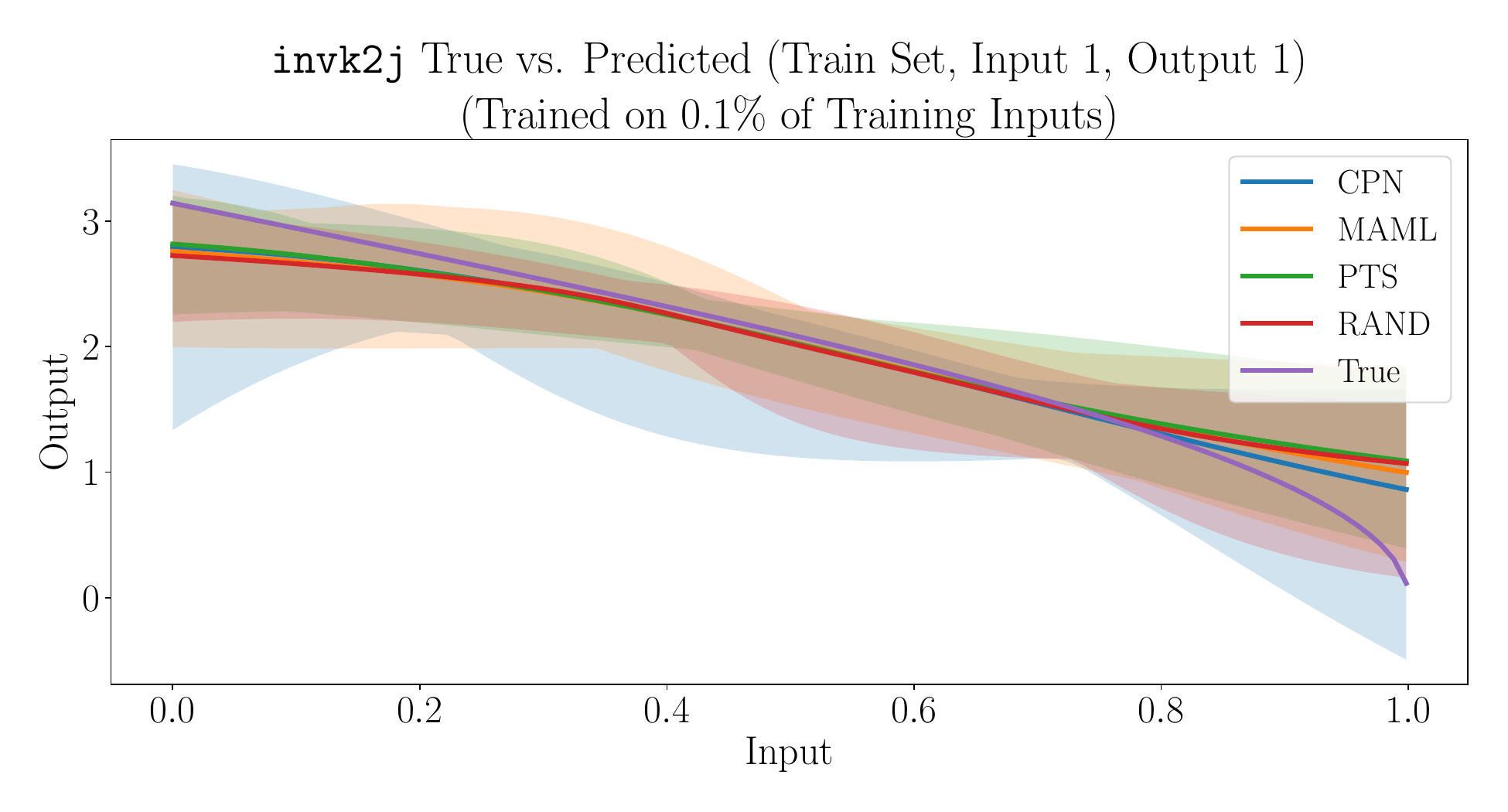}
\\
\vspace{-0.5em}
\includegraphics[width=0.49\textwidth]{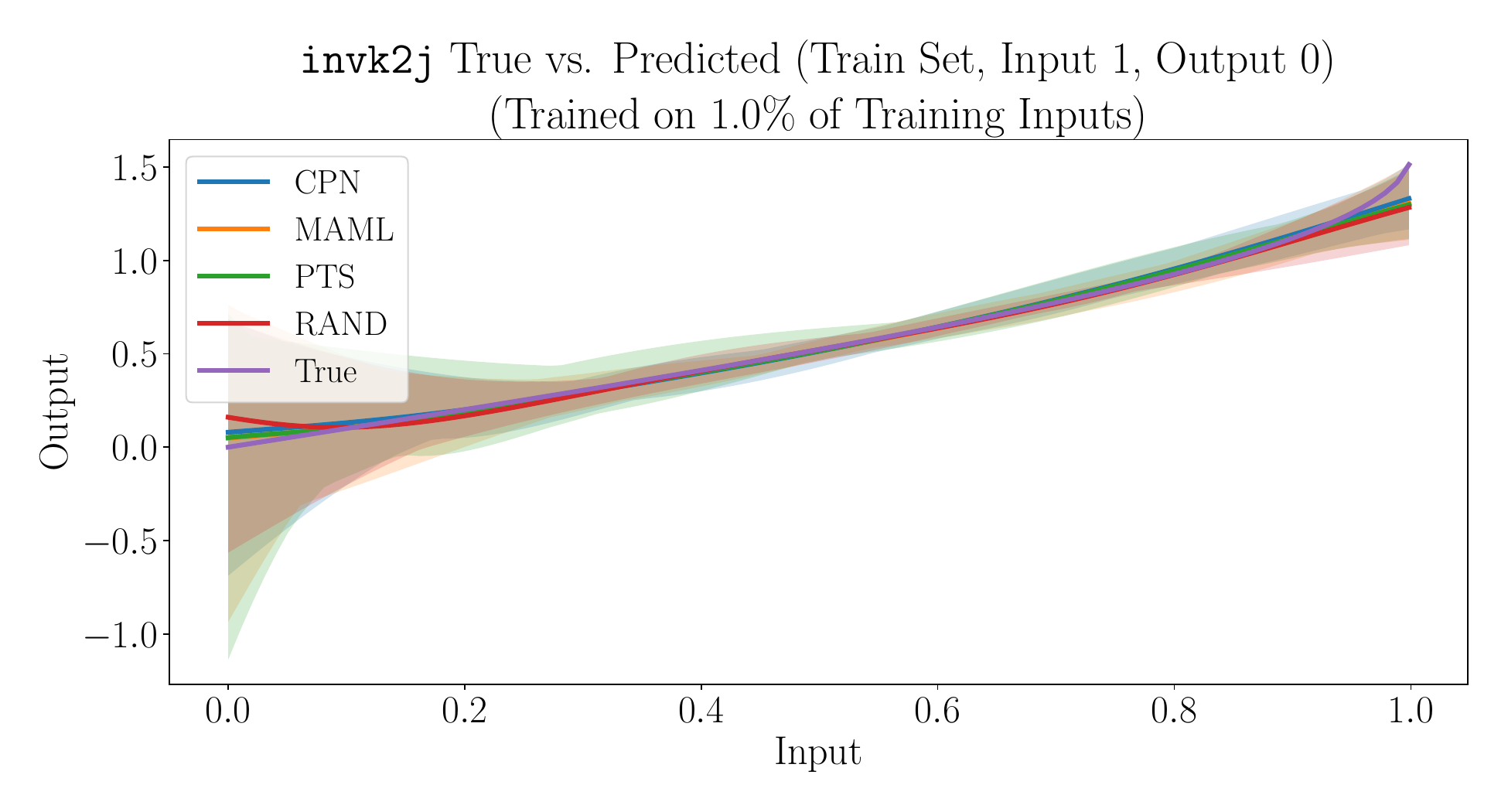}
\includegraphics[width=0.49\textwidth]{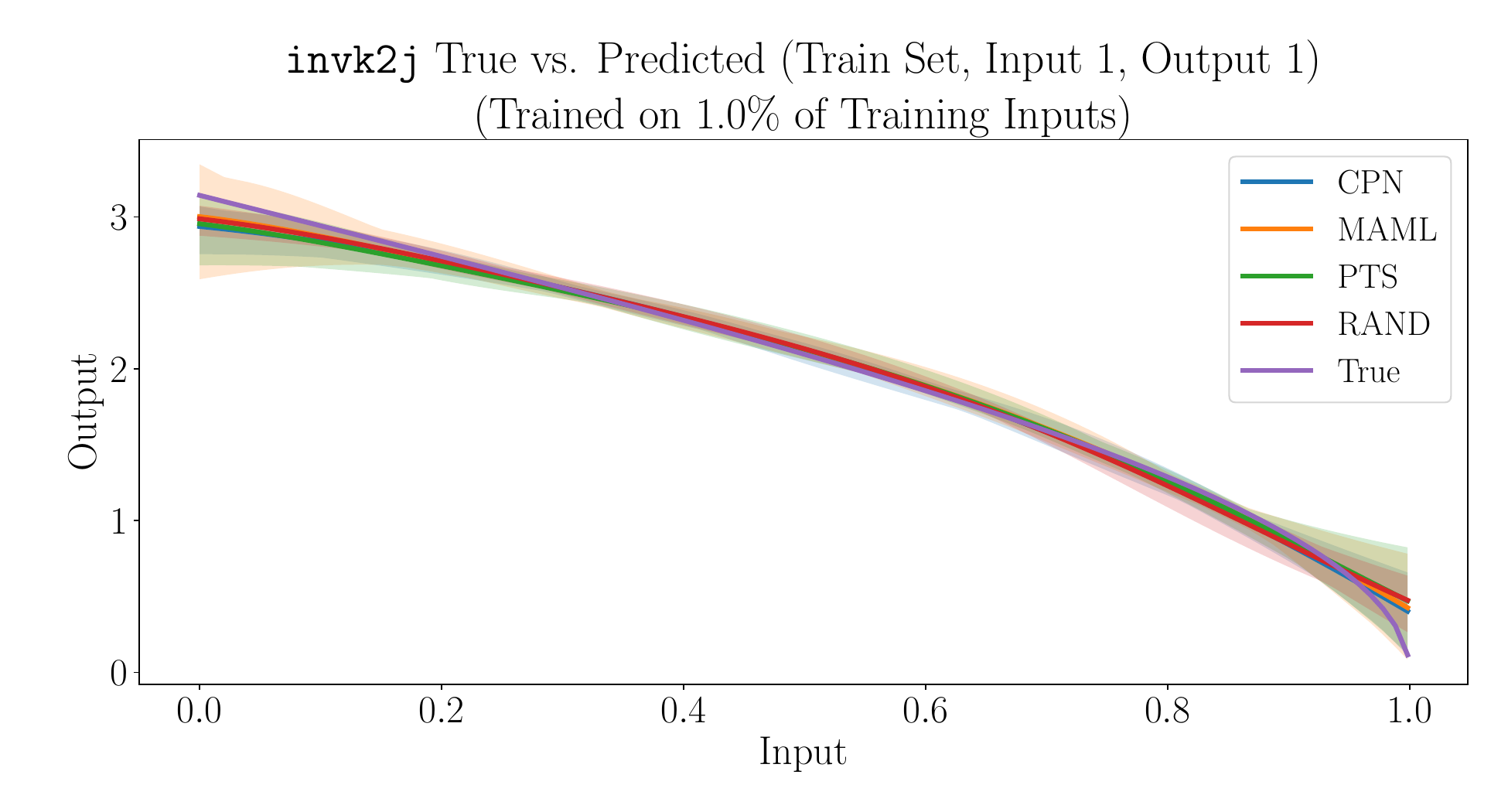}
\\
\vspace{-0.5em}
\includegraphics[width=0.49\textwidth]{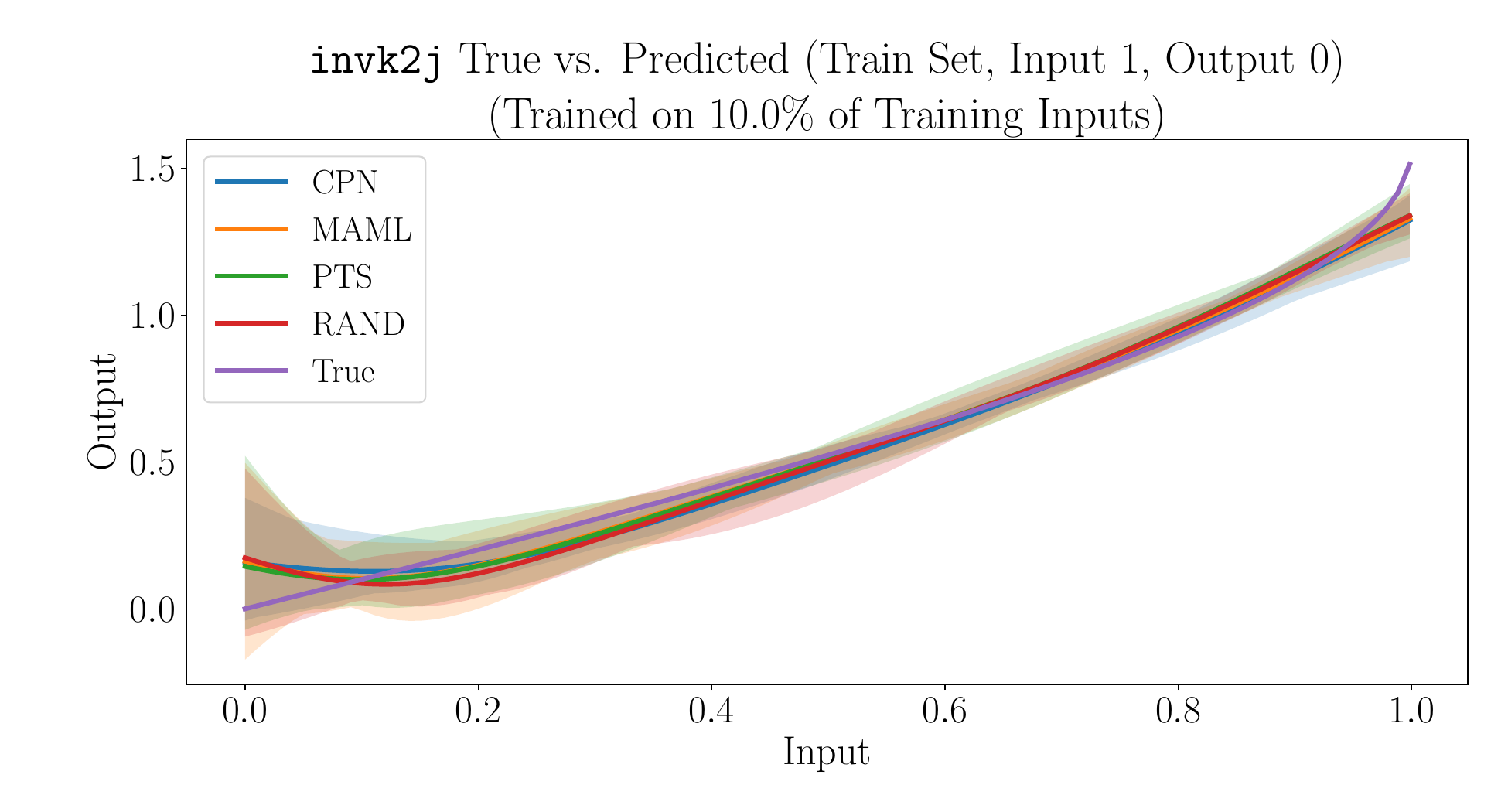}
\includegraphics[width=0.49\textwidth]{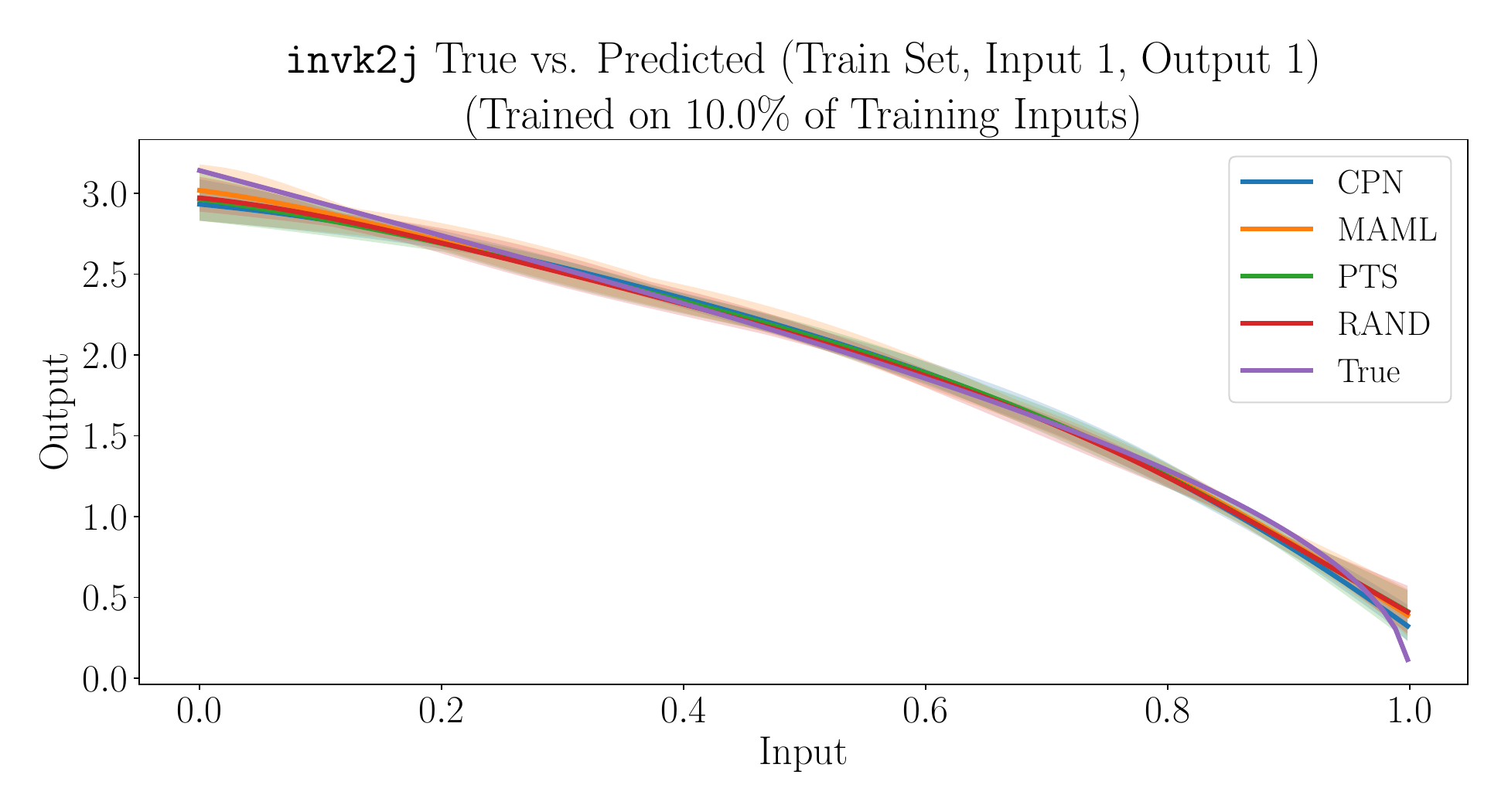}
\\
\vspace{-0.5em}
\includegraphics[width=0.49\textwidth]{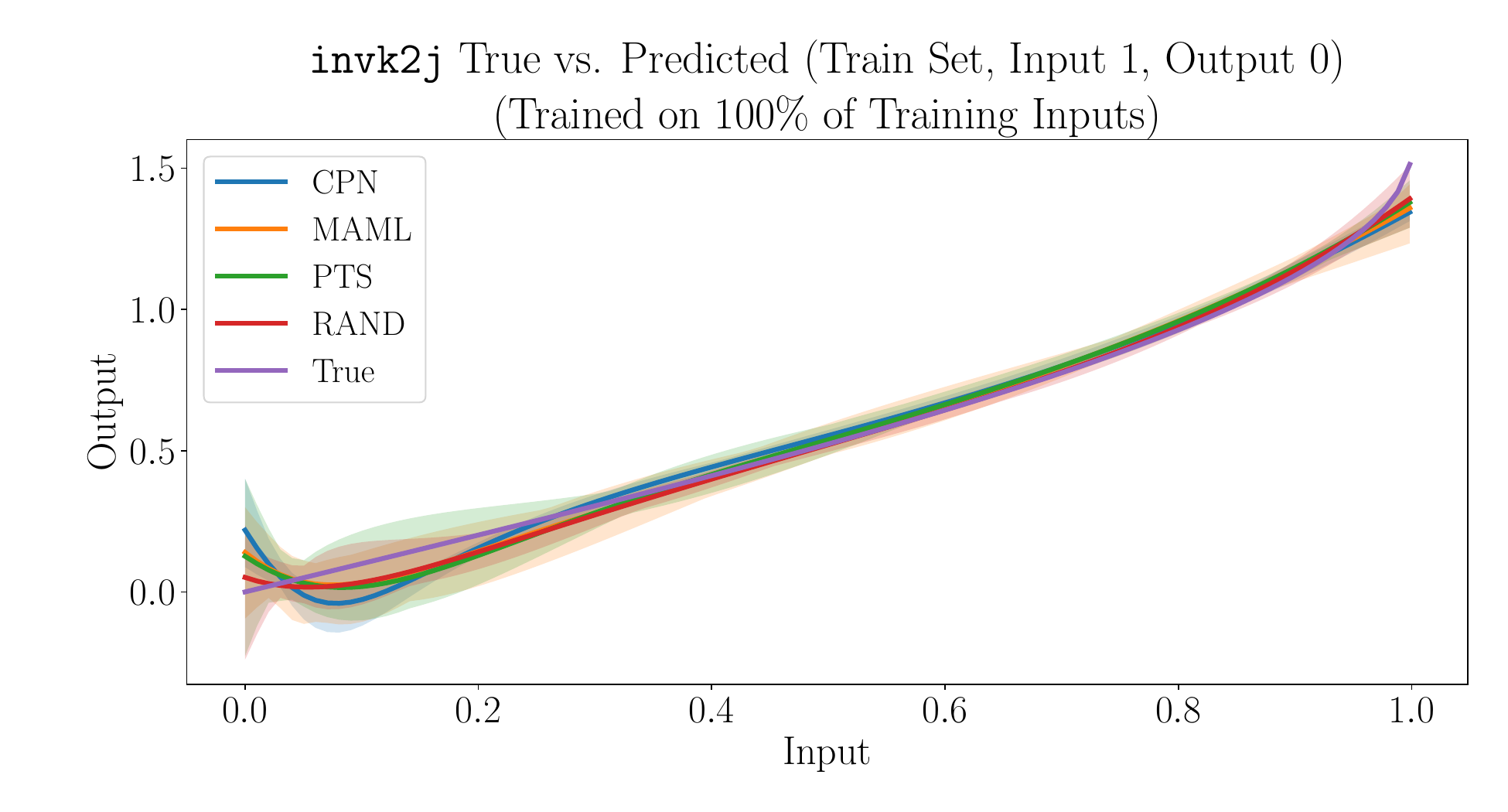}
\includegraphics[width=0.49\textwidth]{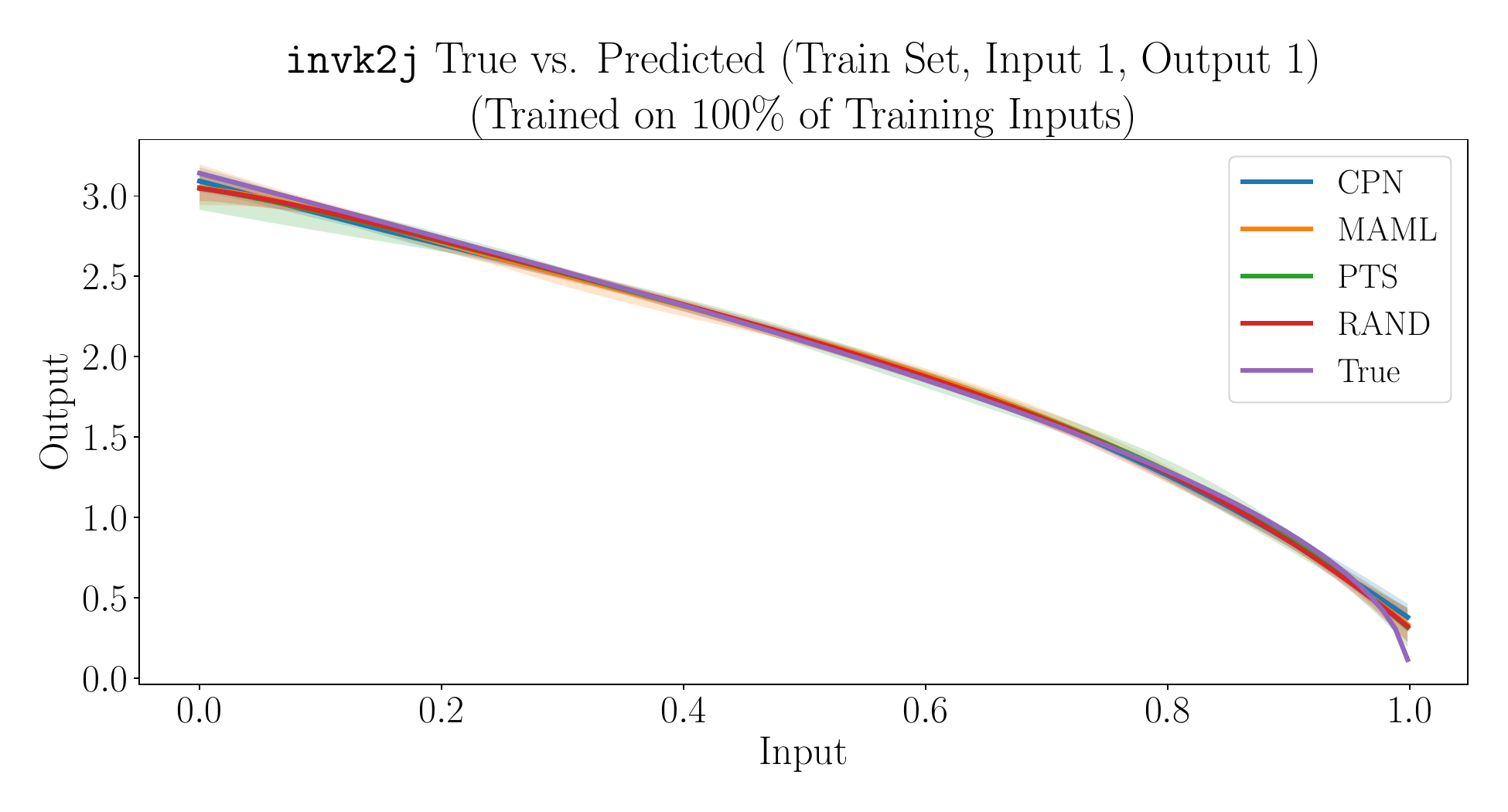}
\\
\vspace*{-1.2em}
  \caption{
    Visual comparisons of the ground-truth \texttt{invk2j} function from \textsc{ParrotBenchCPN} and neural surrogate approximations thereof, when the second input is varied.
    We include results for all dataset sizes evaluated in Section~\ref{sec:data_efficiency}, and we plot each output of the kernel when the input is varied.
  }\label{fig:true_vs_pred_invk2j_input_1}
\end{figure*}

\begin{figure*}
\centering
\vspace*{-0.7em}
\includegraphics[width=0.54\textwidth]{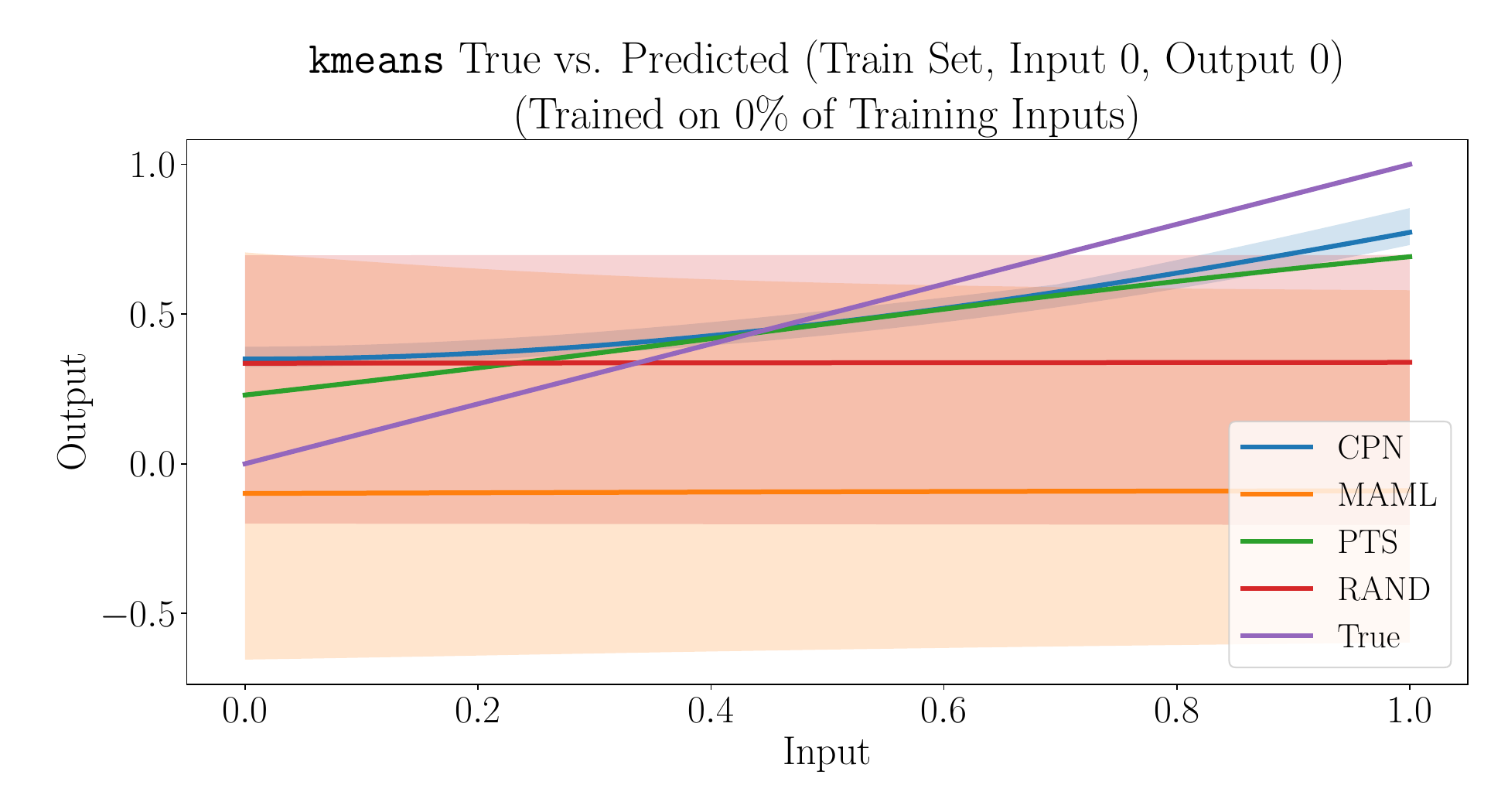}
\\
\vspace*{-1.6em}
\includegraphics[width=0.54\textwidth]{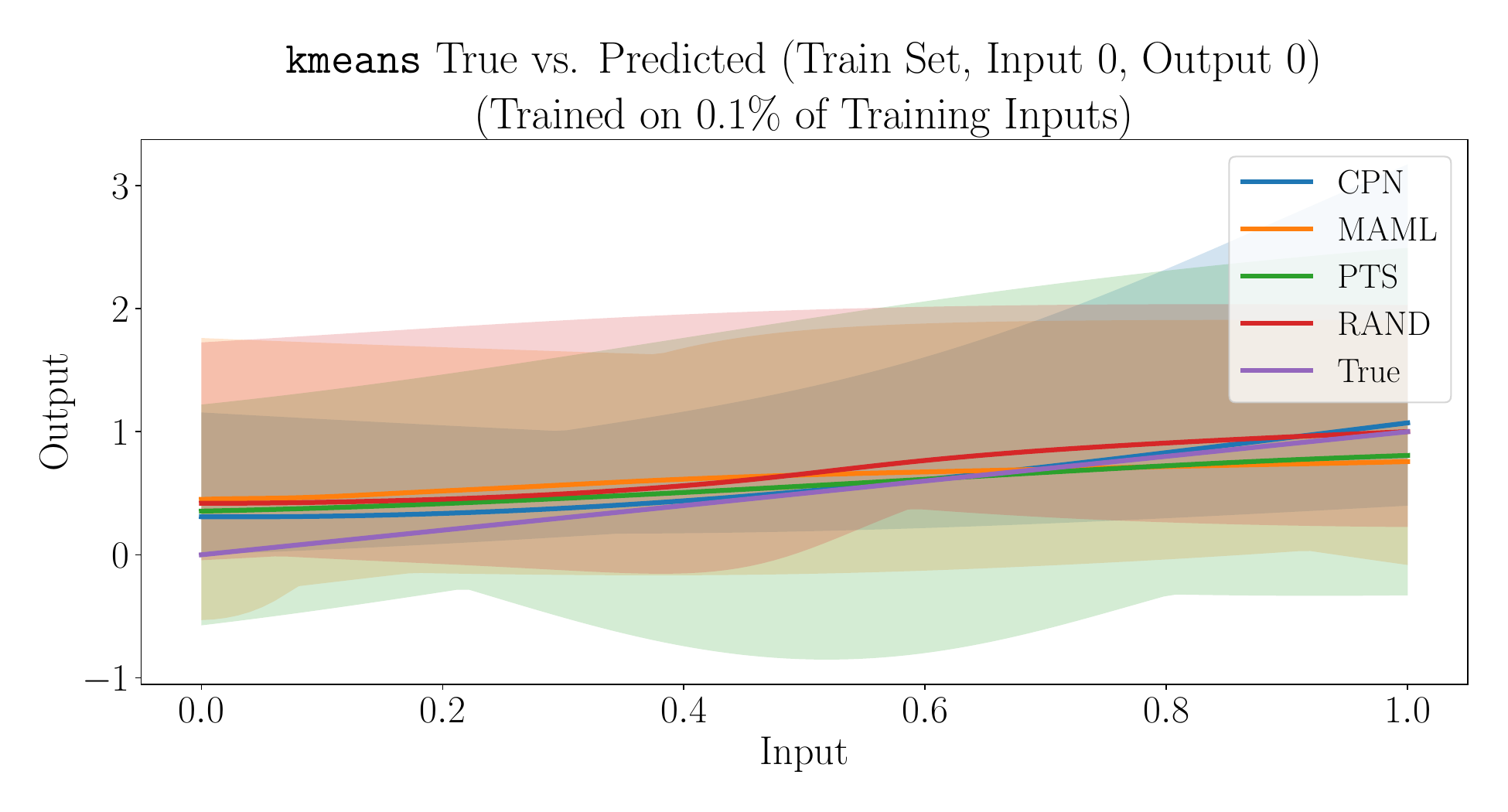}
\\
\vspace*{-1.6em}
\includegraphics[width=0.54\textwidth]{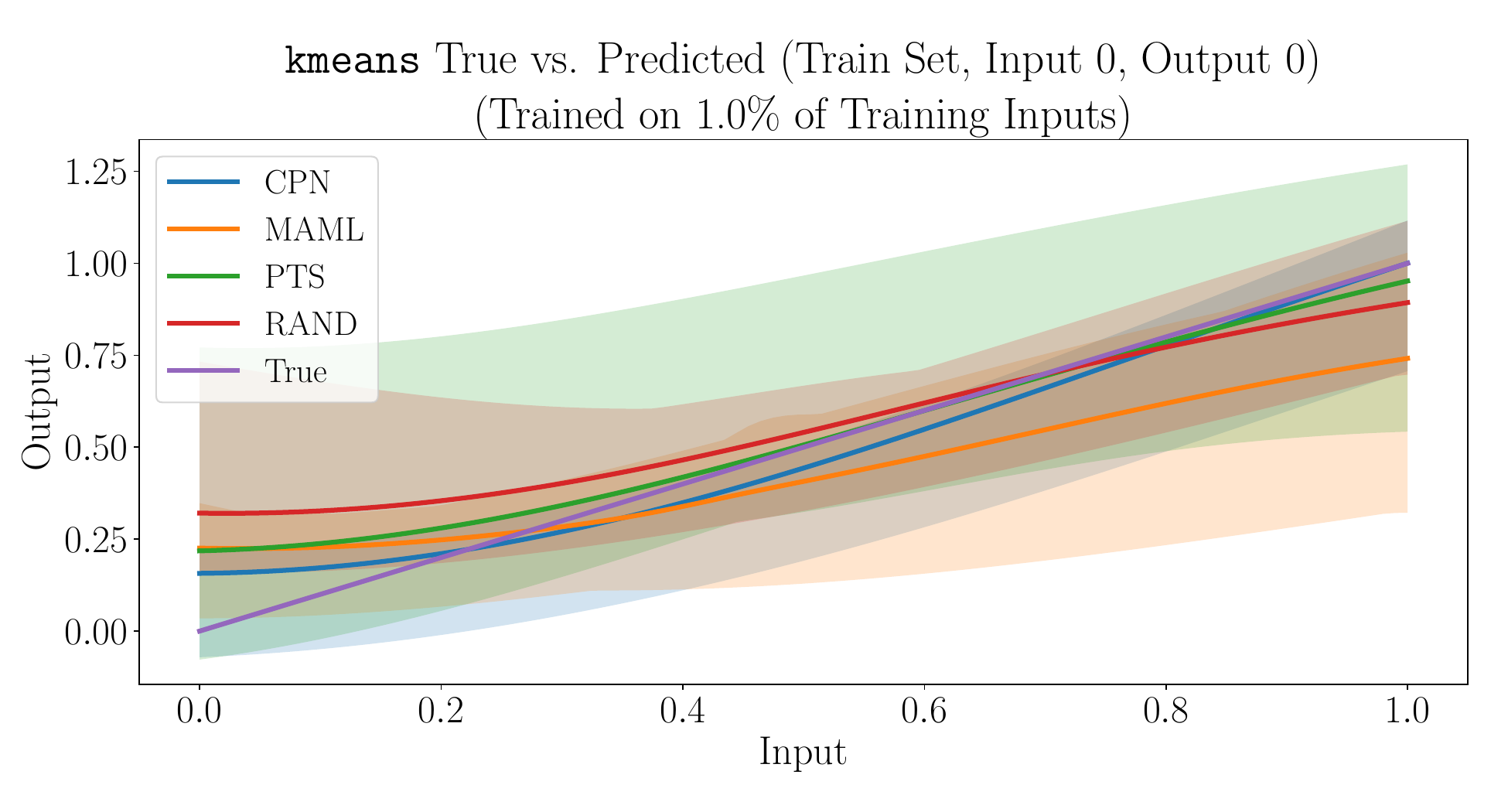}
\\
\vspace*{-1.6em}
\includegraphics[width=0.54\textwidth]{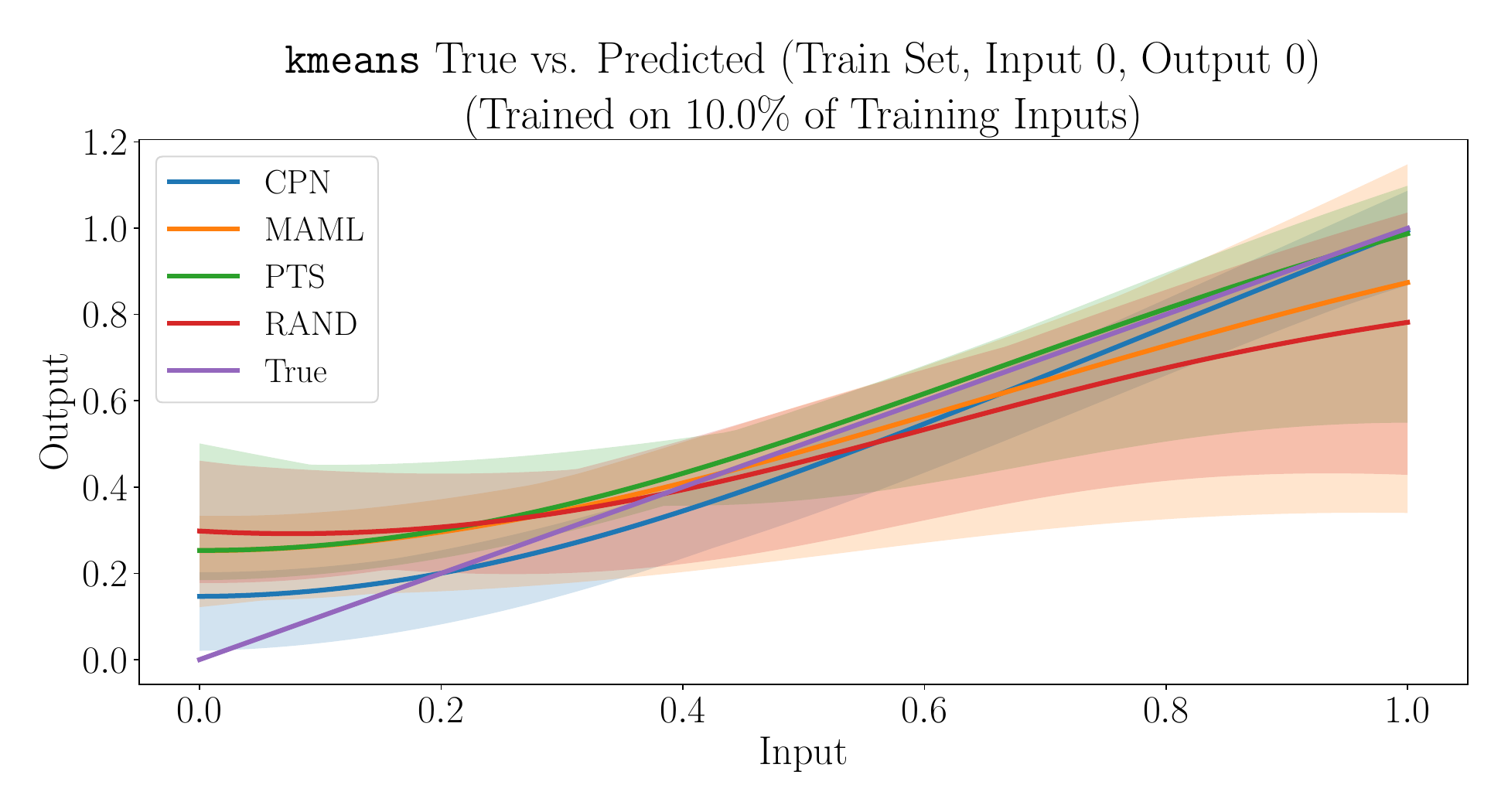}
\\
\vspace*{-1.6em}
\includegraphics[width=0.54\textwidth]{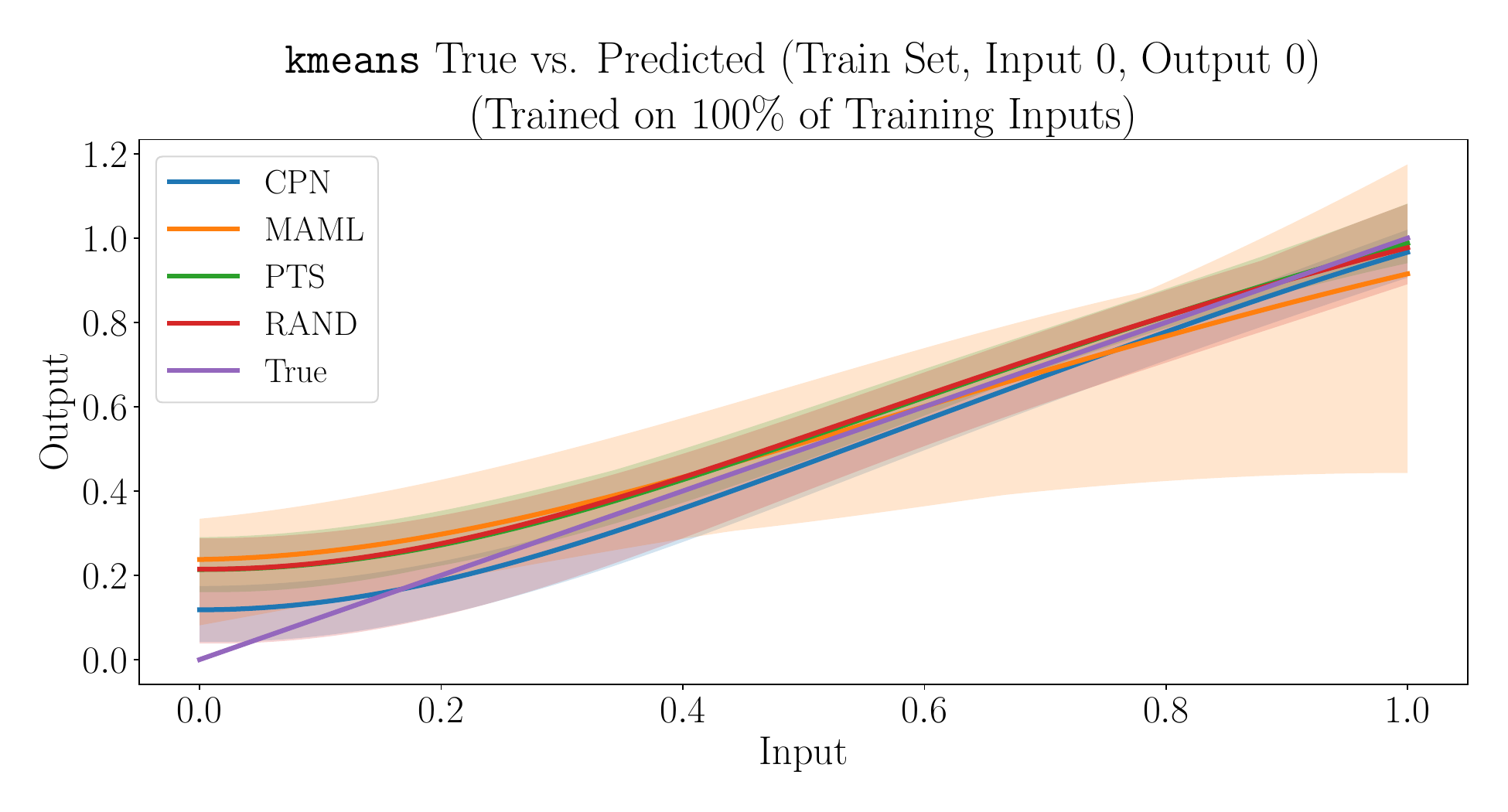}
\\
\vspace*{-1.2em}
\caption{
  Visual comparisons of the ground-truth \texttt{kmeans} function from \textsc{ParrotBenchCPN} and neural surrogate approximations thereof, when the first input is varied.
  We include results for all dataset sizes evaluated in Section~\ref{sec:data_efficiency}.
}\label{fig:true_vs_pred_kmeans_input_0}
\end{figure*}

\begin{figure*}
\centering
\vspace*{-0.7em}
\includegraphics[width=0.54\textwidth]{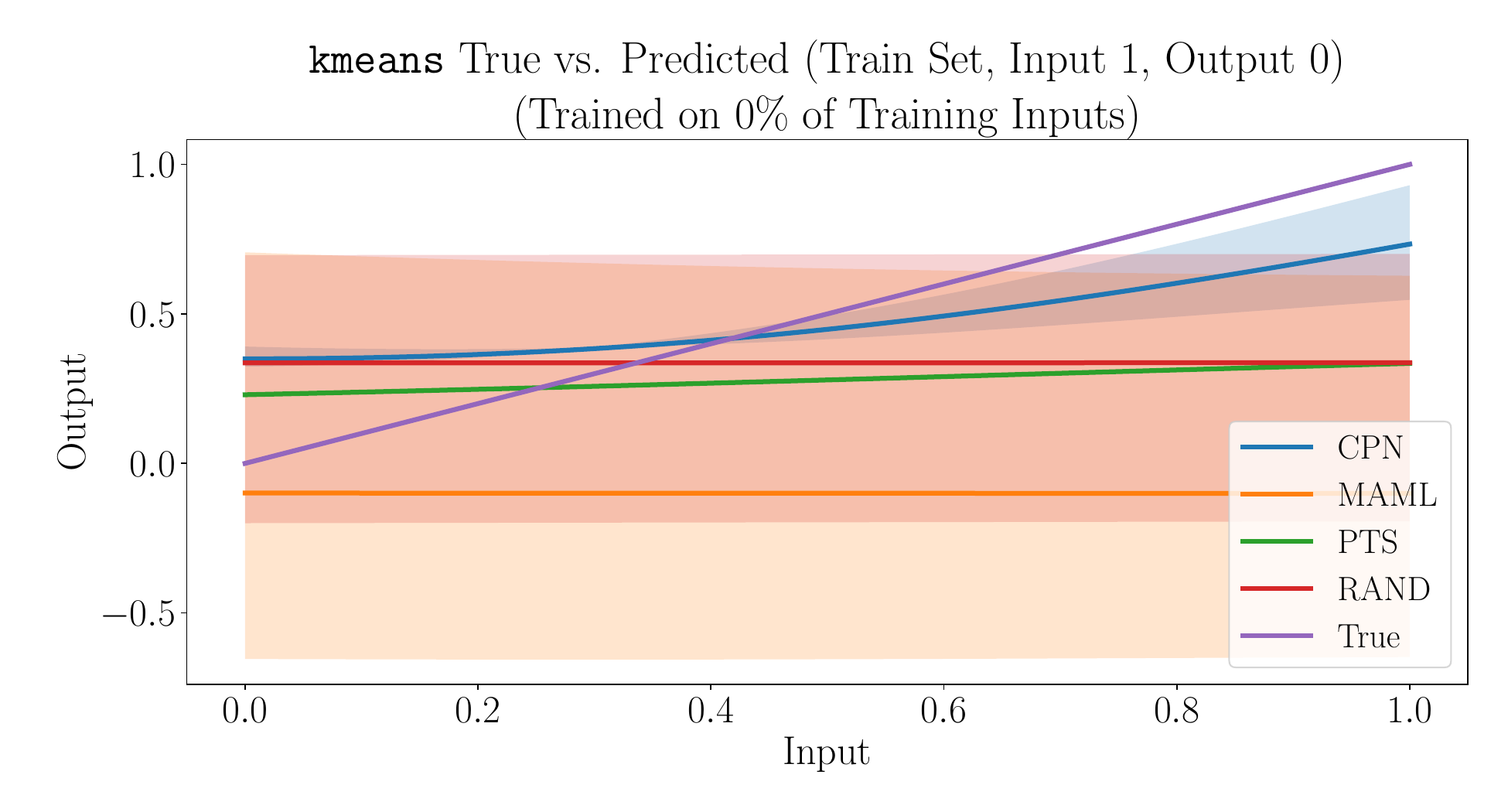}
\\
\vspace*{-1.6em}
\includegraphics[width=0.54\textwidth]{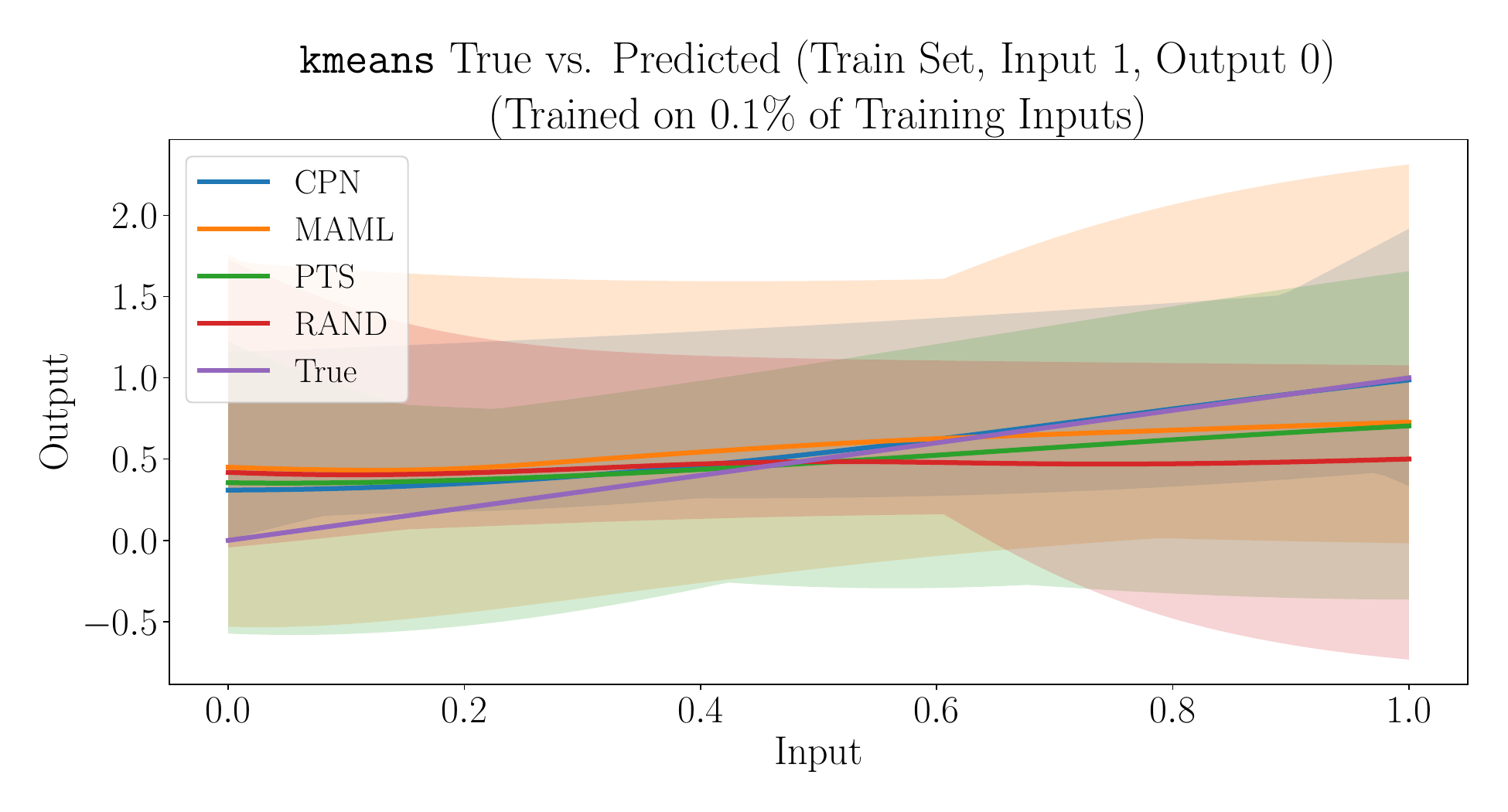}
\\
\vspace*{-1.6em}
\includegraphics[width=0.54\textwidth]{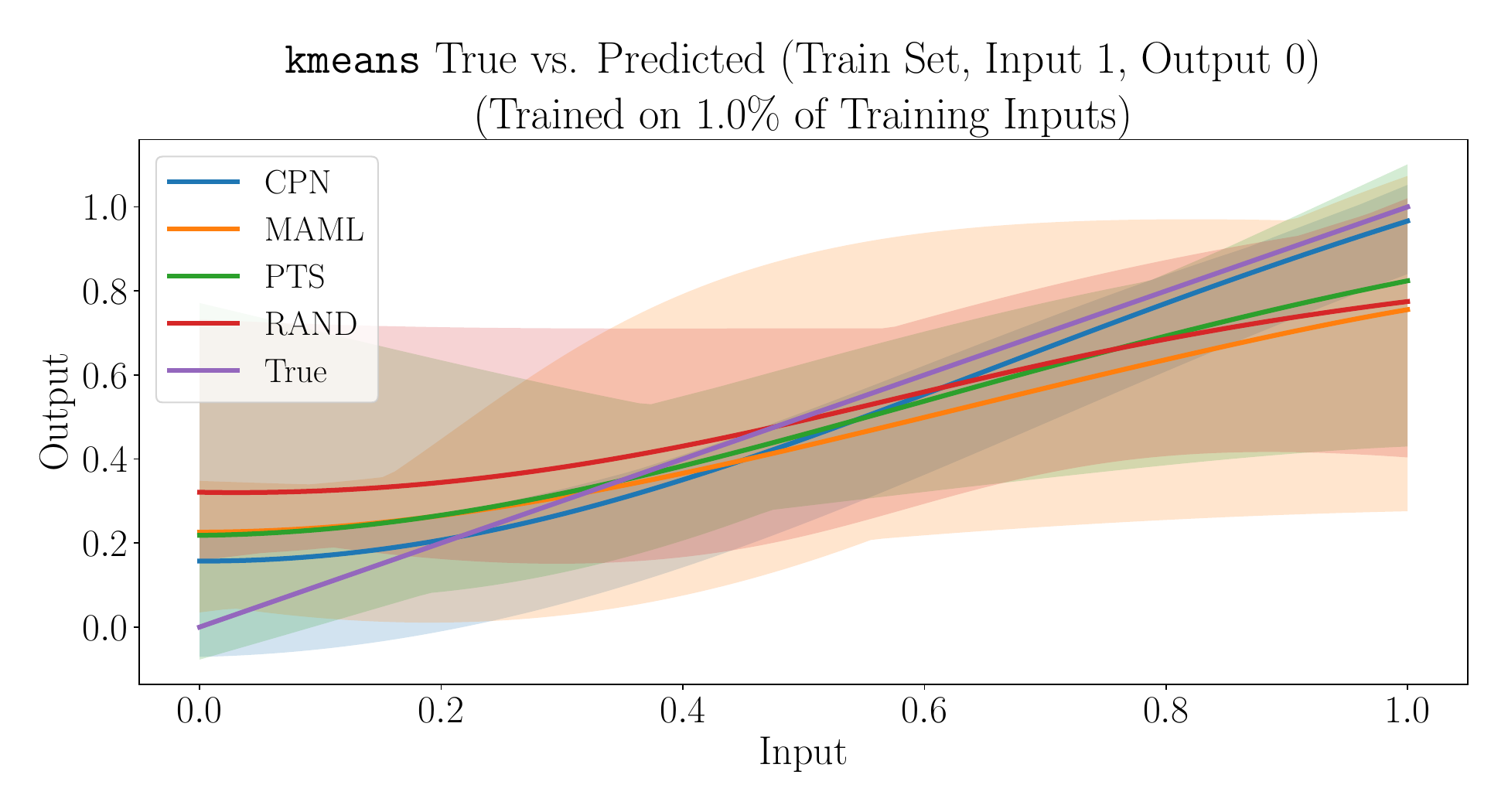}
\\
\vspace*{-1.6em}
\includegraphics[width=0.54\textwidth]{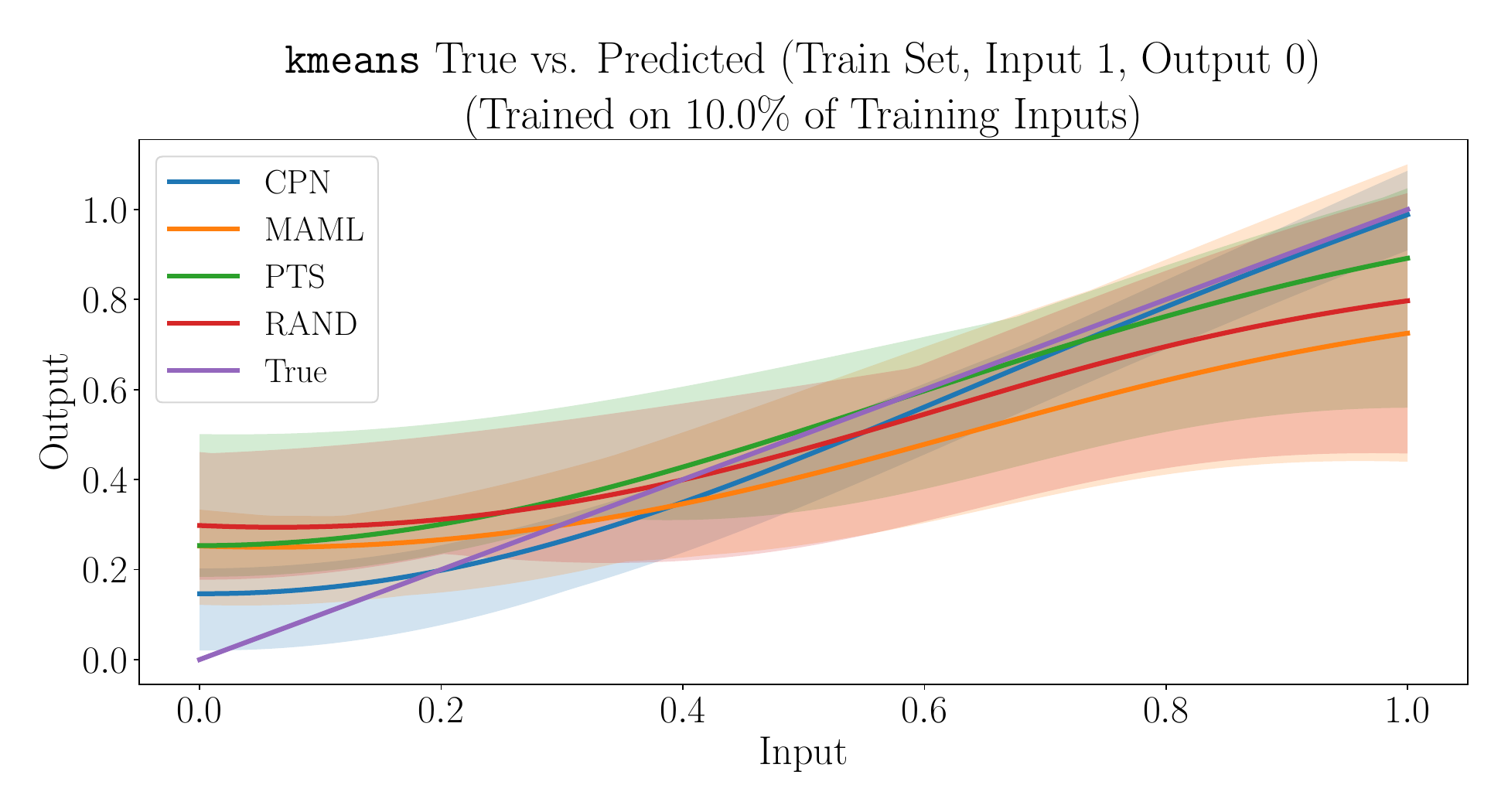}
\\
\vspace*{-1.6em}
\includegraphics[width=0.54\textwidth]{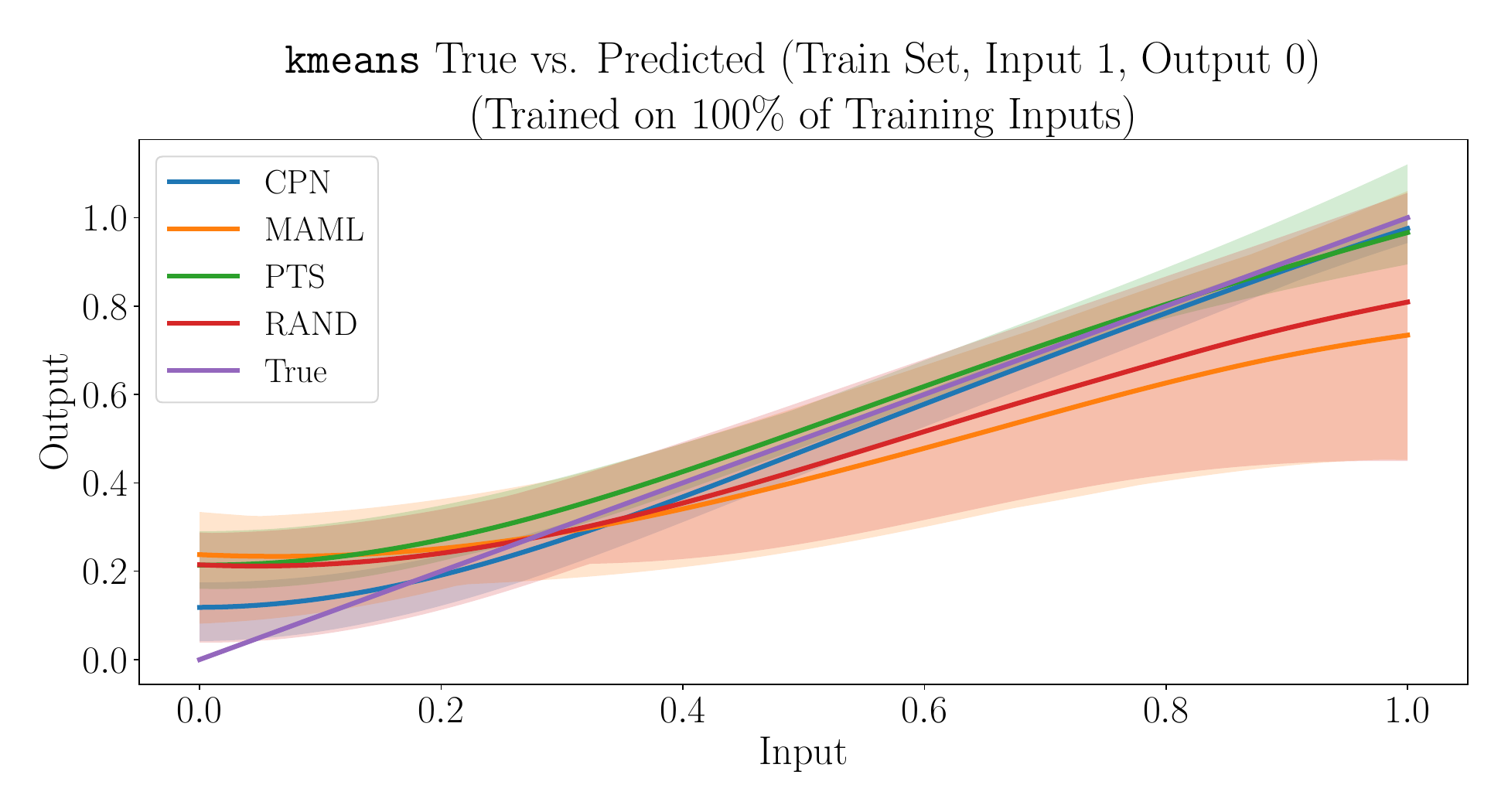}
\\
\vspace*{-1.2em}
\caption{
  Visual comparisons of the ground-truth \texttt{kmeans} function from \textsc{ParrotBenchCPN} and neural surrogate approximations thereof, when the second input is varied.
  We include results for all dataset sizes evaluated in Section~\ref{sec:data_efficiency}.
}\label{fig:true_vs_pred_kmeans_input_1}
\end{figure*}

\begin{figure*}
\centering
\vspace*{-0.7em}
\includegraphics[width=0.54\textwidth]{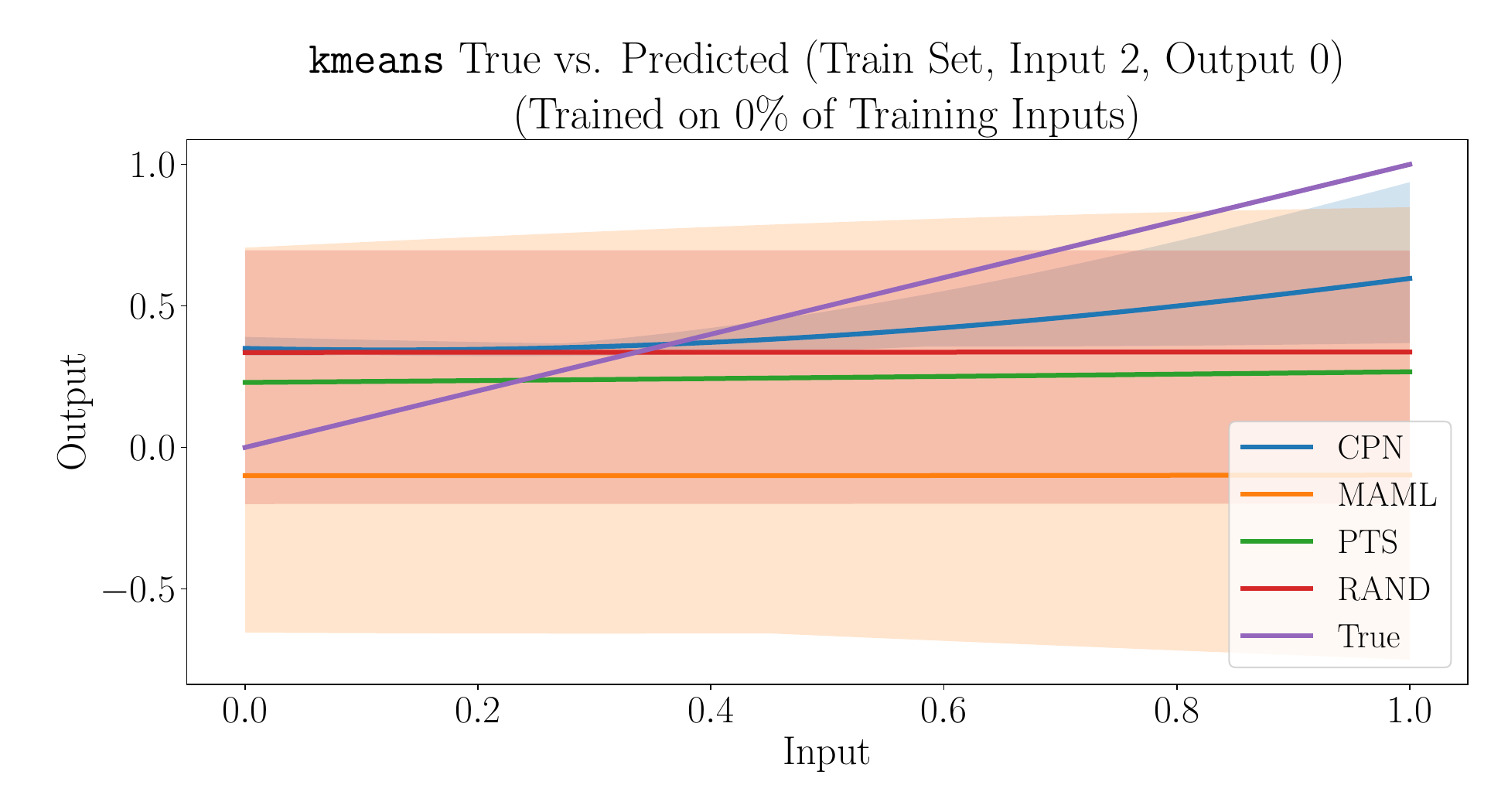}
\\
\vspace*{-1.6em}
\includegraphics[width=0.54\textwidth]{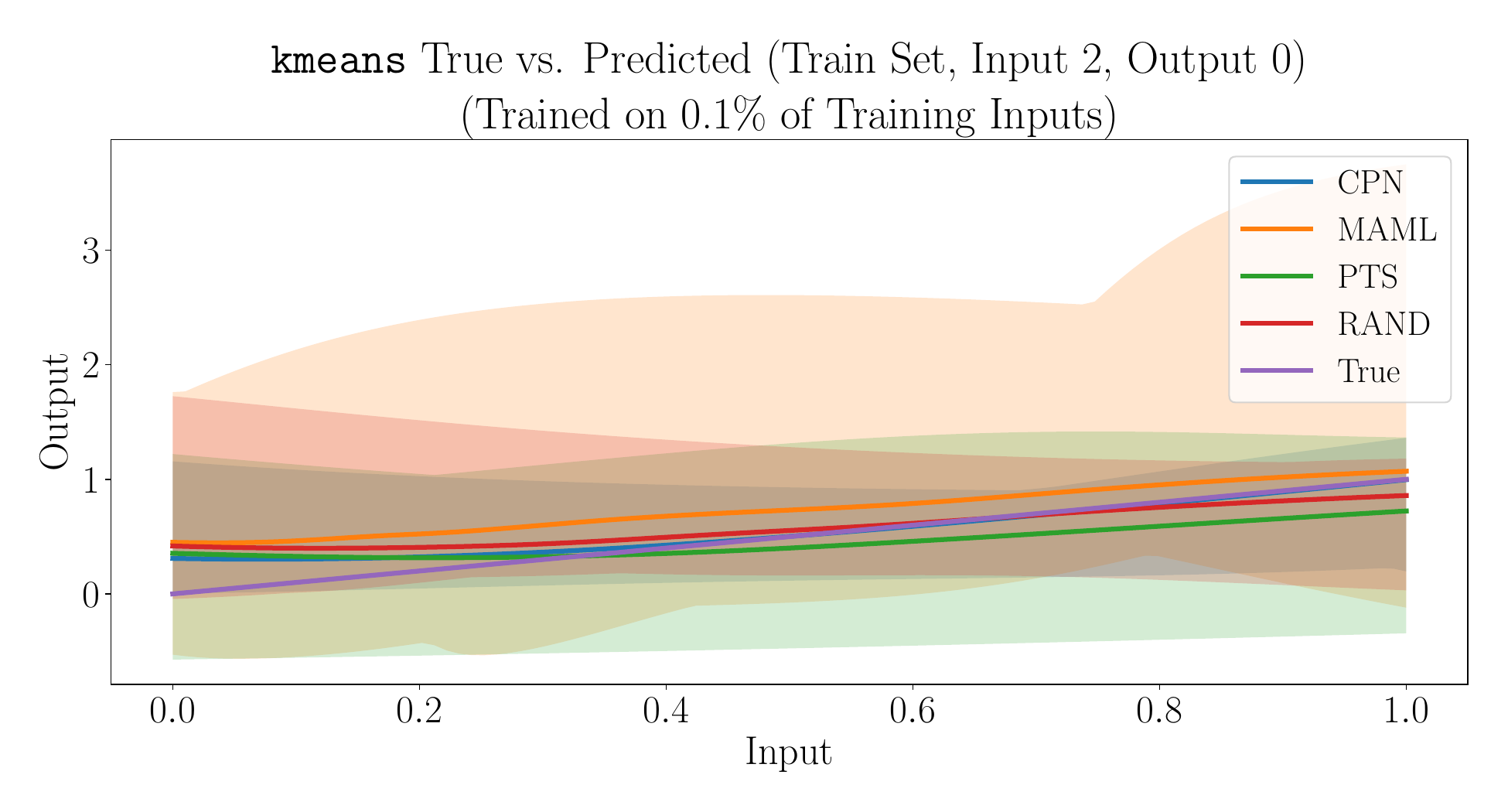}
\\
\vspace*{-1.6em}
\includegraphics[width=0.54\textwidth]{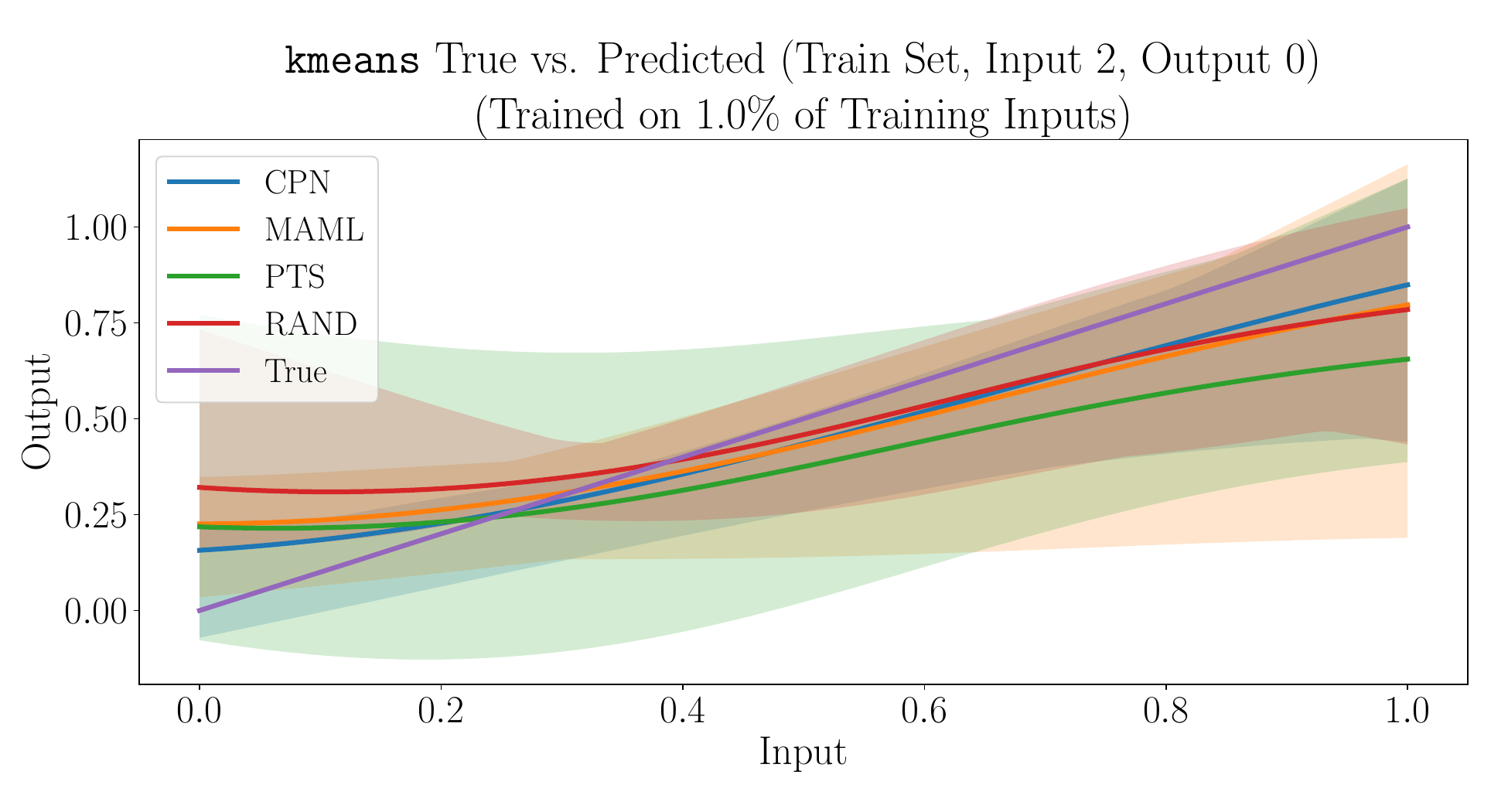}
\\
\vspace*{-1.6em}
\includegraphics[width=0.54\textwidth]{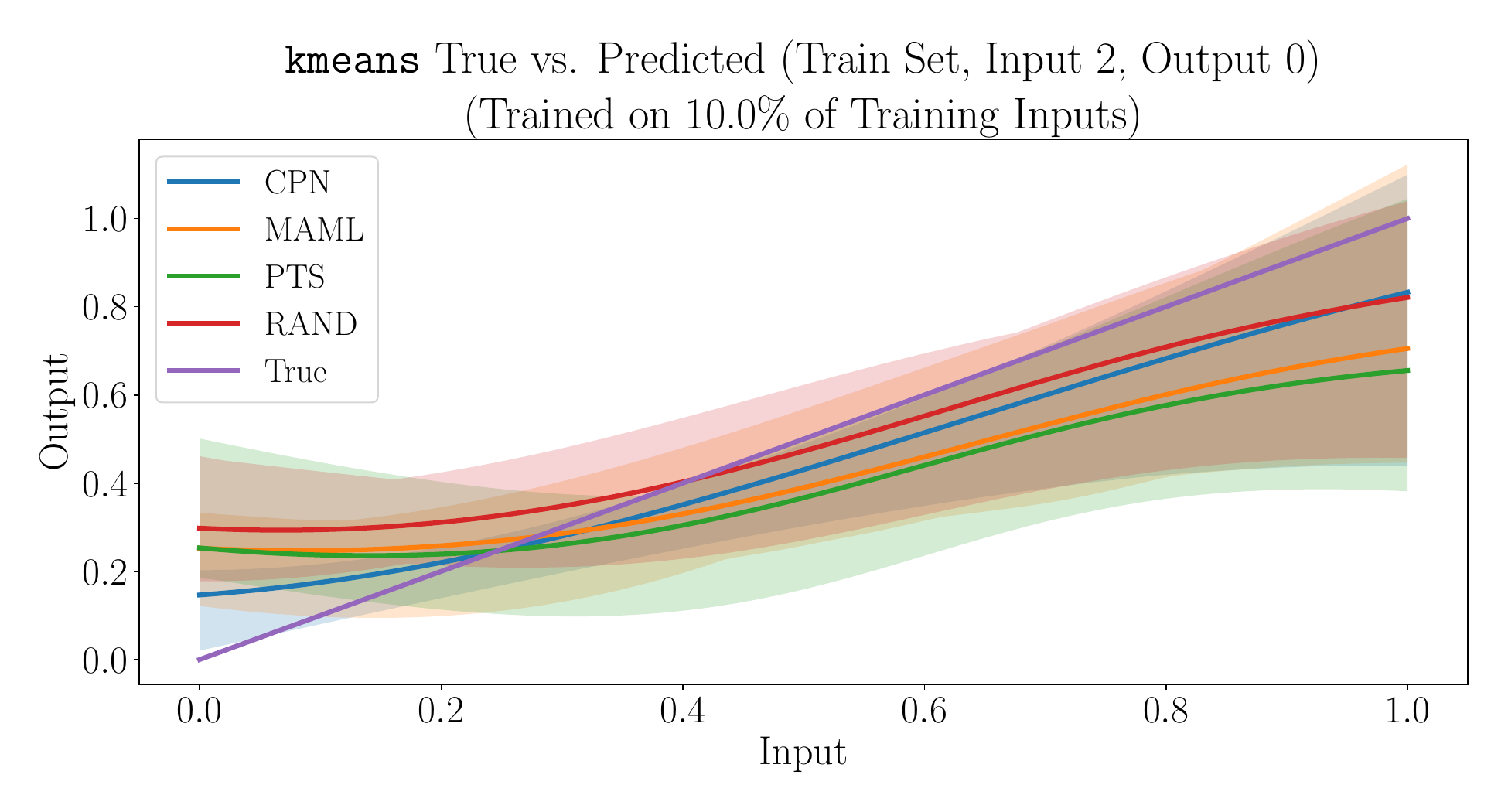}
\\
\vspace*{-1.6em}
\includegraphics[width=0.54\textwidth]{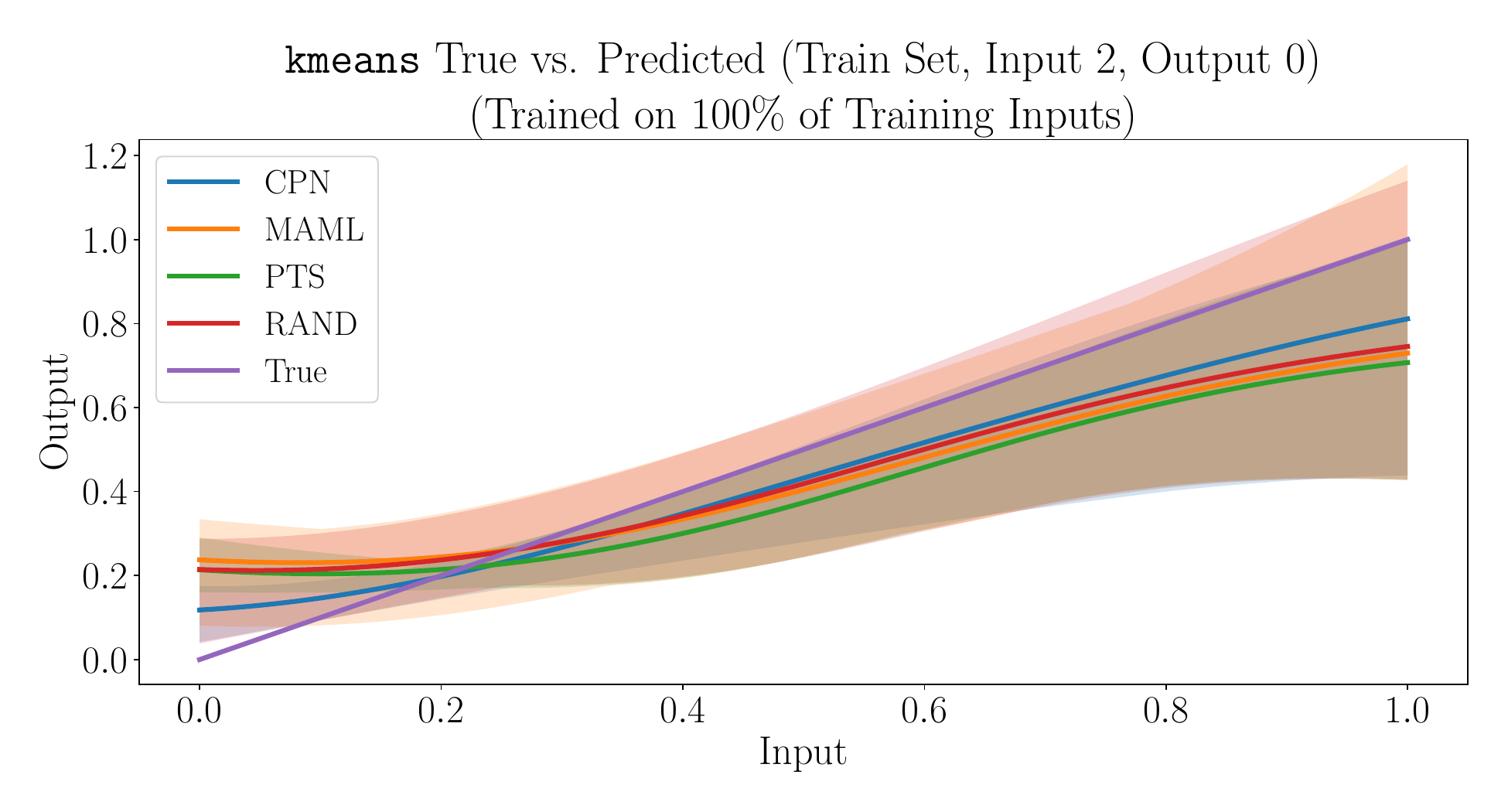}
\\
\vspace*{-1.2em}
\caption{
  Visual comparisons of the ground-truth \texttt{kmeans} function from \textsc{ParrotBenchCPN} and neural surrogate approximations thereof, when the third input is varied.
  We include results for all dataset sizes evaluated in Section~\ref{sec:data_efficiency}.
}\label{fig:true_vs_pred_kmeans_input_2}
\end{figure*}

\begin{figure*}
\centering
\vspace*{-0.7em}
\includegraphics[width=0.54\textwidth]{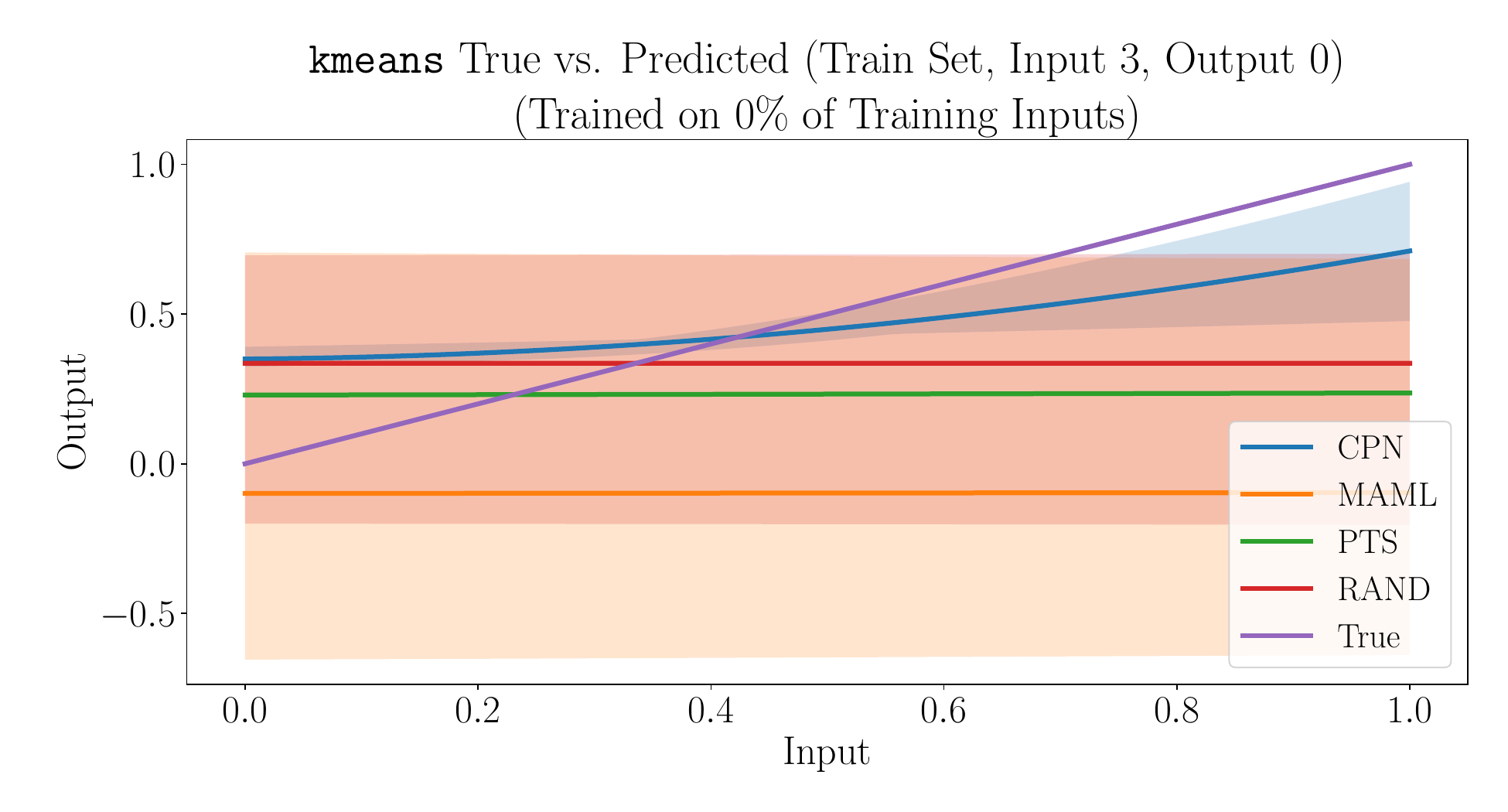}
\\
\vspace*{-1.6em}
\includegraphics[width=0.54\textwidth]{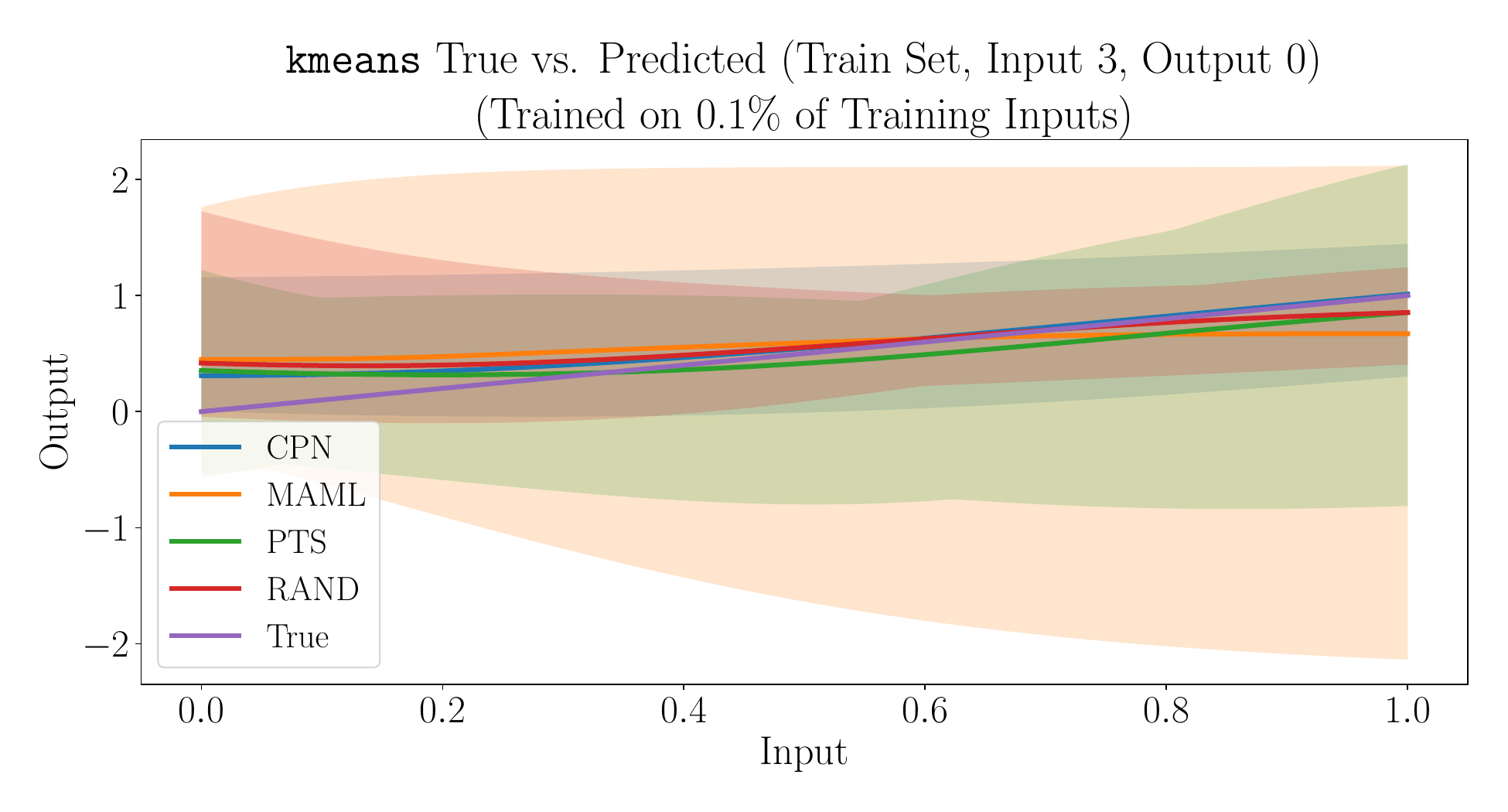}
\\
\vspace*{-1.6em}
\includegraphics[width=0.54\textwidth]{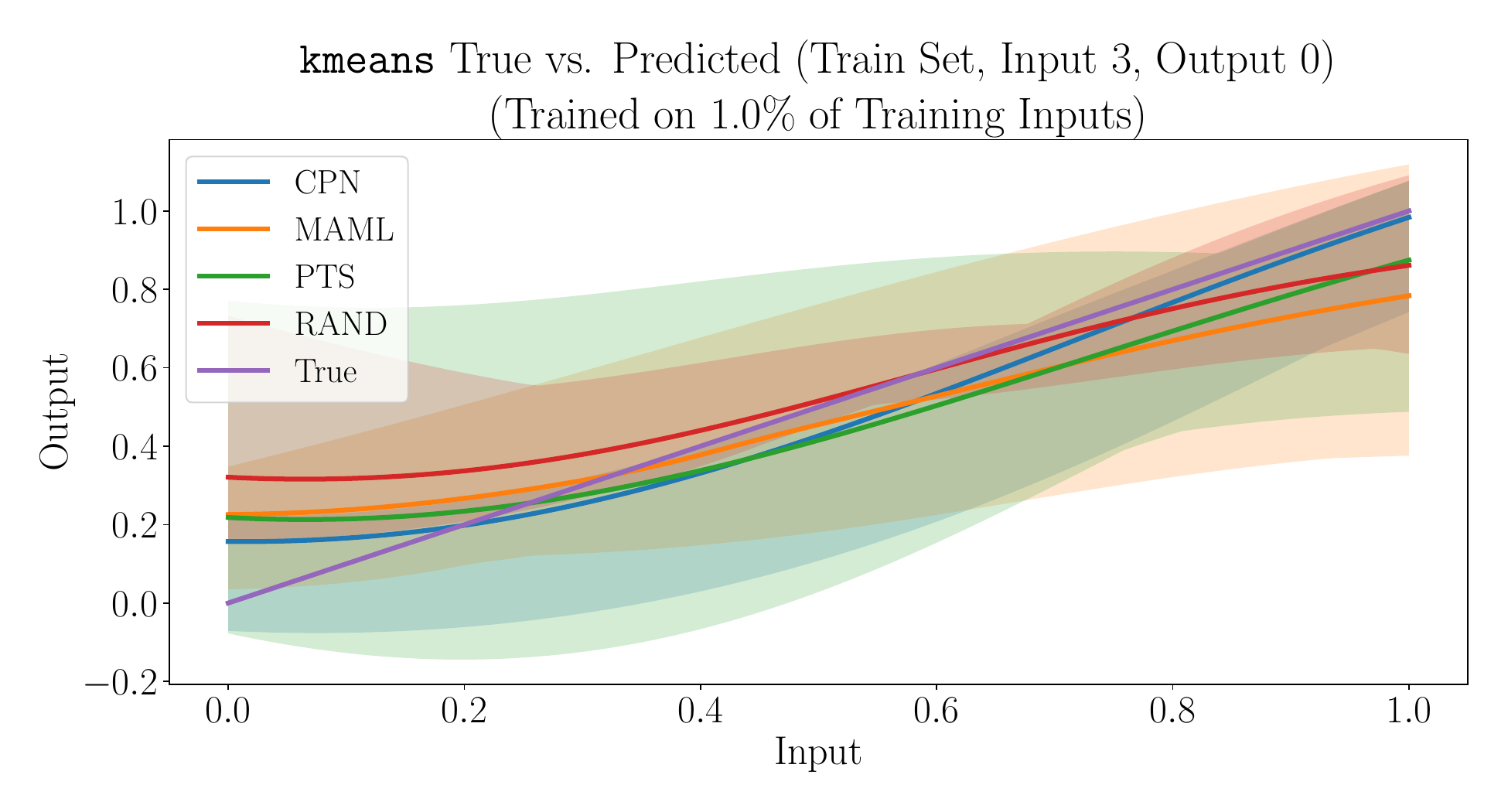}
\\
\vspace*{-1.6em}
\includegraphics[width=0.54\textwidth]{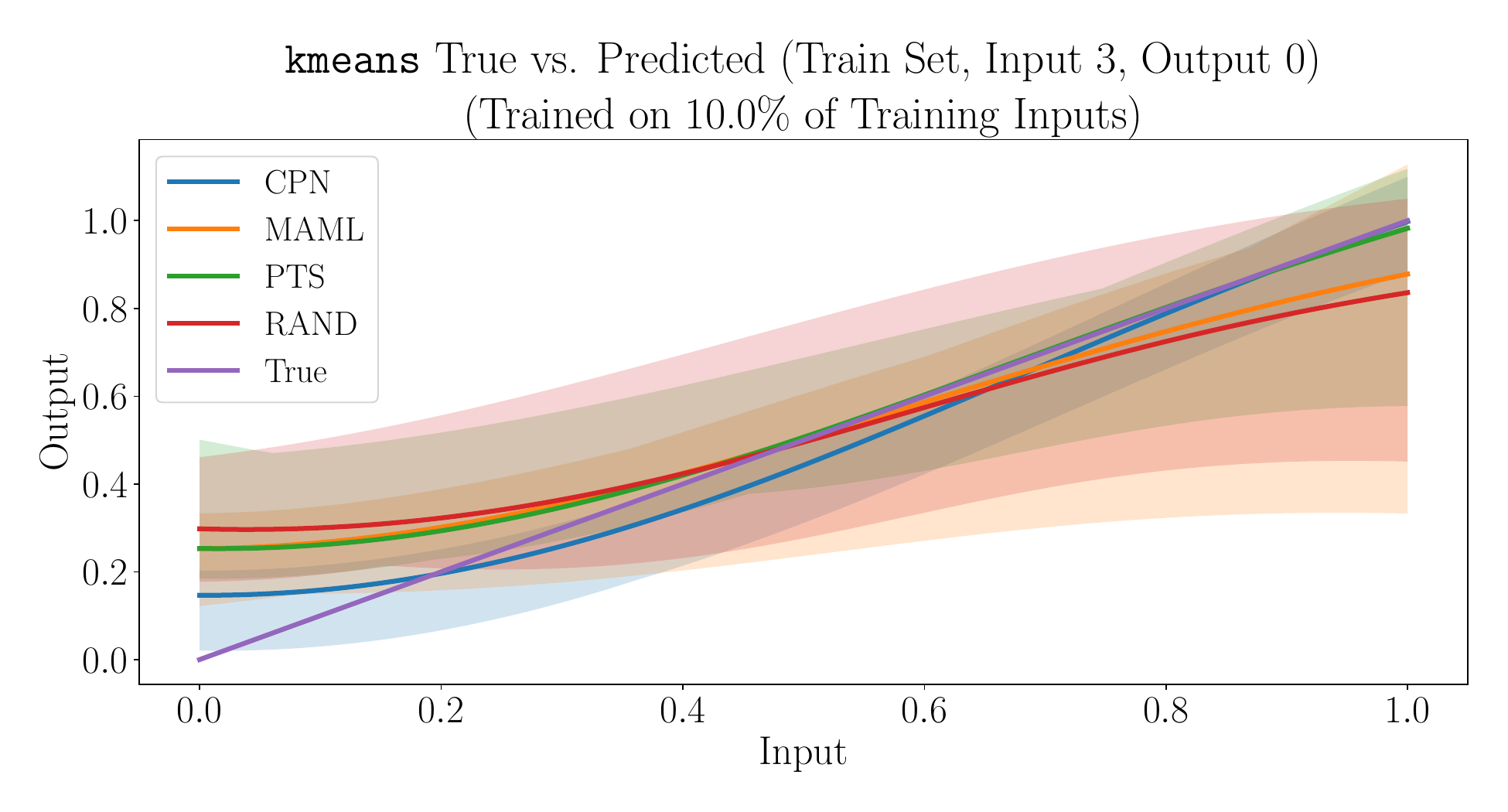}
\\
\vspace*{-1.6em}
\includegraphics[width=0.54\textwidth]{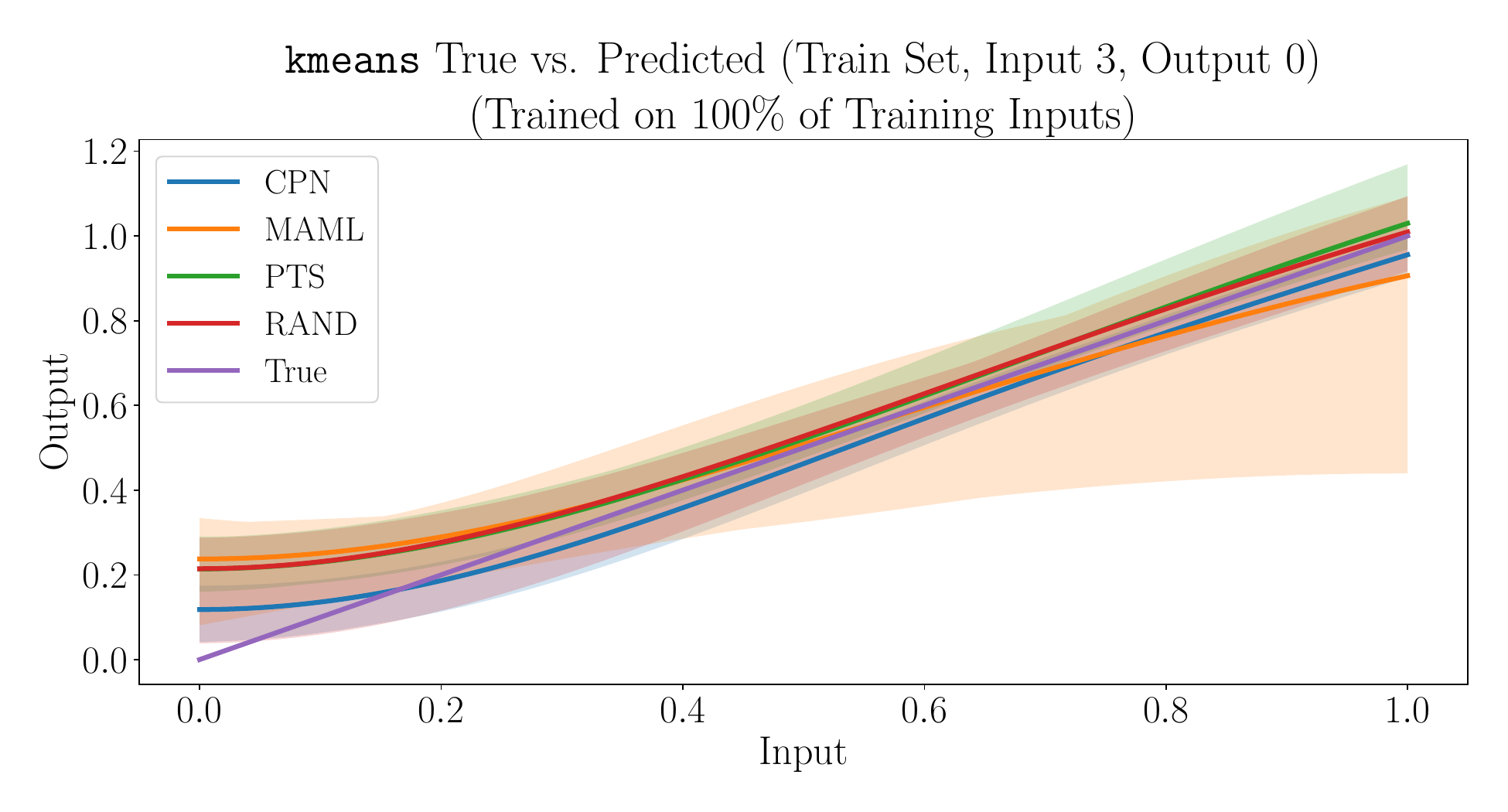}
\\
\vspace*{-1.2em}
\caption{
  Visual comparisons of the ground-truth \texttt{kmeans} function from \textsc{ParrotBenchCPN} and neural surrogate approximations thereof, when the fourth input is varied.
  We include results for all dataset sizes evaluated in Section~\ref{sec:data_efficiency}.
}\label{fig:true_vs_pred_kmeans_input_3}
\end{figure*}

\begin{figure*}
\centering
\vspace*{-0.7em}
\includegraphics[width=0.54\textwidth]{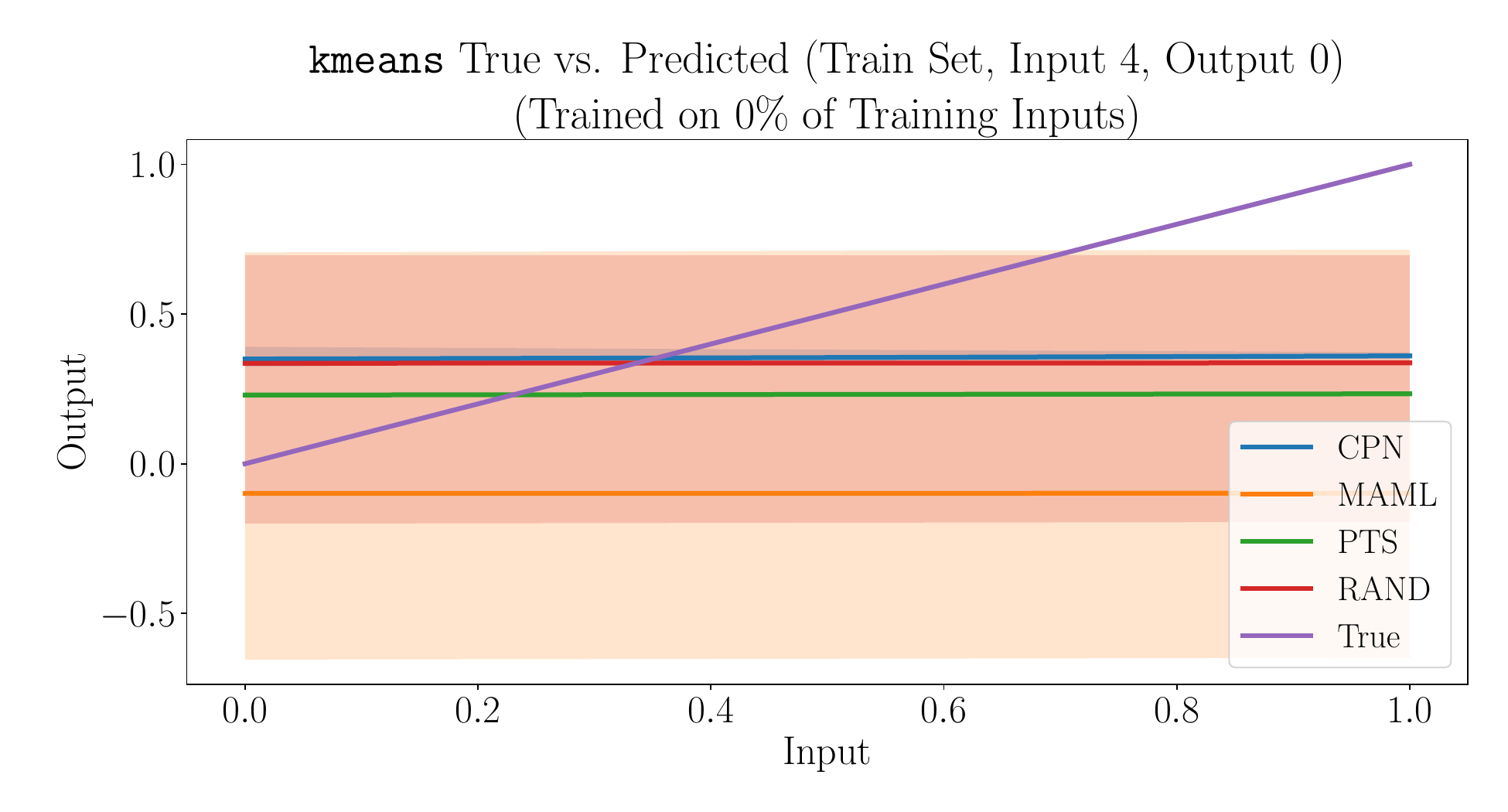}
\\
\vspace*{-1.6em}
\includegraphics[width=0.54\textwidth]{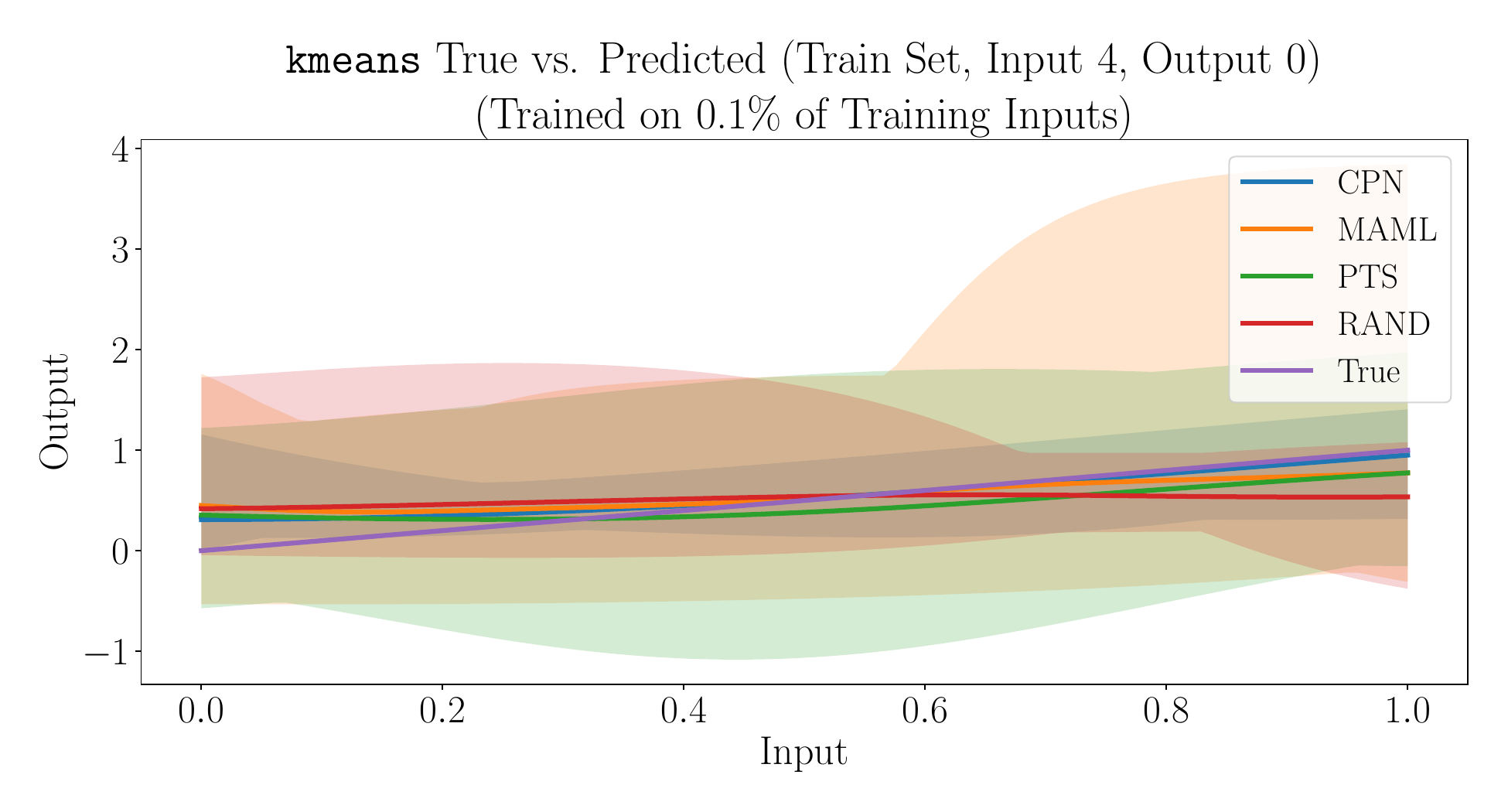}
\\
\vspace*{-1.6em}
\includegraphics[width=0.54\textwidth]{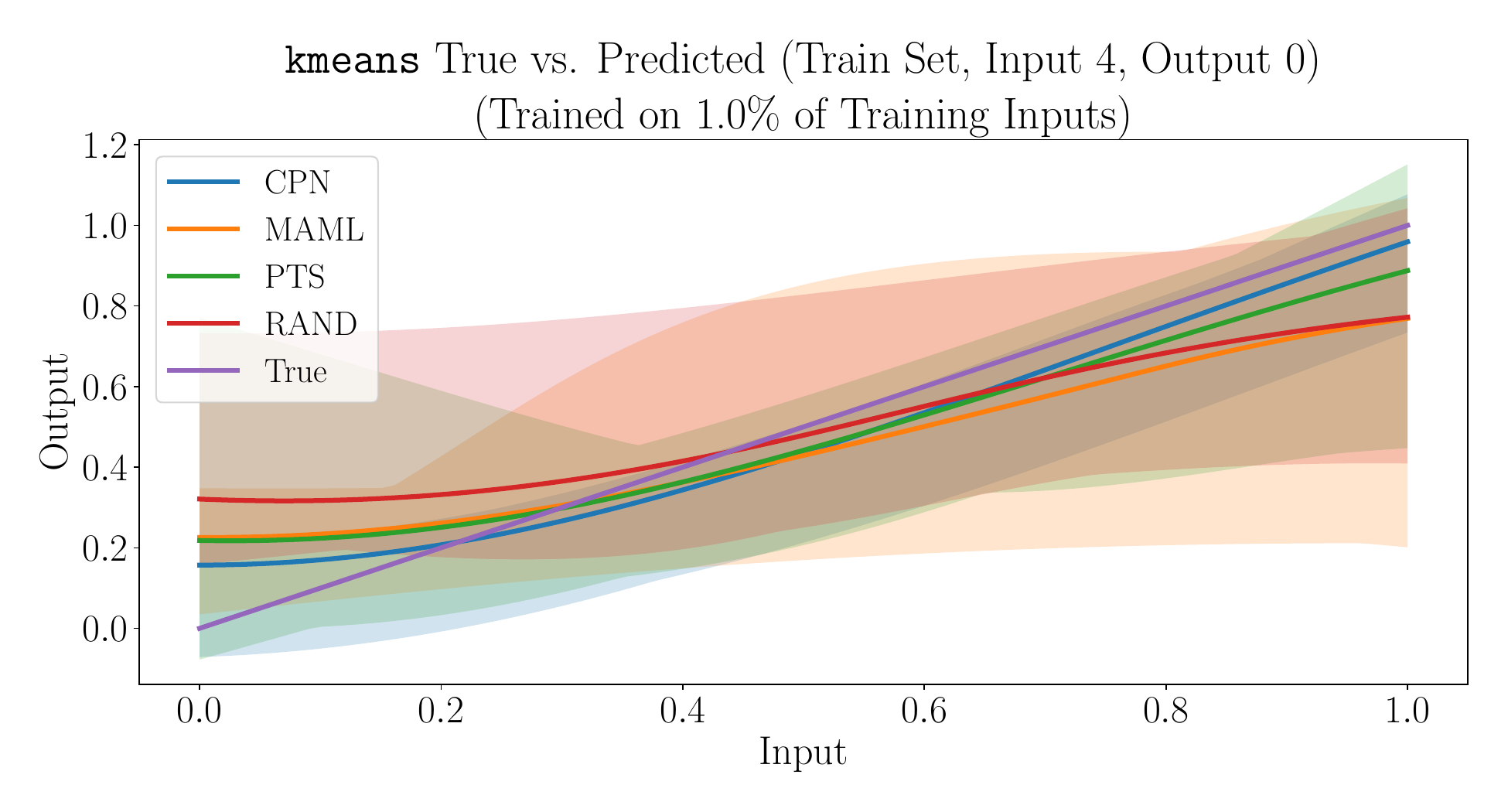}
\\
\vspace*{-1.6em}
\includegraphics[width=0.54\textwidth]{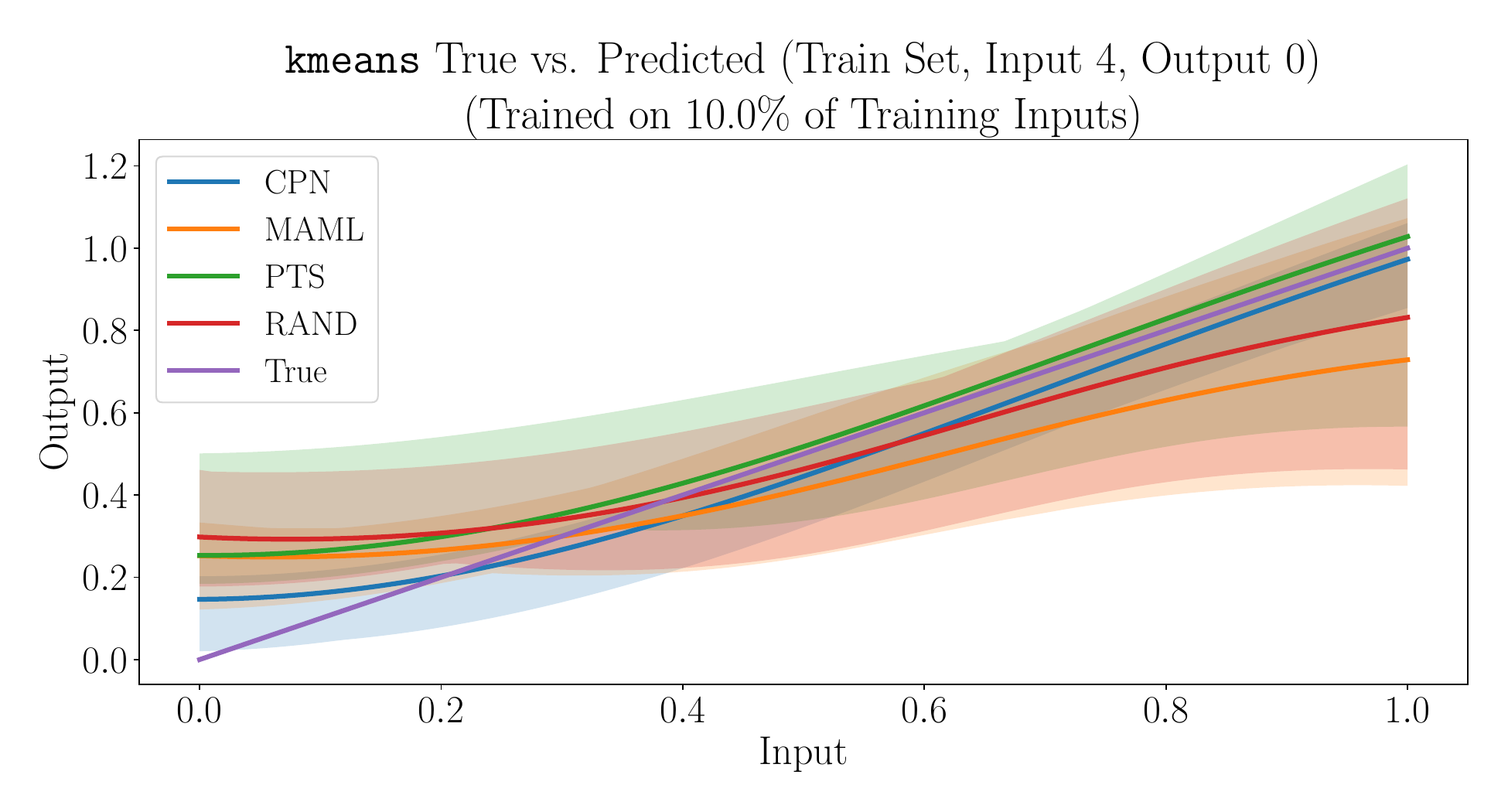}
\\
\vspace*{-1.6em}
\includegraphics[width=0.54\textwidth]{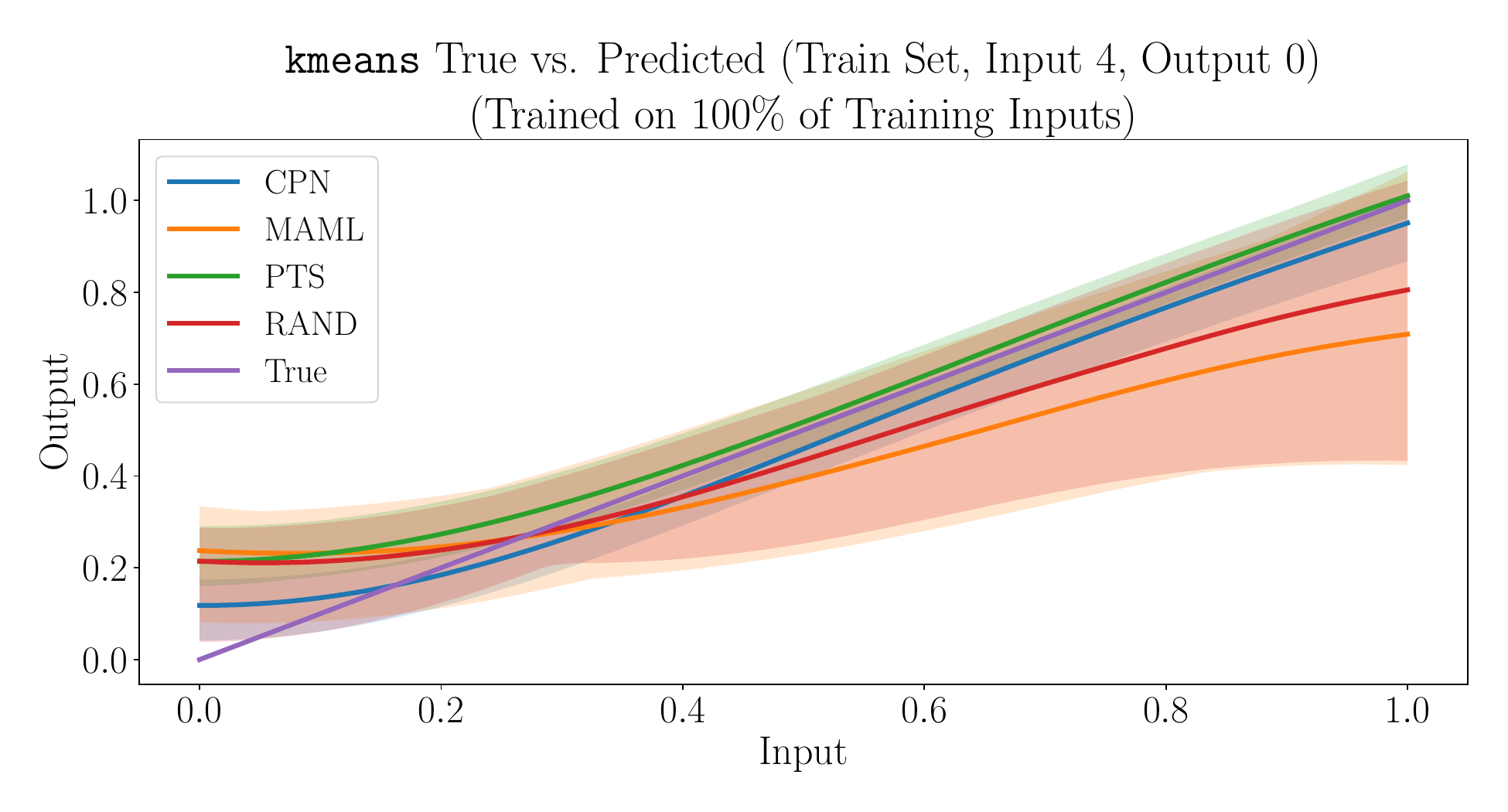}
\\
\vspace*{-1.2em}
\caption{
  Visual comparisons of the ground-truth \texttt{kmeans} function from \textsc{ParrotBenchCPN} and neural surrogate approximations thereof, when the fifth input is varied.
  We include results for all dataset sizes evaluated in Section~\ref{sec:data_efficiency}.
}\label{fig:true_vs_pred_kmeans_input_4}
\end{figure*}

\begin{figure*}
\centering
\vspace*{-0.7em}
\includegraphics[width=0.54\textwidth]{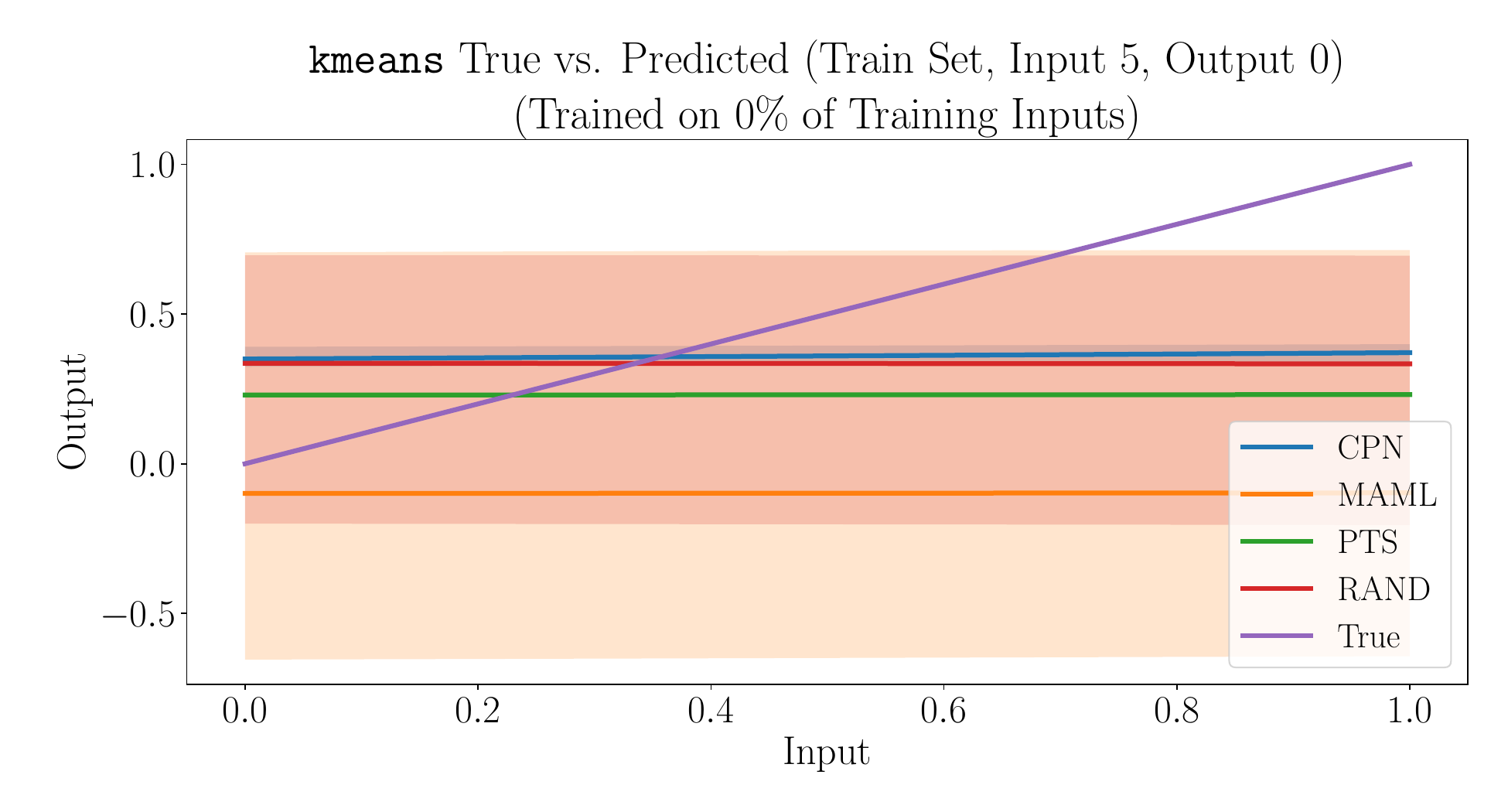}
\\
\vspace*{-1.6em}
\includegraphics[width=0.54\textwidth]{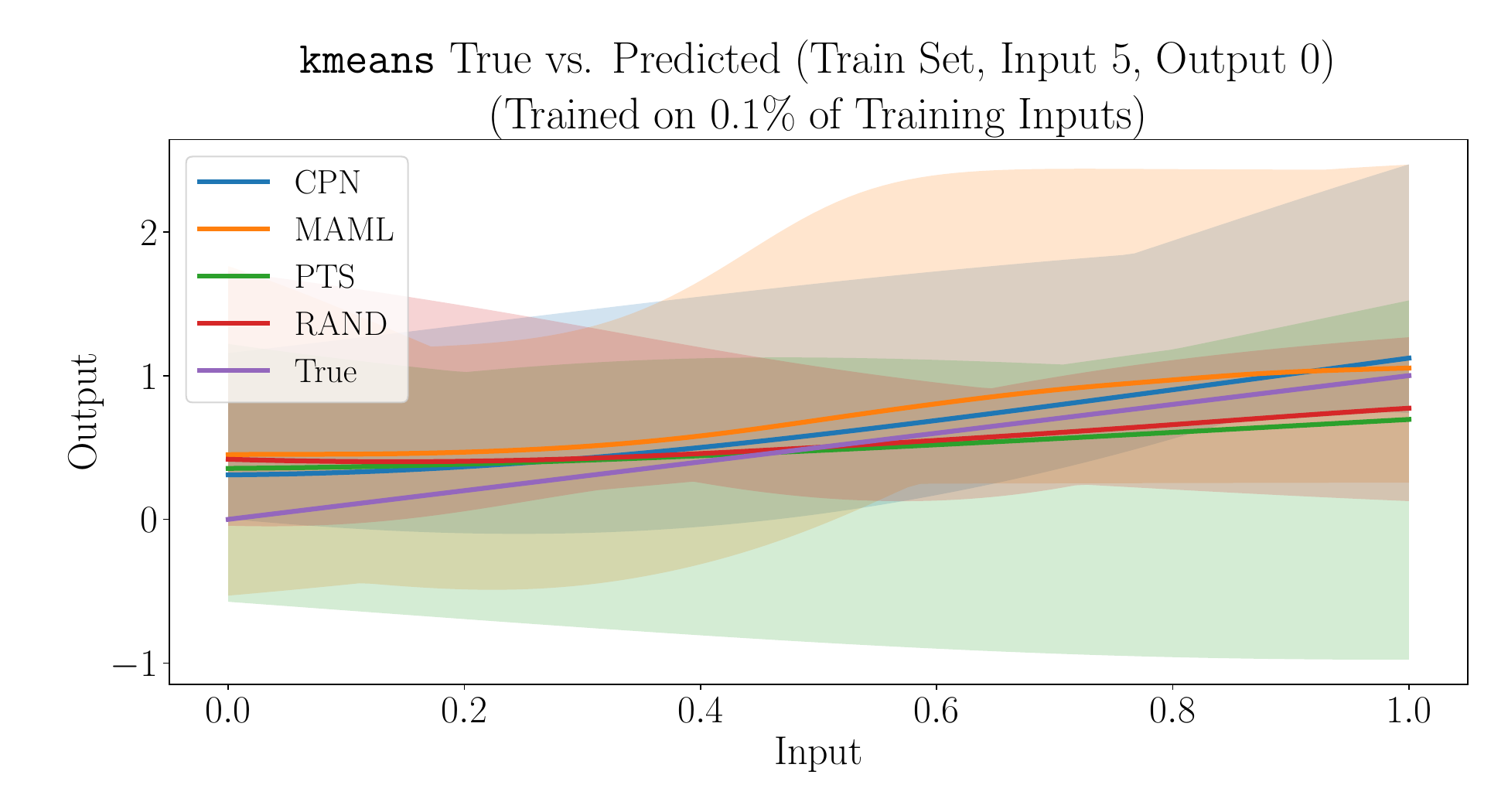}
\\
\vspace*{-1.6em}
\includegraphics[width=0.54\textwidth]{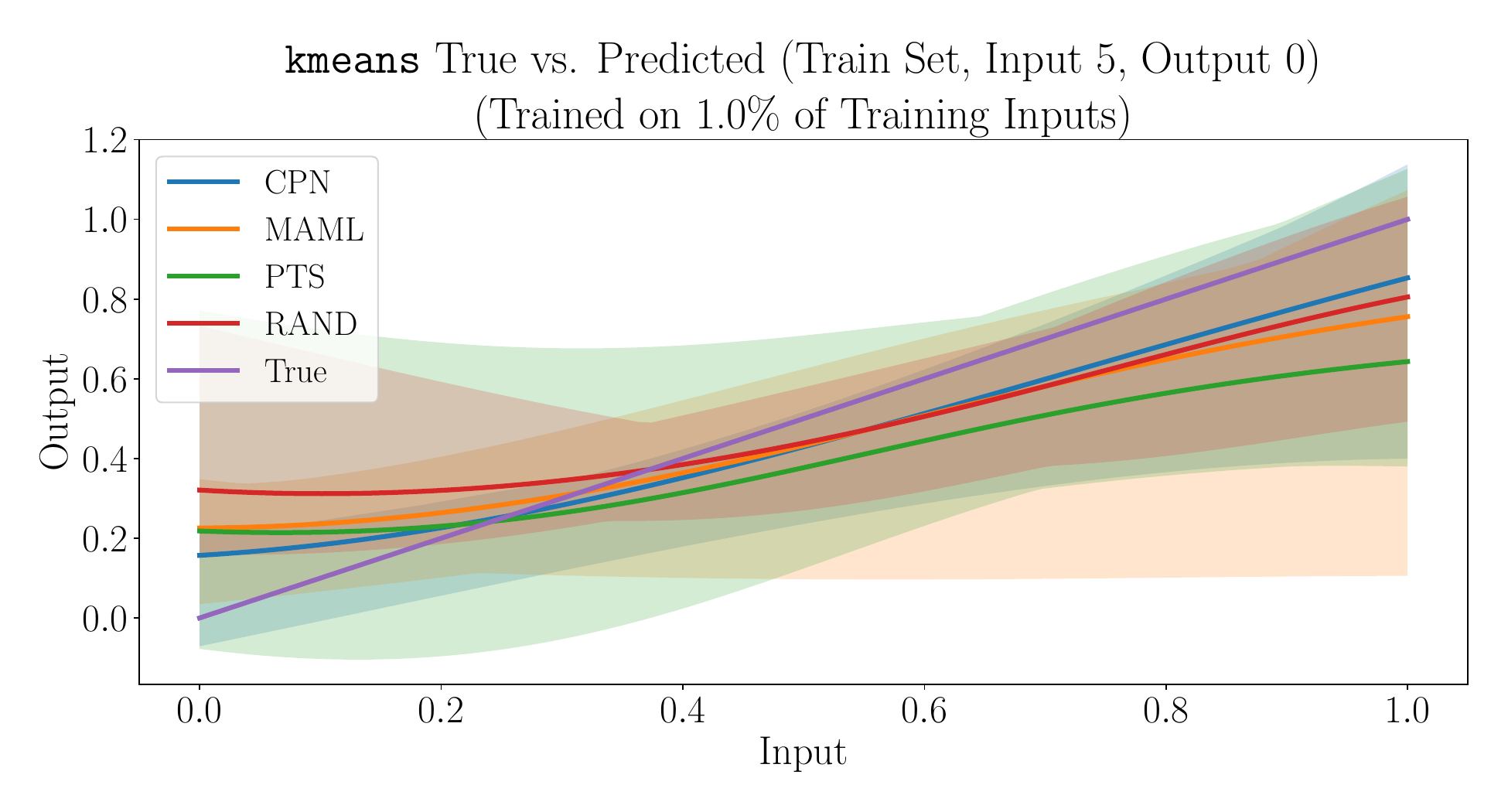}
\\
\vspace*{-1.6em}
\includegraphics[width=0.54\textwidth]{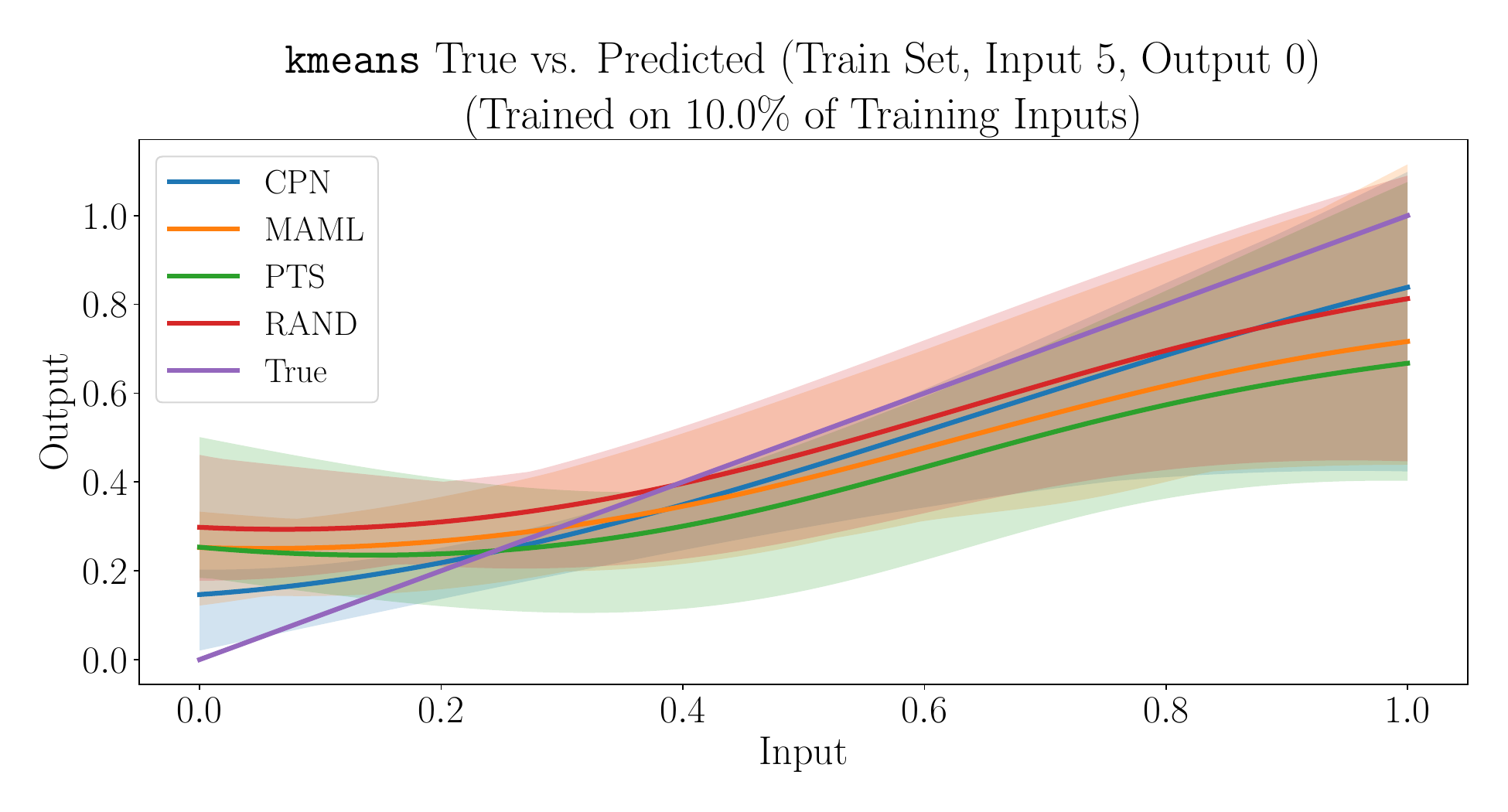}
\\
\vspace*{-1.6em}
\includegraphics[width=0.54\textwidth]{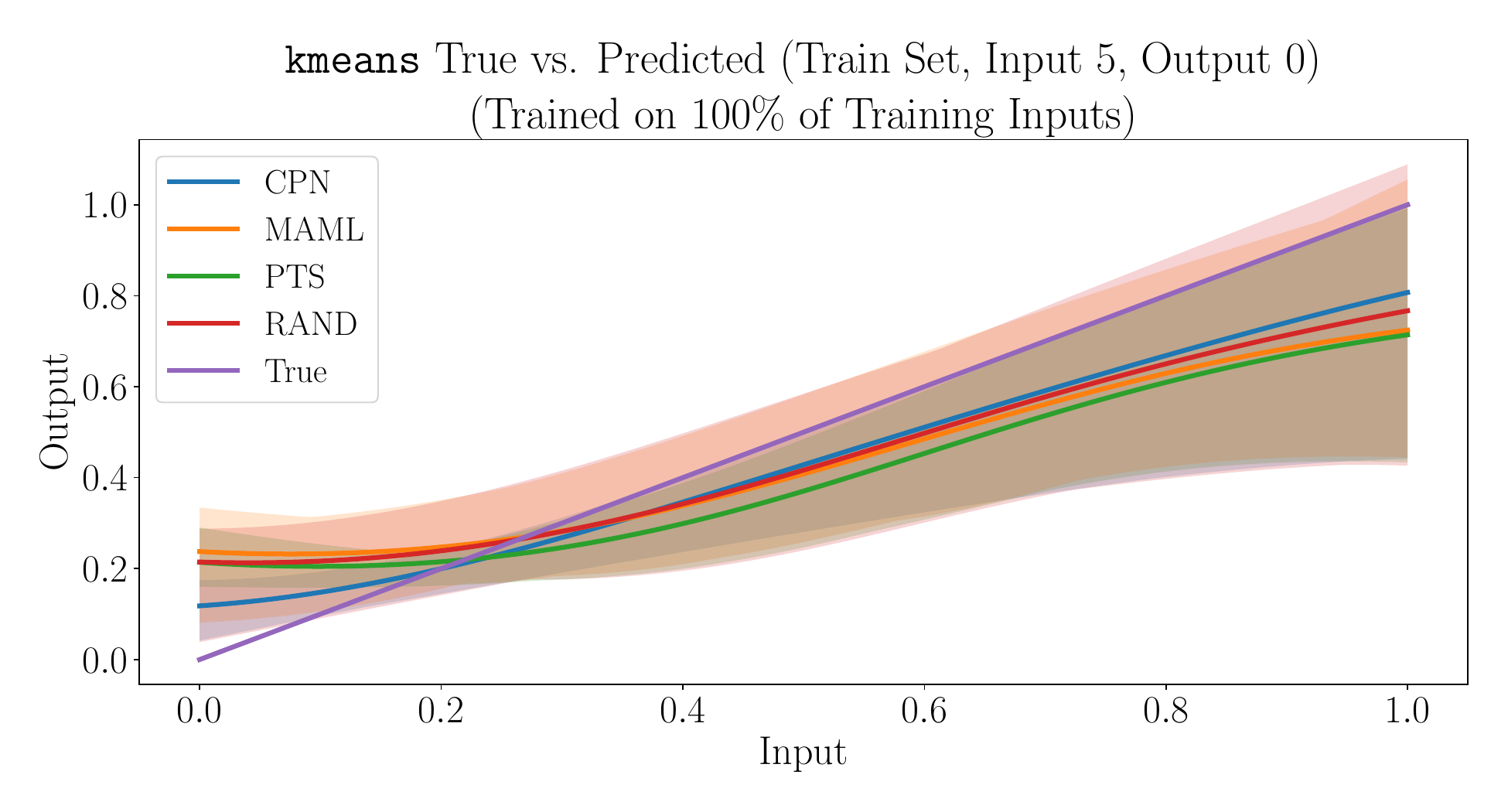}
\\
\vspace*{-1.2em}
\caption{
  Visual comparisons of the ground-truth \texttt{kmeans} function from \textsc{ParrotBenchCPN} and neural surrogate approximations thereof, when the sixth input is varied.
  We include results for all dataset sizes evaluated in Section~\ref{sec:data_efficiency}.
}\label{fig:true_vs_pred_kmeans_input_5}
\end{figure*}

\begin{figure*}
\centering
\vspace*{-0.7em}
\includegraphics[width=0.54\textwidth]{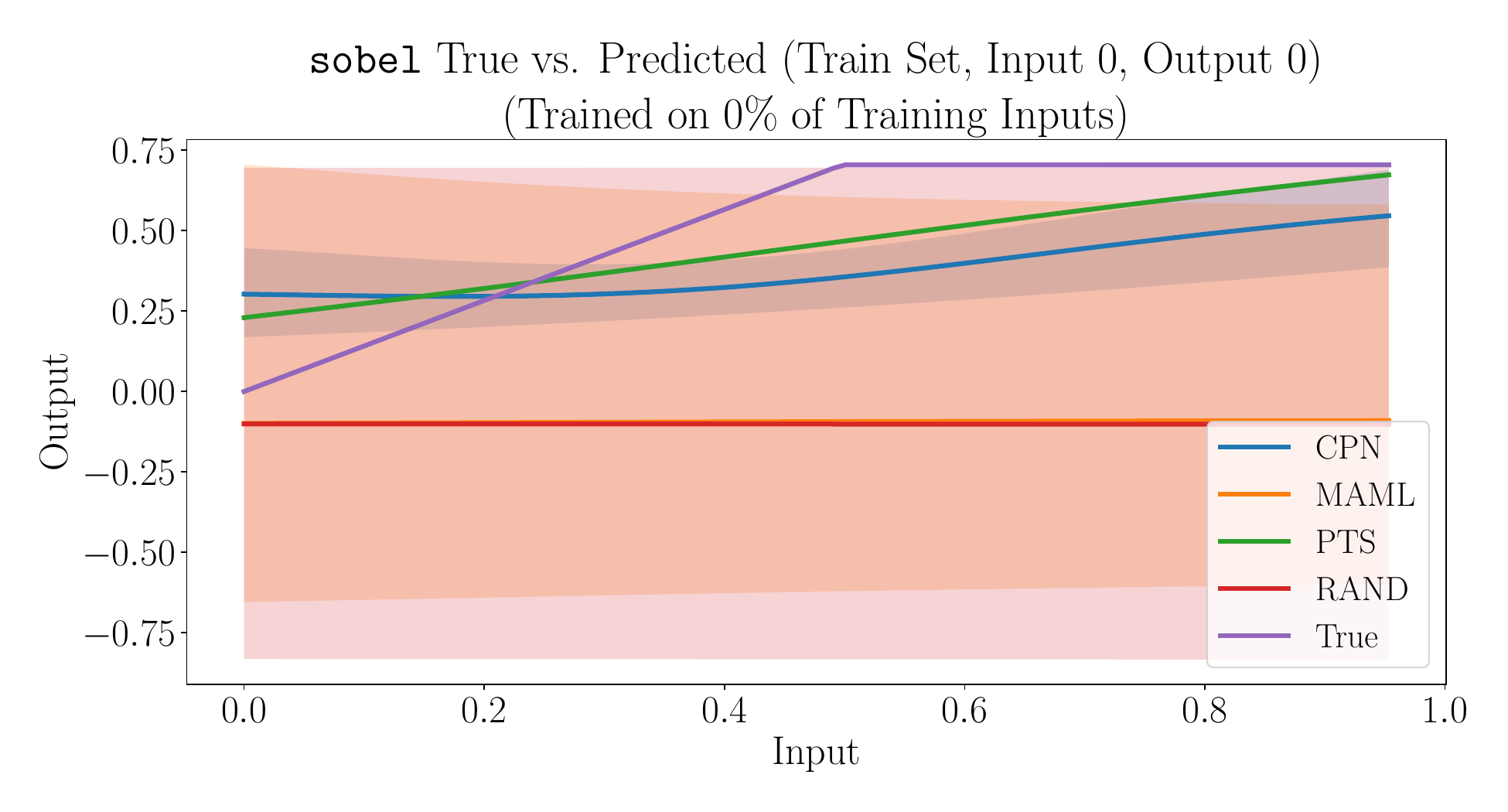}
\\
\vspace*{-1.6em}
\includegraphics[width=0.54\textwidth]{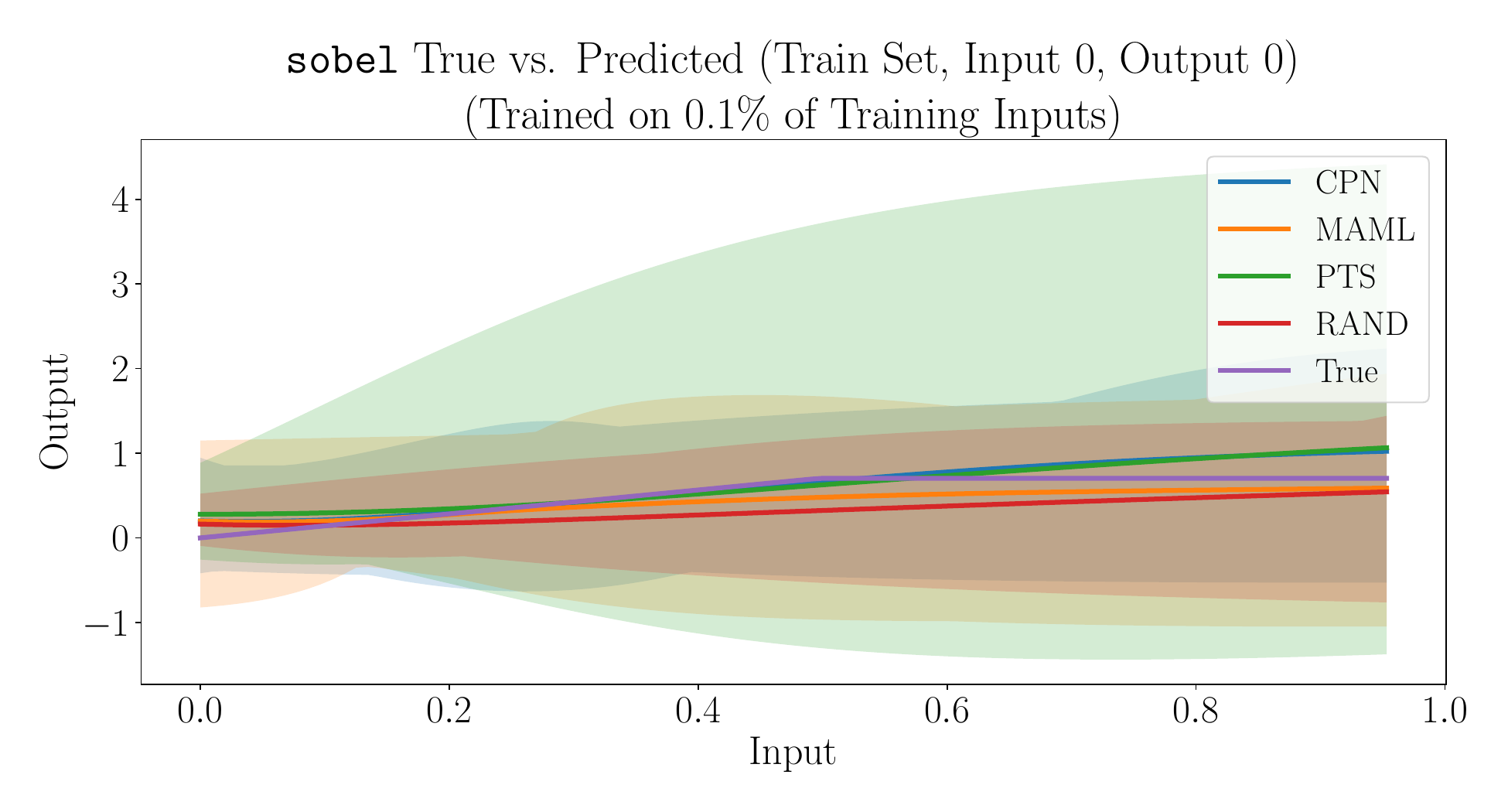}
\\
\vspace*{-1.6em}
\includegraphics[width=0.54\textwidth]{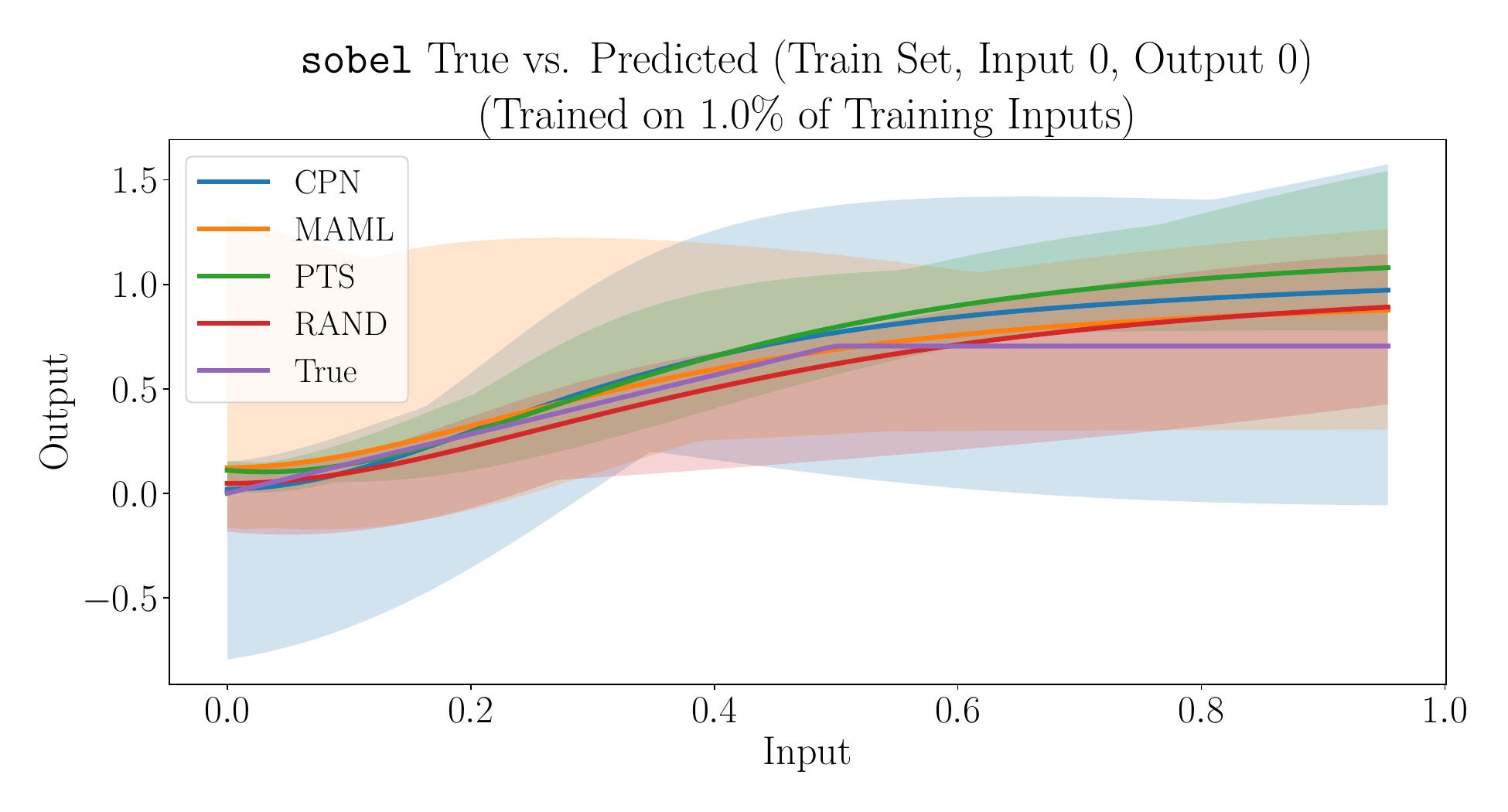}
\\
\vspace*{-1.6em}
\includegraphics[width=0.54\textwidth]{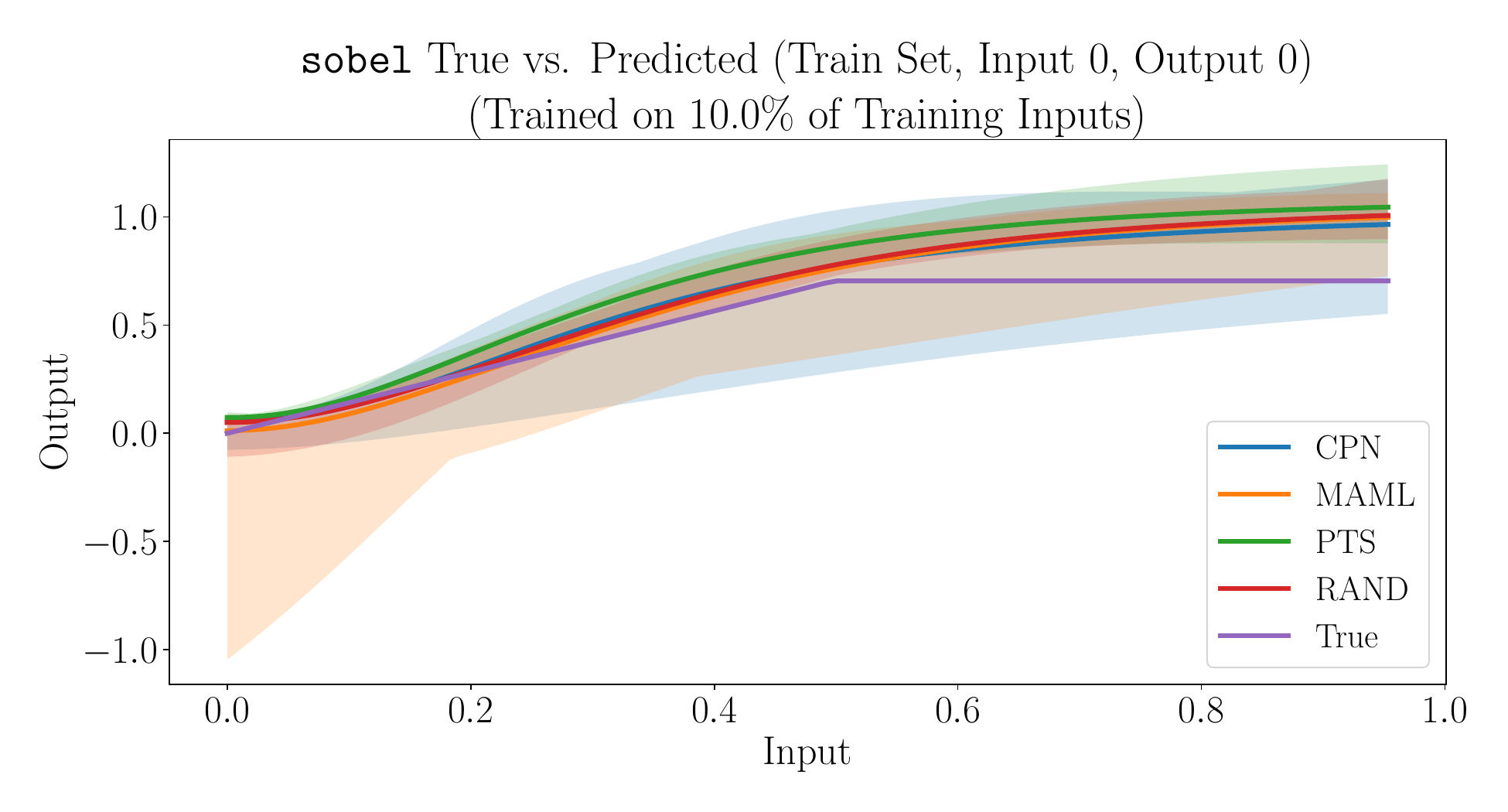}
\\
\vspace*{-1.6em}
\includegraphics[width=0.54\textwidth]{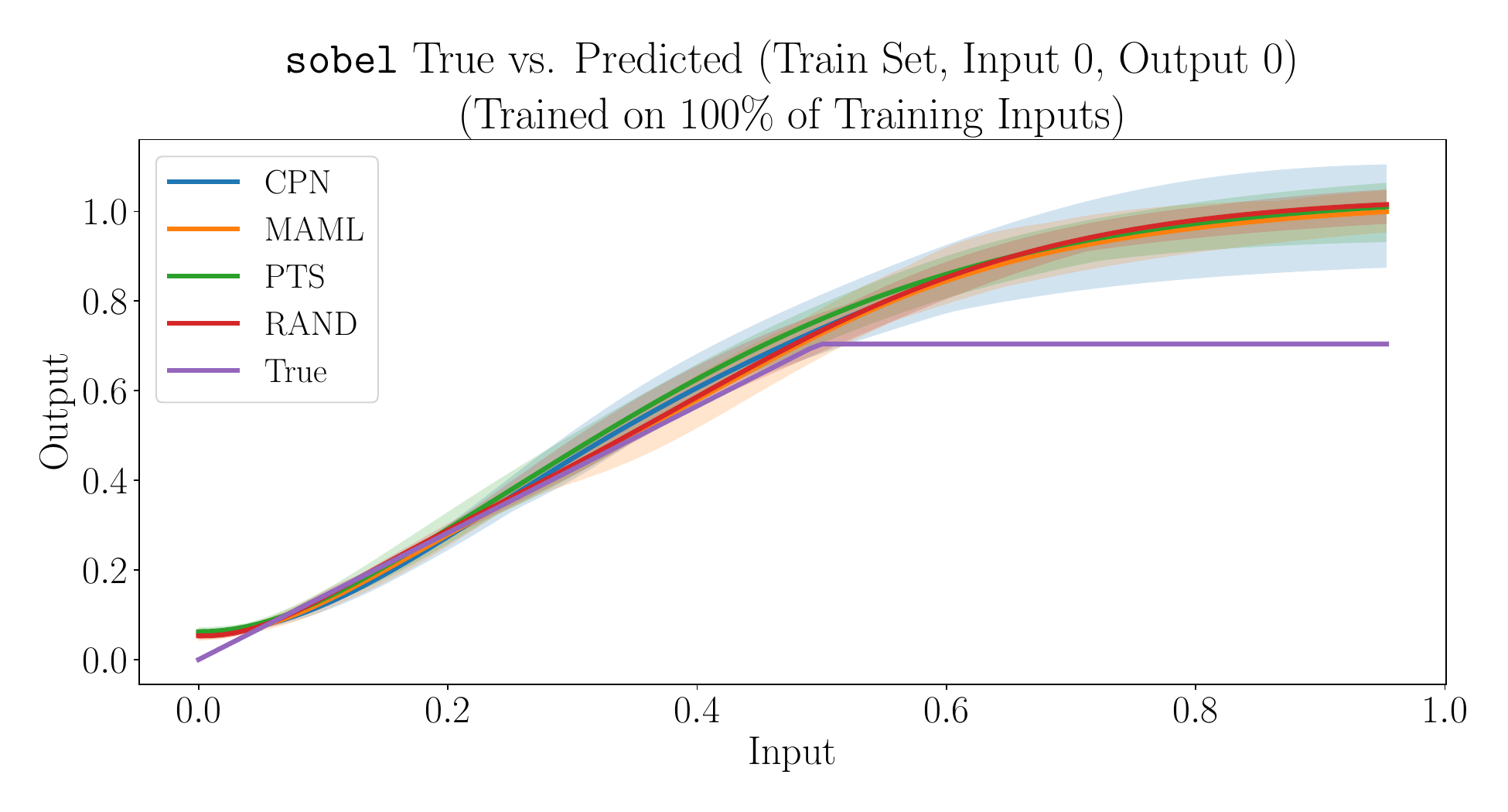}
\\
\vspace*{-1.2em}
\caption{
  Visual comparisons of the ground-truth \texttt{sobel} function from \textsc{ParrotBenchCPN} and neural surrogate approximations thereof, when the first input is varied.
  We include results for all dataset sizes evaluated in Section~\ref{sec:data_efficiency}.
}\label{fig:true_vs_pred_sobel_input_0}
\end{figure*}

\begin{figure*}
\centering
\vspace*{-0.7em}
\includegraphics[width=0.54\textwidth]{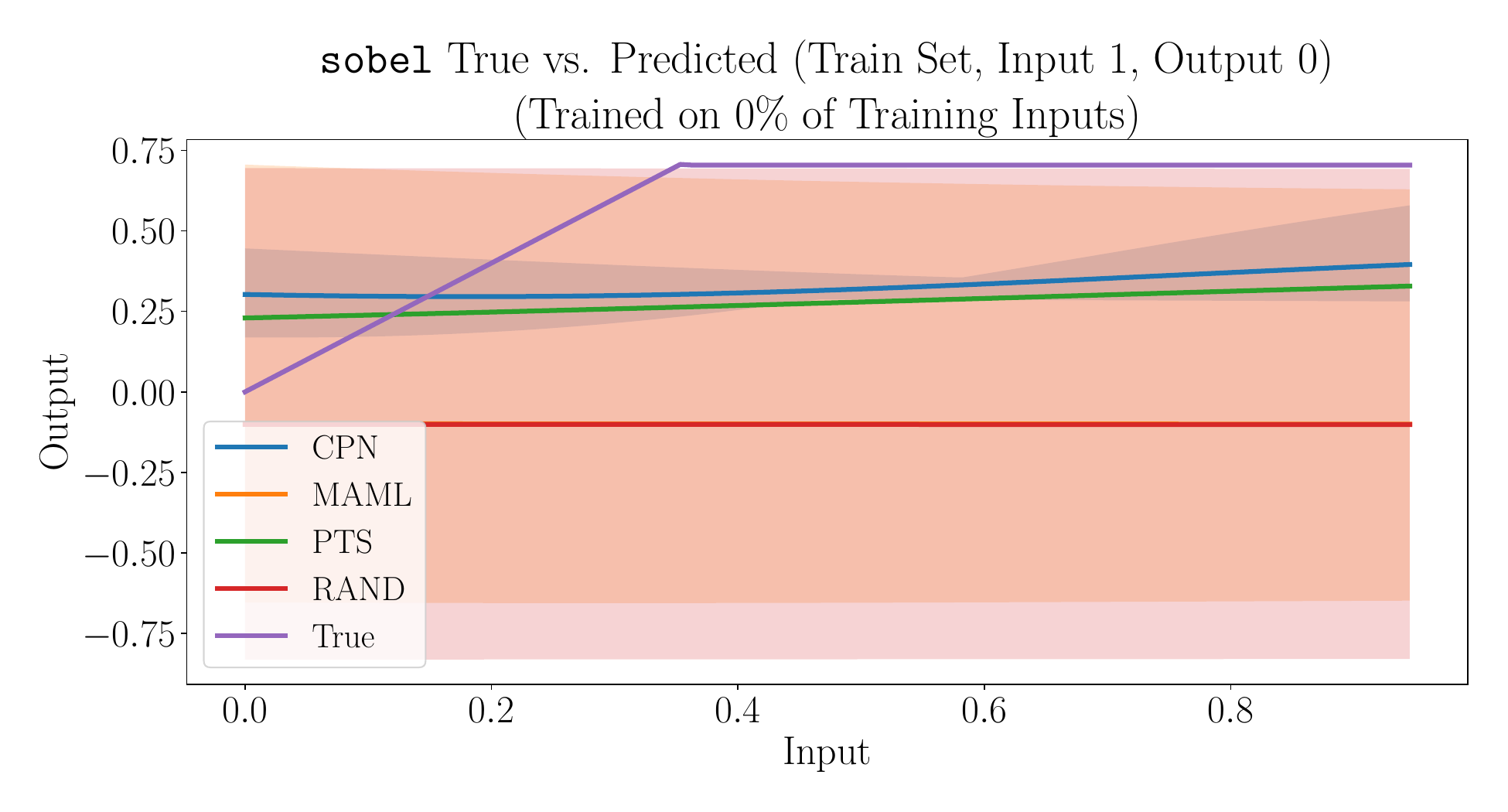}
\\
\vspace*{-1.6em}
\includegraphics[width=0.54\textwidth]{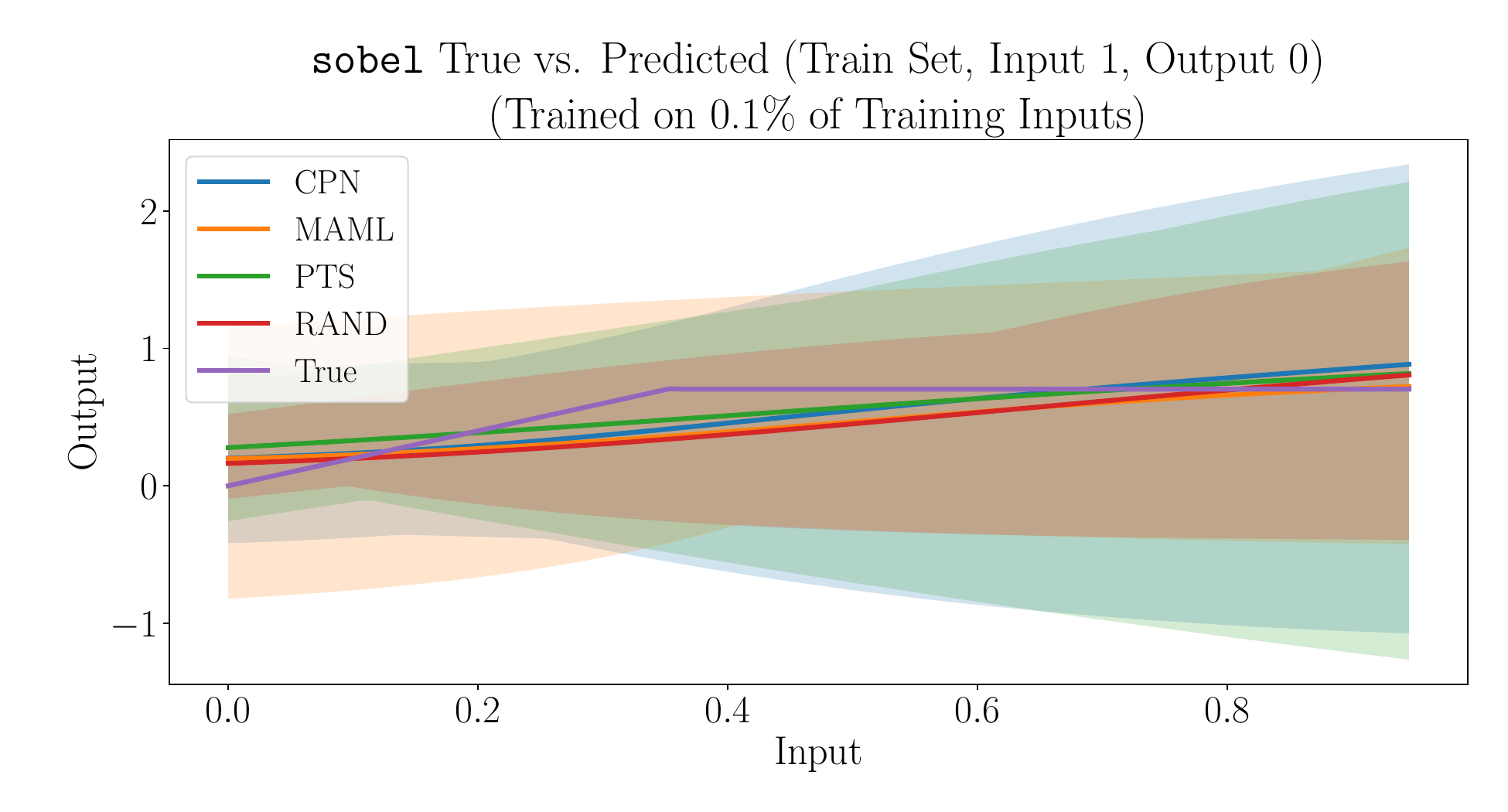}
\\
\vspace*{-1.6em}
\includegraphics[width=0.54\textwidth]{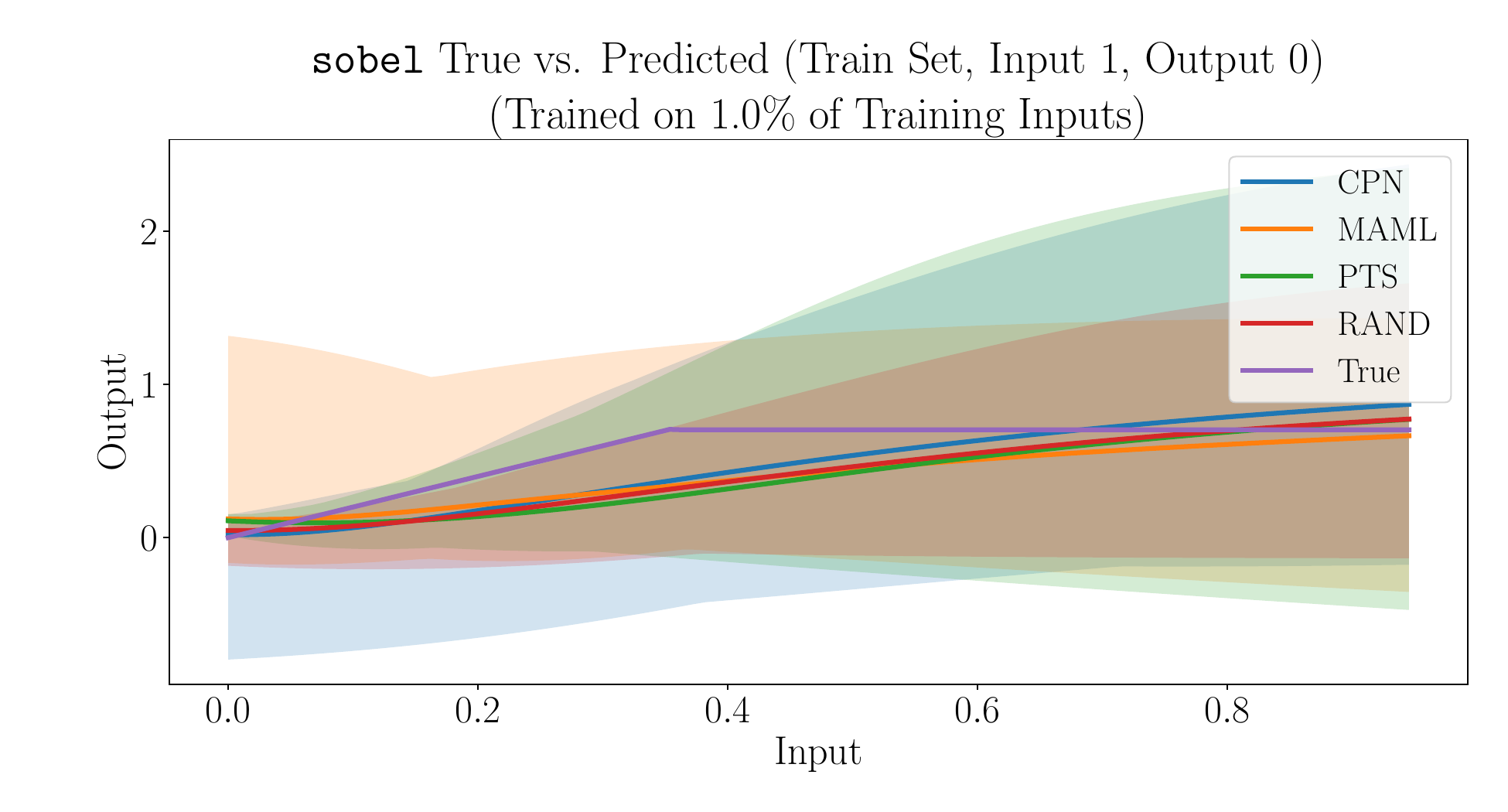}
\\
\vspace*{-1.6em}
\includegraphics[width=0.54\textwidth]{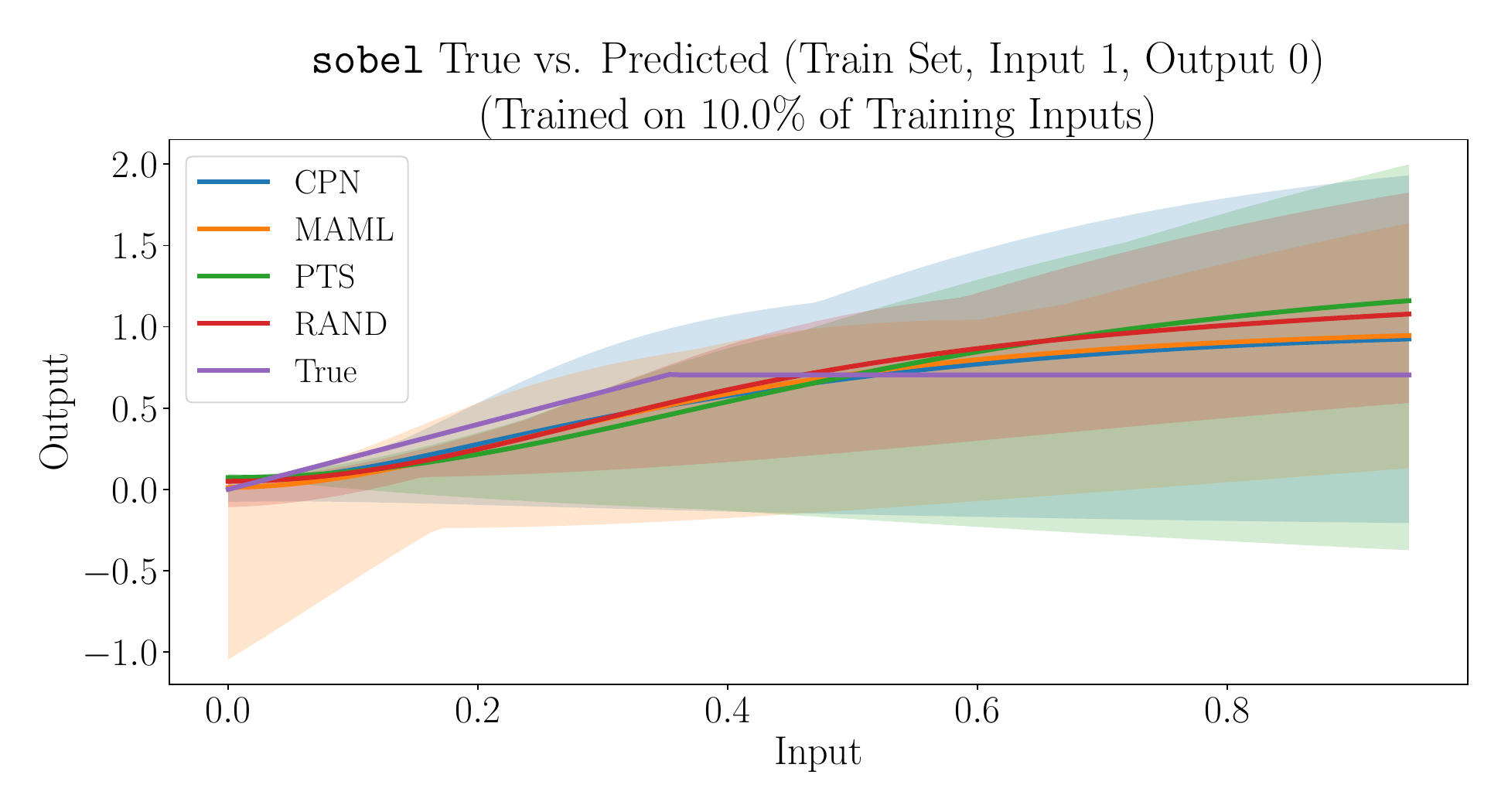}
\\
\vspace*{-1.6em}
\includegraphics[width=0.54\textwidth]{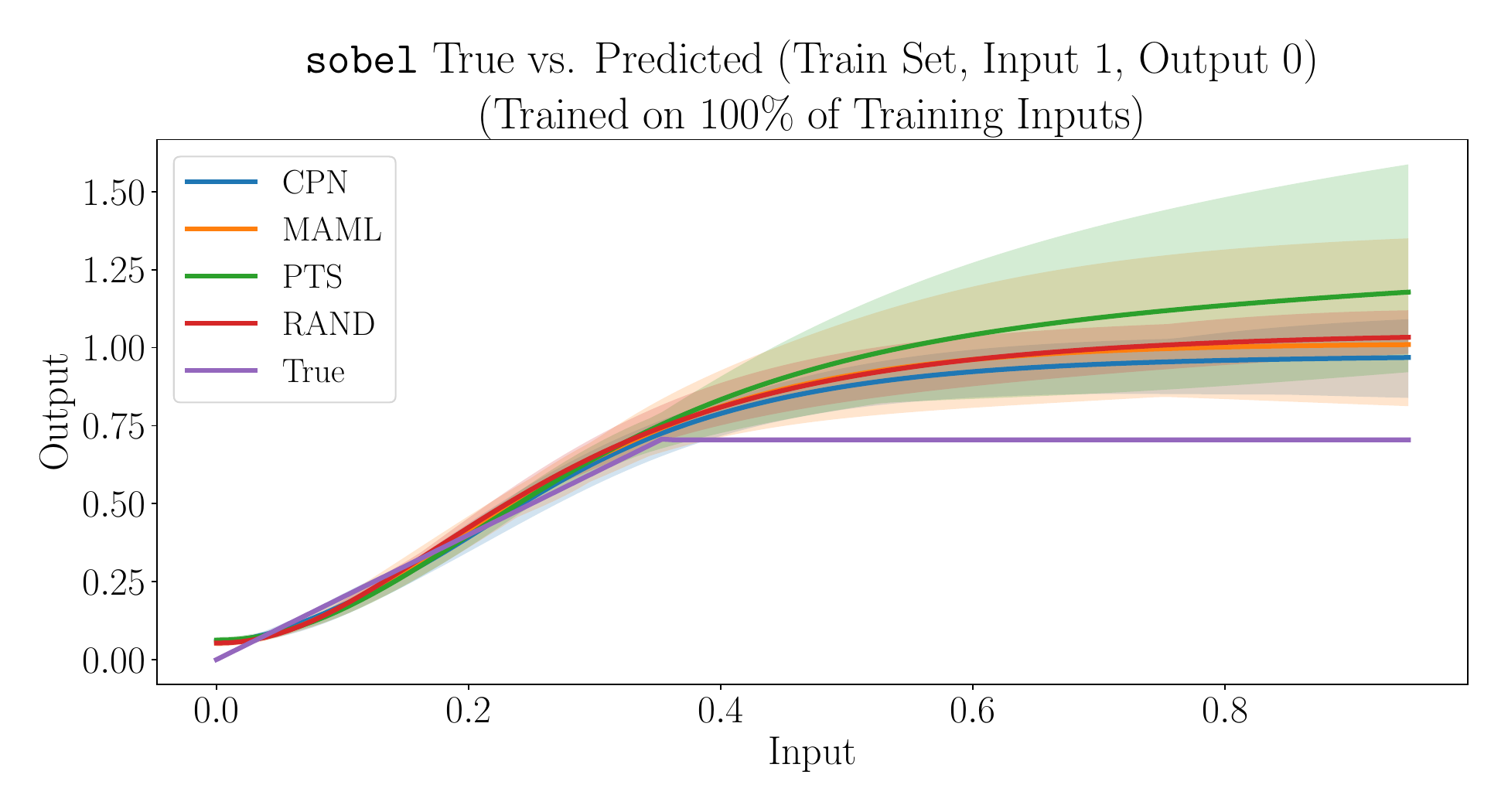}
\\
\vspace*{-1.2em}
\caption{
  Visual comparisons of the ground-truth \texttt{sobel} function from \textsc{ParrotBenchCPN} and neural surrogate approximations thereof, when the second input is varied.
  We include results for all dataset sizes evaluated in Section~\ref{sec:data_efficiency}.
}\label{fig:true_vs_pred_sobel_input_1}
\end{figure*}

\begin{figure*}
\centering
\vspace*{-0.7em}
\includegraphics[width=0.54\textwidth]{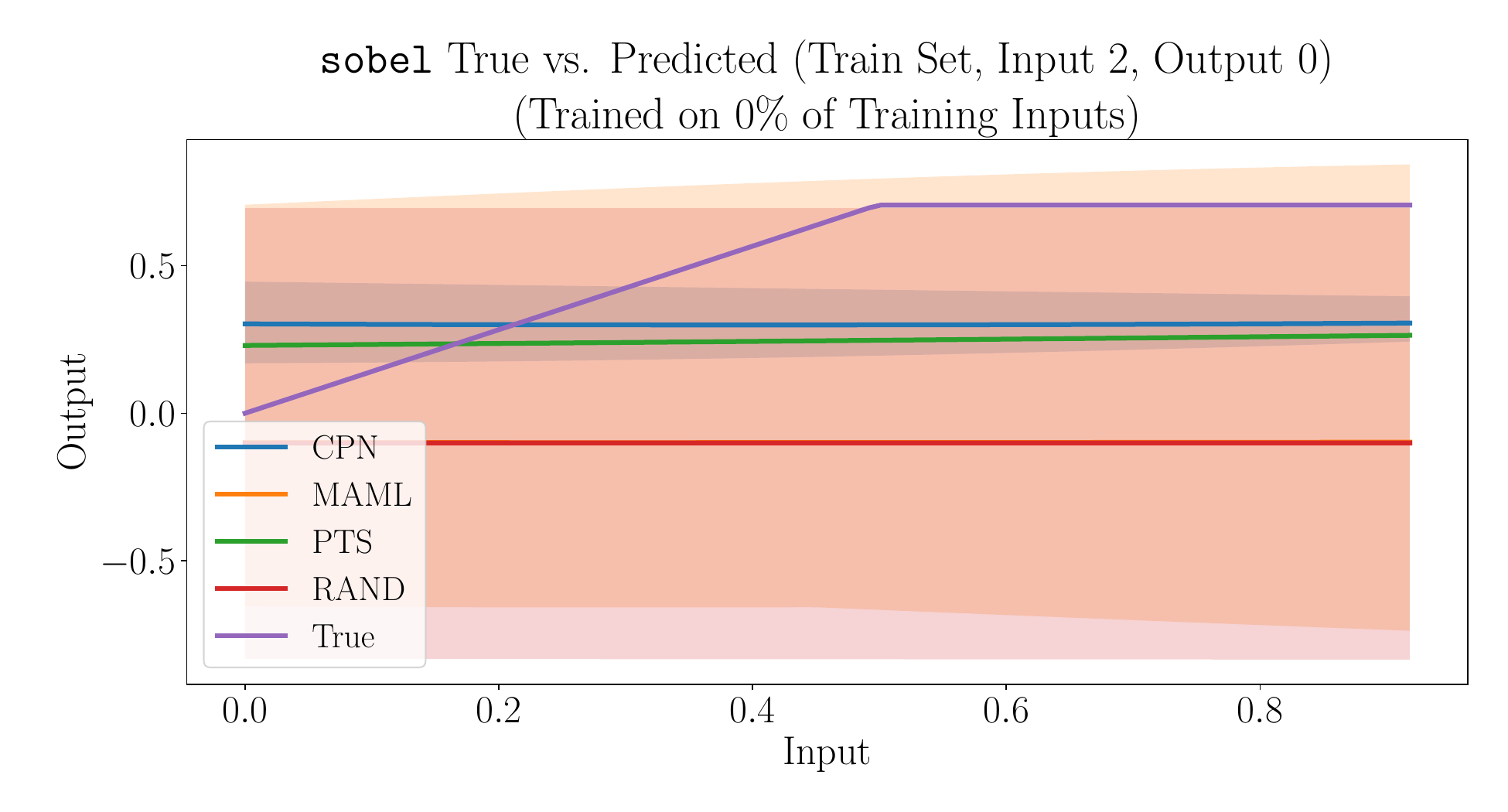}
\\
\vspace*{-1.6em}
\includegraphics[width=0.54\textwidth]{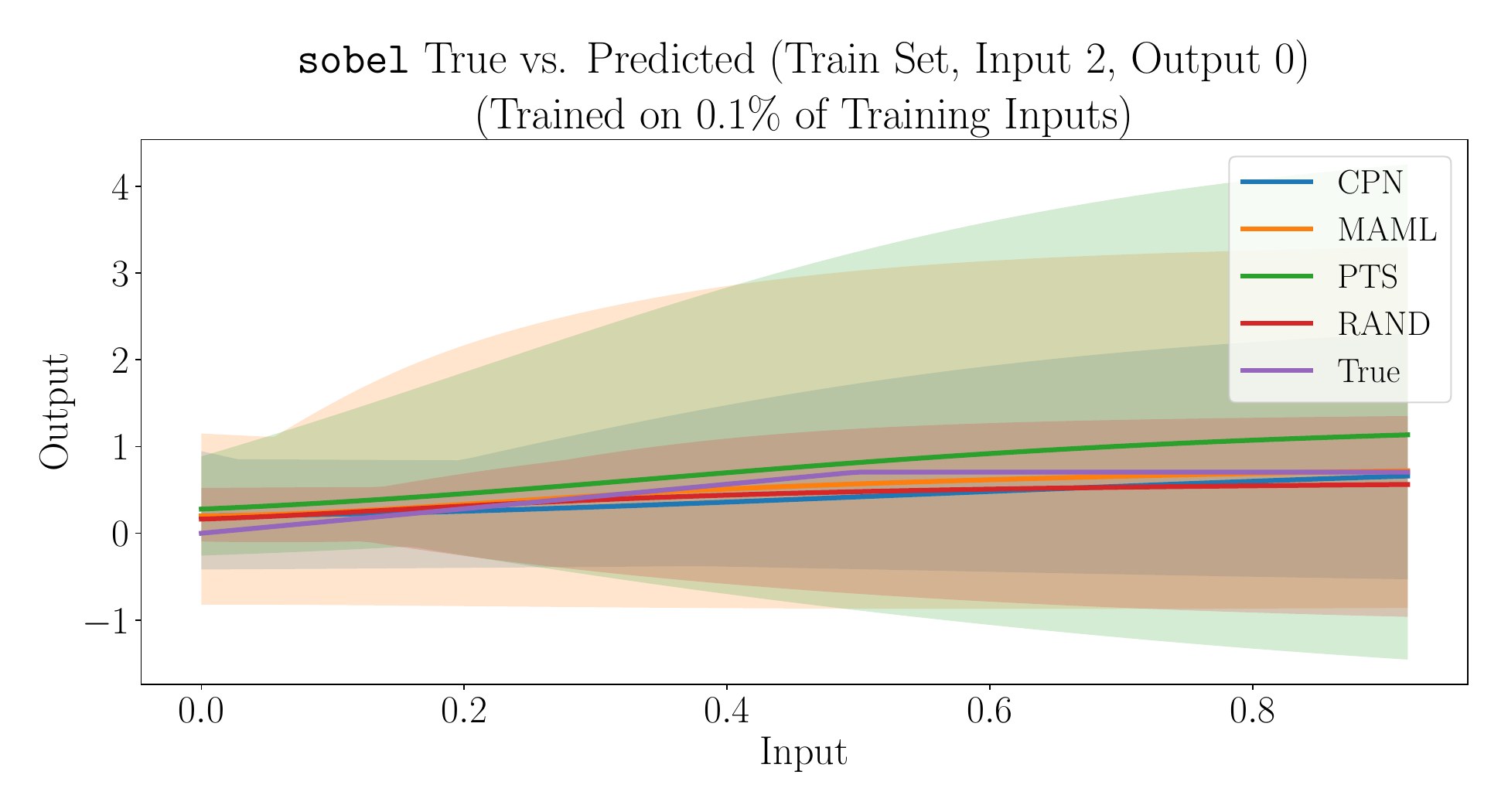}
\\
\vspace*{-1.6em}
\includegraphics[width=0.54\textwidth]{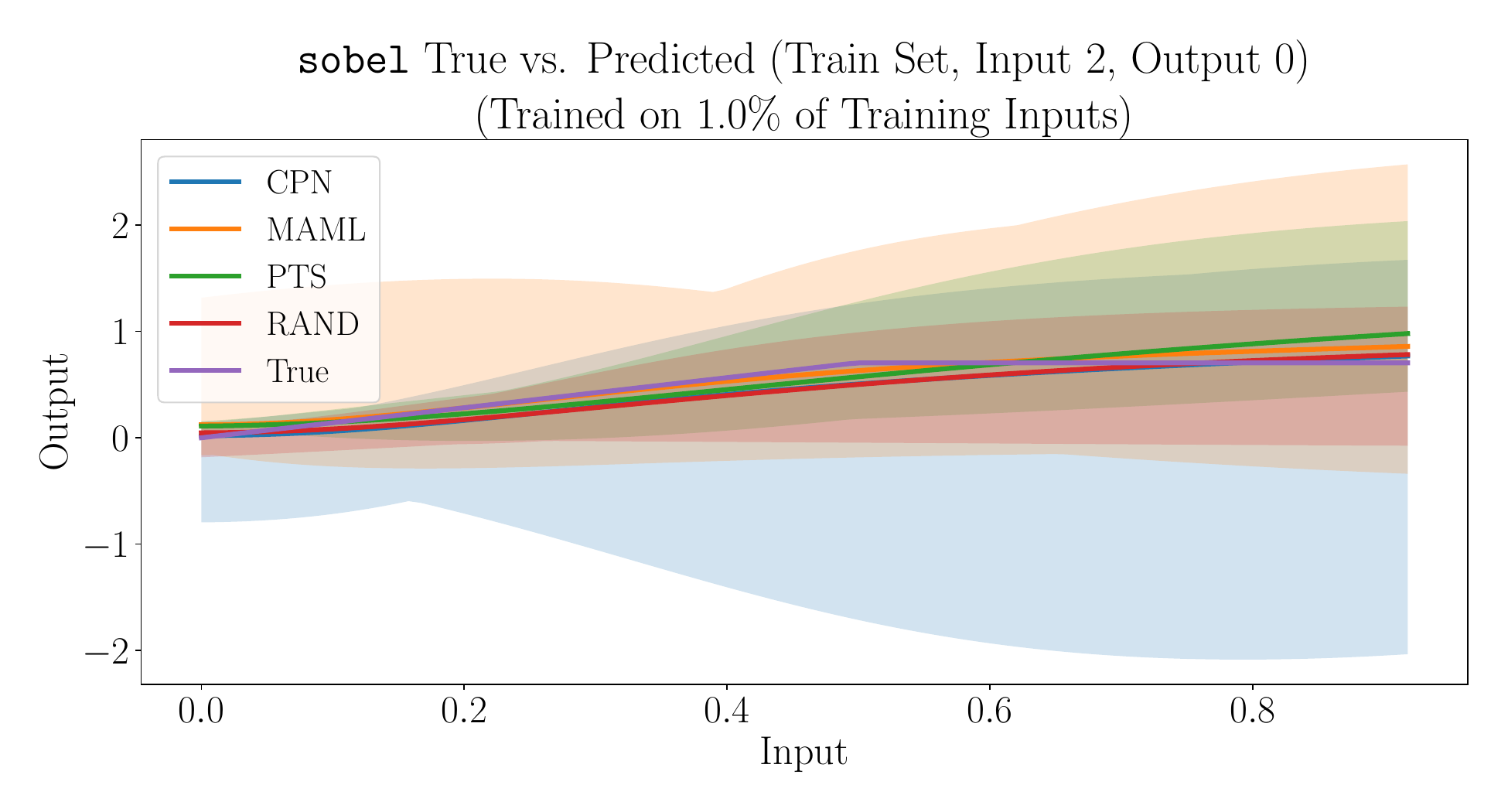}
\\
\vspace*{-1.6em}
\includegraphics[width=0.54\textwidth]{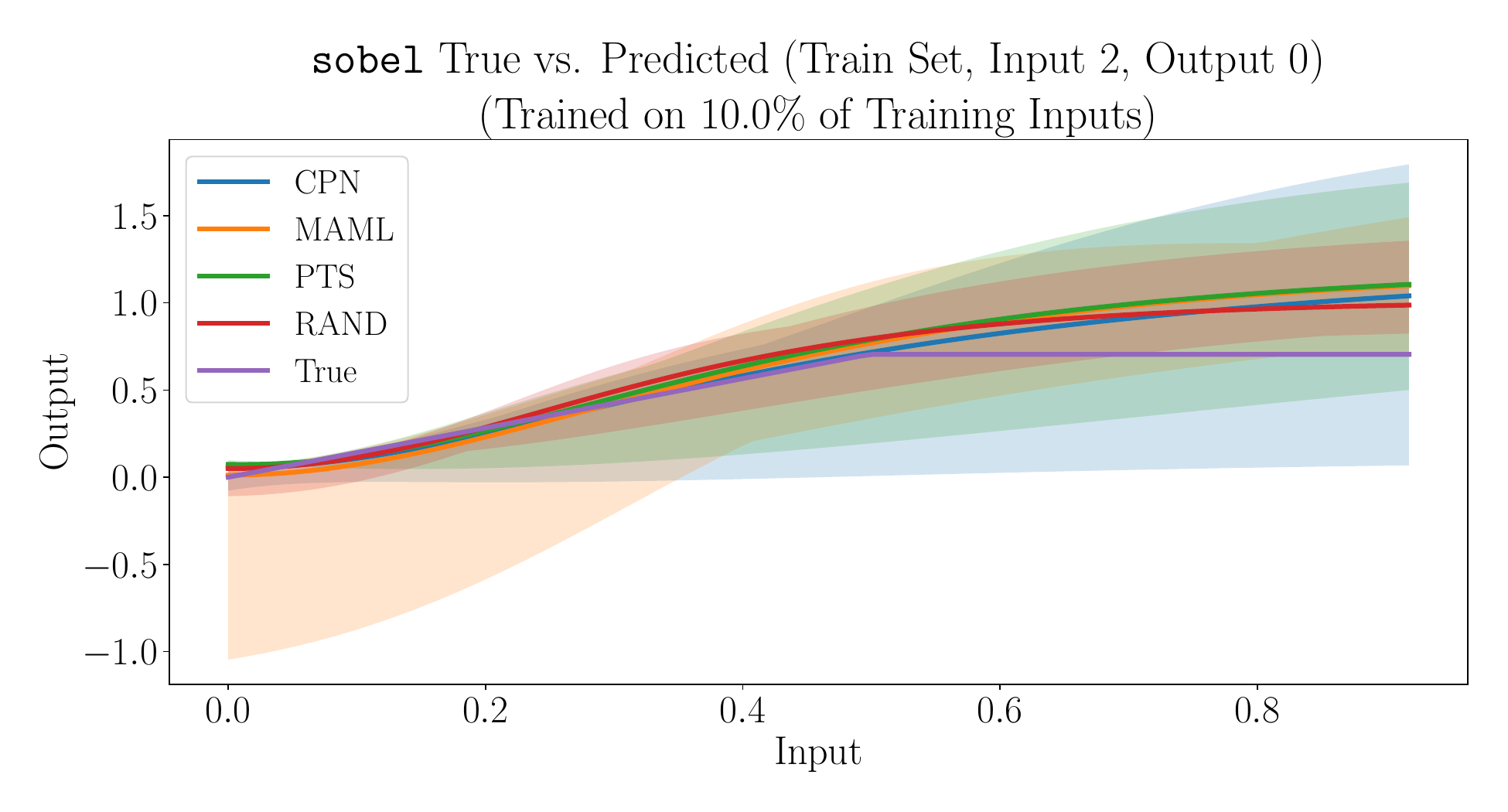}
\\
\vspace*{-1.6em}
\includegraphics[width=0.54\textwidth]{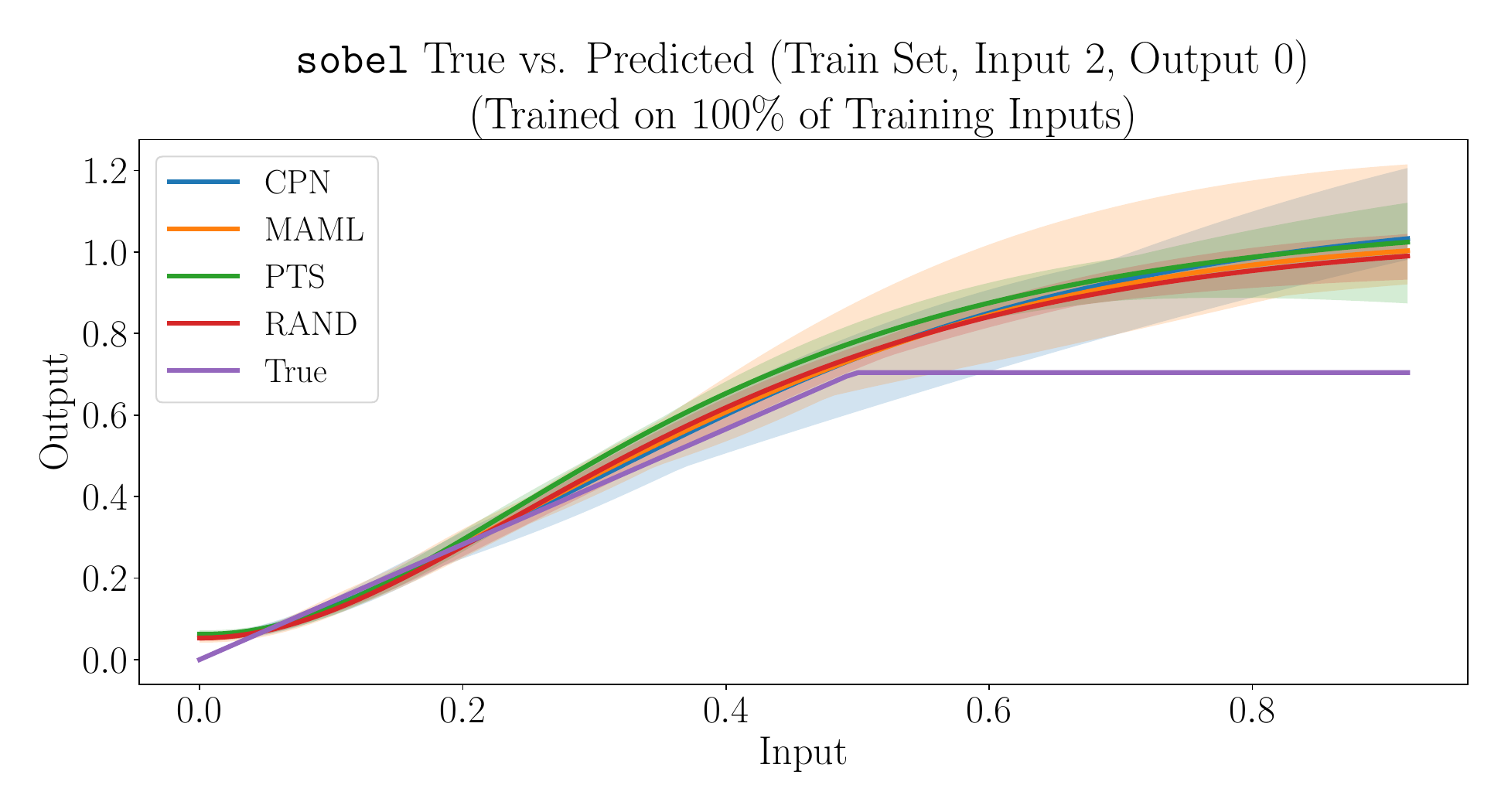}
\\
\vspace*{-1.2em}
\caption{
  Visual comparisons of the ground-truth \texttt{sobel} function from \textsc{ParrotBenchCPN} and neural surrogate approximations thereof, when the third input is varied.
  We include results for all dataset sizes evaluated in Section~\ref{sec:data_efficiency}.
}\label{fig:true_vs_pred_sobel_input_2}
\end{figure*}

\begin{figure*}
\centering
\vspace*{-0.7em}
\includegraphics[width=0.54\textwidth]{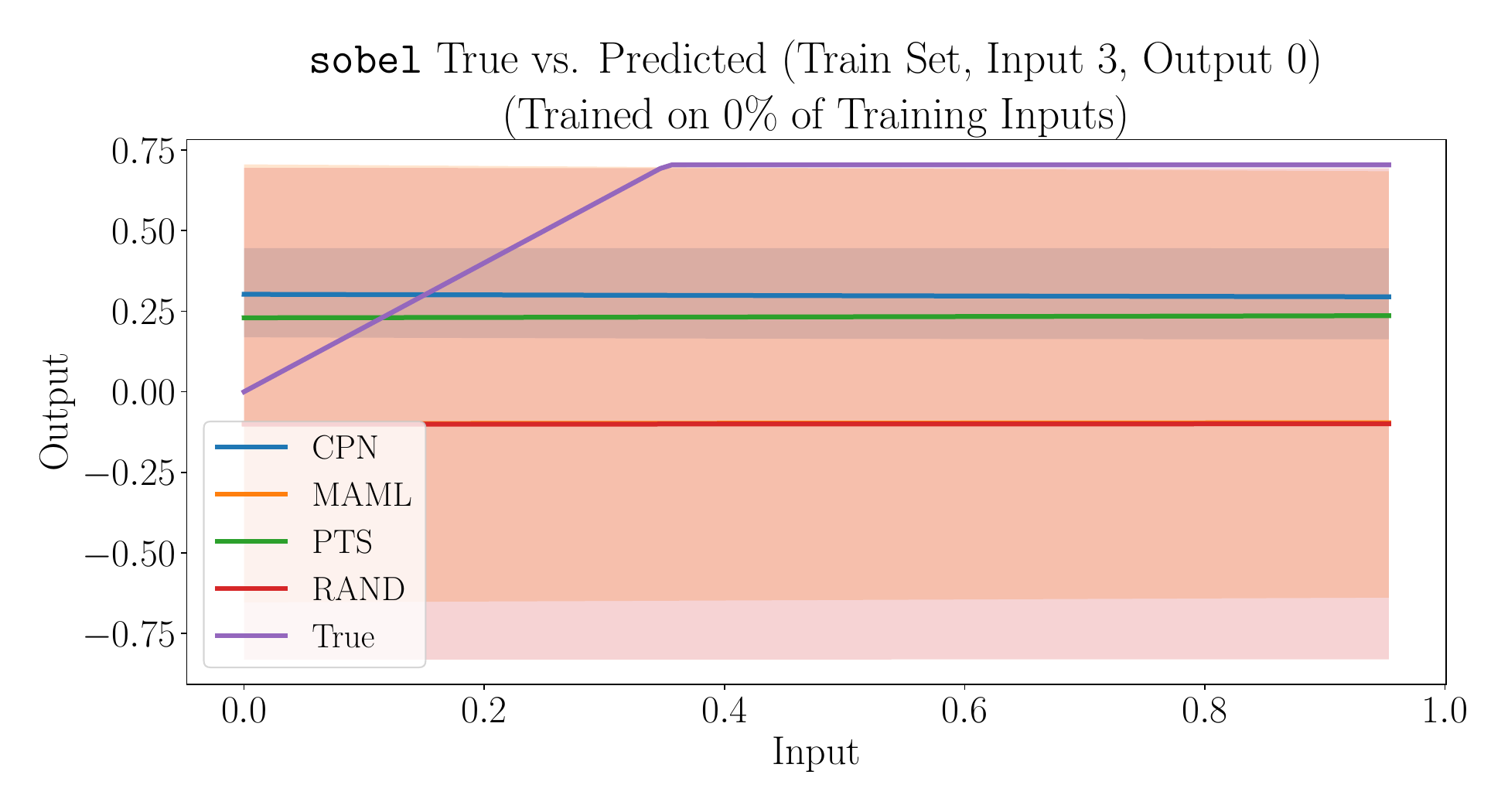}
\\
\vspace*{-1.6em}
\includegraphics[width=0.54\textwidth]{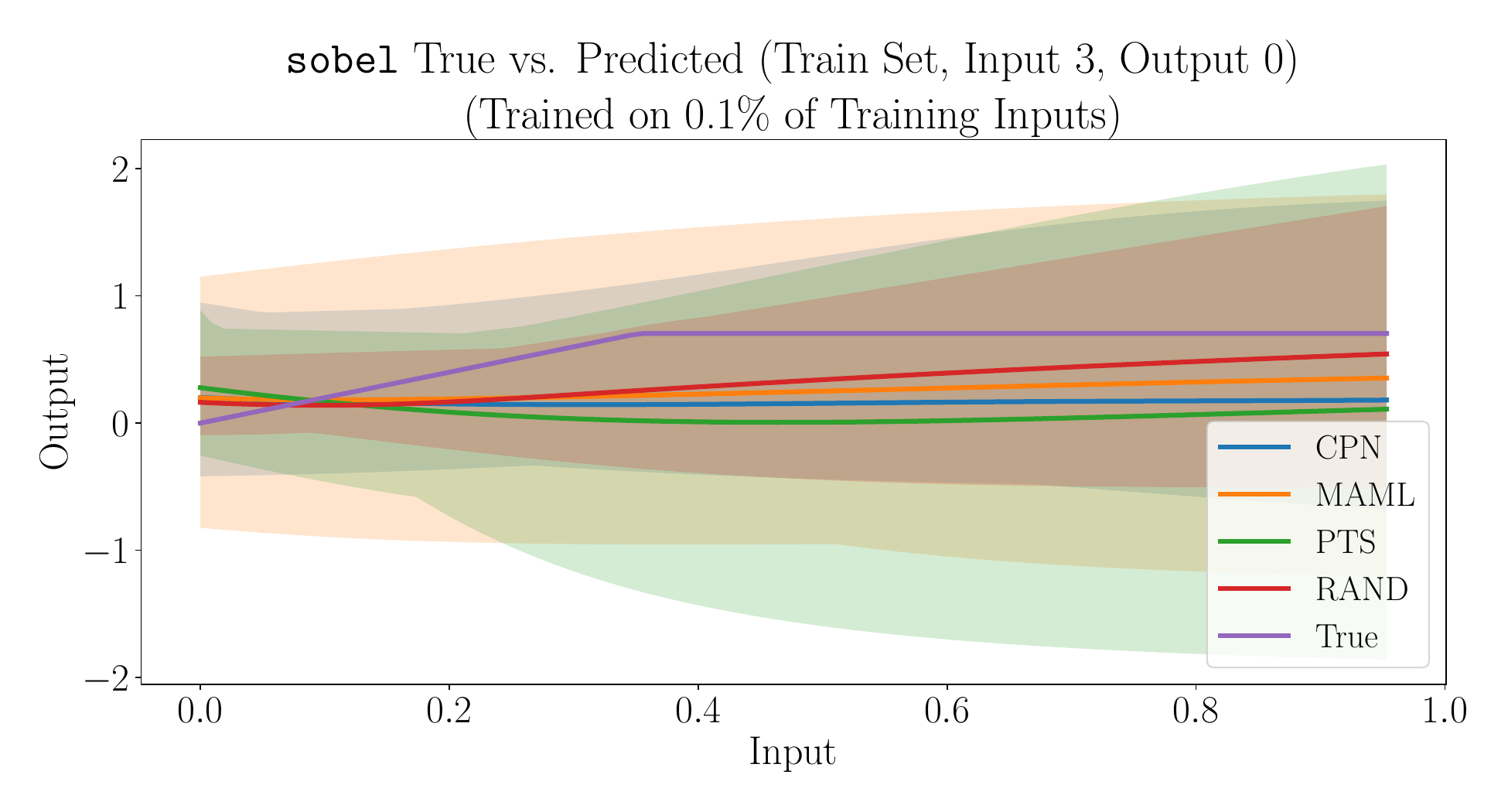}
\\
\vspace*{-1.6em}
\includegraphics[width=0.54\textwidth]{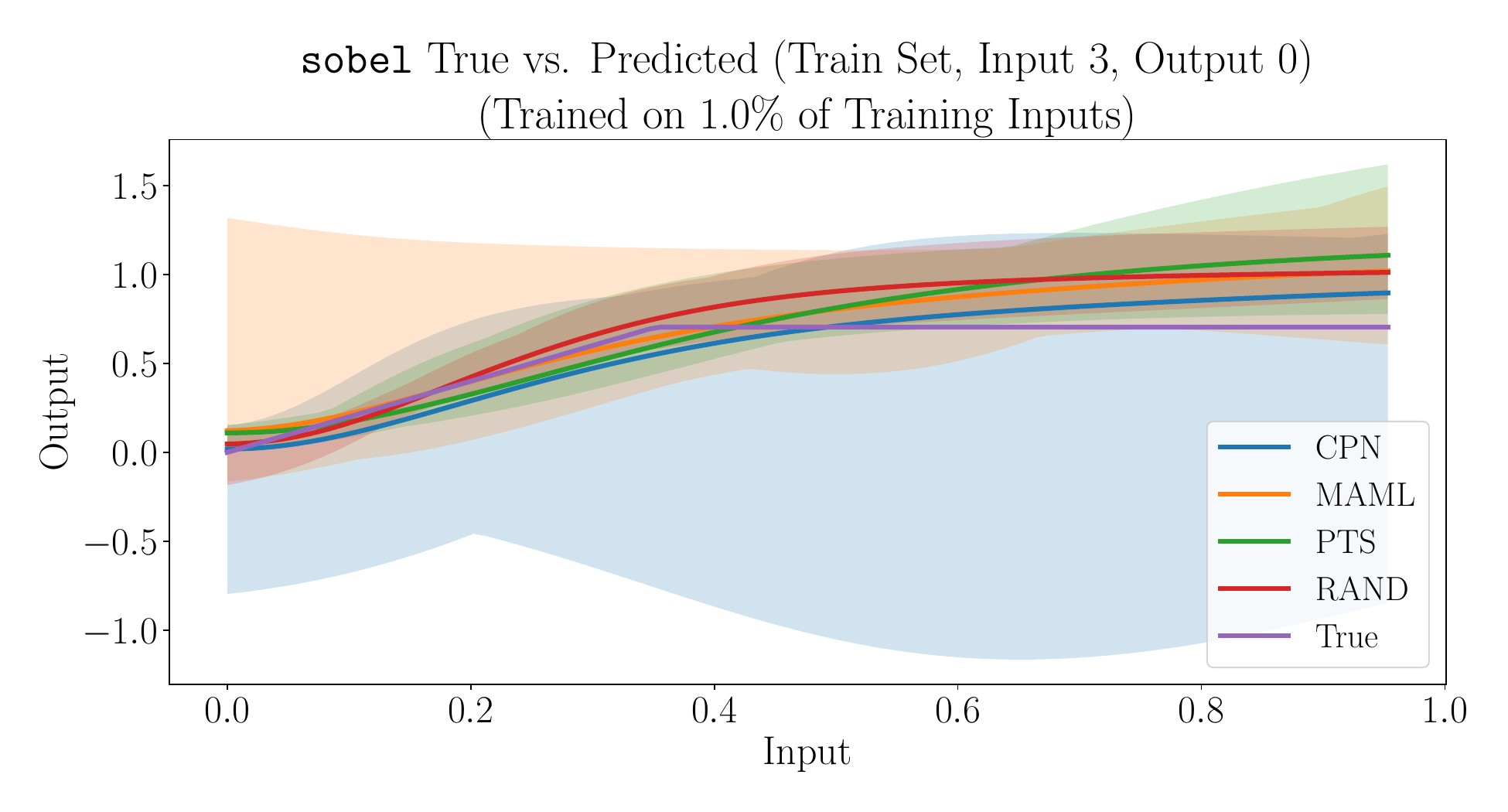}
\\
\vspace*{-1.6em}
\includegraphics[width=0.54\textwidth]{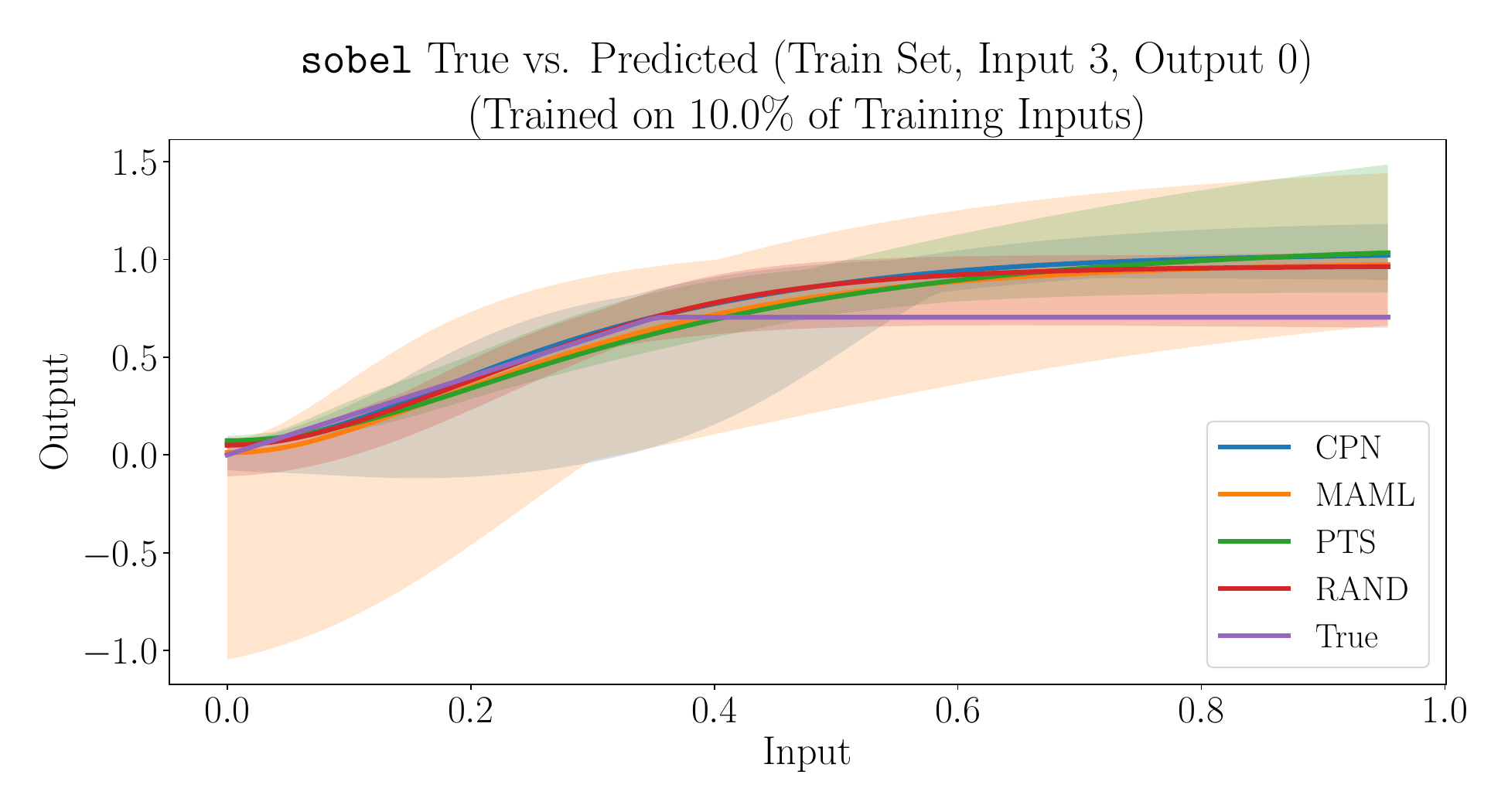}
\\
\vspace*{-1.6em}
\includegraphics[width=0.54\textwidth]{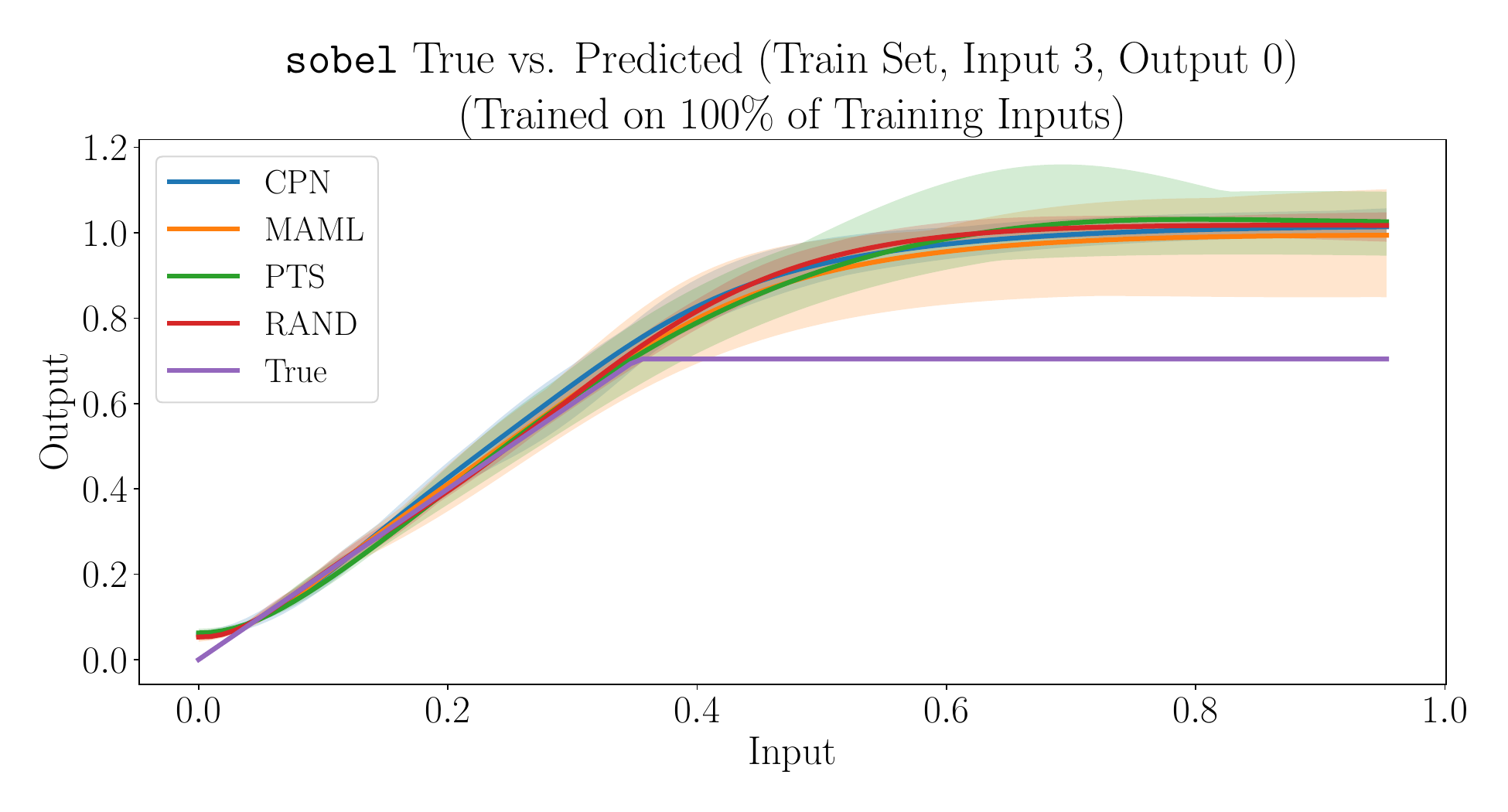}
\\
\vspace*{-1.2em}
\caption{
  Visual comparisons of the ground-truth \texttt{sobel} function from \textsc{ParrotBenchCPN} and neural surrogate approximations thereof, when the fourth input is varied.
  We include results for all dataset sizes evaluated in Section~\ref{sec:data_efficiency}.
}\label{fig:true_vs_pred_sobel_input_3}
\end{figure*}

\begin{figure*}
\centering
\vspace*{-0.7em}
\includegraphics[width=0.54\textwidth]{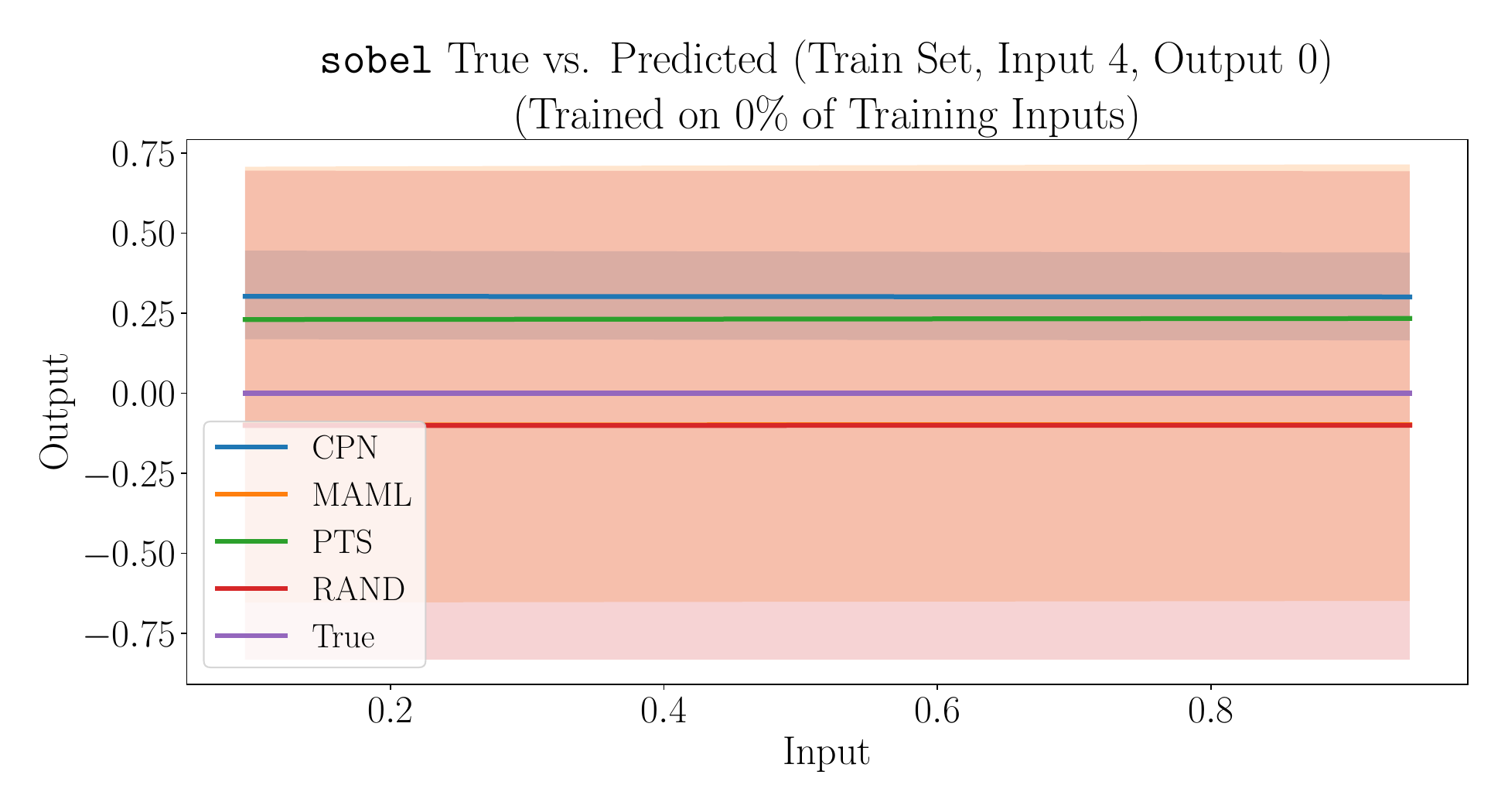}
\\
\vspace*{-1.6em}
\includegraphics[width=0.54\textwidth]{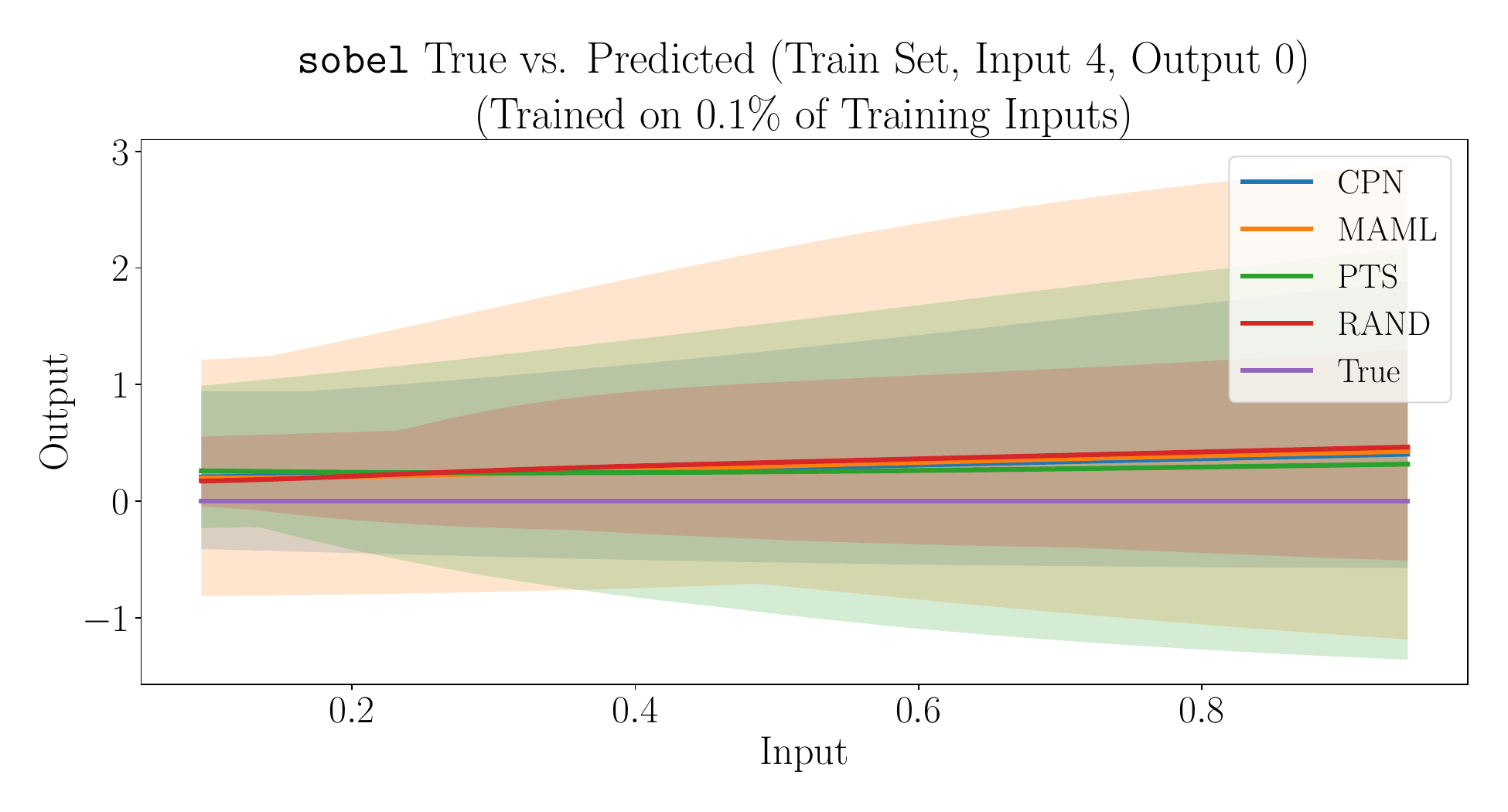}
\\
\vspace*{-1.6em}
\includegraphics[width=0.54\textwidth]{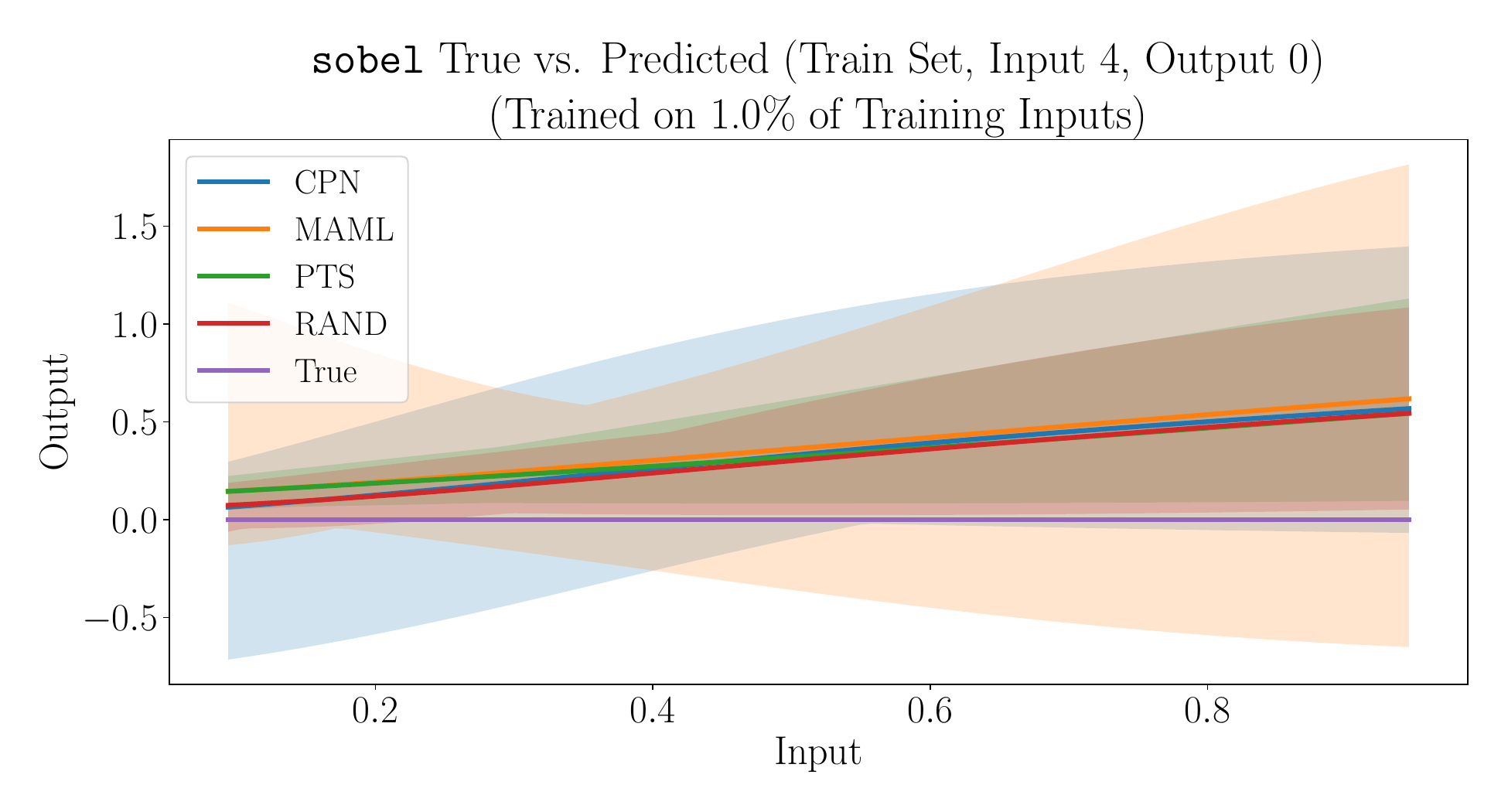}
\\
\vspace*{-1.6em}
\includegraphics[width=0.54\textwidth]{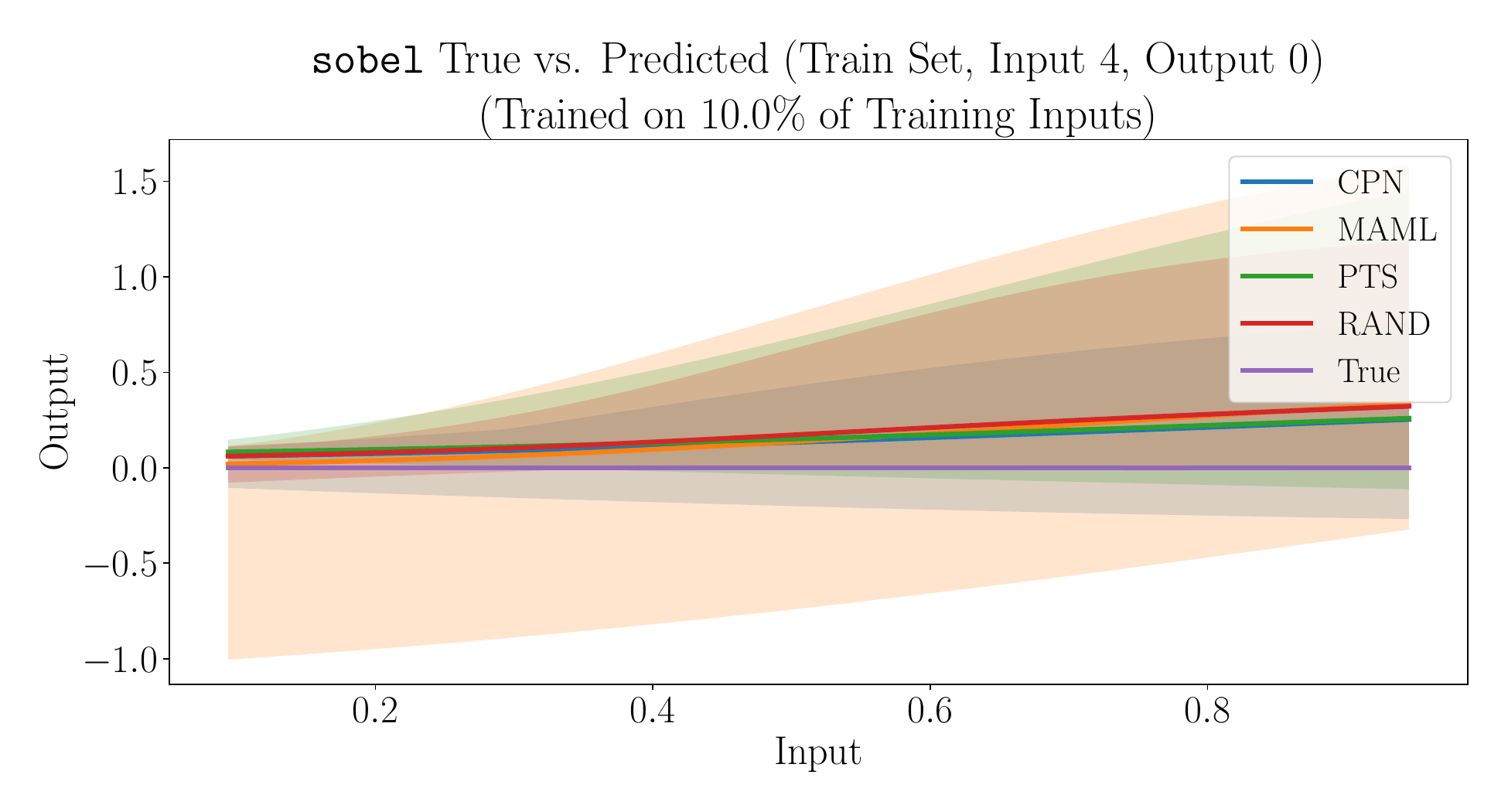}
\\
\vspace*{-1.6em}
\includegraphics[width=0.54\textwidth]{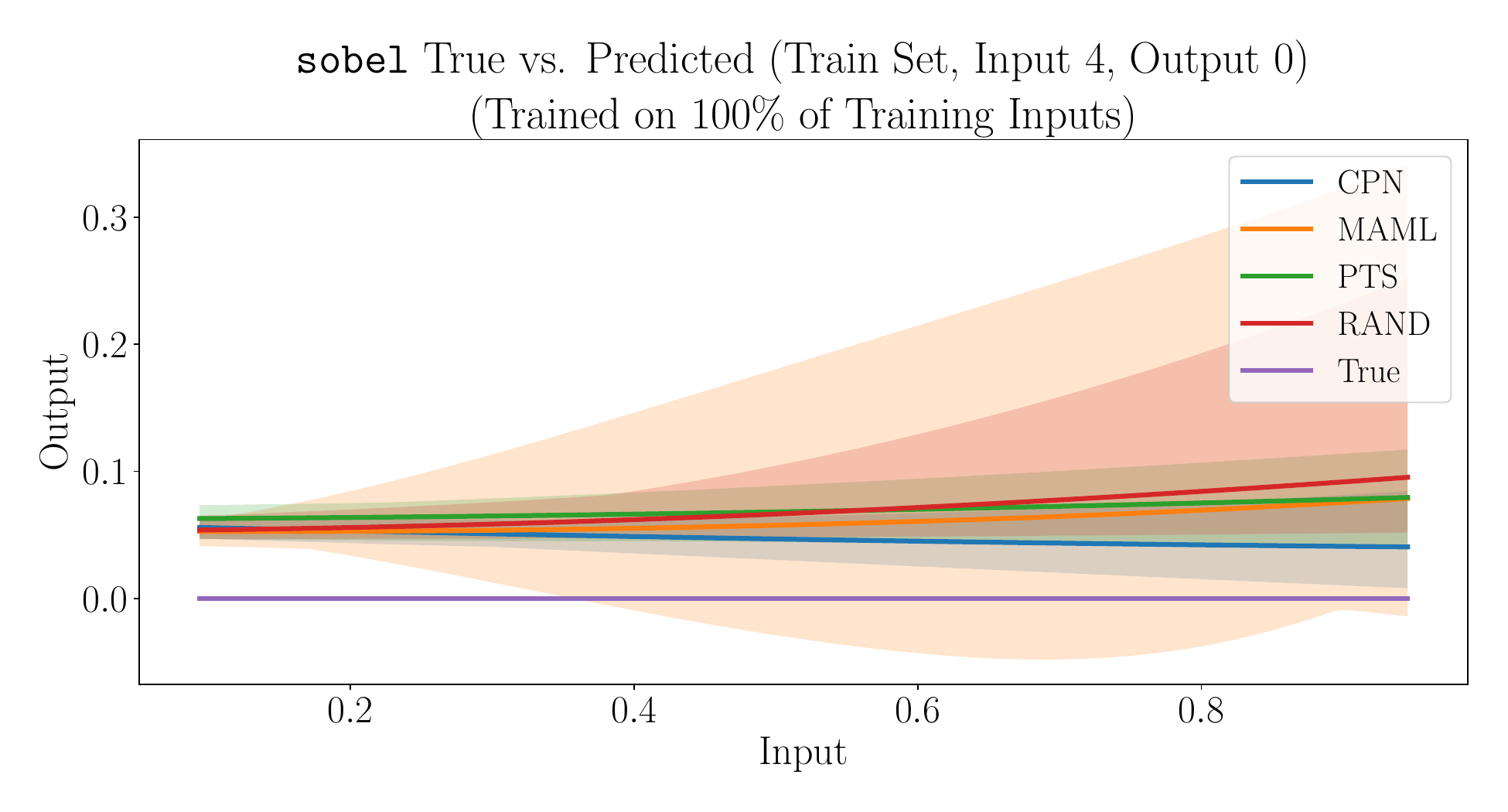}
\\
\vspace*{-1.2em}
\caption{
  Visual comparisons of the ground-truth \texttt{sobel} function from \textsc{ParrotBenchCPN} and neural surrogate approximations thereof, when the fifth input is varied.
  We include results for all dataset sizes evaluated in Section~\ref{sec:data_efficiency}.
}\label{fig:true_vs_pred_sobel_input_4}
\end{figure*}

\begin{figure*}
\centering
\vspace*{-0.7em}
\includegraphics[width=0.54\textwidth]{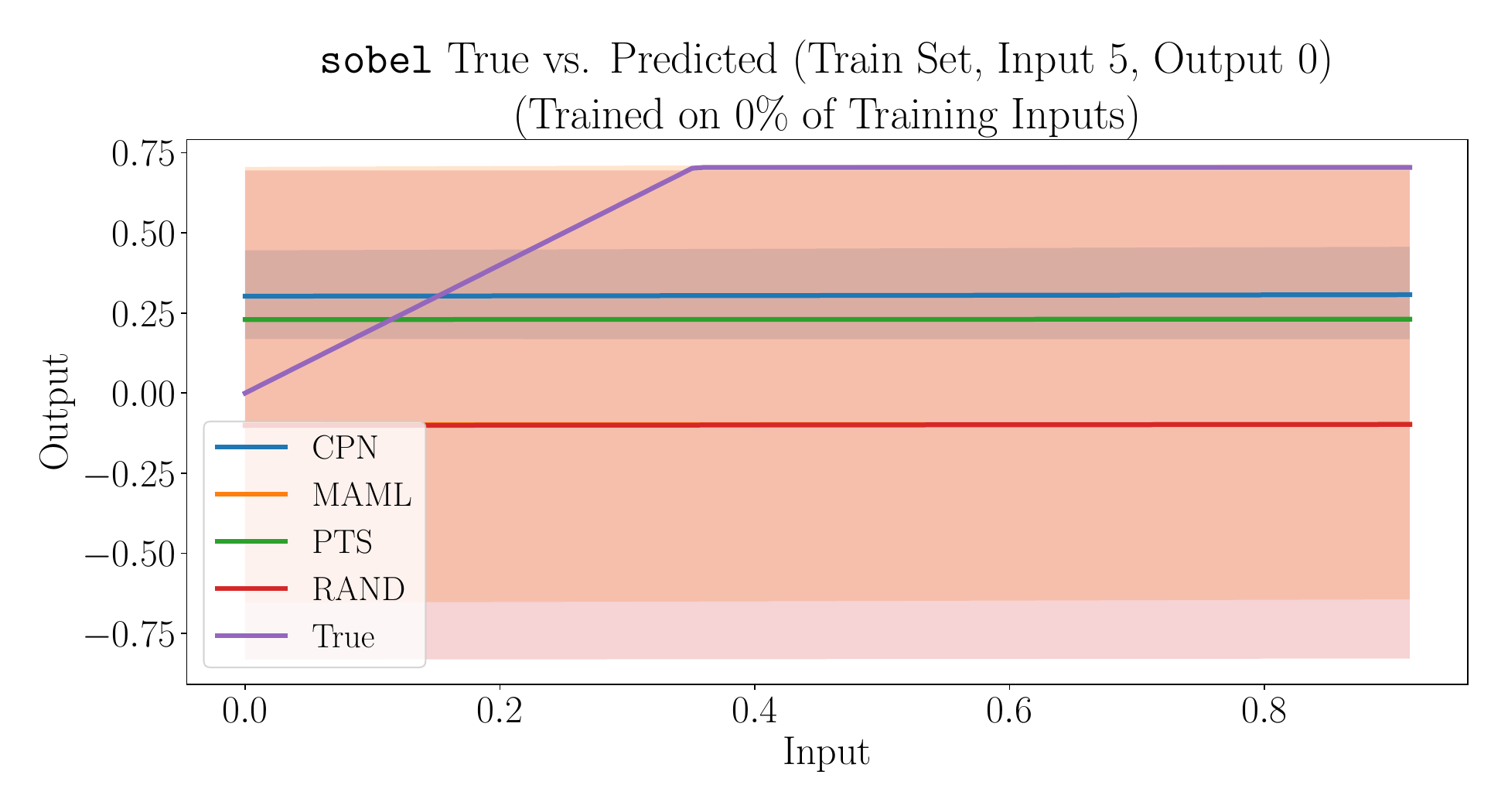}
\\
\vspace*{-1.6em}
\includegraphics[width=0.54\textwidth]{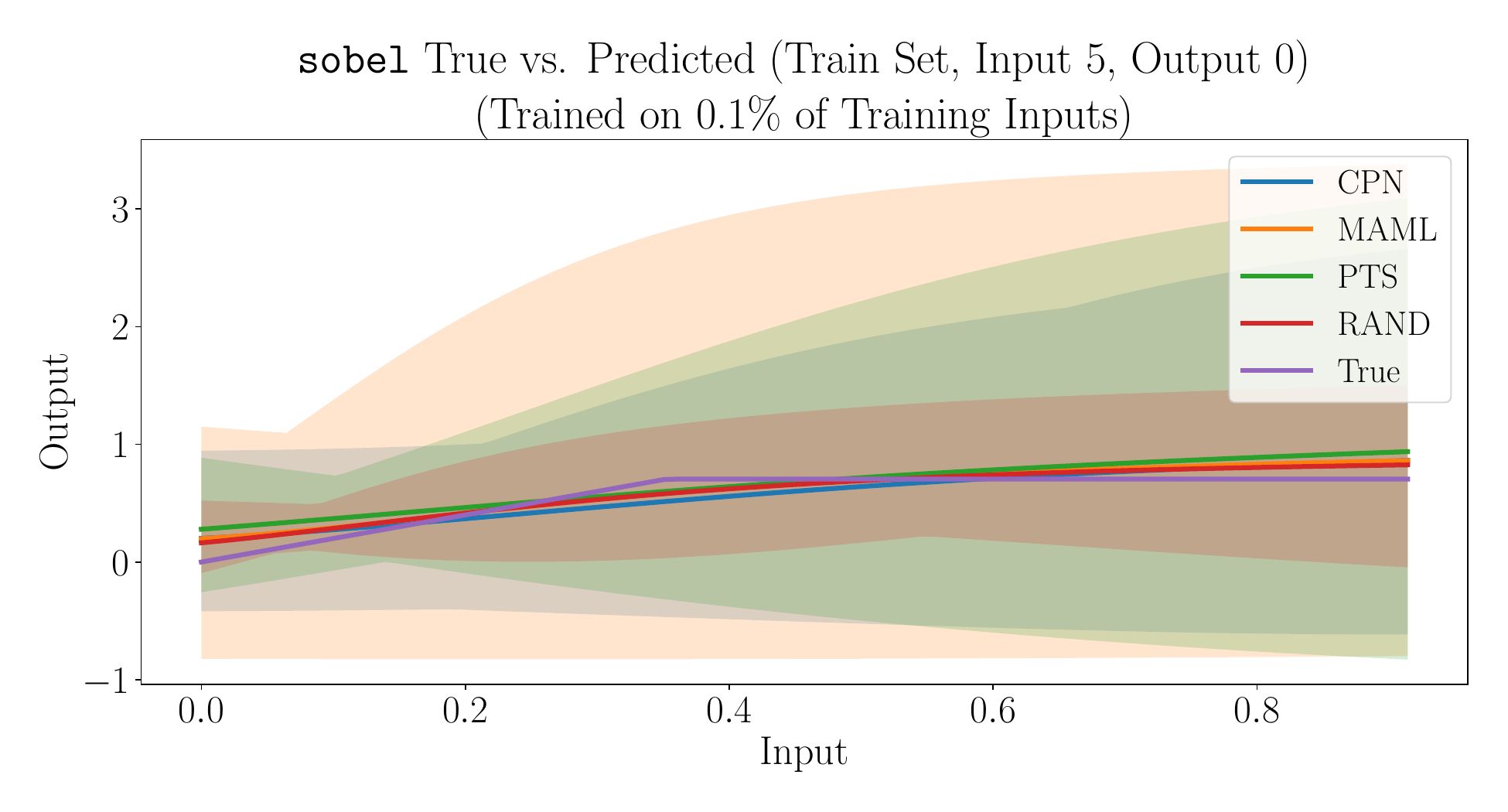}
\\
\vspace*{-1.6em}
\includegraphics[width=0.54\textwidth]{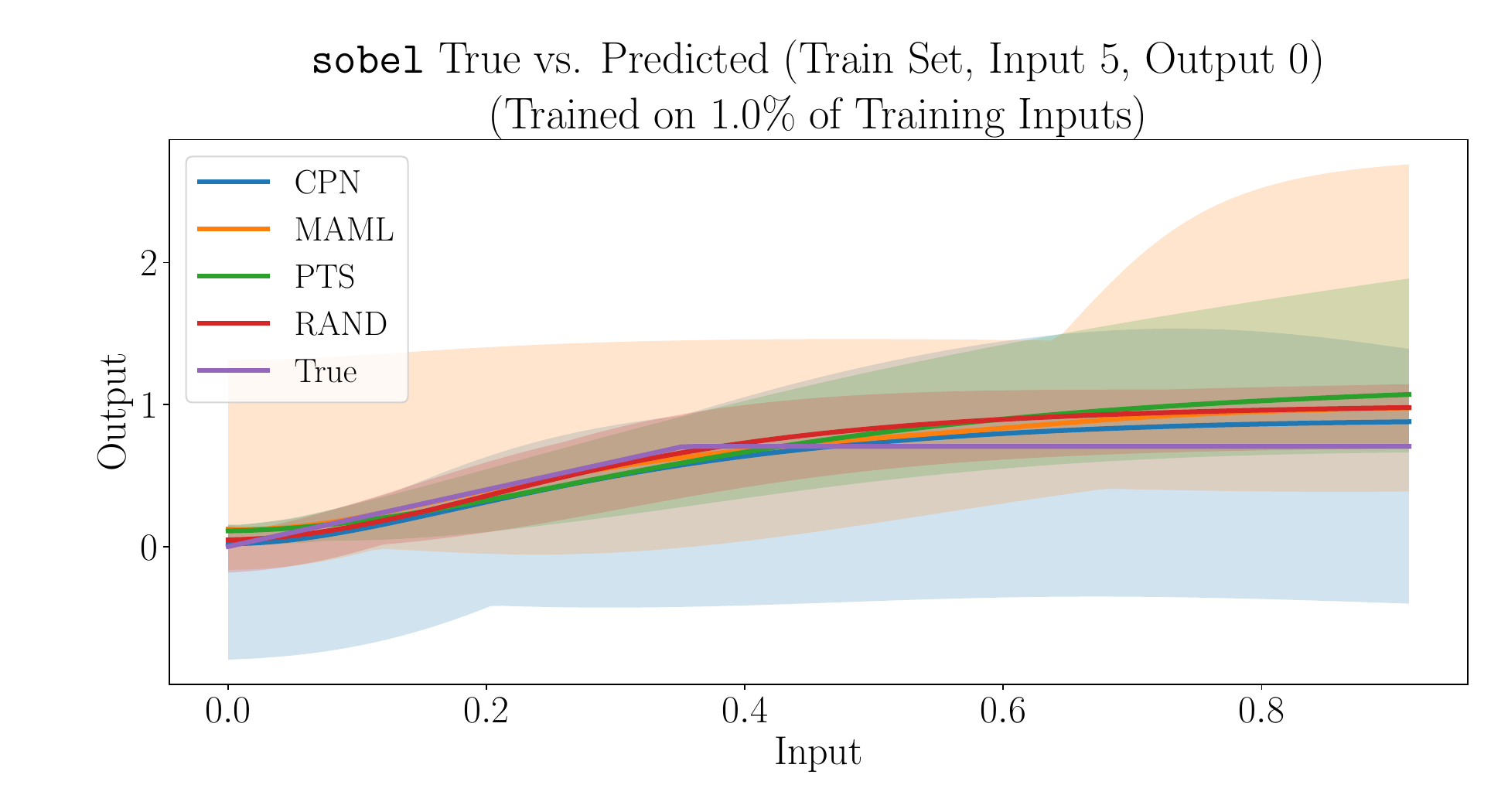}
\\
\vspace*{-1.6em}
\includegraphics[width=0.54\textwidth]{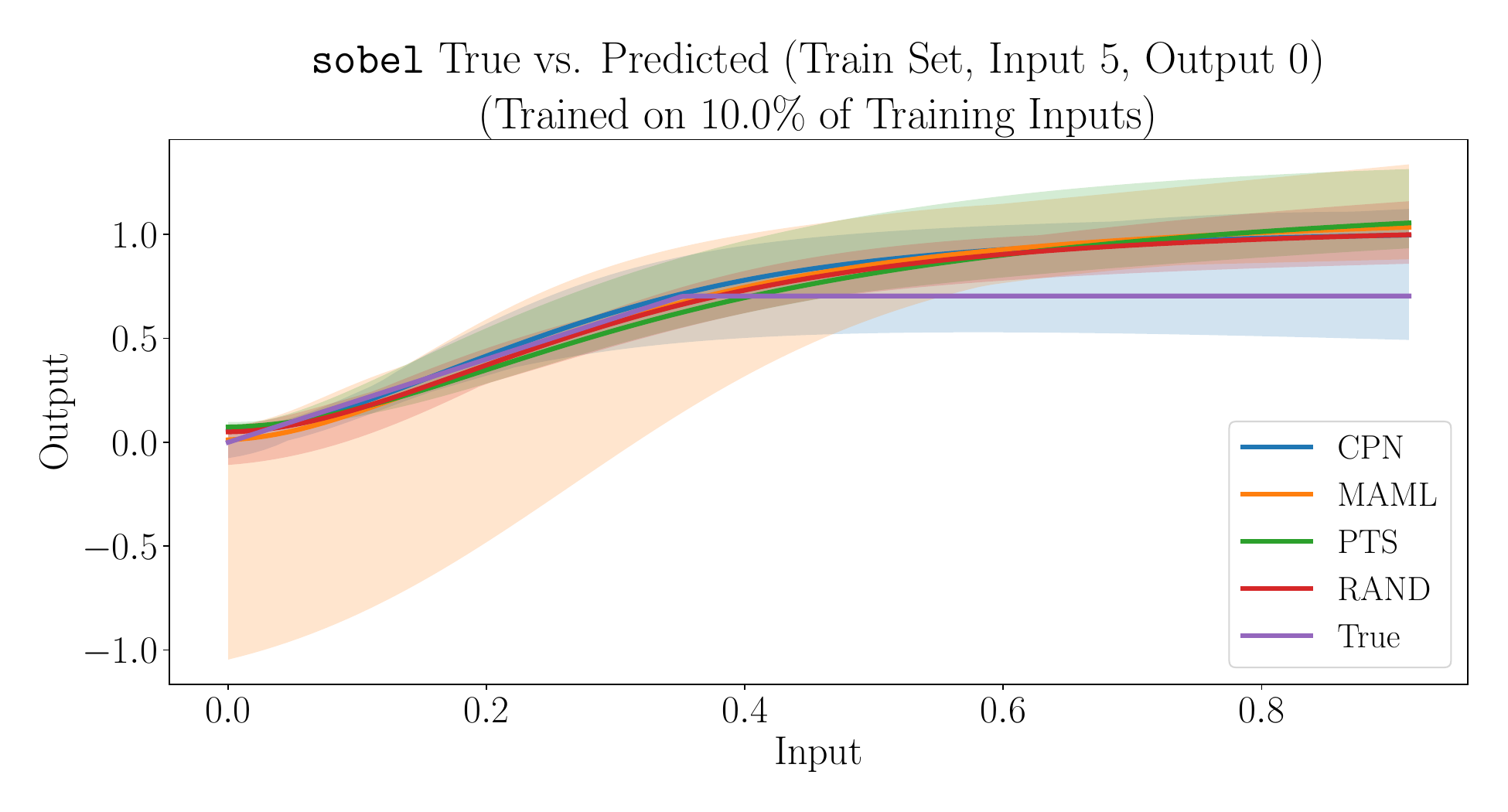}
\\
\vspace*{-1.6em}
\includegraphics[width=0.54\textwidth]{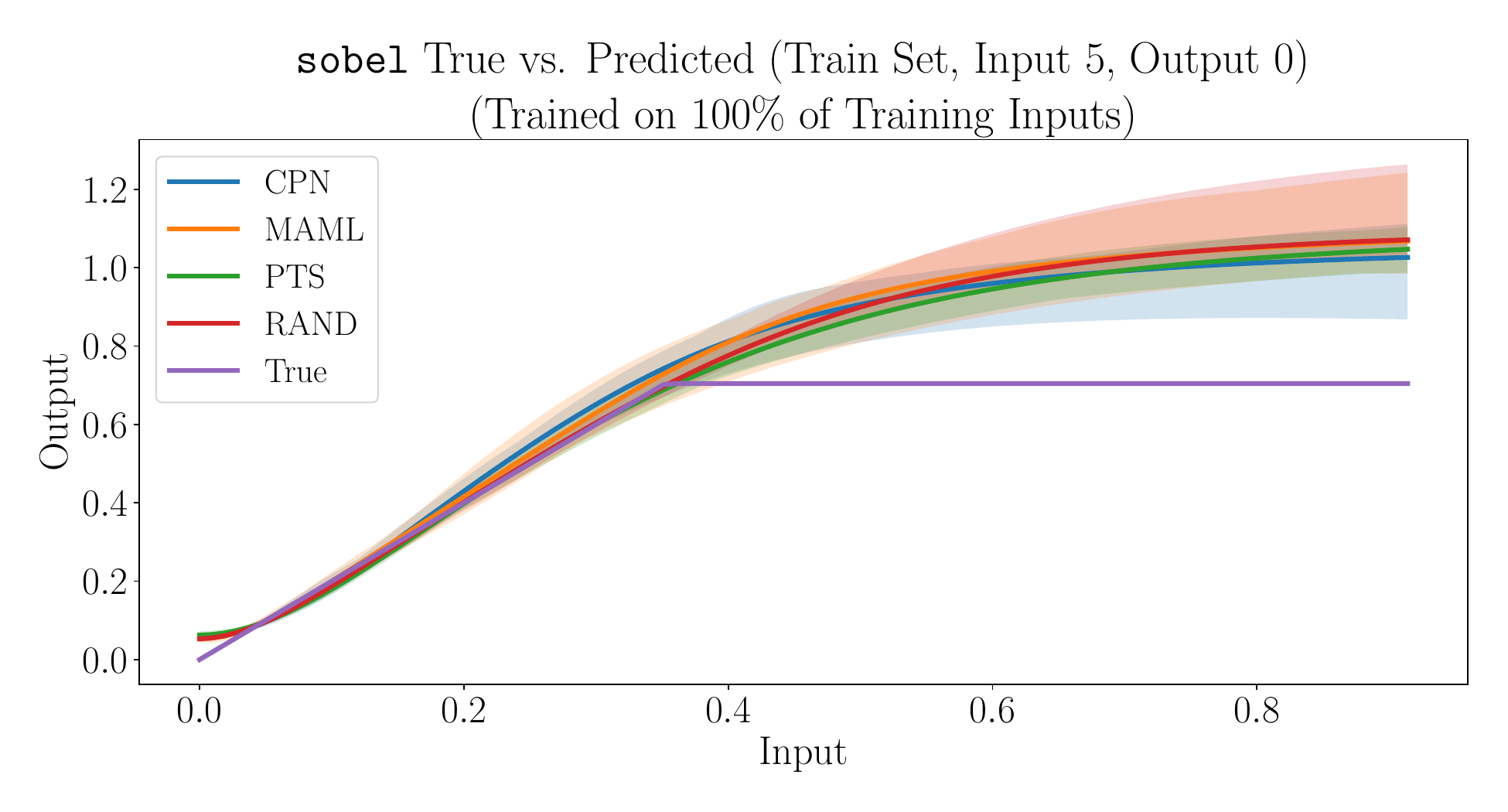}
\\
\vspace*{-1.2em}
\caption{
  Visual comparisons of the ground-truth \texttt{sobel} function from \textsc{ParrotBenchCPN} and neural surrogate approximations thereof, when the sixth input is varied.
  We include results for all dataset sizes evaluated in Section~\ref{sec:data_efficiency}.
}\label{fig:true_vs_pred_sobel_input_5}
\end{figure*}

\begin{figure*}
\centering
\vspace*{-0.7em}
\includegraphics[width=0.54\textwidth]{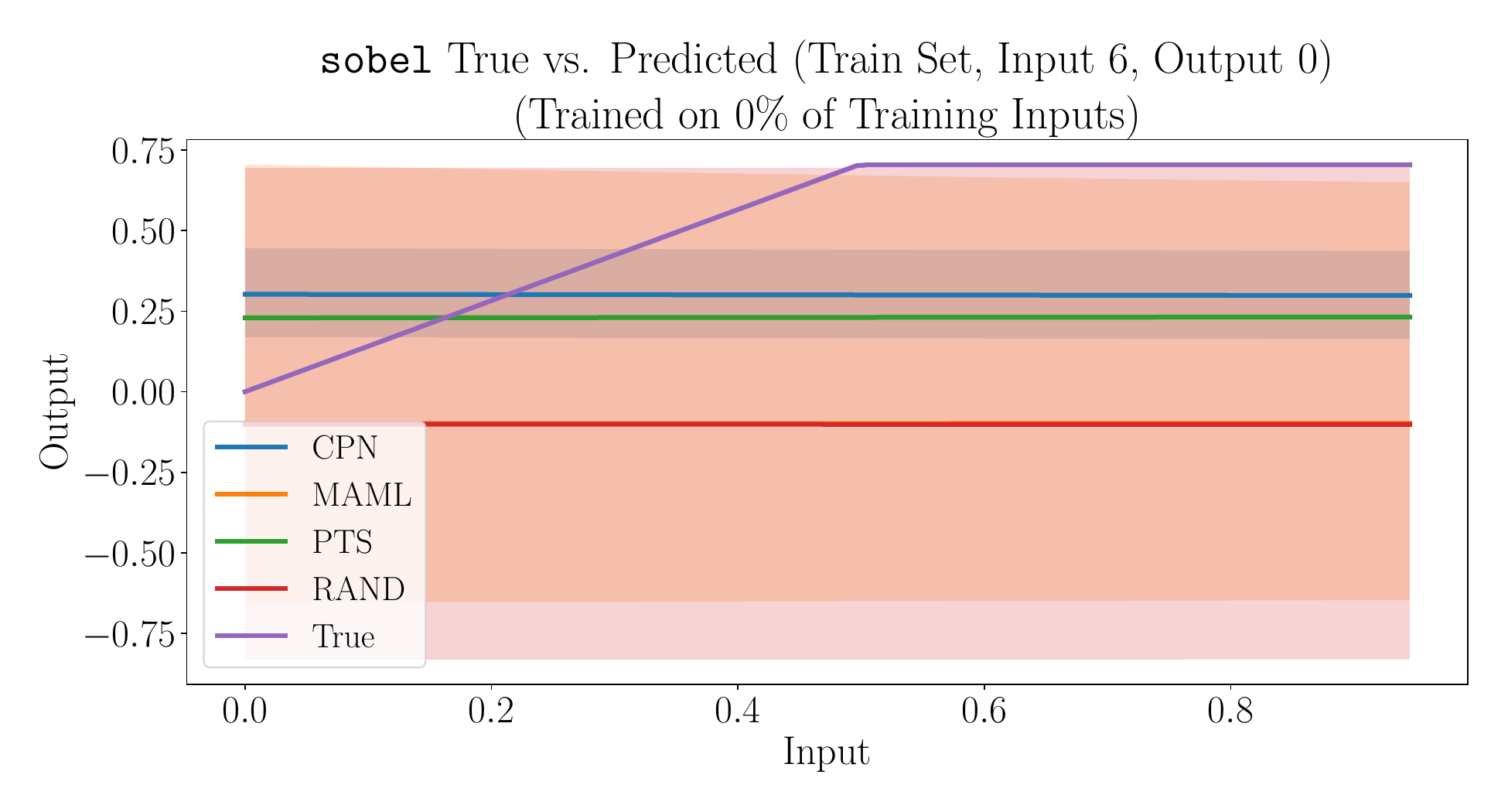}
\\
\vspace*{-1.6em}
\includegraphics[width=0.54\textwidth]{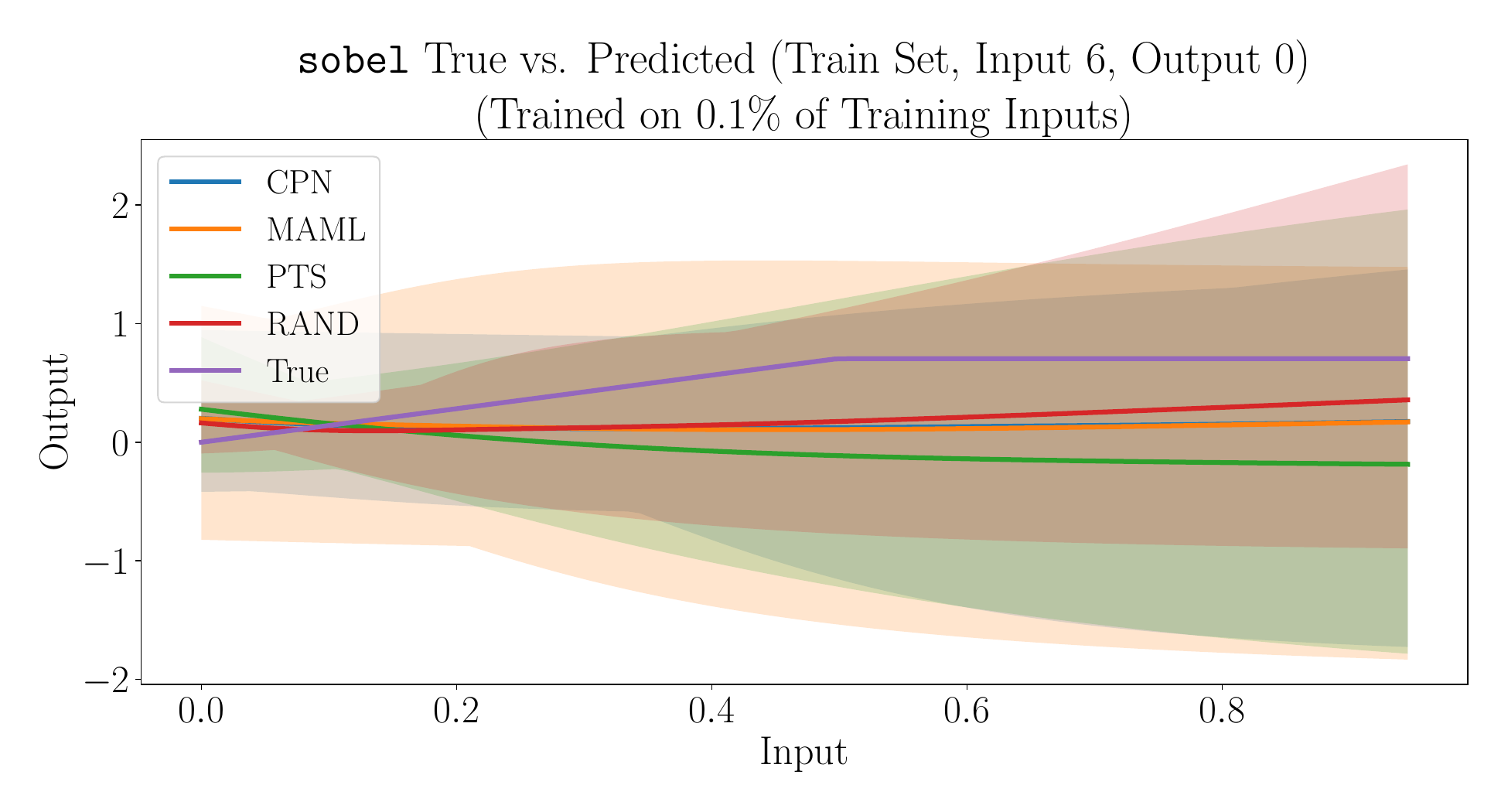}
\\
\vspace*{-1.6em}
\includegraphics[width=0.54\textwidth]{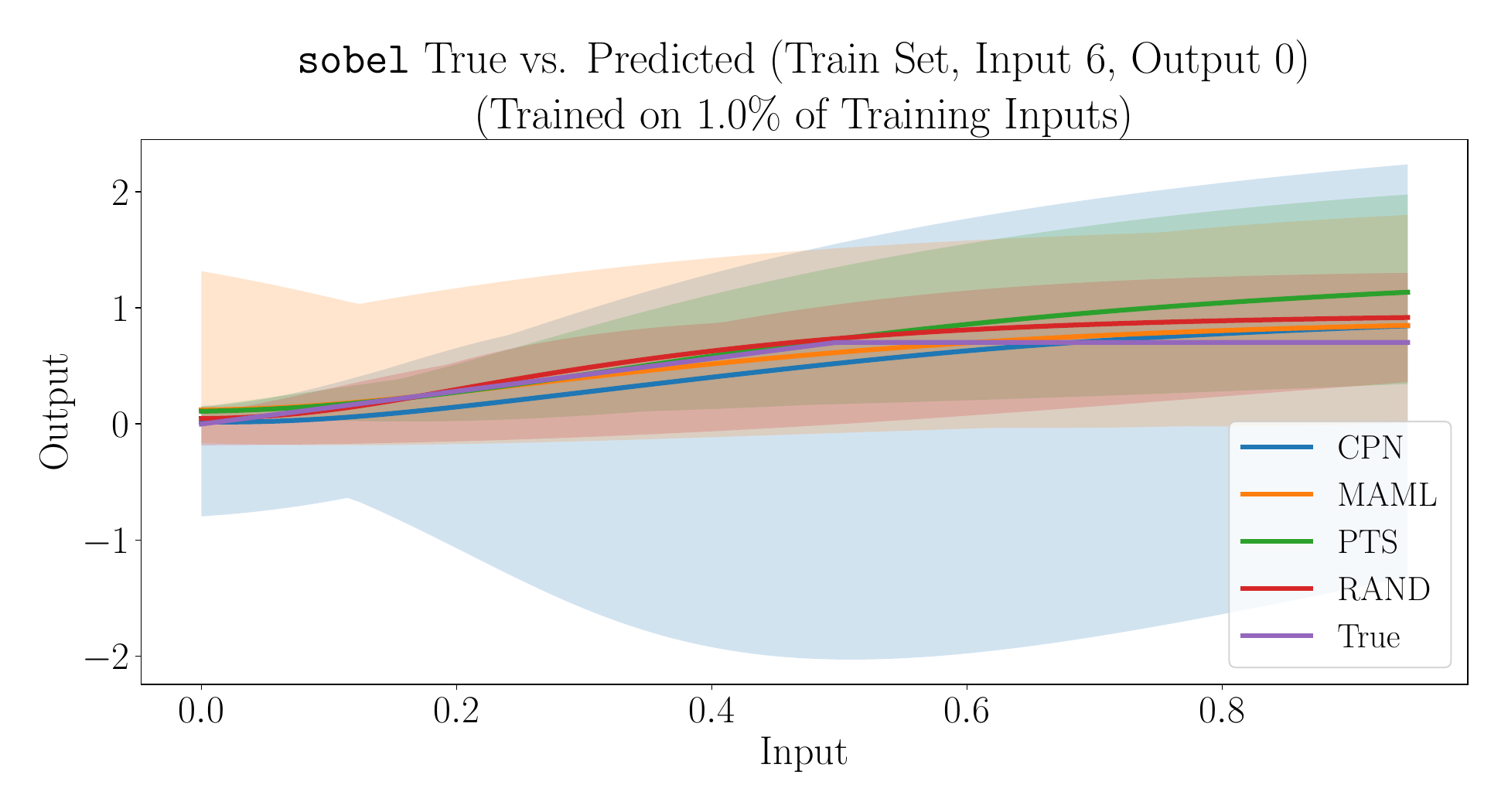}
\\
\vspace*{-1.6em}
\includegraphics[width=0.54\textwidth]{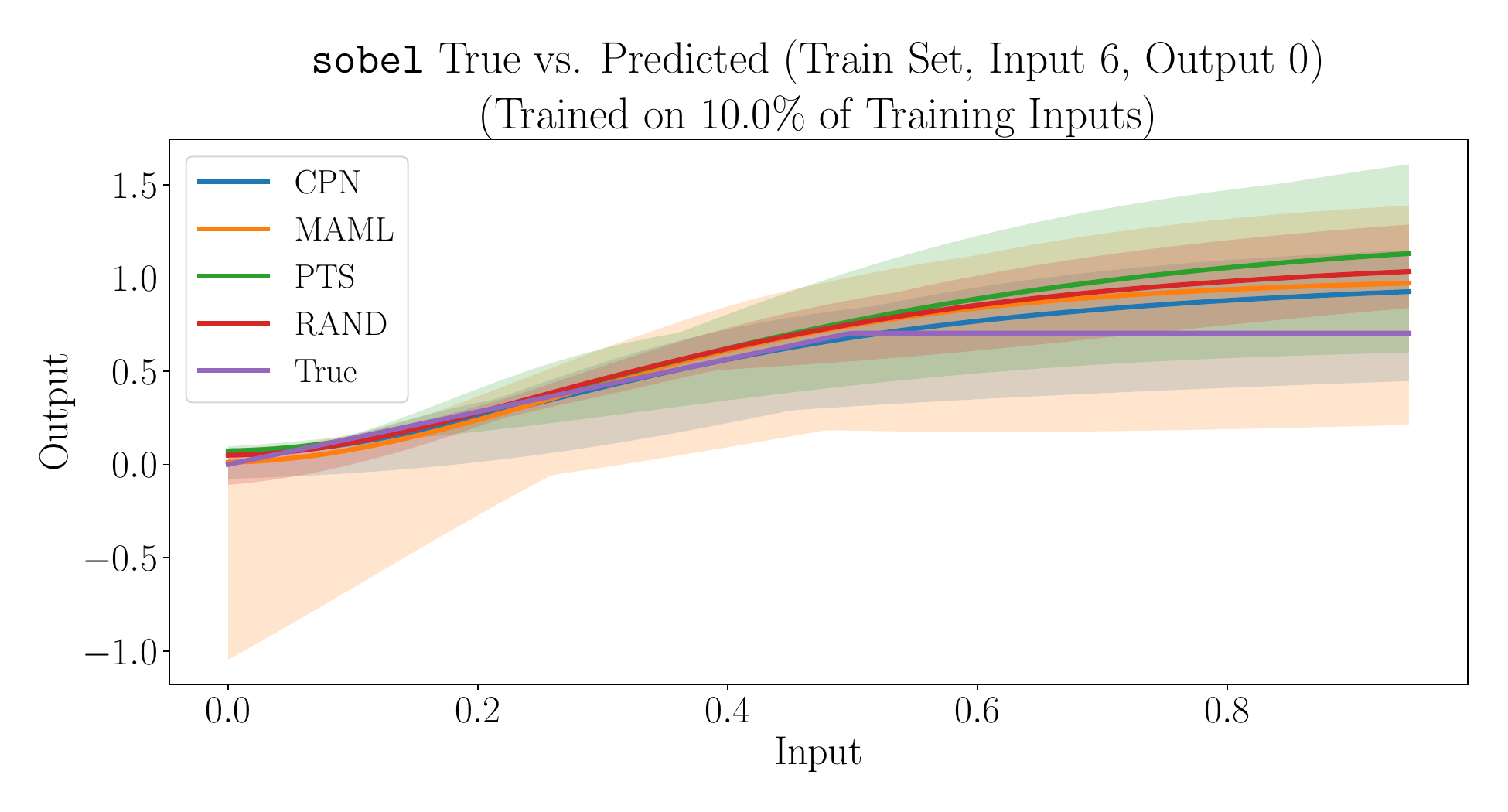}
\\
\vspace*{-1.6em}
\includegraphics[width=0.54\textwidth]{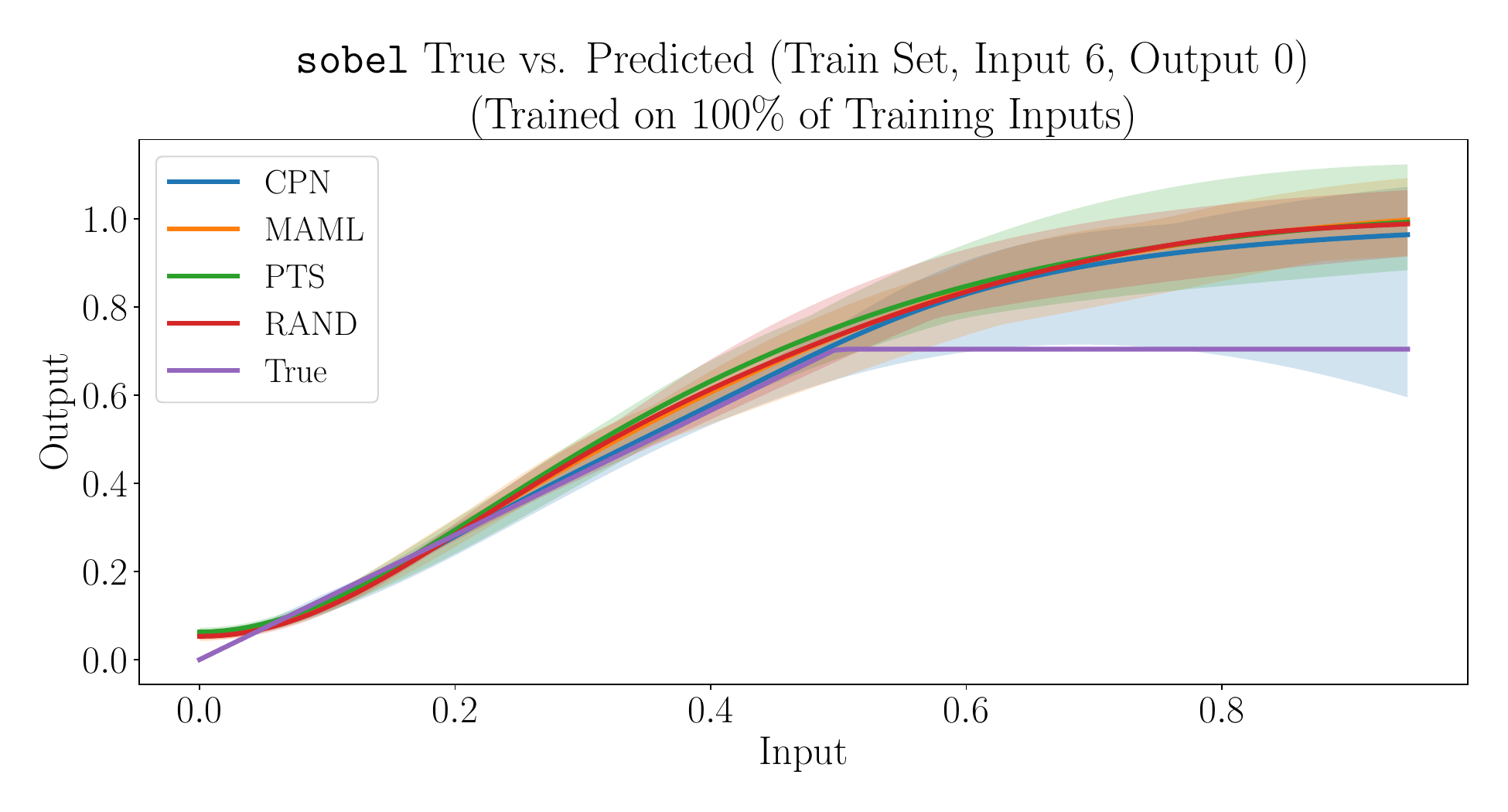}
\\
\vspace*{-1.2em}
\caption{
  Visual comparisons of the ground-truth \texttt{sobel} function from \textsc{ParrotBenchCPN} and neural surrogate approximations thereof, when the seventh input is varied.
  We include results for all dataset sizes evaluated in Section~\ref{sec:data_efficiency}.
}\label{fig:true_vs_pred_sobel_input_6}
\end{figure*}

\begin{figure*}
\centering
\vspace*{-0.7em}
\includegraphics[width=0.54\textwidth]{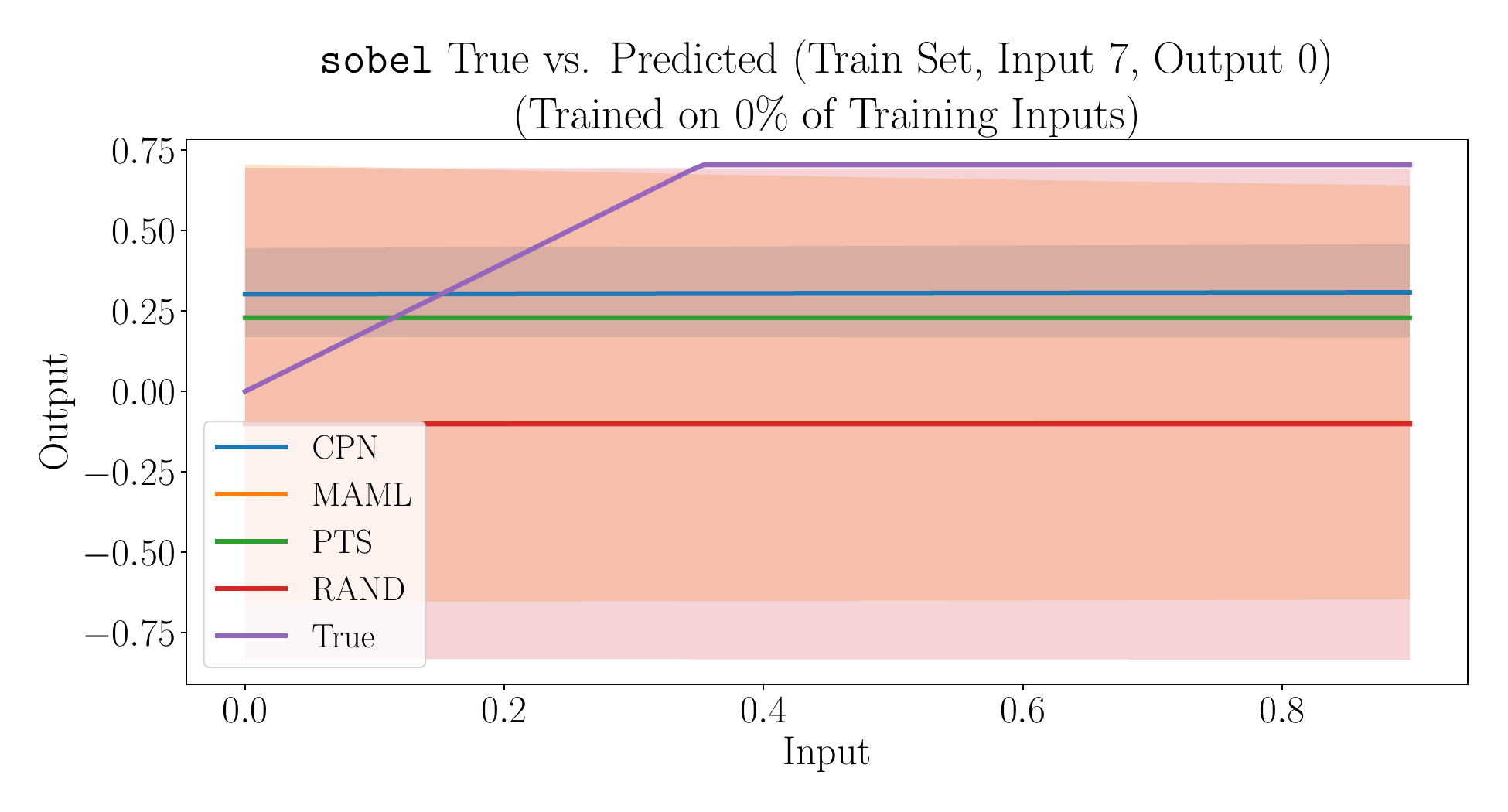}
\\
\vspace*{-1.6em}
\includegraphics[width=0.54\textwidth]{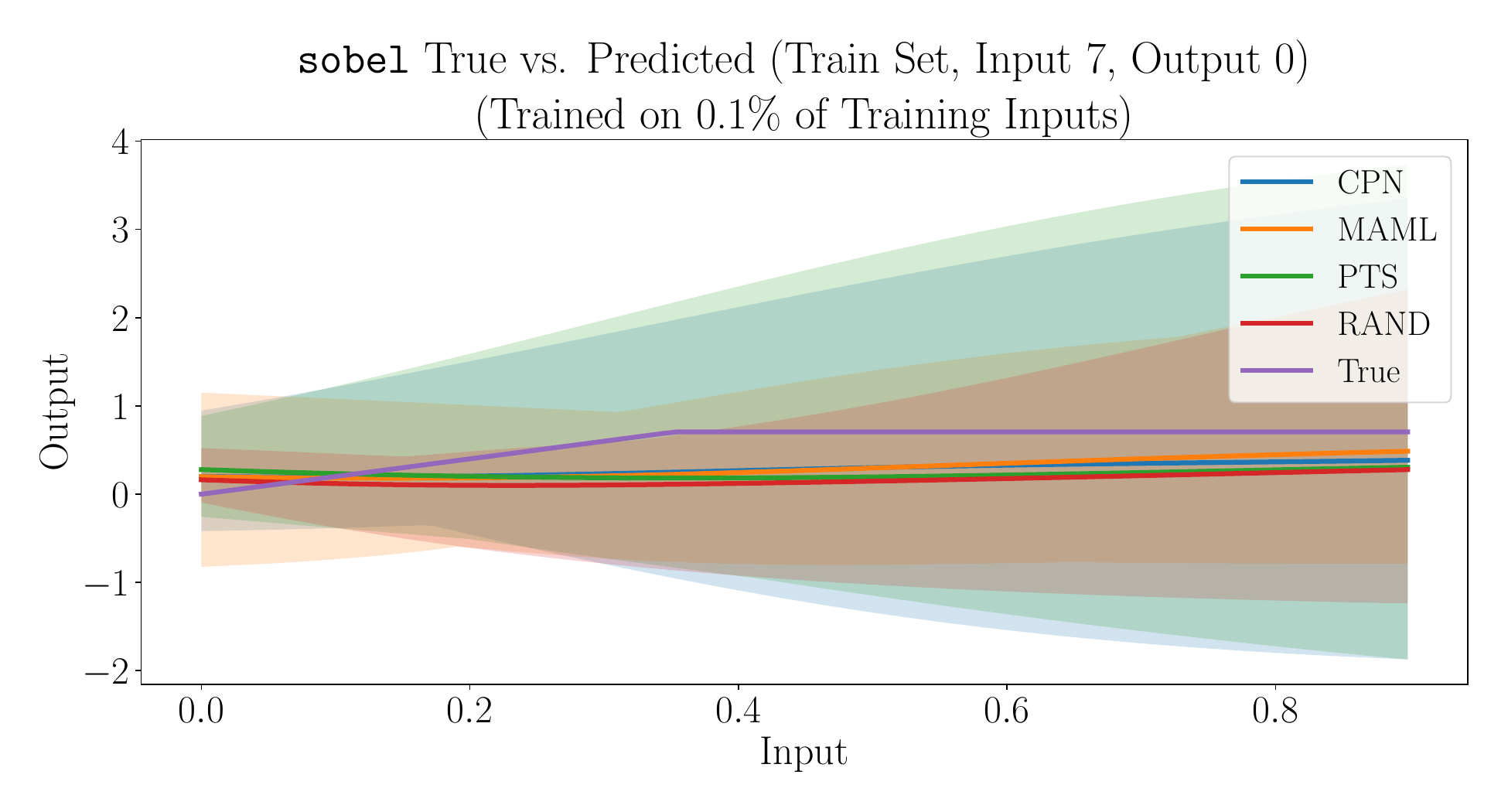}
\\
\vspace*{-1.6em}
\includegraphics[width=0.54\textwidth]{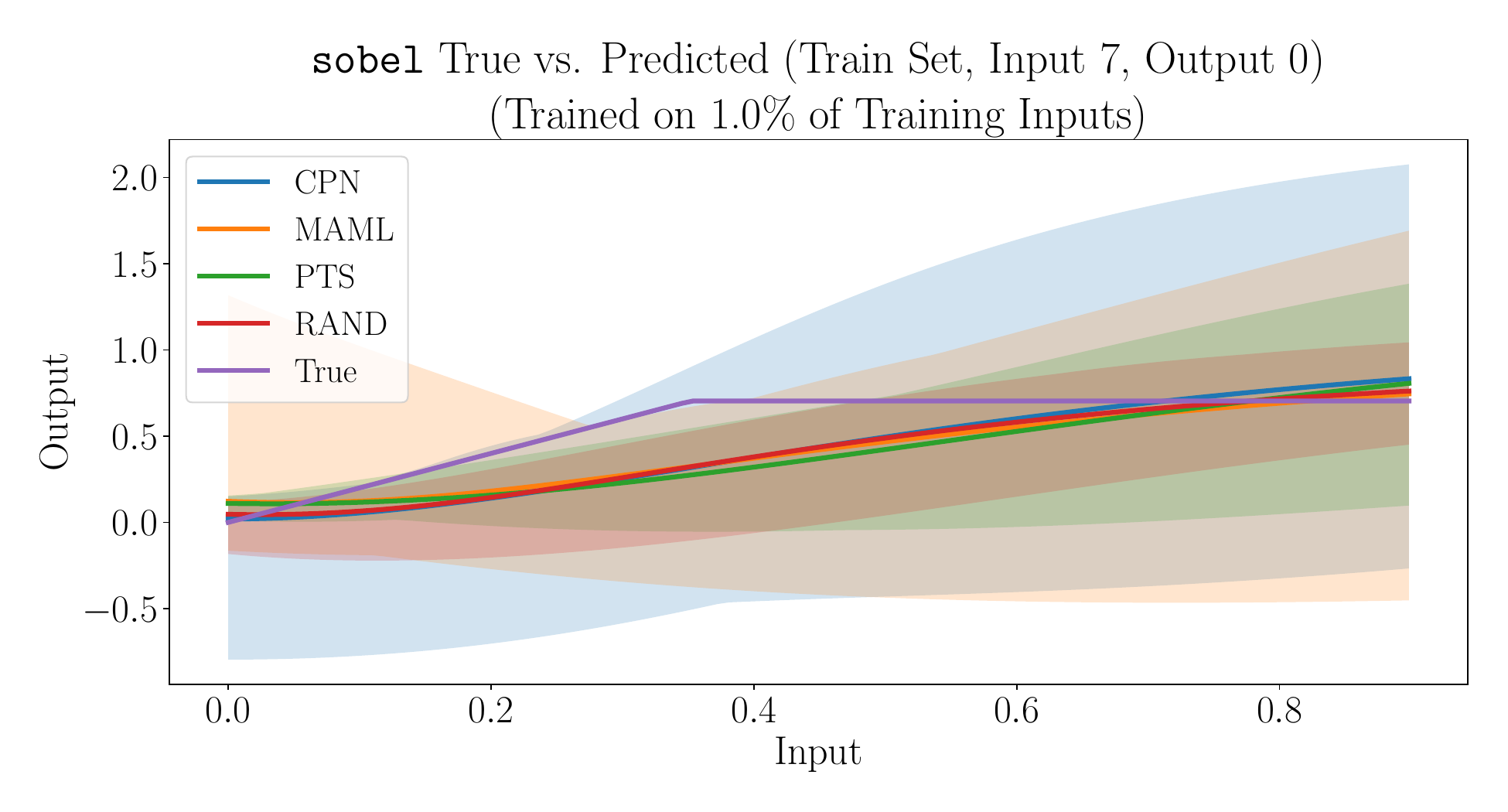}
\\
\vspace*{-1.6em}
\includegraphics[width=0.54\textwidth]{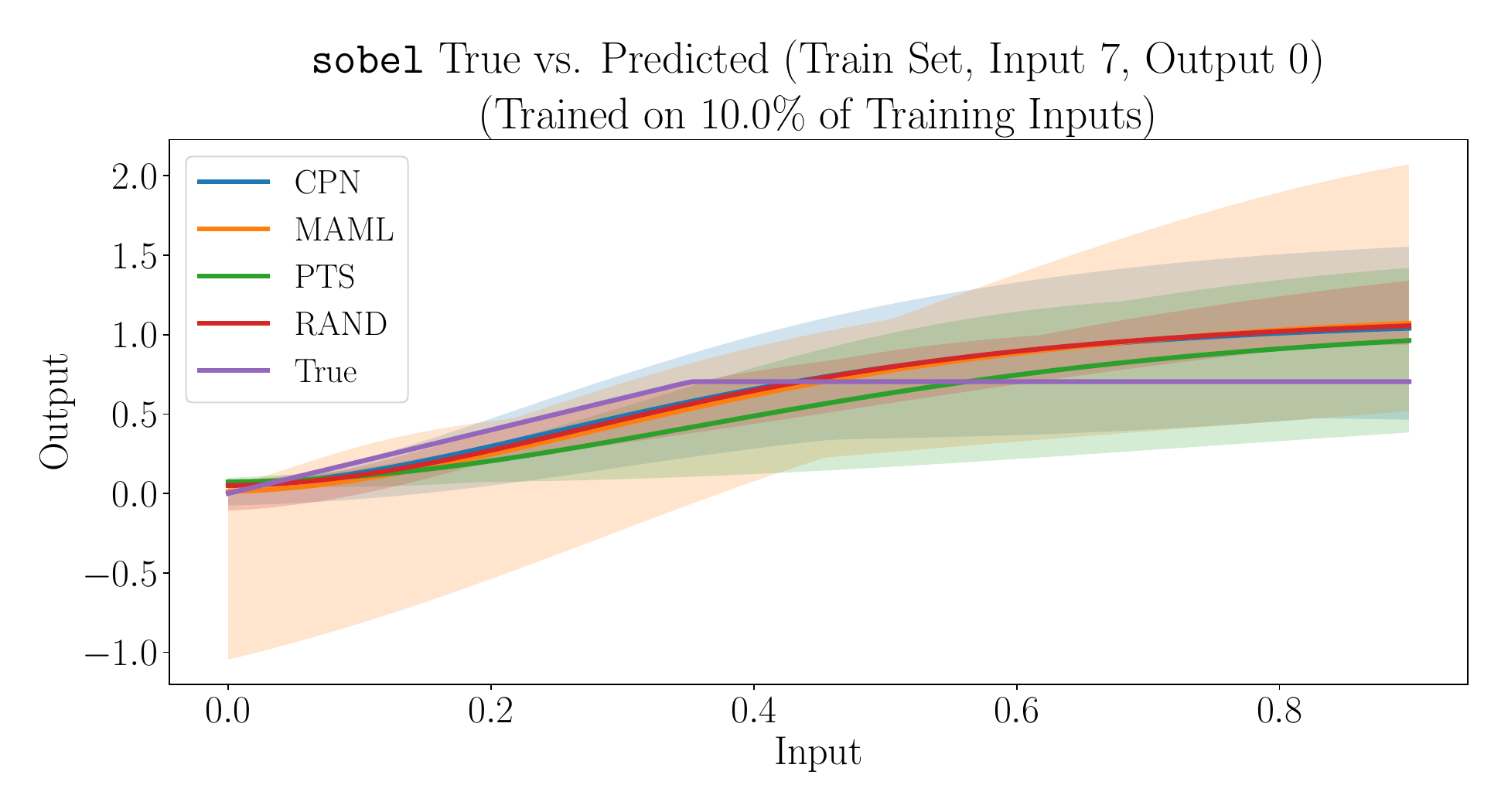}
\\
\vspace*{-1.6em}
\includegraphics[width=0.54\textwidth]{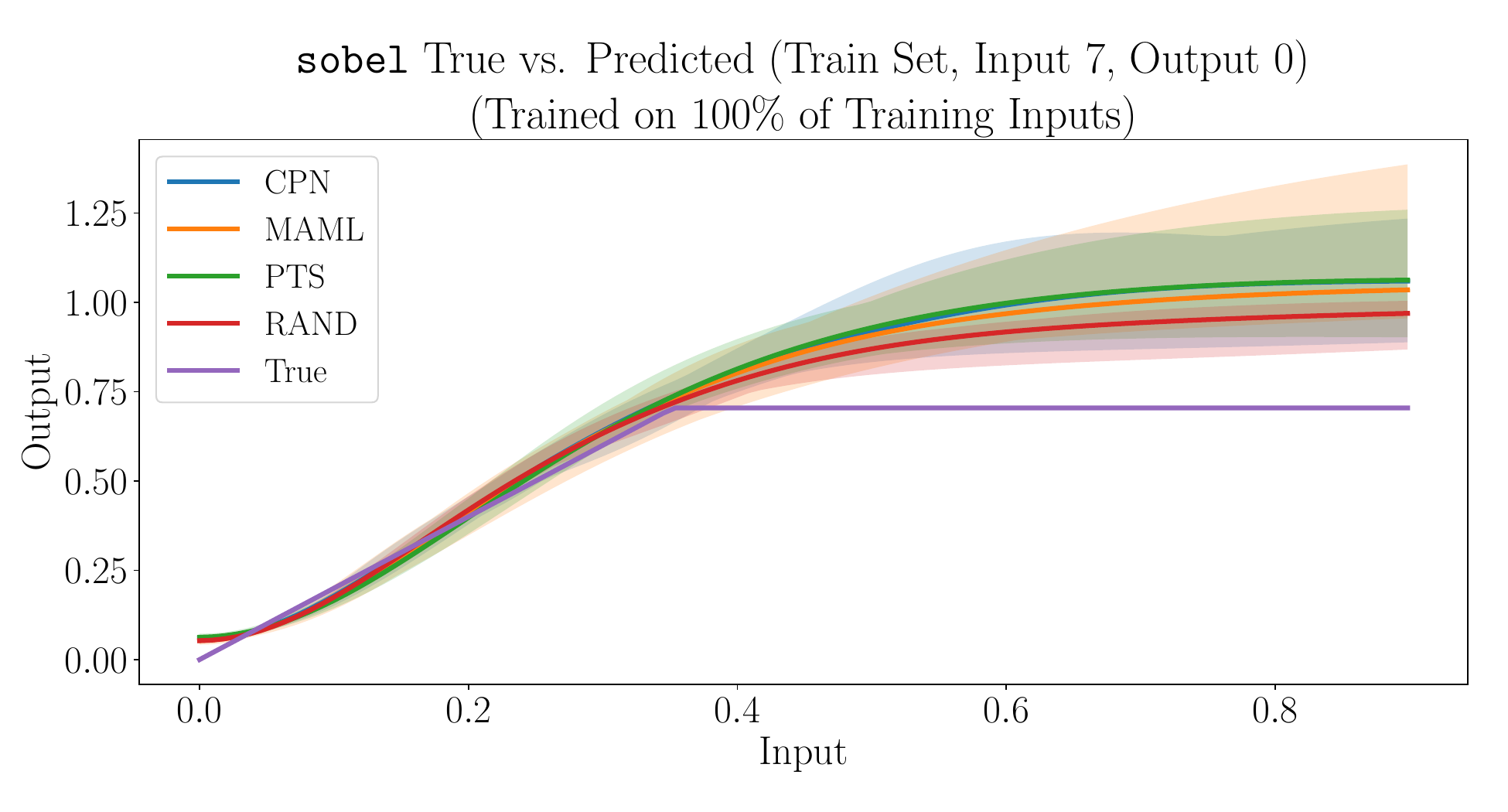}
\\
\vspace*{-1.2em}
\caption{
  Visual comparisons of the ground-truth \texttt{sobel} function from \textsc{ParrotBenchCPN} and neural surrogate approximations thereof, when the eighth input is varied.
  We include results for all dataset sizes evaluated in Section~\ref{sec:data_efficiency}.
}\label{fig:true_vs_pred_sobel_input_7}
\end{figure*}

\begin{figure*}
\centering
\vspace*{-0.7em}
\includegraphics[width=0.54\textwidth]{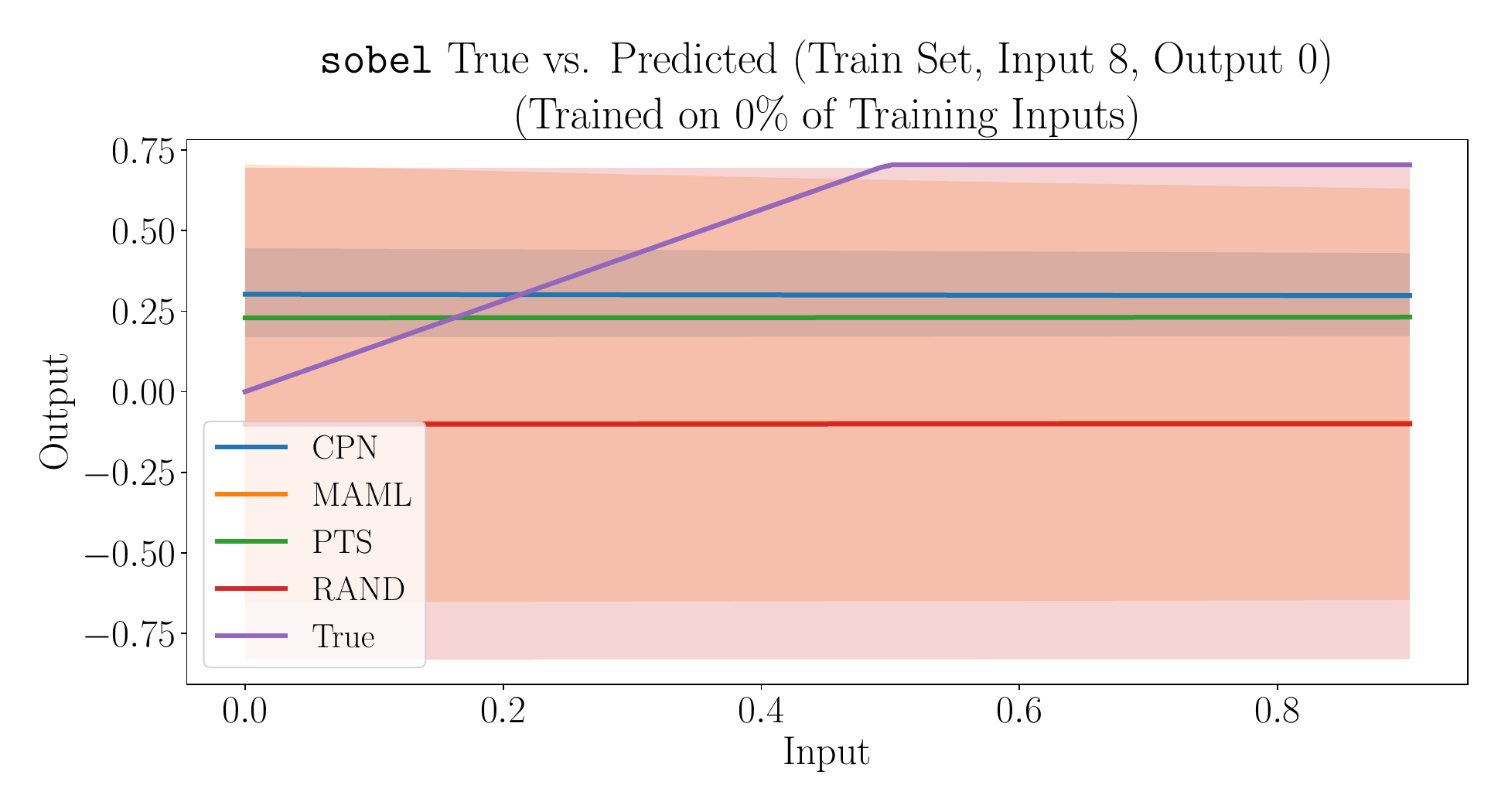}
\\
\vspace*{-1.6em}
\includegraphics[width=0.54\textwidth]{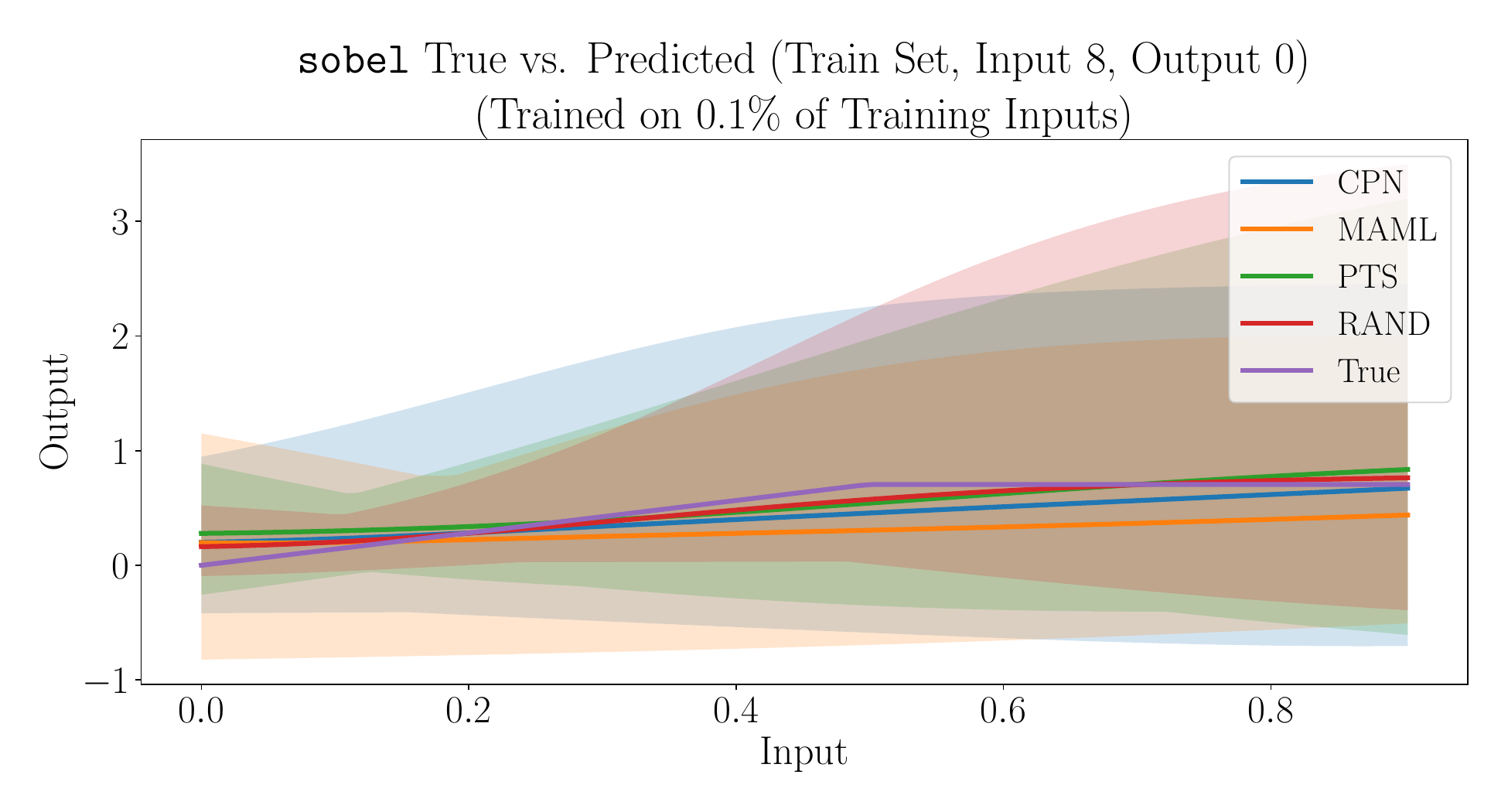}
\\
\vspace*{-1.6em}
\includegraphics[width=0.54\textwidth]{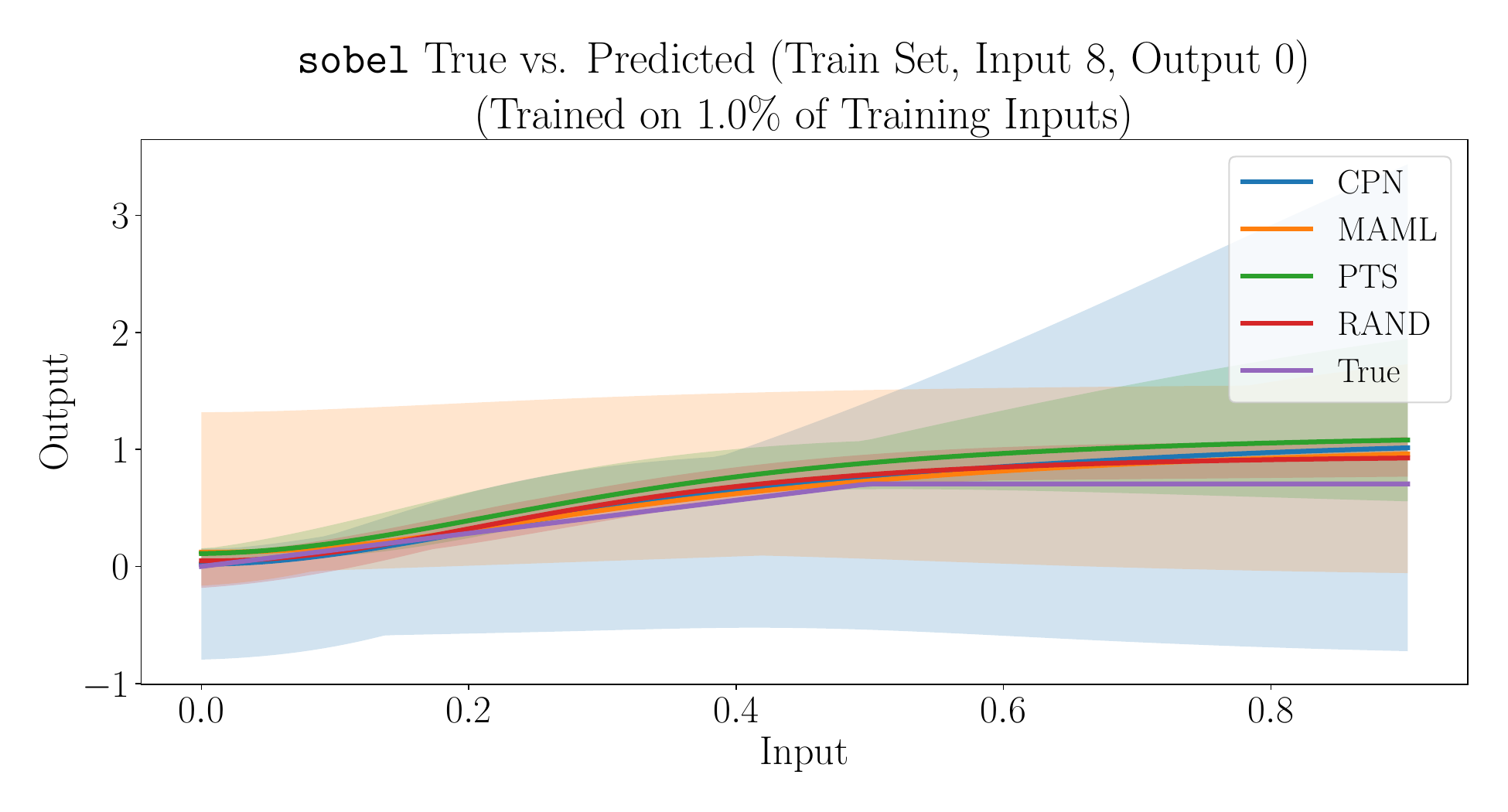}
\\
\vspace*{-1.6em}
\includegraphics[width=0.54\textwidth]{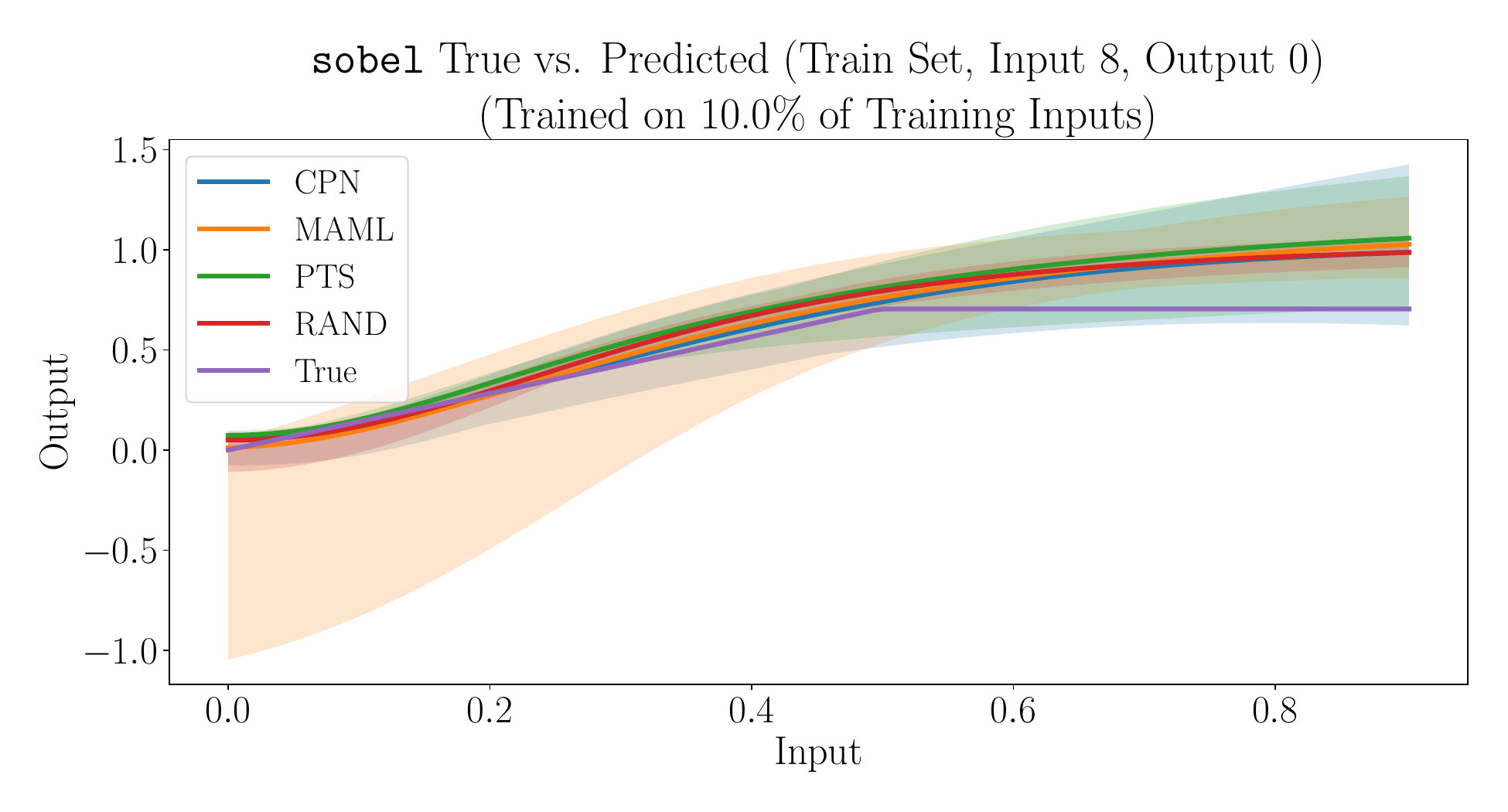}
\\
\vspace*{-1.6em}
\includegraphics[width=0.54\textwidth]{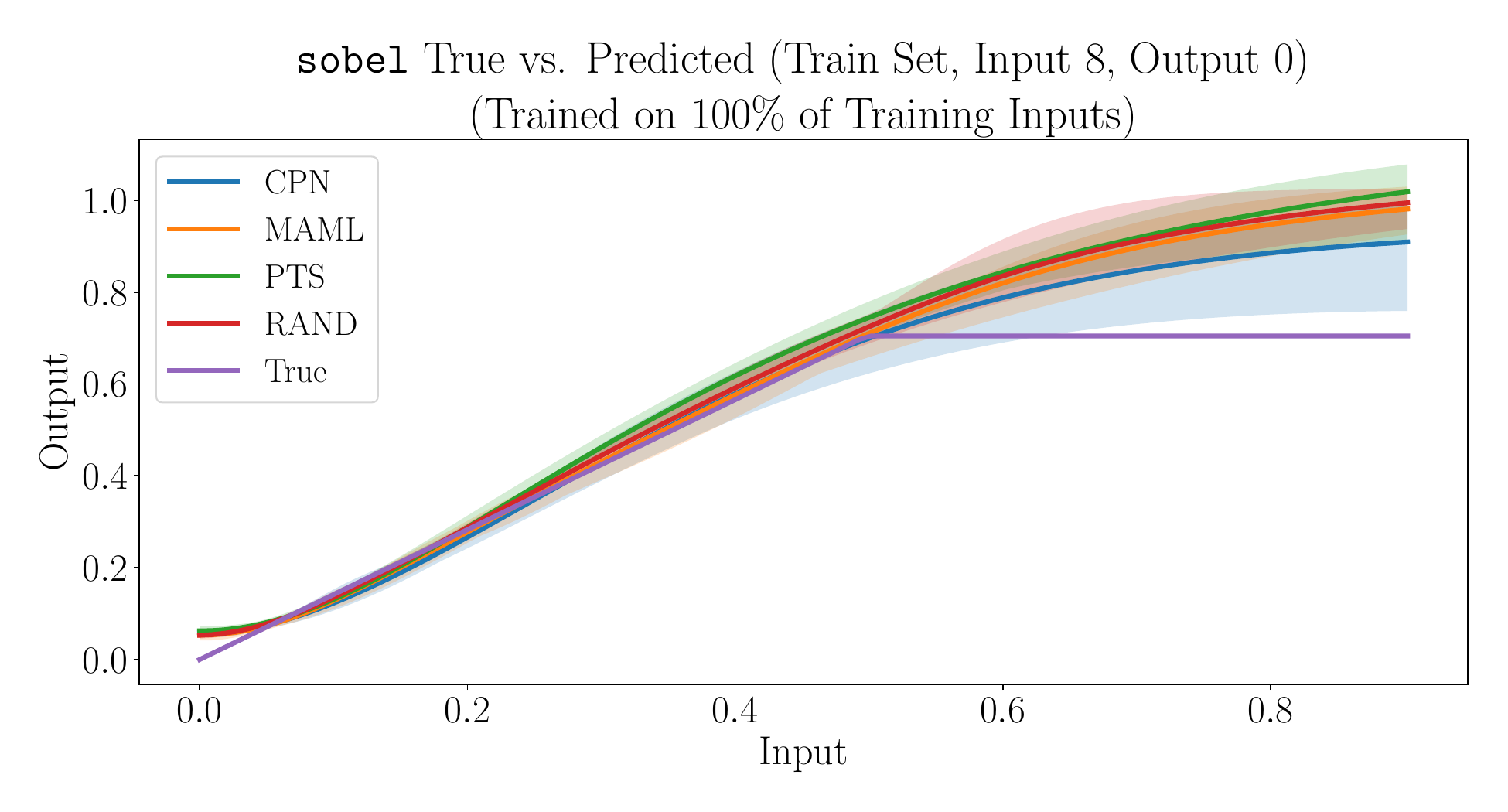}
\\
\vspace*{-1.2em}
\caption{
  Visual comparisons of the ground-truth \texttt{sobel} function from \textsc{ParrotBenchCPN} and neural surrogate approximations thereof, when the ninth input is varied.
  We include results for all dataset sizes evaluated in Section~\ref{sec:data_efficiency}.
}\label{fig:true_vs_pred_sobel_input_8}
\end{figure*}

Figure~\ref{fig:true_vs_pred_fft} shows results for \texttt{fft}, Figures~\ref{fig:true_vs_pred_invk2j_input_0} and \ref{fig:true_vs_pred_invk2j_input_1} show results for both inputs of \texttt{invk2j}, Figures~\ref{fig:true_vs_pred_kmeans_input_0}, \ref{fig:true_vs_pred_kmeans_input_1}, \ref{fig:true_vs_pred_kmeans_input_2}, \ref{fig:true_vs_pred_kmeans_input_3}, \ref{fig:true_vs_pred_kmeans_input_4}, and \ref{fig:true_vs_pred_kmeans_input_5} show results for each input of \texttt{kmeans}, and Figures~\ref{fig:true_vs_pred_sobel_input_0}, \ref{fig:true_vs_pred_sobel_input_1}, \ref{fig:true_vs_pred_sobel_input_2}, \ref{fig:true_vs_pred_sobel_input_3}, \ref{fig:true_vs_pred_sobel_input_4}, \ref{fig:true_vs_pred_sobel_input_5}, \ref{fig:true_vs_pred_sobel_input_6}, \ref{fig:true_vs_pred_sobel_input_7}, and \ref{fig:true_vs_pred_sobel_input_8} show results for each input of \texttt{sobel}.


\section{Neural Surrogates Achieve Acceptable Error}\label{sec:absolute_error_justification}
In this section, we show that, in the context of our evaluation, the error incurred from using neural surrogates is satisfactory for downstream applications.
We first show that the surrogates of \citet{Parrot2012} achieve acceptable end-to-end error on \textsc{ParrotBenchCPN} programs, then we show that our surrogates achieve commensurate or lower error than their surrogates.

\begin{figure*}
  \centering
  \begin{tabular}{lllll|ll}
    \toprule
    Benchmark & CPN & MAML & PTS & RND & PRT & E2E Error \\
    \midrule
    \texttt{fft} & \num{4.3e-06} & \num{3.1e-06} & \num{5.3e-06} & \num{3.2e-06} & \num{2.0e-5} & $7.22\%$ \\
    \texttt{invk2j} & \num{3.3e-03} & \num{3.3e-03} & \num{3.4e-03} & \num{3.1e-03} & \num{5.6e-3} & $7.50\%$\\
    \texttt{kmeans} & \num{3.4e-03} & \num{1.3e-02} & \num{5.2e-03} & \num{8.3e-03} & \num{1.7e-3} & $6.18\%$ \\
    \texttt{sobel} & \num{5.4e-04} & \num{4.5e-04} & \num{6.3e-04} & \num{4.2e-04} & \num{2.3e-3} & $3.44\%$ \\
    \bottomrule
  \end{tabular}
    \caption{
      MSE on \textsc{ParrotBenchCPN} testing set for each initialization method, MSE of the neural surrogates \citet{Parrot2012} train (PRT), and end-to-end error achieved by the surrogates of Esmaeilzadeh et al. (E2E Error).
    }\label{fig:absolute_error_justification}
\end{figure*}

\subsection{End-to-End Error}
Esmaeilzadeh et al. calculate end-to-end error for the benchmarks we consider from \textsc{ParrotBench} as follows:
\begin{itemize}
  \item \textbf{\texttt{fft}.} Apply the fast Fourier transform to a sequence of 2,048 values, where the value at the $i$th index is $i$, and measure the average relative error between the output of the original \texttt{fft} implementation and the approximate \texttt{fft} implementation.
  \item \textbf{\texttt{invk2j}.} Generate 1,000 pairs of joint angles $(\theta_1, \theta_2)$, with both angles sampled uniformly at random from $[0, \pi / 2]$.
  Run forward kinematics on these angles, to obtain $(x, y)$ coordinates for the tip of the joint arm.
  Run inverse kinematics on these $(x, y)$ coordinates, to obtain joint angles $(\tilde{\theta_1}, \tilde{\theta_2})$ that place the tip of the joint arm at $(x, y)$.
  Measure the average relative error between the joint angles recovered by the original \texttt{invk2j} implementation and the approximate \texttt{invk2j} implementation.
  \item \textbf{\texttt{kmeans}.} Apply one iteration of k-means clustering to each pixel of the image in Figure~\ref{fig:parrot_peppers}, then set each pixel's color to the color of the closest centroid.
  Measure the average root mean squared error between the image produced by the original \texttt{kmeans} implementation and the approximate \texttt{kmeans} implementation.
  \item \textbf{\texttt{sobel}.}
  Convert the image in Figure~\ref{fig:parrot_peppers} to grayscale using a weighted average of $30\%$ red, $59\%$ green, and $11\%$ blue.
  Apply the \texttt{sobel} filter to the first row of the image, the first column, and the last row.
  Measure the average root mean squared error between the image produced by the original \texttt{sobel} implementation and the approximate \texttt{sobel} implementation.
\end{itemize}

Figure~\ref{fig:absolute_error_justification} shows the end-to-end error of neural surrogates of \textsc{ParrotBenchCPN} programs, which are semantically equivalent to a subset of the benchmarks Esmaeilzadeh et al. evaluated on.
The end-to-end error for each benchmark is $\leq 7.5\%$, and an end-to end quality loss of $10\%$ or more is common in the approximate computing literature~\cite{Parrot2012,Truffle2012,EnerJ2011,Green2010,QoSProfiling2010}.
For example, Park et al. develop neural surrogates of programs for image processing, audio processing, and speech processing, and they collect user feedback on the perceptual quality of the approximate programs~\cite{AxGames2016}.
Their results show that, on a majority of the benchmarks they consider, a quality loss of $\geq 10\%$ is deemed acceptable by $\geq 80\%$ of users.
The neural surrogates we train achieve commensurate and often lower test error than the surrogates of \citet{Parrot2012}.
Thus, the neural surrogates we train achieve an acceptable level of approximation.

\section{Downcasting Incurs Negligible Error}\label{sec:downcasting_error_justification}
The neural surrogate architectures we target uses a single-precision floating-point data type, but many of the programs we compile in Section~\ref{sec:evaluation} use double-precision data types.
For example, $59\%$ of \textsc{ExeStackCPN} programs use at least one double-precision datatype and $56\%$ of \textsc{ExeStackCPN} programs use exclusively double-precision datatypes.
We now show this implicit downcasting incurs low error relative to overall neural surrogate approximation error.

\begin{figure*}
  \centering
  \begin{tabular}{ll}
    \toprule
    Benchmark & \texttt{float} vs. \texttt{double} MSE \\
    \midrule
    \texttt{fft} & \num{1.2e-14} \\
    \texttt{invk2j} & \num{1.6e-11} \\
    \texttt{kmeans} & \num{0.0} \\
    \texttt{sobel} & \num{6.51e-8} \\
    \bottomrule
  \end{tabular}
    \caption{
      MSE between \textsc{ParrotBenchCPN} implementations that solely use the \texttt{float} datatype and implementations that solely use the \texttt{double} datatype.
      To calculate MSE, each program is evaluated on all inputs from double-precision versions of the training and testing set of \textsc{ParrotBenchCPN}, and MSE is computed using the programs' outputs.
    }\label{fig:downcast_error_justification}
\end{figure*}


\paragraph{Methodology.}
We generate two versions of each \textsc{ParrotBenchCPN} program: one using only the \texttt{float} type and one using only the \texttt{double} type.
This replacement includes arguments, internal variables, and any casts.
We then generate random double-precision inputs according to the methodology in Section~\ref{sec:parrot_input_generation} and execute each version of each program.
We report the mean squared error (MSE) between the outputs of the single- and double-precision versions of each program in Figure~\ref{fig:absolute_error_justification}.
We deem a downcasting error acceptable if it is an order of magnitude smaller than the error incurred by using neural surrogates at all, compared to the original implementation (see Appendix~\ref{sec:absolute_error_justification}).

\paragraph{Results.}
Figure~\ref{fig:downcast_error_justification} shows the downcasting error for each \textsc{ParrotBenchCPN} program, which we compare to Figure~\ref{sec:absolute_error_justification} from Appendix~\ref{sec:absolute_error_justification}.
The downcasting error of \texttt{fft} is significantly smaller than the surrogate error ($\num{1.12e-14}$ vs. $\num{2.0e-5}$).
The downcasting error of \texttt{invk2j} is significantly smaller than the surrogate error ($\num{1.6e-11}$ vs. $\num{5.6e-3}$).
Surprisingly, the downcasting error of \texttt{kmeans} is too small to be captured by a floating-point data type, so it registers as $0.0$.
The downcasting error of \texttt{sobel} is significantly smaller than the surrogate error ($\num{6.51e-8}$ vs. $\num{2.3e-3}$).
We conclude that downcasting from double-precision data types does not significantly affect the overall neural surrogate approximation, error, which we have already shown is acceptable in Appendix~\ref{sec:absolute_error_justification}.

\section{Variable-Input Support for Initialization Methods}\label{sec:padding_for_variable_inputs}
\textsc{CompNet}s, MAML, and pretrained surrogates each produce a fixed-size weight vector for initializing surrogates.
However, programs in \textsc{ExeStackCPN} and \textsc{ParrotBenchCPN} have various numbers of inputs.
To support programs in these datasets, we develop strategies for adapting these initialization methods, and we present a methodology for choosing the best of these strategies.

\paragraph{Variable-Input Strategies.}
To develop variable-input initialization methods, we chose a vector size with as many parameters as the architecture with the largest number of inputs we wish to support, defined as the \textit{covering architecture} in Section~\ref{sec:compnet}, and we developed strategies for supplying data to unused inputs.

There are two types of padding data we considered: randomly distributed and constantly zero.
With random padding, any excess inputs are supplied with values from the same distribution as the primary inputs.
With zero padding, any excess inputs are supplied with zeroes.

There are three phases in which data is supplied to an initialization method: training the initialization method, finetuning surrogates initialized by the method, and evaluating surrogates initialized by the method.
Thus, we categorize the strategies we consider by the type of data the initialization method is trained on, the type of data the initialized surrogates are finetuned on, and the type of data the initialized surrogates are evaluated on.
We considered most permutations of random and zero padding for each of these three phases.
Notably, however, we did not consider the family of strategies where one finetunes on zero-padded inputs and evaluates on random-padded inputs because it seemed unlikely that adding a new source of noise at inference time would lead to any improvement.



\paragraph{Methodology.}
To decide which strategy to use for each initialization method, we performed the \textsc{ParrotBenchCPN} data efficiency evaluation of Section~\ref{sec:data_efficiency} with a set of padding strategies applied to each initialization method. 
For each initialization method, we chose the strategy that achieved the greatest overall test loss improvement over random initialization.
Note that these experiments were performed prior to adding variable-output support, so we split the \texttt{fft} and \texttt{invk2j} benchmarks in \textsc{ParrotBenchCPN} into multiple programs---one for each output.

\paragraph{Results.}
We present the results in separate figures for random initialization (Figure~\ref{fig:rnd_ft_ev_modes}), pretrained surrogates trained on random-padded and zero-padded inputs (Figures~\ref{fig:pts_r_ft_ev_modes} and \ref{fig:pts_z_ft_ev_modes}), MAML initializations trained on random-padded and zero-padded inputs (Figures~\ref{fig:maml_r_ft_ev_modes} and \ref{fig:maml_z_ft_ev_modes}), and \textsc{CompNets} trained on random-padded and zero-padded inputs (Figures~\ref{fig:cpn_r_ft_ev_modes} and \ref{fig:cpn_z_ft_ev_modes}).

Random initialization sees performance degradation with every padding strategy.
One explanation for this degradation is that the baseline is random initialization with an architecture that has exactly as many inputs as needed, whereas each of the padding strategies operates on the covering architecture.
Since the magnitude of weights in the He initialization is inversely proportional to the fan-in and fan-out of a neuron~\cite{HeInit2015}, the magnitude of weights in the first layer of the network will be smaller, potentially slowing convergence.

Surrogates that are pretrained on random-padded inputs perform approximately as well as surrogates pretrained on zero-padded inputs for all finetuning and evaluation variants.
Among the finetuning and evaluation variants, finetuning and evaluating on zero-padded inputs performs the best.
The best configuration by a small margin is pretraining on random-padded inputs and finetuning and evaluating on zero-padded inputs; this configuration achieves a geometric mean test loss improvement of $1.13\times$.

Across all finetuning and evaluation modes, MAML initializations trained on zero-padded inputs outperform MAML initializations trained on random-padded inputs.
When MAML initializations are trained on zero-padded inputs, finetuning and evaluating on zero-padded inputs leads to the greatest geometric mean test loss improvement of $1.29\times$ over random initialization.

Across all finetuning and evaluation modes, \textsc{CompNet} initializations trained on random-padded inputs outperform \textsc{CompNet} initializations trained on zero-padded inputs.
When \textsc{CompNet} initializations are trained on random-padded inputs, finetuning and evaluating on zero-padded inputs leads to the greatest geometric mean test loss improvement of $1.96\times$ over random initialization.

\paragraph{Conclusion.}
In light of these results, we make the following decisions.
We choose \textsc{CompNet}s that are trained on random-padded inputs and the surrogates they produce are finetuned and evaluated on zero-padded inputs.
We choose MAML initializations that are trained, finetuned, and evaluated on zero-padded inputs.
We choose pretrained surrogates that are pretrained on random-padded inputs and finetuned and evaluated on zero-padded inputs.
We choose standard random initialization over any of the padded variants (i.e., we make the topology match the program's input-output signature).

The reason why some initialization methods perform better when training on random-padded inputs and others perform better when training on zero-padded inputs is unclear and we believe deserves further study.

\begin{figure*}
    \centering
    \begin{tabular}{lllll}
    \toprule
    Program & RND & RND FT-R EV-R & RND FT-R EV-Z & RND FT-Z EV-Z \\
    \midrule
    \texttt{fft} (0) & $\mathbf{1.00}\times$ & $0.02\times$ & $0.02\times$ & $0.19\times$ \\
    \texttt{fft} (1) & $1.00\times$ & $0.18\times$ & $0.23\times$ & $\mathbf{1.06}\times$ \\
    \texttt{invk2j} (0) & $\mathbf{1.00}\times$ & $0.45\times$ & $0.53\times$ & $\mathbf{1.00}\times$ \\
    \texttt{invk2j} (1) & $\mathbf{1.00}\times$ & $0.31\times$ & $0.32\times$ & $0.82\times$ \\
    \texttt{kmeans} & $\mathbf{1.00}\times$ & $0.64\times$ & $0.65\times$ & $0.83\times$ \\
    \texttt{sobel} & $\mathbf{1.00}\times$ & $\mathbf{1.00}\times$ & $\mathbf{1.00}\times$ & $\mathbf{1.00}\times$ \\
    \bottomrule
    \end{tabular}
    \vspace{1em}
    
    \begin{tabular}{lllll}
    \toprule
    Dataset Size & RND & RND FT-R EV-R & RND FT-R EV-Z & RND FT-Z EV-Z \\
    \midrule
    $0\%$ & $\mathbf{1.00}\times$ & $0.80\times$ & $0.80\times$ & $0.80\times$ \\
    $0.1\%$ & $\mathbf{1.00}\times$ & $0.05\times$ & $0.08\times$ & $0.51\times$ \\
    $1\%$ & $\mathbf{1.00}\times$ & $0.10\times$ & $0.10\times$ & $0.39\times$ \\
    $10\%$ & $1.00\times$ & $0.22\times$ & $0.23\times$ & $\mathbf{1.01}\times$ \\
    $100\%$ & $1.00\times$ & $1.22\times$ & $\mathbf{1.29}\times$ & $1.18\times$ \\
    \bottomrule
    \end{tabular}
    \vspace{1em}
    
    \begin{tabular}{lllll}
    \toprule
    Statistic & RND & RND FT-R EV-R & RND FT-R EV-Z & RND FT-Z EV-Z \\
    \midrule
    0th & $\mathbf{1.00}\times$ & $\num{4.46e-04}\times$ & $\num{6.03e-04}\times$ & $0.01\times$ \\
    25th & $\mathbf{1.00}\times$ & $0.27\times$ & $0.30\times$ & $0.92\times$ \\
    50th & $\mathbf{1.00}\times$ & $0.83\times$ & $0.86\times$ & $\mathbf{1.00}\times$ \\
    75th & $1.00\times$ & $1.00\times$ & $1.00\times$ & $\mathbf{1.09}\times$ \\
    100th & $1.00\times$ & $5.64\times$ & $\mathbf{6.48}\times$ & $1.66\times$ \\
    \midrule
    MPI & \textbf{0}th & 68th & 63rd & 50th \\
    \midrule
    GM & $\mathbf{1.00}\times$ & $0.26\times$ & $0.28\times$ & $0.72\times$ \\
    \bottomrule
    \end{tabular}
    \caption{
        Data efficiency results for \textsc{ParrotBenchCPN} programs using variants of random initialization.
        FT-R and FT-Z mean the surrogate initialization was finetuned using random-padded and zero-padded inputs, respectively.
        EV-R and EV-Z mean the surrogate initialization was evaluated using random-padded and zero-padded inputs, respectively.
    }\label{fig:rnd_ft_ev_modes}
\end{figure*}

\begin{figure*}
\centering
\begin{tabular}{llll}
    \toprule
    Program & PTS-R FT-R EV-R & PTS-R FT-R EV-Z & PTS-R FT-Z EV-Z \\ 
    \midrule
    \texttt{fft} (0) & $0.06\times$ & $0.06\times$ & $\mathbf{1.53}\times$ \\
    \texttt{fft} (1) & $0.17\times$ & $0.19\times$ & $\mathbf{0.76}\times$ \\
    \texttt{invk2j} (0) & $0.50\times$ & $0.59\times$ & $\mathbf{1.18}\times$ \\
    \texttt{invk2j} (1) & $0.28\times$ & $0.29\times$ & $\mathbf{0.77}\times$ \\
    \texttt{kmeans} & $1.77\times$ & $1.80\times$ & $\mathbf{2.28}\times$ \\
    \texttt{sobel} & $\mathbf{0.85}\times$ & $\mathbf{0.85}\times$ & $\mathbf{0.85}\times$ \\
    \bottomrule
    \end{tabular}
        \vspace{1em}
    
    \begin{tabular}{llll}
    \toprule
    Dataset Size & PTS-R FT-R EV-R & PTS-R FT-R EV-Z & PTS-R FT-Z EV-Z \\ 
    \midrule
    $0\%$ & $\mathbf{1.38}\times$ & $\mathbf{1.38}\times$ & $\mathbf{1.38}\times$ \\
    $0.1\%$ & $0.05\times$ & $0.06\times$ & $\mathbf{1.26}\times$ \\
    $1\%$ & $0.11\times$ & $0.12\times$ & $\mathbf{1.03}\times$ \\
    $10\%$ & $0.60\times$ & $0.62\times$ & $\mathbf{0.94}\times$ \\
    $100\%$ & $1.33\times$ & $\mathbf{1.45}\times$ & $1.07\times$ \\
    \bottomrule
    \end{tabular}
        \vspace{1em}
    
    \begin{tabular}{llll}
    \toprule
    Statistic & PTS-R FT-R EV-R & PTS-R FT-R EV-Z & PTS-R FT-Z EV-Z \\ 
    \midrule
    0th & $\num{3.50e-04}\times$ & $\num{4.76e-04}\times$ & $\mathbf{0.15}\times$ \\
    25th & $0.26\times$ & $0.35\times$ & $\mathbf{0.75}\times$ \\
    50th & $0.73\times$ & $0.77\times$ & $\mathbf{1.07}\times$ \\
    75th & $1.38\times$ & $1.39\times$ & $\mathbf{1.65}\times$ \\
    100th & $28.05\times$ & $\mathbf{28.24}\times$ & $28.03\times$  \\
    \midrule
    MPI & 66th & 65th & \textbf{47}th \\ 
    \midrule
    GM & $0.36\times$ & $0.39\times$ & $\mathbf{1.13}\times$ \\ 
    \bottomrule
    \end{tabular}
    \caption{
        Data efficiency results for \textsc{ParrotBenchCPN} using pretrained surrogates trained with random-padded inputs.
        FT-R and FT-Z mean the surrogate initialization was finetuned using random-padded and zero-padded inputs, respectively.
        EV-R and EV-Z mean the surrogate initialization was evaluated using random-padded and zero-padded inputs, respectively.
    }\label{fig:pts_r_ft_ev_modes}
\end{figure*}

\begin{figure*}
    \centering
    \begin{tabular}{llll}
        \toprule
        Program & PTS-Z FT-R EV-R & PTS-Z FT-R EV-Z & PTS-Z FT-Z EV-Z \\ 
        \midrule
        \texttt{fft} (0) & $0.06\times$ & $0.06\times$ & $\mathbf{0.95}\times$ \\
        \texttt{fft} (1) & $0.24\times$ & $0.33\times$ & $\mathbf{1.04}\times$ \\
        \texttt{invk2j} (0) & $0.53\times$ & $0.62\times$ & $\mathbf{1.25}\times$ \\
        \texttt{invk2j} (1) & $0.35\times$ & $0.37\times$ & $\mathbf{1.08}\times$ \\
        \texttt{kmeans} & $0.95\times$ & $0.93\times$ & $\mathbf{1.31}\times$ \\
        \texttt{sobel} & $\mathbf{1.07}\times$ & $\mathbf{1.07}\times$ & $\mathbf{1.07}\times$ \\
        \bottomrule
        \end{tabular}
        \vspace{1em}
        
        \begin{tabular}{llll}
        \toprule
        Dataset Size & PTS-Z FT-R EV-R & PTS-Z FT-R EV-Z & PTS-Z FT-Z EV-Z \\ 
        \midrule
        $0\%$ & $\mathbf{1.58}\times$ & $\mathbf{1.58}\times$ & $\mathbf{1.58}\times$ \\
        $0.1\%$ & $0.06\times$ & $0.09\times$ & $\mathbf{1.17}\times$ \\
        $1\%$ & $0.10\times$ & $0.10\times$ & $\mathbf{1.09}\times$ \\
        $10\%$ & $0.80\times$ & $0.84\times$ & $\mathbf{1.12}\times$ \\
        $100\%$ & $0.91\times$ & $\mathbf{0.96}\times$ & $0.74\times$ \\
        \bottomrule
        \end{tabular}
        \vspace{1em}
        
        \begin{tabular}{llll}
        \toprule
        Statistic & PTS-Z FT-R EV-R & PTS-Z FT-R EV-Z & PTS-Z FT-Z EV-Z \\ 
        \midrule
        0th & $\num{5.52e-04}\times$ & $\num{6.16e-04}\times$ & $\mathbf{0.07}\times$ \\
        25th & $0.37\times$ & $0.47\times$ & $\mathbf{0.87}\times$ \\
        50th & $0.84\times$ & $0.84\times$ & $\mathbf{1.10}\times$ \\
        75th & $1.13\times$ & $1.19\times$ & $\mathbf{1.41}\times$ \\
        100th & $10.94\times$ & $\mathbf{12.86}\times$ & $7.41\times$ \\
        \midrule
        MPI & 66th & 65th & \textbf{41}st \\ 
        \midrule
        GM & $0.37\times$ & $0.41\times$ & $\mathbf{1.11}\times$ \\ 
        \bottomrule
        \end{tabular}
    \caption{
        Data efficiency results for \textsc{ParrotBenchCPN} using pretrained surrogates trained with zero-padded inputs.
        FT-R and FT-Z mean the surrogate initialization was finetuned using random-padded and zero-padded inputs, respectively.
        EV-R and EV-Z mean the surrogate initialization was evaluated using random-padded and zero-padded inputs, respectively.
    }\label{fig:pts_z_ft_ev_modes}
\end{figure*}

\begin{figure*}
    \centering
    \begin{tabular}{llll}
        \toprule
        Program & MAML-R FT-R EV-R & MAML-R FT-R EV-Z & MAML-R FT-Z EV-Z \\ 
        \midrule
        \texttt{fft} (0) & $0.06\times$ & $0.07\times$ & $\mathbf{1.30}\times$ \\
        \texttt{fft} (1) & $0.32\times$ & $0.45\times$ & $\mathbf{1.11}\times$ \\
        \texttt{invk2j} (0) & $0.39\times$ & $0.48\times$ & $\mathbf{0.82}\times$ \\
        \texttt{invk2j} (1) & $0.64\times$ & $0.75\times$ & $\mathbf{1.13}\times$ \\
        \texttt{kmeans} & $0.63\times$ & $\mathbf{0.65}\times$ & $\mathbf{0.65}\times$ \\
        \texttt{sobel} & $\mathbf{0.44}\times$ & $\mathbf{0.44}\times$ & $\mathbf{0.44}\times$ \\
        \bottomrule
        \end{tabular}
        \vspace{1em}
        
        \begin{tabular}{llll}
        \toprule
        Dataset Size & MAML-R FT-R EV-R & MAML-R FT-R EV-Z & MAML-R FT-Z EV-Z \\ 
        \midrule
        $0\%$ & $\mathbf{0.97}\times$ & $\mathbf{0.97}\times$ & $\mathbf{0.97}\times$ \\
        $0.1\%$ & $0.06\times$ & $0.08\times$ & $\mathbf{1.12}\times$ \\
        $1\%$ & $0.12\times$ & $0.15\times$ & $\mathbf{0.82}\times$ \\
        $10\%$ & $0.51\times$ & $0.54\times$ & $\mathbf{0.56}\times$ \\
        $100\%$ & $1.11\times$ & $\mathbf{1.33}\times$ & $0.90\times$ \\
        \bottomrule
        \end{tabular}
        \vspace{1em}
        
        \begin{tabular}{llll}
        \toprule
        Statistic & MAML-R FT-R EV-R & MAML-R FT-R EV-Z & MAML-R FT-Z EV-Z \\ 
        \midrule
        0th & $\num{7.16e-04}\times$ & $\num{7.54e-04}\times$ & $\mathbf{0.07}\times$ \\
        25th & $0.27\times$ & $0.32\times$ & $\mathbf{0.52}\times$ \\
        50th & $0.54\times$ & $0.58\times$ & $\mathbf{0.87}\times$ \\
        75th & $0.95\times$ & $0.99\times$ & $\mathbf{1.39}\times$ \\
        100th & $12.03\times$ & $\mathbf{16.94}\times$ & $15.18\times$ \\
        \midrule
        MPI & 79th & 76th & \textbf{65}th \\ 
        \midrule
        GM & $0.33\times$ & $0.39\times$ & $\mathbf{0.85}\times$ \\ 
        \bottomrule
        \end{tabular}
    \caption{
        Data efficiency results for \textsc{ParrotBenchCPN} using MAML initializations trained with random-padded inputs.
        FT-R and FT-Z mean the surrogate initialization was finetuned using random-padded and zero-padded inputs, respectively.
        EV-R and EV-Z mean the surrogate initialization was evaluated using random-padded and zero-padded inputs, respectively.
    }\label{fig:maml_r_ft_ev_modes}
\end{figure*}

\begin{figure*}
    \centering
    \begin{tabular}{llll}
        \toprule
        Program & MAML-Z FT-R EV-R & MAML-Z FT-R EV-Z & MAML-Z FT-Z EV-Z \\ 
        \midrule
        \texttt{fft} (0) & $0.18\times$ & $0.21\times$ & $\mathbf{2.85}\times$ \\
        \texttt{fft} (1) & $0.54\times$ & $0.69\times$ & $\mathbf{1.16}\times$ \\
        \texttt{invk2j} (0) & $0.62\times$ & $0.77\times$ & $\mathbf{1.35}\times$ \\
        \texttt{invk2j} (1) & $0.50\times$ & $0.56\times$ & $\mathbf{1.11}\times$ \\
        \texttt{kmeans} & $0.88\times$ & $0.91\times$ & $\mathbf{1.04}\times$ \\
        \texttt{sobel} & $\mathbf{0.89}\times$ & $\mathbf{0.89}\times$ & $\mathbf{0.89}\times$ \\
        \bottomrule
        \end{tabular}
        \vspace{1em}
        
        \begin{tabular}{llll}
        \toprule
        Dataset Size & MAML-Z FT-R EV-R & MAML-Z FT-R EV-Z & MAML-Z FT-Z EV-Z \\ 
        \midrule
        $0\%$ & $\mathbf{1.35}\times$ & $1.34\times$ & $1.34\times$ \\
        $0.1\%$ & $0.06\times$ & $0.08\times$ & $\mathbf{1.40}\times$ \\
        $1\%$ & $0.14\times$ & $0.17\times$ & $\mathbf{1.38}\times$ \\
        $10\%$ & $1.34\times$ & $\mathbf{1.44}\times$ & $1.16\times$ \\
        $100\%$ & $2.60\times$ & $\mathbf{2.89}\times$ & $1.19\times$ \\
        \bottomrule
        \end{tabular}
        \vspace{1em}
        
        \begin{tabular}{llll}
        \toprule
        Statistic & MAML-Z FT-R EV-R & MAML-Z FT-R EV-Z & MAML-Z FT-Z EV-Z \\ 
        \midrule
        0th & $\num{7.14e-04}\times$ & $\num{1.08e-03}\times$ & $\mathbf{0.29}\times$ \\
        25th & $0.41\times$ & $0.46\times$ & $\mathbf{0.86}\times$ \\
        50th & $0.88\times$ & $0.91\times$ & $\mathbf{1.10}\times$ \\
        75th & $1.42\times$ & $1.43\times$ & $\mathbf{1.50}\times$ \\
        100th & $57.67\times$ & $\mathbf{76.44}\times$ & $13.85\times$ \\
        \midrule
        MPI & 56th & 55th & \textbf{39}th \\ 
        \midrule
        GM & $0.53\times$ & $0.61\times$ & $\mathbf{1.29}\times$ \\ 
        \bottomrule
        \end{tabular}
    \caption{
        Data efficiency results for \textsc{ParrotBenchCPN} using MAML initializations trained with zero-padded inputs.
        FT-R and FT-Z mean the surrogate initialization was finetuned using random-padded and zero-padded inputs, respectively.
        EV-R and EV-Z mean the surrogate initialization was evaluated using random-padded and zero-padded inputs, respectively.
    }\label{fig:maml_z_ft_ev_modes}
\end{figure*}

\begin{figure*}
\centering
\begin{tabular}{llll}
    \toprule
    Program & CPN-R FT-R EV-R & CPN-R FT-R EV-Z & CPN-R FT-Z EV-Z \\ 
    \midrule
    \texttt{fft} (0) & $0.88\times$ & $0.99\times$ & $\mathbf{7.18}\times$ \\
    \texttt{fft} (1) & $0.48\times$ & $0.73\times$ & $\mathbf{1.17}\times$ \\
    \texttt{invk2j} (0) & $0.56\times$ & $0.73\times$ & $\mathbf{1.09}\times$ \\
    \texttt{invk2j} (1) & $0.55\times$ & $0.62\times$ & $\mathbf{1.04}\times$ \\
    \texttt{kmeans} & $4.12\times$ & $4.26\times$ & $\mathbf{5.22}\times$ \\
    \texttt{sobel} & $\mathbf{1.14}\times$ & $\mathbf{1.14}\times$ & $\mathbf{1.14}\times$ \\
    \bottomrule
    \end{tabular}
    \vspace{1em}
    
    \begin{tabular}{llll}
    \toprule
    Dataset Size & CPN-R FT-R EV-R & CPN-R FT-R EV-Z & CPN-R FT-Z EV-Z \\ 
    \midrule
    $0\%$ & $\mathbf{1.42}\times$ & $\mathbf{1.42}\times$ & $\mathbf{1.42}\times$ \\
    $0.1\%$ & $0.09\times$ & $0.14\times$ & $\mathbf{2.33}\times$ \\
    $1\%$ & $1.11\times$ & $1.44\times$ & $\mathbf{2.56}\times$ \\
    $10\%$ & $1.89\times$ & $1.99\times$ & $\mathbf{2.18}\times$ \\
    $100\%$ & $2.42\times$ & $\mathbf{2.57}\times$ & $1.57\times$ \\
    \bottomrule
    \end{tabular}
        \vspace{1em}
    
    \begin{tabular}{llll}
    \toprule
    Statistic & CPN-R FT-R EV-R & CPN-R FT-R EV-Z & CPN-R FT-Z EV-Z \\ 
    \midrule
    0th & $\num{1.18e-03}\times$ & $\num{1.47e-03}\times$ & $\mathbf{0.19}\times$ \\
    25th & $0.49\times$ & $0.60\times$ & $\mathbf{0.86}\times$ \\
    50th & $0.99\times$ & $1.01\times$ & $\mathbf{1.22}\times$ \\
    75th & $2.17\times$ & $2.20\times$ & $\mathbf{2.31}\times$ \\
    100th & $171.49\times$ & $191.02\times$ & $\mathbf{1478.96}\times$ \\
    \midrule
    MPI & 51st & 48th & \textbf{35}th \\ 
    \midrule
    GM & $0.92\times$ & $1.08\times$ & $\mathbf{1.96}\times$ \\ 
    \bottomrule
    \end{tabular}
    \caption{
        Data efficiency results for \textsc{ParrotBenchCPN} programs using \textsc{CompNet}s trained on random-padded inputs.
        FT-R and FT-Z mean the surrogate initialization was finetuned using random-padded and zero-padded inputs, respectively.
        EV-R and EV-Z mean the surrogate initialization was evaluated using random-padded and zero-padded inputs, respectively.
    }\label{fig:cpn_r_ft_ev_modes}
\end{figure*}

\begin{figure*}
\centering
\begin{tabular}{llll}
    \toprule
    Program & CPN-Z FT-R EV-R & CPN-Z FT-R EV-Z & CPN-Z FT-Z EV-Z \\ 
    \midrule
    \texttt{fft} (0) & $0.02\times$ & $0.02\times$ & $\mathbf{0.58}\times$ \\
    \texttt{fft} (1) & $0.24\times$ & $0.28\times$ & $\mathbf{1.06}\times$ \\
    \texttt{invk2j} (0) & $0.47\times$ & $0.61\times$ & $\mathbf{1.07}\times$ \\
    \texttt{invk2j} (1) & $0.60\times$ & $0.67\times$ & $\mathbf{1.62}\times$ \\
    \texttt{kmeans} & $1.45\times$ & $1.51\times$ & $\mathbf{1.78}\times$ \\
    \texttt{sobel} & $\mathbf{0.91}\times$ & $\mathbf{0.91}\times$ & $\mathbf{0.91}\times$ \\
    \bottomrule
    \end{tabular}
        \vspace{1em}
    
    \begin{tabular}{llll}
    \toprule
    Dataset Size & CPN-Z FT-R EV-R & CPN-Z FT-R EV-Z & CPN-Z FT-Z EV-Z \\ 
    \midrule
    $0\%$ & $1.50\times$ & $\mathbf{1.51}\times$ & $\mathbf{1.51}\times$ \\
    $0.1\%$ & $0.05\times$ & $0.07\times$ & $\mathbf{0.88}\times$ \\
    $1\%$ & $0.10\times$ & $0.12\times$ & $\mathbf{1.33}\times$ \\
    $10\%$ & $0.40\times$ & $0.45\times$ & $\mathbf{0.66}\times$ \\
    $100\%$ & $1.65\times$ & $\mathbf{1.75}\times$ & $1.36\times$ \\
    \bottomrule
    \end{tabular}
        \vspace{1em}
    
    \begin{tabular}{llllll}
    \toprule
    Statistic & CPN-Z FT-R EV-R & CPN-Z FT-R EV-Z & CPN-Z FT-Z EV-Z \\ 
    \midrule
    0th & $\num{3.05e-04}\times$ & $\mathbf{5.88 \cdot 10^{-4}}\times$ & $\num{1.52e-04}\times$ \\
    25th & $0.39\times$ & $0.42\times$ & $\mathbf{0.77}\times$ \\
    50th & $0.79\times$ & $0.82\times$ & $\mathbf{1.14}\times$ \\
    75th & $1.31\times$ & $1.37\times$ & $\mathbf{1.60}\times$ \\
    100th & $70.29\times$ & $\mathbf{70.97}\times$ & $69.18\times$ \\
    \midrule
    MPI & 63rd & 59th & \textbf{38}th \\ 
    \midrule
    GM & $0.35\times$ & $0.39\times$ & $\mathbf{1.10}\times$ \\ 
    \bottomrule
    \end{tabular}
    \caption{
        Data efficiency results for \textsc{ParrotBenchCPN} programs using \textsc{CompNet}s trained on zero-padded inputs.
        FT-R and FT-Z mean the surrogate initialization was finetuned using random-padded and zero-padded inputs, respectively.
        EV-R and EV-Z mean the surrogate initialization was evaluated using random-padded and zero-padded inputs, respectively.
    }\label{fig:cpn_z_ft_ev_modes}
\end{figure*}

\section{Variable-Output Support for Initialization Methods}\label{sec:variable_output_strategies}
Recall, all programs in \textsc{ExeStackCPN} have a single output (Section~\ref{sec:exe_stack}).
However, the \texttt{fft} and \texttt{invk2j} benchmarks in \textsc{ParrotBenchCPN} have multiple outputs.
In this appendix, we propose and evaluate a set of strategies to adapt initialization methods trained on \textsc{ExeStackCPN} to support variable-output programs.

\paragraph{Methodology.}
For each initialization method, we produce a neural surrogate initialization, then we apply one of the following strategies:
\begin{itemize}
    \item \textbf{Grow:} Use the initialization produced by the method and extend the final layer with randomly initialized weights to reach the target number of outputs.
    \item \textbf{Reinitialize:} Use the initialization produced by the initialization method but randomly initialize the final layer, sized to match the target number of outputs.
    \item \textbf{Clone:} Use the initialization produced by the initialization method but duplicate the weights for the one active output in the final layer of the initialization, to generate weights for the target number of outputs.
\end{itemize}

To decide which strategy to use for each initialization method, we performed the \textsc{ParrotBenchCPN} data efficiency evaluation of Section~\ref{sec:data_efficiency}, and we swept over a set of variable-output strategies applied to each initialization method.
We used initialization methods that support variable-input programs, using the best strategies from Appendix~\ref{sec:padding_for_variable_inputs}.
For each initialization method, we choose the strategy that achieves the greatest overall test loss improvement over random initialization.

\paragraph{Results.}
We present the results for \textsc{CompNet}s, MAML, and pretrained surrogates in Figures~\ref{fig:compnet_variable_output}, \ref{fig:maml_variable_output}, and \ref{fig:pts_variable_output}.

\begin{figure*}
    \centering
    \begin{tabular}{llll}
        \toprule
        Program & CPN-R Z/Z (Grow) & CPN-R Z/Z (Reinit) & CPN-R Z/Z (Clone) \\
        \midrule
        \texttt{fft} & $0.95\times$ & $\mathbf{1.49}\times$ & $1.47\times$ \\
        \texttt{invk2j} & $0.86\times$ & $\mathbf{1.01}\times$ & $\mathbf{1.01}\times$ \\
        \texttt{kmeans} & $\mathbf{7.85}\times$ & $1.77\times$ & $\mathbf{7.85}\times$ \\
        \texttt{sobel} & $\mathbf{1.14}\times$ & $1.12\times$ & $\mathbf{1.14}\times$ \\
        \bottomrule
        \end{tabular}
        \vspace{1em}
        
        \begin{tabular}{llll}
        \toprule
        Dataset Size & CPN-R Z/Z (Grow) & CPN-R Z/Z (Reinit) & CPN-R Z/Z (Clone) \\
        \midrule
        $0\%$ & $\mathbf{1.86}\times$ & $0.95\times$ & $1.81\times$ \\
        $0.1\%$ & $1.61\times$ & $1.46\times$ & $\mathbf{1.98}\times$ \\
        $1\%$ & $1.49\times$ & $1.40\times$ & $\mathbf{1.77}\times$ \\
        $10\%$ & $2.13\times$ & $1.93\times$ & $\mathbf{2.38}\times$ \\
        $100\%$ & $1.26\times$ & $1.05\times$ & $\mathbf{1.68}\times$ \\
        \bottomrule
        \end{tabular}
        \vspace{1em}
        
        \begin{tabular}{llll}
        \toprule
        Statistic & CPN-R Z/Z (Grow) & CPN-R Z/Z (Reinit) & CPN-R Z/Z (Clone) \\
        \midrule
        0th & $0.17\times$ & $\mathbf{0.33}\times$ & $0.22\times$ \\
        25th & $0.79\times$ & $0.85\times$ & $\mathbf{0.88}\times$ \\
        50th & $1.05\times$ & $1.13\times$ & $\mathbf{1.23}\times$ \\
        75th & $1.96\times$ & $1.76\times$ & $\mathbf{2.96}\times$ \\
        100th & $\mathbf{106.91}\times$ & $31.55\times$ & $\mathbf{106.91}\times$ \\
        \midrule
        MPI & 42nd & \textbf{33}rd & 36th \\
        \midrule
        GM & $1.64\times$ & $1.31\times$ & $\mathbf{1.91}\times$ \\
        \bottomrule
        \end{tabular}
    \caption{
        Data efficiency results for \textsc{ParrotBenchCPN} programs using \textsc{CompNet}s trained on various variable-output strategies.
        CPN-R means we train the \textsc{CompNet}s on random-padded inputs.
        Z/Z means we finetune and evaluate \textsc{CompNet}-initialized surrogates on zero-padded inputs (see Appendix~\ref{sec:padding_for_variable_inputs}).
    }\label{fig:compnet_variable_output}
\end{figure*}

\begin{figure*}
    \centering
    \begin{tabular}{llll}
        \toprule
        Program & MAML-Z Z/Z (Grow) & MAML-Z Z/Z (Reinit) & MAML-Z Z/Z (Clone) \\
        \midrule
        \texttt{fft} & $0.65\times$ & $\mathbf{0.98}\times$ & $0.63\times$ \\
        \texttt{invk2j} & $1.06\times$ & $\mathbf{1.07}\times$ & $0.88\times$ \\
        \texttt{kmeans} & $\mathbf{0.94}\times$ & $0.68\times$ & $\mathbf{0.94}\times$ \\
        \texttt{sobel} & $0.89\times$ & $\mathbf{1.06}\times$ & $0.89\times$ \\
        \bottomrule
        \end{tabular}
        
        \vspace{1em}
        \begin{tabular}{llll}
        \toprule
        Dataset Size & MAML-Z Z/Z (Grow) & MAML-Z Z/Z (Reinit) & MAML-Z Z/Z (Clone) \\
        \midrule
        $0\%$ & $\mathbf{1.42}\times$ & $0.90\times$ & $\mathbf{1.42}\times$ \\
        $0.1\%$ & $0.92\times$ & $\mathbf{0.94}\times$ & $0.73\times$ \\
        $1\%$ & $0.73\times$ & $\mathbf{0.93}\times$ & $0.51\times$ \\
        $10\%$ & $0.88\times$ & $\mathbf{1.11}\times$ & $0.94\times$ \\
        $100\%$ & $0.60\times$ & $\mathbf{0.81}\times$ & $0.75\times$ \\
        \bottomrule
        \end{tabular}
        
        \vspace{1em}
        \begin{tabular}{llll}
        \toprule
        Statistic & MAML-Z Z/Z (Grow) & MAML-Z Z/Z (Reinit) & MAML-Z Z/Z (Clone) \\
        \midrule
        0th & $0.15\times$ & $\mathbf{0.28}\times$ & $0.05\times$ \\
        25th & $0.64\times$ & $\mathbf{0.82}\times$ & $0.64\times$ \\
        50th & $0.92\times$ & $\mathbf{0.97}\times$ & $0.85\times$ \\
        75th & $1.14\times$ & $1.14\times$ & $\mathbf{1.16}\times$ \\
        100th & $4.01\times$ & $1.99\times$ & $\mathbf{8.00}\times$ \\
        \midrule
        MPI & 58th & \textbf{54}th & 66th \\
        \midrule
        GM & $0.87\times$ & $\mathbf{0.93}\times$ & $0.82\times$ \\
        \bottomrule
        \end{tabular}

    \caption{
        Data efficiency results for \textsc{ParrotBenchCPN} programs using MAML initializations trained on various variable-output strategies.
        MAML-Z means we train the MAML initializations on zero-padded inputs.
        Z/Z means we finetune and evaluate MAML-initialized surrogates on zero-padded inputs (see Appendix~\ref{sec:padding_for_variable_inputs}).
    }\label{fig:maml_variable_output}
\end{figure*}

\begin{figure*}
    \centering
    \begin{tabular}{llll}
        \toprule
        Program & PTS-R Z/Z (Grow) & PTS-R Z/Z (Reinit) & PTS-R Z/Z (Clone) \\
        \midrule
        \texttt{fft} & $0.61\times$ & $\mathbf{0.88}\times$ & $0.46\times$ \\
        \texttt{invk2j} & $1.05\times$ & $0.95\times$ & $\mathbf{1.12}\times$ \\
        \texttt{kmeans} & $\mathbf{2.24}\times$ & $0.65\times$ & $\mathbf{2.24}\times$ \\
        \texttt{sobel} & $0.85\times$ & $\mathbf{0.92}\times$ & $0.85\times$ \\
        \bottomrule
        \end{tabular}
        
        \vspace{1em}
        \begin{tabular}{llll}
        \toprule
        Dataset Size & PTS-R Z/Z (Grow) & PTS-R Z/Z (Reinit) & PTS-R Z/Z (Clone) \\
        \midrule
        $0\%$ & $1.56\times$ & $0.79\times$ & $\mathbf{1.65}\times$ \\
        $0.1\%$ & $\mathbf{0.98}\times$ & $0.93\times$ & $0.81\times$ \\
        $1\%$ & $0.79\times$ & $\mathbf{0.87}\times$ & $0.75\times$ \\
        $10\%$ & $\mathbf{1.23}\times$ & $1.00\times$ & $1.00\times$ \\
        $100\%$ & $0.86\times$ & $0.67\times$ & $\mathbf{0.98}\times$ \\
        \bottomrule
        \end{tabular}
        
        \vspace{1em}
        \begin{tabular}{llll}
        \toprule
        Statistic & PTS-R Z/Z (Grow) & PTS-R Z/Z (Reinit) & PTS-R Z/Z (Clone) \\
        \midrule
        0th & $\mathbf{0.23}\times$ & $0.22\times$ & $0.21\times$ \\
        25th & $0.75\times$ & $\mathbf{0.77}\times$ & $0.65\times$ \\
        50th & $\mathbf{0.97}\times$ & $0.93\times$ & $0.85\times$ \\
        75th & $1.26\times$ & $1.04\times$ & $\mathbf{1.40}\times$ \\
        100th & $\mathbf{38.18}\times$ & $1.86\times$ & $\mathbf{38.18}\times$ \\
        \midrule
        MPI & \textbf{54}th & 69th & 63rd \\
        \midrule
        GM & $\mathbf{1.05}\times$ & $0.84\times$ & $1.00\times$ \\
        \bottomrule
        \end{tabular}

    \caption{
        Data efficiency results for \textsc{ParrotBenchCPN} programs using pretrained initializations trained on various variable-output strategies.
        PTS-R means we train the pretrained initializations on random-padded inputs.
        Z/Z means we finetune and evaluate pretrain-initialized surrogates on zero-padded inputs (see Appendix~\ref{sec:padding_for_variable_inputs}).
    }\label{fig:pts_variable_output}
\end{figure*}

The best-performing strategy for \textsc{CompNet}s is cloning, with a geometric mean test loss improvement of $1.91\times$, the best-performing strategy for MAML is reinitialization, with a geometric mean test loss improvement of $0.93\times$, and the best-performing strategy for pretrained surrogates is growing, with a geometric mean test loss improvement of $1.05\times$.
Note that the \texttt{fft} and \texttt{invk2j} benchmarks are the only programs where the variable-output strategies are necessary, but we perform each strategy indiscriminately.
This indiscriminate application harms performance for the reinitialization strategy on \texttt{kmeans} and \texttt{sobel} when using \textsc{CompNet}s and pretrained surrogates.
For \textsc{CompNet}s in particular, if we only applied each strategy where necessary, reinitialization would have outperformed cloning by a small margin.

\paragraph{Conclusion.}
In light of these results, we make the following decisions.
We choose the cloning strategy for \textsc{CompNet}s, the reinitialization strategy for MAML, and the growing strategy for pretrained surrogates.

\end{document}